\documentclass{article} %
\usepackage{iclr2027_conference,times}

\usepackage{amsmath,amsfonts,bm}

\def\eqref#1{equation~\ref{#1}}

\def\1{\bm{1}}

\DeclareMathAlphabet{\mathsfit}{\encodingdefault}{\sfdefault}{m}{sl}
\SetMathAlphabet{\mathsfit}{bold}{\encodingdefault}{\sfdefault}{bx}{n}

\usepackage{hyperref}
\usepackage{url}

\usepackage{graphicx}
\usepackage{microtype}
\usepackage{subcaption}
\usepackage{booktabs}
\usepackage{wrapfig}

\usepackage{amsthm}
\newtheorem{theorem}{Theorem}[section]
\newtheorem{corollary}[theorem]{Corollary}

\iclrfinalcopy

\title{Don't Forget! Decomposing the Training Dynamics of Memorization in Language Models}

\makeatletter
\def\@fnsymbol#1{\ensuremath{\ifcase#1\or *\or \dagger\or \ddagger\or
   \mathsection\or \mathparagraph\or \|\or **\or \dagger\dagger
   \or \ddagger\ddagger \else\@ctrerr\fi}}

\renewcommand{\author}[1]{%
  \gdef\@author{%
    \begin{minipage}{\textwidth}
      \centering
      \vspace{0.5cm}
      #1
    \end{minipage}%
  }%
}
\makeatother

\author{\textbf{\centering Florian Eichin$^1$ \ \ \ \ Philipp Mondorf$^1$ \ \ \ \ Andrei Mircea$^2$ \ \ \ \ Yupei Du$^3$ \\ Barbara Plank$^1$ \ \ \ \ \ \ \ \ \  Michael A. Hedderich$^1$} \\ \vspace{8pt} 
\textnormal{$^1$MaiNLP, Center for Information and Language Processing, LMU Munich \\
\& Munich Center for Machine Learning (MCML), Germany \\
$^2$Mila – Quebec AI Institute \& University of Montreal, Canada \\
$^3$Saarland University, Germany \\
Corresponcence to \texttt{feichin@cis.lmu.de}}}

\begin{document}

\maketitle

\begin{abstract}
    Memorization has been proposed as a mechanism to explain how language models fit the tail of their training distributions, but its training dynamics are not understood well. In this work, we take a fine-grained look at memorization by decomposing the loss trajectory of memorized sequences over training and model parameters. Across the Pythia family, we study memorization of duplicated training sequences (recitation) and rare ones (recollection). We find that memorization in both cases is characterized by sequence-level gradient alignment, though recitation suffers from misalignment with other training influences which causes forgetting, explaining the necessity for higher duplication of these examples. We further show that the lower model layers are the most involved in memorization and forgetting. Predicting memorization, our decomposition improves over a cross-entropy baseline, especially in larger models and early in training. Intervening on a small set of highly influential parameters we are able to ablate memorization in the final model. Together, these findings advance our understanding of how memorization develops during training and offer insights for predicting and intervening on it.
\end{abstract}

\section{Introduction}

Language models can reproduce training sequences verbatim, a behavior known as \emph{memorization} \citep{feldman-2020-memorization}. Despite a growing body of work on this phenomenon, our understanding of how and where memorization develops in the model during training remains limited. This restricts our ability to predict and intervene on memorization, which is desirable for various applications, for instance where privacy or data security are of concern \citep{biderman2026position}. In this study, we look at memorization as a training dynamic. %
In particular, we ask which learning contributions along the dimensions of parameters, data, and training process characterize memorization and whether such understanding is actionable for prediction and intervention.

We extend the recently proposed ExPLAIND decomposition \citep{eichin2026explaind}, %
to map a model's loss trajectory onto its influences along training, parameters, and data. From this decomposition, we derive an intuitive mathematical interpretation of how learning mechanisms like memorization are the result of three aspects: The alignment of a prediction's gradients to the model parameter update, the magnitude of the training updates, and the sensitivity of a prediction to such updates. We apply ExPLAIND across the pretraining trajectories of the Pythia model family, %
where we examine recitation (memorization of highly frequent training sequences) and recollection (memorization of rare training sequences). Previous work has shown recitation to be the dominant mode of verbatim memorization \citep{prashanth2025recite}, but it is unclear why their memorization requires duplication, while recollected sequences do not.

We study the training dynamics of memorization in Pythia and show that a large fraction of memorization happens in very early pretraining before basic generalization has manifested. Our decomposition suggests two aspects of memorization: a sequence's alignment with parameter updates from its own occurrences in training as well as its interaction with the remaining training influences. Memorized sequences exhibit high gradient alignment with their own occurrences, particularly in lower model layers. Highly duplicated, recited sequences tend to exhibit stronger negative alignment with both other training data and weight-decay updates throughout training, suggesting that repeated exposure helps avoid forgetting. %
On the other hand, recollected sequences with few training occurrences exhibit less negative alignment, consistent with their memorization despite low duplication.

To validate our understanding, we test these explanations through early prediction and post-hoc intervention. Features derived from the decomposition improve memorization prediction over a cross-entropy baseline, particularly in larger models and early in training. Interventions on a small set of parameters identified as influential early in pretraining, mostly in lower attention and MLP layers, ablate memorization in the final model.

\section{Related Work}

Memorization has been proposed as a mechanism for deep learning models to fit long-tail distributions \citep{pmlr-v70-arpit17a,pmlr-v80-chatterjee18a,feldman-2020-memorization}, and has been validated and studied in various works \citep{feldmanWhatNeuralNetworks2020,zhengEmpiricalStudyMemorization2022}. Memorization in Pythia has been studied by \citet{bidermanEmergentPredictableMemorization2023}, who find that memorization is predictable during pretraining and scales with model size. %
Further, \citet{prashanth2025recite} study Pythia's memorization through the lens of training corpus statistics and further argue for a taxonomy for memorization based on duplication, which has been shown to be one factor of memorization \citet{lee-etal-2022-deduplicating}. We follow their approach and investigate memorization with high (recitation) and low duplication (recollection) but we predict memorization in much earlier training phases and with features derived from our loss decomposition. \citet{tirumala2022memorization} also study the training dynamics of memorization and find larger models to memorize faster, which our results contradict, and forget less, which we also find and explain. 

We also localize memorization, i.e., find the model parameters or layers implementing recall of memories, which has been another popular focus \citep{mengLocatingEditingFactual2022,maini-localization-2023,l2m2-ws-2025-1,ortuCompetitionMechanismsTracing2024}, though the faithfulness of existing methods has been questioned \citep{changLocalizationMethodsActually2024}. Previous work has argued that both MLP layers \citep{gevaTransformerFeedForwardLayers2021,daiKnowledgeNeuronsPretrained2022} and attention heads \citep{stoehrLocalizingParagraphMemorization2024} are involved in the mechanisms for recall, which our results confirm. Further, \citet{mireshghallah-etal-2022-empirical} identifies the model embeddings as an important factor for recall. %
Memorization has also been contrasted with generalization by various works \citep{zhengEmpiricalStudyMemorization2022,dankersGeneralisationFirstMemorisation2024}, which show that memorization and generalization are entangled \citep{du-etal-2025-reason,dankers-2026-memorisation}. Our results also indicate an interdependence of generalization and memorization, with both emerging roughly at the same time in early pretraining. %

As regards methodology, gradient-based data attribution relates training examples to the evaluation loss \citep{kohUnderstandingBlackboxPredictions2017,pruthi2020tracin}, while loss change allocation maps it across parameters and training \citep{lanLCALossChange2019,kangaslahtiHiddenBreakthroughsLanguage2025}. Our study is based on \citet{eichin2026explaind}'s decomposition, which builds on work by \citet{bellExactKernelEquivalence2023}, and is the only unified and exact approach combining both. We extend it to an explicit notion of alignment between train influences and the prediction. Further, we validate our findings using prediction and causal intervention over training.

\section{Methodology}
\label{sec:acc_influences}

Our study is based on the ExPLAIND decomposition of gradient descent models by \citet{eichin2026explaind}. Using their central Theorem, we decompose a model's loss into per-sample and per-parameter contributions across training. We extend it by further decomposing their influence scores into parts that can be interpreted as the \emph{sensitivity} of the prediction, the \emph{magnitude} of the parameter update, and the \emph{alignment} of a parameter update with the prediction's gradient. We apply their Theorem 3.1 and Corollary 3.3 to the evaluation loss and state the accumulated result below (full proof in App.~\ref{app:methodology}).

Let $f_\theta$ be a model with parameters $\theta\in\mathbb R^D$ and per-sample loss $L$. For a sequence $x$, define the sequence loss $\ell(\theta;x):=N^{-1}\sum_{x_i \in x}L(f_\theta(x_{<i}),x_{i})$ for some $N \in \mathbb{N}$, used for both training and evaluation. At step $s$, let $\theta_s\to\theta_{s+1}$ be the AdamW update with training batch $B_s$, learning rate $\alpha_s>0$, weight decay $\lambda\geq0$, $\beta_1,\beta_2\in[0,1)$, gradient clipping factor $c_s$, and $\epsilon>0$. Let $m_s$ be the first moment and $\hat v_{s+1}$ the bias-corrected second moment. Then, the loss change decomposes exactly
\begin{equation}
\label{eq:acc_decomp}
\textstyle
    \ell(\theta_{s+1};x)-\ell(\theta_s;x) = \sum_{z\in B_s}\mathcal I_s^z(x) +\mathcal I_s^{\mathrm{fm}}(x)+\mathcal I_s^{\mathrm{reg}}(x),
\end{equation}
where, for each training sequence or update component $\rho\in B_s \cup\{\mathrm{fm},\mathrm{reg}\}$,
\begin{equation}
\label{method:accumulated_influence_terms}
    \mathcal I_s^\rho(x):=-\Phi_s(x)^\top u_s^\rho,
\end{equation}
and we define the path-integrated test gradient and update terms with $z \in B_s$ as
\begin{equation}
\label{eq:path_gradient}
    \Phi_s(x):=\int_0^1\nabla_\theta\ell(\theta_s(t);x)\,dt,\ \ \mathrm{where}\ \  \theta_s(t):=\theta_s+t(\theta_{s+1}-\theta_s),
\end{equation}
\begin{equation}
\label{eq:update_components}
    u_s^z :=\frac{\alpha_s(1-\beta_1)c_s}{1-\beta_1^{s+1}}
       \frac{\nabla_\theta\ell(\theta_s;z)}
            {\sqrt{\hat v_{s+1}}+\epsilon},\ \ \ 
    u_s^{\mathrm{fm}} :=\frac{\alpha_s\beta_1}{1-\beta_1^{s+1}}
       \frac{m_s}{\sqrt{\hat v_{s+1}}+\epsilon},\ \  \ 
    u_s^{\mathrm{reg}}:=\alpha_s\lambda\theta_s.
\end{equation}
All gradients and update components $u_s^\rho$ are column vectors in $\mathbb R^D$; the square root and division are coordinatewise. The vectors $u_s^\rho$ are the update components subtracted from the parameters in a single AdamW parameter update.
Intuitively, the influence terms $\mathcal I_s^\rho$ decompose the training dynamics of a model into interpretable influences that shape the parameter trajectory of the model. For each example $z$, we have  $I_s^z(x)$, for the first moment and weight decay terms, we have $\mathcal I_s^{\mathrm{fm}}(x)$ and $\mathcal I_s^{\mathrm{reg}}(x)$, all quantifying the respective loss changes on $x$ they introduce in training.

Note that each influence term $\mathcal I_s^\rho$ in Eq.~\ref{method:accumulated_influence_terms} is an inner product between the path-integrated test gradient $\Phi_s(x)$ and a component of the parameter update $u_s^\rho$. Writing the inner product in polar form,
\begin{equation}
\label{eq:alignment_polar}
\textstyle
    \mathcal I_s^\rho(x) =-\lVert\Phi_s(x)\rVert\,\lVert u_s^\rho\rVert \cdot\cos\bigl(\Phi_s(x),u_s^\rho\bigr),
\end{equation}
separates each influence into three factors:

\setlength{\leftmargini}{14pt}
\begin{itemize} 
    \item \textbf{Test sensitivity} $\lVert \Phi_s(x) \rVert$: how strongly the test loss reacts to parameter movement. For a test point whose loss landscape is flat, %
    $\Phi_s(x)$ is small and thus ``forgetting" the instance through further updates becomes more difficult.
    
    \item \textbf{Update magnitude} $\lVert u_s^\rho \rVert$: how much the respective update component moves the parameters. Note that due to the division by the second moment, a component is not influential merely because its gradients are large, but because they are large \emph{relative to the recent gradient history}.
    
    \item \textbf{Directional alignment} $\cos(\Phi_s(x), u_s^\rho)$: how much the direction of a parameter update aligns with the direction of the integrated loss gradient of the predicted example $x$. Positive alignment yields negative influence, i.e., the component decreases the test loss; negative alignment increases loss; values close to zero imply that the model is changed without affecting the prediction of $x$.
\end{itemize}

We use this decomposition to analyze how training occurrences of the memorized sequences influence their predictions over the course of training, and how the rest of the training influences work towards or against memorization solutions.

An update may be aligned with a prediction in one layer but not in another and large opposing contributions may cancel at the model level. %
Since we are interested in uncovering such effects and localizing model internals that implement memorization, we follow \citet{eichin2026explaind} and further decompose each influence over model layers. Let $\Theta=\{\Theta_1,\ldots,\Theta_R\}$ be a partitioning of the parameter coordinates. Restricting the vectors of our decomposition to these coordinates gives
\begin{equation}
\label{eq:layerwise_influences}
\textstyle
    \mathcal I_s^\rho(x)=\sum_{r=1}^R\mathcal I_s^\rho(x;\Theta_r),
    \ \ \mathrm{\textit{with}} \ \ \ 
    \mathcal I_s^\rho(x;\Theta_r):=-\bigl(\Phi_s(x)|_{\Theta_r}\bigr)^\top \bigl(u_s^\rho|_{\Theta_r}\bigr). \\
\end{equation}
Thus Eq.~\ref{eq:acc_decomp} also decomposes over parameter groups. Grouping to the layer level as we do below, these influences show where learning happens in the model. Analogous definitions for the parameter-wise test sensitivity $\lVert\Phi_s(x)|_{\Theta_r}\rVert$ and update magnitude $\lVert u_s^\rho|_{\Theta_r}\rVert$ follow. For alignment, we are interested in the contribution to the model-level alignment so we decompose the cosine as 
\begin{equation}
\label{eq:cosine_decomposition}
{\textstyle
    \cos(\Phi_s(x), u_s^\rho) = \sum_{r=1}^{R} \cos(\Phi_s(x), u_s^\rho; \Theta_r),}
    \ \mathrm{\textit{with}} \ 
    \cos(\Phi_s(x), u_s^\rho; \Theta_r) := \frac{\Phi_s(x)|_{\Theta_r} \cdot u_s^\rho|_{\Theta_r}}{\lVert \Phi_s(x) \rVert \lVert u_s^\rho \rVert}.
\end{equation}

\textbf{Experiment settings.} To study the training dynamics of verbatim memorization in LLMs, we use our decomposition to analyze LLMs across the Pythia family \citep{bidermanPythiaSuiteAnalyzing2023}, which provide fine-grained access to checkpoints and training data and for which an exhaustive searches of duplicated and memorized sequences exists. 
We focus on Pythia-$1$B as our object of study, though we perform all experiments for models of size $160$m, $410$m, $1$B, $1.4$B, $2.8$B, and $6.9$B, unless stated otherwise. For all numeric results and figures we present, we link to a version in the Appendix containing the same data for the other models. 
We study memorized sequences over training and decompose their cross entropy (CE) loss, which we sample from an exhaustive dataset of memorized sequences by \citet{bidermanEmergentPredictableMemorization2023} for each model. We use their definition of memorization as $32$-extractability, i.e., whether, given the first $32$ tokens of the sequence, the model greedily generates the next $32$ tokens verbatim. We filter out sequences with repeating subsequences ("reconstruction" in \citet{prashanth2025recite}'s taxonomy) and samples of code, which we found to often contain boilerplate sequences, resulting in memorized examples resembling natural language. We follow \citet{prashanth2025recite}'s taxonomy and further discriminate between \textbf{recollected} (memorized sequences that appear up to five times in the training set) and \textbf{recited} sequences (memorized sequences that appear more than five times in the training set). To contextualize our results, we include two control groups, \textbf{control} and \textbf{control-duplicated}, whose samples also occur up to and more than five times, respectively, but are not 32-extractable in the final model. For each of the four classes, we sample 32 sequences and study their decomposition. Details on datasets and implementations in App.~\ref{app:experiment_settings}.

To our surprise, we could not verify $32$-extractability for around $10-20\%$ sequences sampled from \citet{bidermanEmergentPredictableMemorization2023}'s dataset across model sizes. %
To investigate this discrepancy, we plot the average fraction of the correct continuation token being the maximum logit prediction across all sequences in Fig.~\ref{fig:pythia-extractability-continuous}. The problem only shows for a small fraction of tokens, which indicates the problem is likely due to small numeric differences of our hardware and implementation for obtaining the logits. While we do not consider this a problem for our analysis, it calls into question whether demanding strict 32-extractable yields a robust definition of verbatim memorization.

\textbf{Scaling to LLMs.} We adapt our decomposition to Pythia and to its large training data as follows. First, following \citet{eichin2026explaind}, we do not require access to the full training trajectory: we subsample the training steps to the set $S$ of available checkpoints and decompose a simulated update per checkpoint $\theta_s$. Since the Pythia models do not provide optimizer states, we omit the first-moment term $u_s^{\mathrm{fm}}$ and estimate the second-moment from independent samples of training data (details in App. ~\ref{app:second_moment}). We consider our second moment estimate effective at realistically rescaling the parameter updates, since our simulated parameter updates consistently decrease loss on the memorized and control groups. Second, we sample a training batch from the tail of Pythia's training data, i.e. unseen by any of the checkpoints but the last one, with an original batch size of $2M$ tokens. Using this approach, we also inject our sampled sequences into the batch to simulate their occurrence during training (as proposed by \citealt{körner2026copyfirsttranslatelater}). This way, we can compute the influence a memorized instance would have over training without searching for and decomposing the actual occurrences. %
We compute the simulated parameter update over the resulting batch to obtain $\theta_{s+1}$.

\section{Results}

\label{sec:train_dynamics}

\begin{figure}[t]
    \centering
    \begin{subfigure}{0.2\linewidth}
        \includegraphics[trim=0 0 280 0,clip,width=1\linewidth]{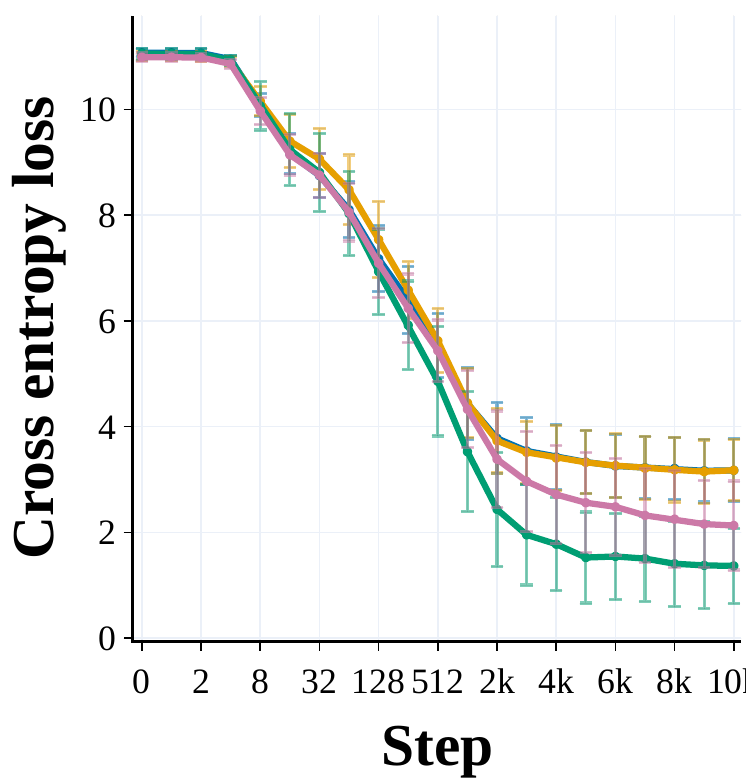}
        \caption{Loss until step 10k}
        \label{fig:pythia-early-loss}
    \end{subfigure}
    \begin{subfigure}{0.2\linewidth}
        \includegraphics[trim=0 0 280 0,clip,width=1\linewidth]{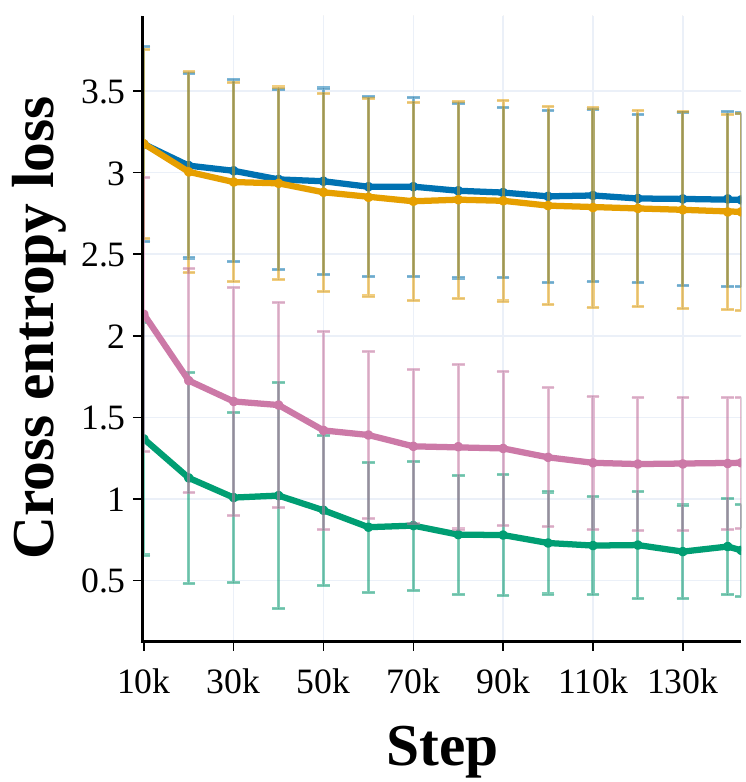}
        \caption{Loss after step 10k}
        \label{fig:pythia-loss}
    \end{subfigure}
    \begin{subfigure}{0.2\linewidth}
        \includegraphics[trim=0 0 280 0,clip,width=1\linewidth]{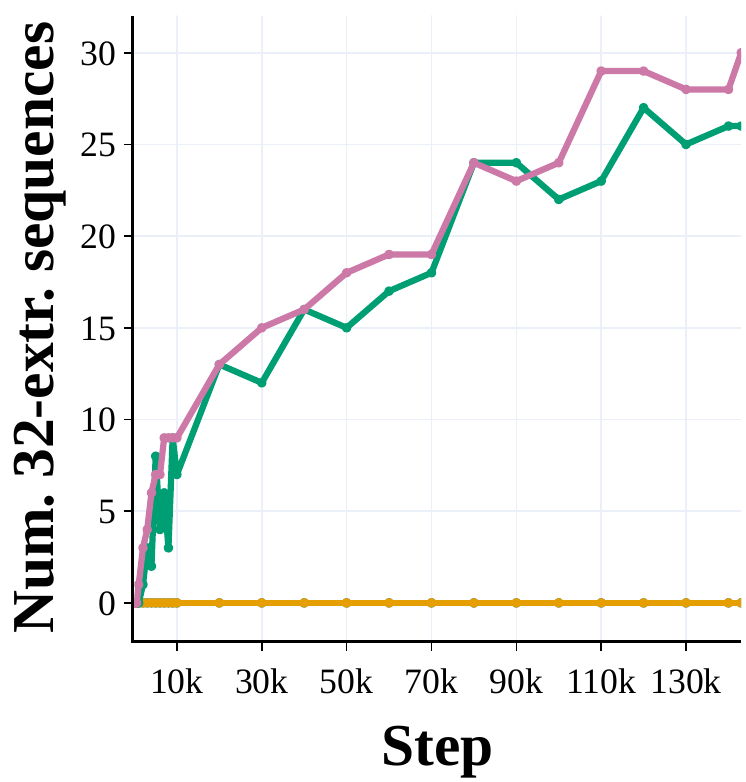}
       \caption{Num. 32-extr.}
        \label{fig:pythia-extractability}
    \end{subfigure}
    \begin{subfigure}{0.37\linewidth}
        \includegraphics[trim=0 0 280 0,clip,width=0.52\linewidth]{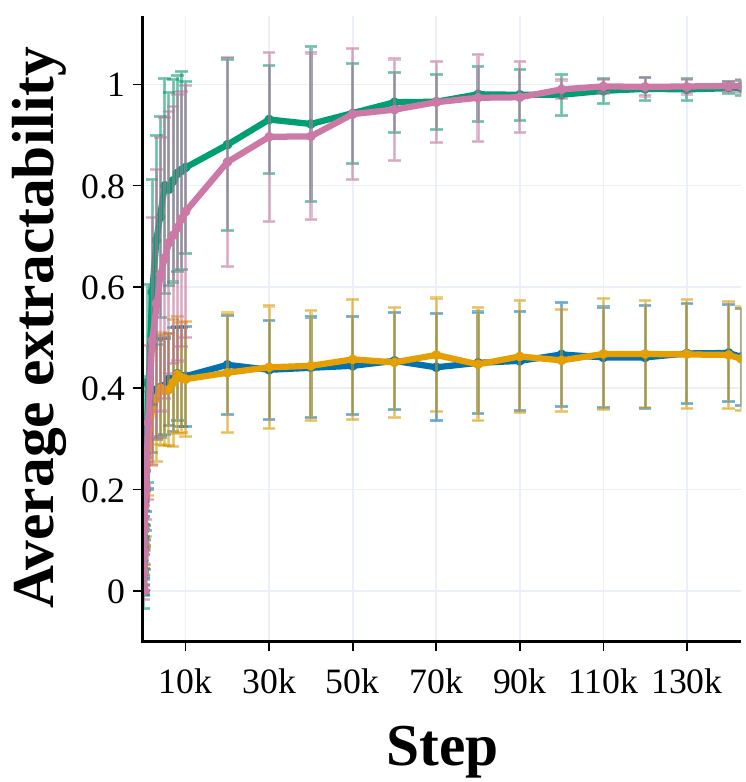}
        \includegraphics[trim=395 50 0 50,clip,width=0.45\linewidth]{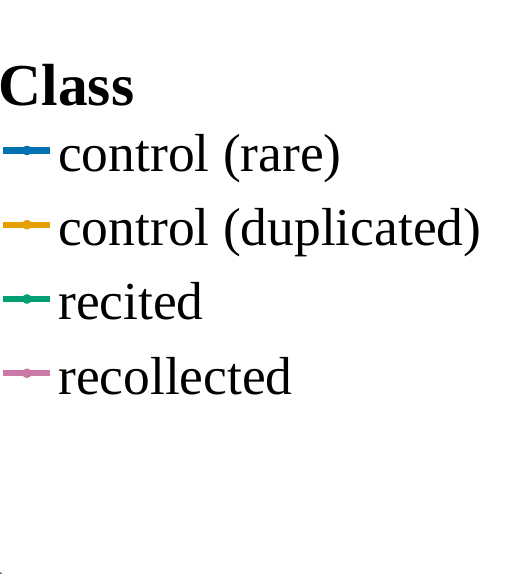}
        \caption{Frac. of extractable tokens}
        \label{fig:pythia-extractability-continuous}
    \end{subfigure}
    \caption{\textbf{Pythia-1B training dynamics.} We plot the loss and extractability of our sampled sequences over time and find that a lot of memorization happens early in training. Other models in App. Fig.~\ref{fig:pythia-all-dynamics}.}
    \label{fig:pythia-1b-dynamics}
\end{figure}

To get a high-level understanding of how memorization behavior in Pythia develops over time, we plot the loss and 32-extractability of the sequences we study in Fig.~\ref{fig:pythia-1b-dynamics}. Surprisingly, the loss trajectory of both the recited and recollected sequences separates from the control groups as early as $256$ and $1,000$ steps respectively, i.e., less than $1\%$ of the training. While previous work shows that memorization in Pythia is predictable from the perplexity in the final checkpoint \citep{prashanth2025recite}, this result suggests that a separation of memorized and not memorized sequences is possible at much earlier training stages than reported by \citet{bidermanEmergentPredictableMemorization2023}, with the peak mean distance of the two classes reached as early as the halfway point of training and stable trajectories throughout. Adding to this, almost half of our memorized samples are already 32-extractable at this point as can be seen from Fig.~\ref{fig:pythia-extractability}, i.e., memorization has already concluded for these instances. We show that this indeed leads to much earlier predictability of memorization in Section~\ref{sec:prediction}.

\textbf{A large fraction of memorized tokens is extractable early in training.} Further, the token-extractability trajectory reveals that a large fraction of the memorization learning happens in early training (see Fig.~\ref{fig:pythia-extractability-continuous}). By $10$k steps, the model already greedily decodes more than 80\% of the memorized sequences, almost double the number of the control groups. This shows that strict 32-extractability as a measure lags behind the continuous, token-level emergence of verbatim memorization, of which a large fraction happens in early training before step $20$k. 

\textbf{Memorization during generalization?} We also track basic linguistic competence of the model through the widely used BLiMP benchmark \citep{warstadt-etal-2020-blimp-benchmark} (see App. Fig.~\ref{fig:blimp}), which measures the model's ability to differentiate between linguistic minimal pairs. Interestingly, this evaluation of basic generalization in the model reaches its first plateau around $10$k. Given that much of memorization develops before this mark and that our samples are mostly composed of natural English language, this suggests that memorization in Pythia happens not after but during or even before generalization in form of basic linguistic capability.

\begin{figure}[h]
    \centering
    \begin{subfigure}{1\linewidth}
    \centering
        \includegraphics[trim=395 203 0 147,clip,height=7pt]{figures/decomp/phases/through_10000/cosines/1b/train_original.pdf}
        \includegraphics[trim=395 238 65 112,clip,height=7pt]{figures/decomp/phases/through_10000/cosines/1b/train_original.pdf}
        \includegraphics[trim=395 168 130 182,clip,height=7pt]{figures/decomp/phases/through_10000/cosines/1b/train_original.pdf}
        \includegraphics[trim=395 133 70 217,clip,height=7pt]{figures/decomp/phases/through_10000/cosines/1b/train_original.pdf}
    \end{subfigure} \\
    \begin{subfigure}{0.195\linewidth}
        \includegraphics[trim=0 34 243 0,clip,width=1\linewidth]{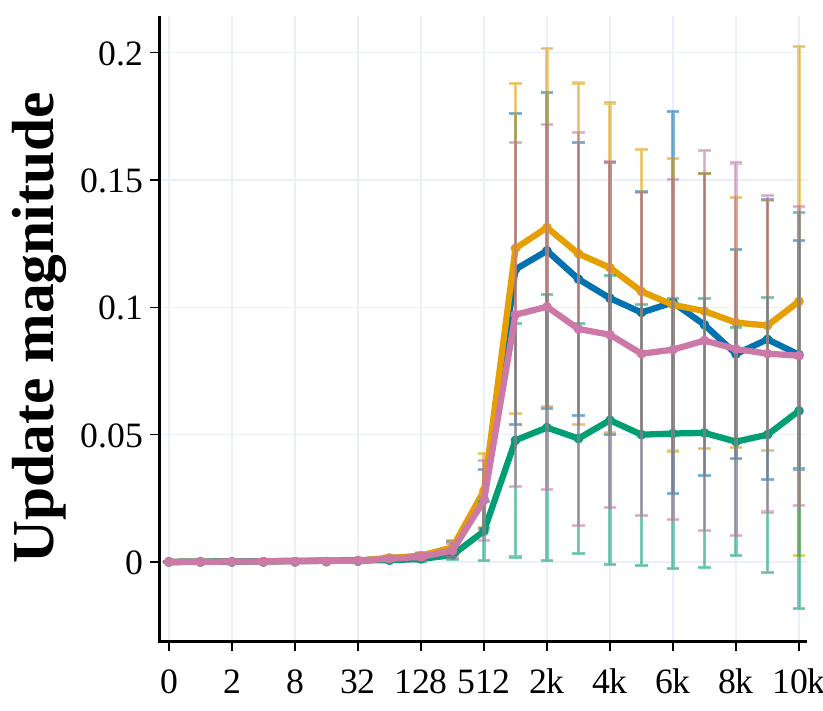}
        \includegraphics[trim=0 0 250 0,clip,width=1\linewidth]{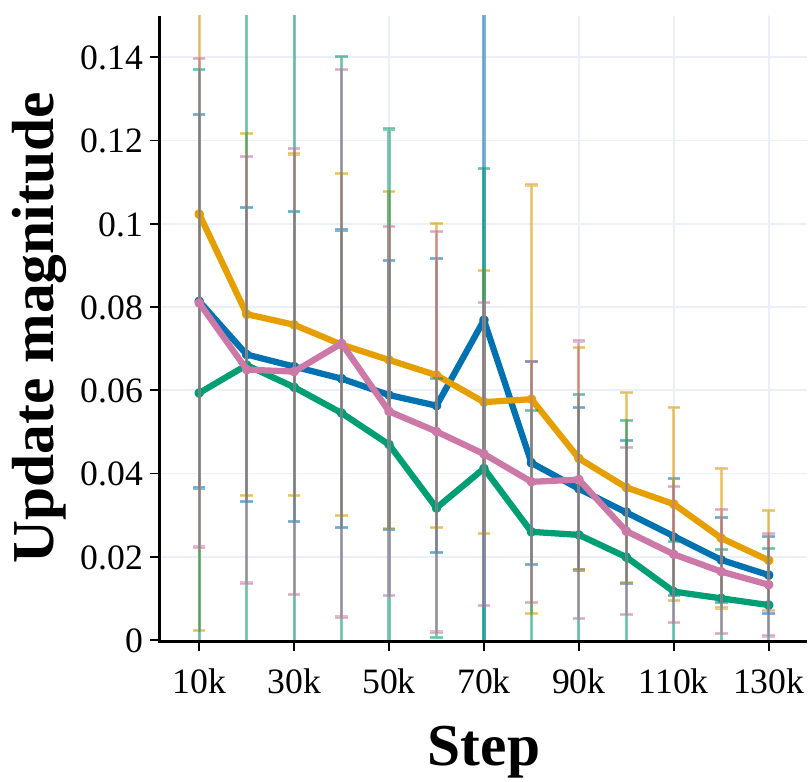}
        \caption{Update magnitude\\ \centering $\lVert u_s^x \rVert$}
        \label{fig:pythia-1b-update-magnitudes}
    \end{subfigure}
    \begin{subfigure}{0.195\linewidth}
        \includegraphics[trim=0 34 243 0,clip,width=1\linewidth]{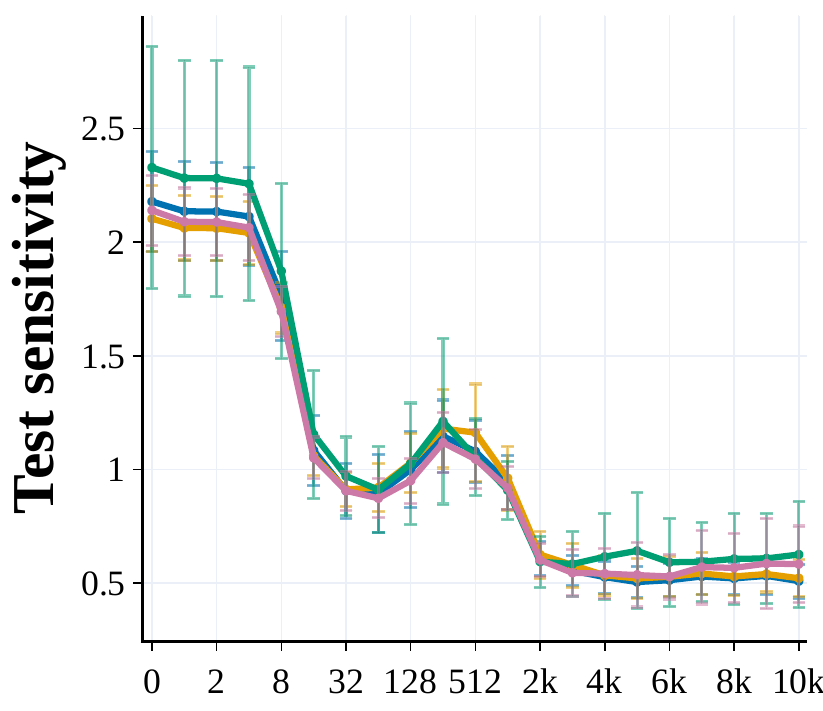}
        \includegraphics[trim=0 0 250 0,clip,width=1\linewidth]{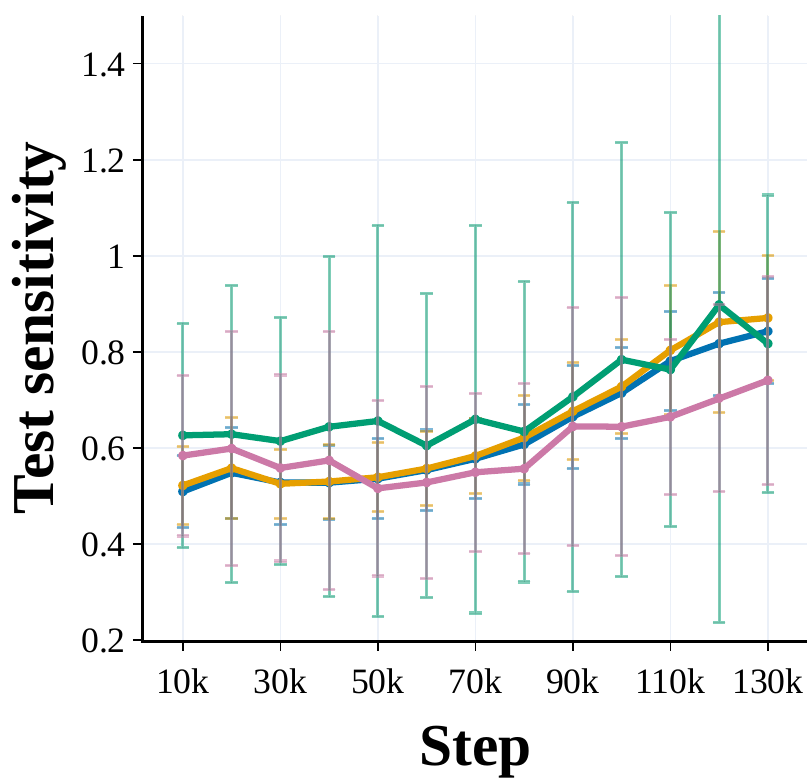}
        \caption{Test sensitivity\\ \centering $\lVert \Phi_s(x) \rVert$}
        \label{fig:pythia-1b-test-sensitivity}
    \end{subfigure}
    \begin{subfigure}{0.195\linewidth}
        \includegraphics[trim=0 34 243 0,clip,width=1\linewidth]{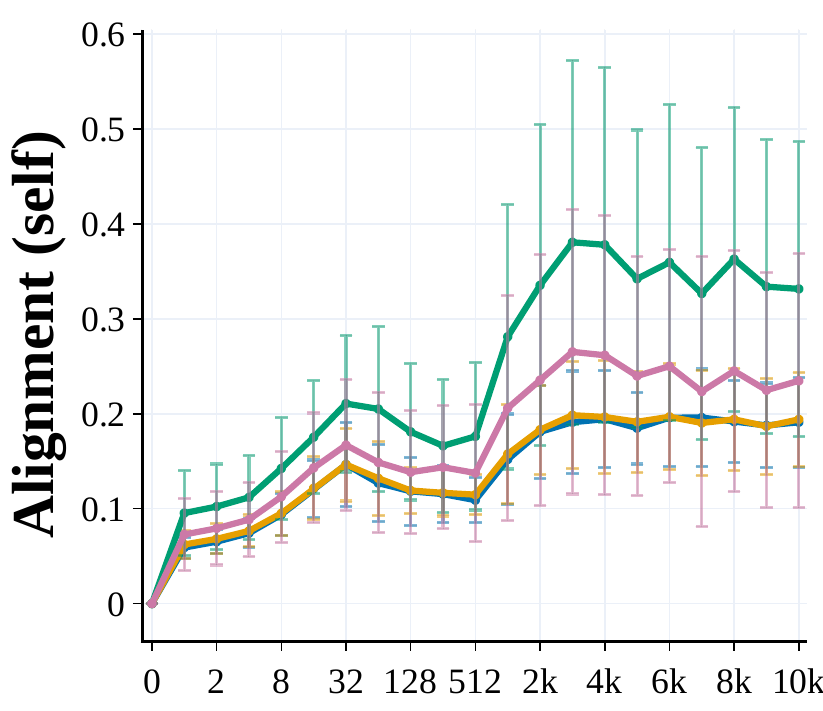}
        \includegraphics[trim=0 0 250 0,clip,width=1\linewidth]{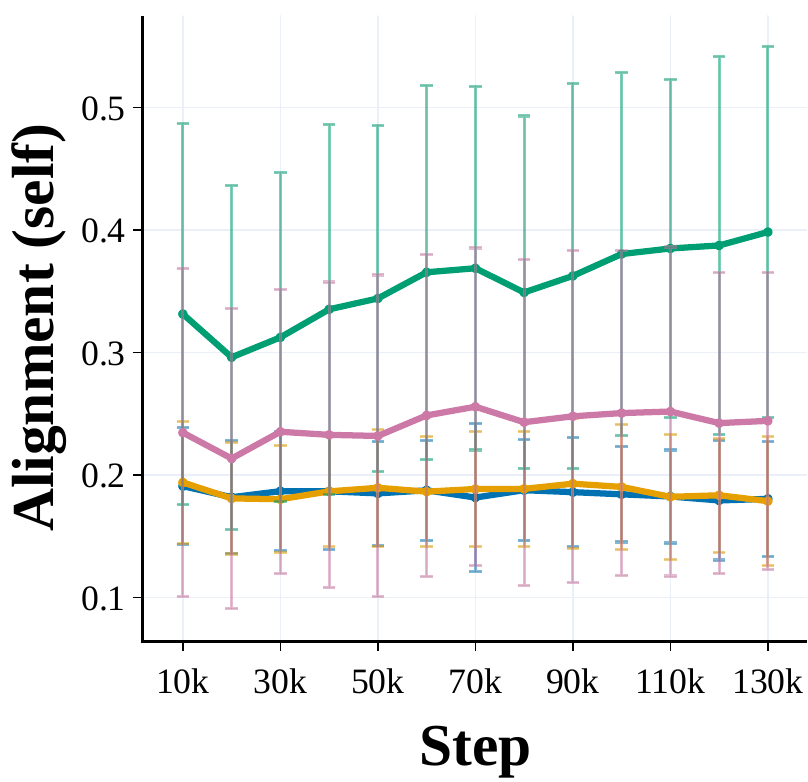}
        \caption{Self alignment\\ \centering $\cos(\Phi_s(x), u_s^x)$}
        \label{fig:pythia-1b-alignment-self}
    \end{subfigure}
    \begin{subfigure}{0.195\linewidth}
        \includegraphics[trim=0 34 243 0,clip,width=1\linewidth]{figures/decomp/phases/through_10000/cosines/1b/train_original.pdf}
        \includegraphics[trim=0 0 250 0,clip,width=1\linewidth]{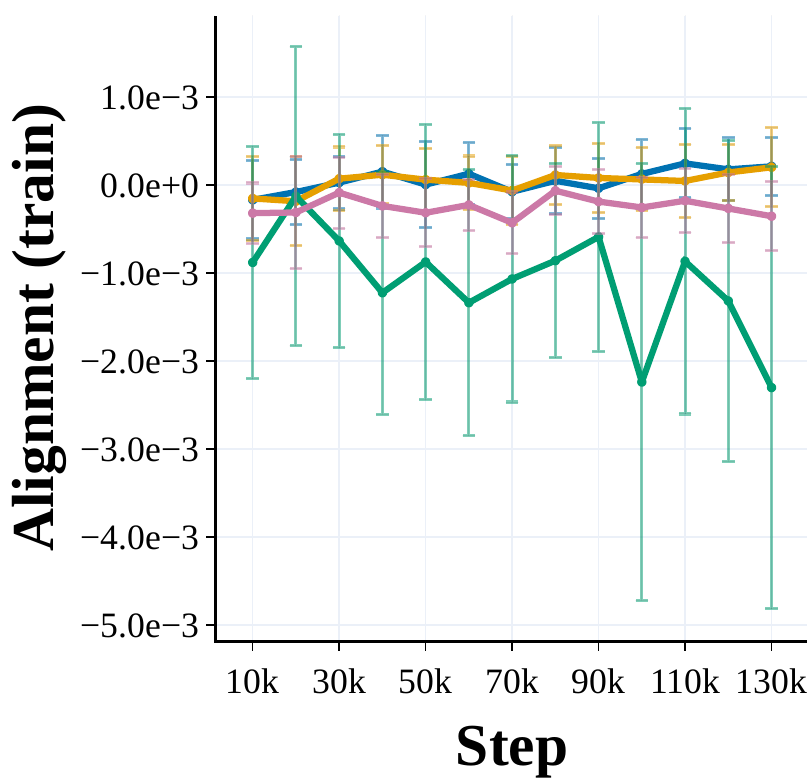}
        \caption{Train alignment\\ \centering $\cos(\Phi_s(x), u_s^{train})$}
        \label{fig:pythia-1b-alignment-train}
    \end{subfigure} 
    \begin{subfigure}{0.195\linewidth}
        \includegraphics[trim=0 34 243 0,clip,width=1\linewidth]{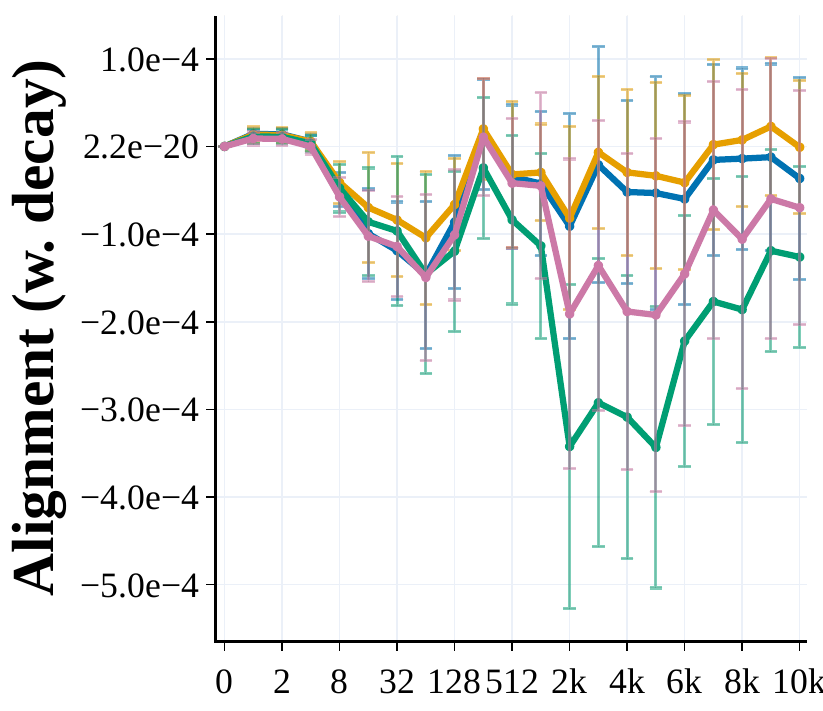}
        \includegraphics[trim=0 0 250 0,clip,width=1\linewidth]{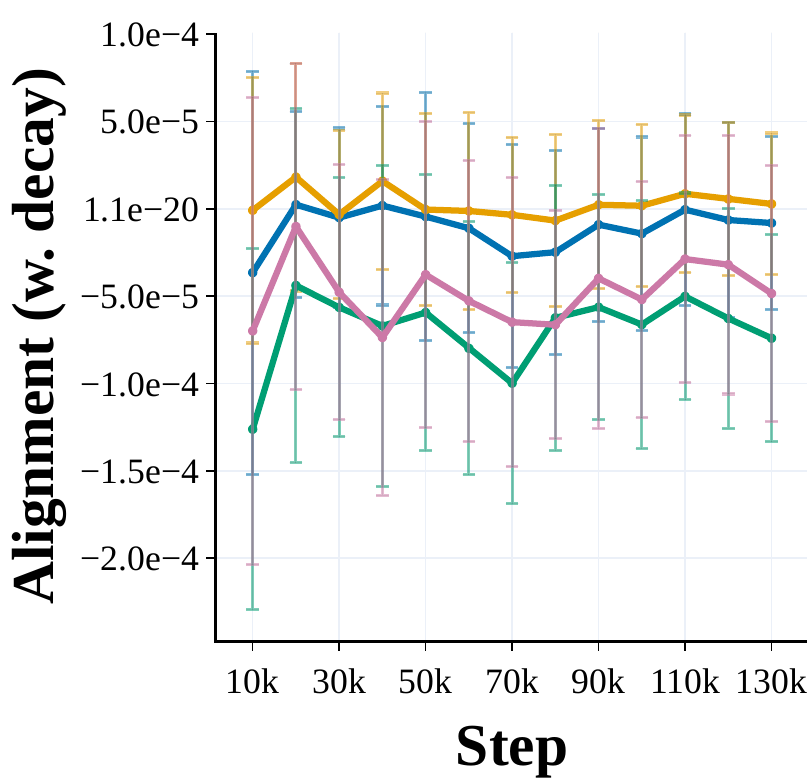}
        \caption{W. decay align.\\ \centering $\cos(\Phi_s(x), u_s^{reg})$}
        \label{fig:pythia-1b-alignment-reg}
    \end{subfigure} \\
    \caption{\textbf{Pythia-1B update magnitude, test sensitivity, and alignment.} We plot the mean and standard deviation over the samples. %
    See Eq.~\ref{eq:alignment_polar} for definitions. 
    Other models in App. Fig.~\ref{fig:pythia-all-norms-layer} and~\ref{fig:pythia-alignment-combined-early-and-late}}
    \label{fig:pythia-1b-alignment}
\end{figure}

\subsection{Decomposition of memorized sequences} 
\label{sec:decomp}

To understand how the above dynamics emerge, we turn to the ExPLAIND decomposition. Interestingly, on the level of update magnitude $\lVert u_s^x \rVert$ and test sensitivity $\lVert \Phi_s(x) \rVert$, the four classes undergo similar trends (see Fig.~\ref{fig:pythia-1b-update-magnitudes},~\ref{fig:pythia-1b-test-sensitivity}). Recited instances, on average, show the lowest update magnitude, while exhibiting the highest test sensitivity. In contrast recollected instances have lower test sensitivity late in training and cluster with the control groups in terms of early sensitivity and update magnitude.

\textbf{Memorization is enabled by self alignment.} Self alignment $\cos(\Phi_s(x), u_s^x)$ is where we find one of the reasons for the diverging loss trajectories. First, consider Fig.~\ref{fig:pythia-1b-alignment-self}, where we plot the alignment of each predicted example with its own training influence. From the beginning of training and with further separation after step 512, the memorized samples show much higher levels of self alignment than the control groups. Recited instances, at $0.3$ to $0.4$ on average, show the highest level of self alignment for the rest of training, followed by the recollected instances around $0.25$. Self alignment in the two control groups follows the same trajectory and is the lowest at around $0.18$ through most of later training. This indicates that the tendency to move prediction towards a memorized solution is on average more than twice as high for recited instances than it is for non-memorized instances, while this gap is smaller for recollected instances.

\textbf{Training causes more forgetting in recited than recollected samples.} Memorization is however not just driven by the sample predictions' tendency to align with their own effect on parameter updates, but by how likely the achieved solution is to survive the other training influences pulling towards generalization. We study the effect of the other training instances on memorized predictions in Fig.~\ref{fig:pythia-1b-alignment-train}. While positively aligned with the training data's updates until step 128, the green recited instances start to show distinctively negative, lower alignment starting from step 256 which keeps decreasing until the end of training. On the other hand, the recollected and control group samples cluster around zero for most of training. This negative alignment suggests that the memorized solution to the recited instances is constantly overwritten by the parameter updates introduced by the other data, while recollected and control examples are, on average, not subject to such influences after initial training. This suggests one core mechanism behind recollection: While recited instances need to be duplicated to reinforce their respective memorized solution against being overwritten by the other training influences, recollected instances, i.e., memorized examples that are not duplicated across training, exist in a part of the parameter space that is, on average, orthogonal to such influences and they do not require continued repair.

\subsection{Memorization prediction} 
\label{sec:prediction}

To validate our explanations of memorization above, we ask to what extent the memorized examples are really separated by the phenomena we describe above. %
We investigate this through a binary linear least squares classifier using our decomposition as features. We standardize the features and use ridge coefficient $10^{-8}$ in a stratified six-fold cross-validation across our $128$ examples, to avoid reliance on hyperparameter tuning. We report average accuracy on the held out fold as well as the average weight assigned to each feature. For an example $x$ and a model checkpoint $\theta_s$ at step $s$, we consider features including \textbf{cross-entropy} $\mathrm{CE}(\theta_s, x)$, a strong predictor for memorization identified by previous work \citep{prashanth2025recite}, and our decomposition features over time. Namely, we consider \textbf{self alignment}, \textbf{train alignment}, and \textbf{weight decay alignment}, i.e., the cosine $\cos(\Phi_s(x), u_s)$ with the update $u_s$ introduced by the example itself, the original training data, and weight decay respectively as defined in Eq.~\ref{method:accumulated_influence_terms} and~\ref{eq:alignment_polar}. We also consider \textbf{test sensitivity} $\lVert \Phi_s(x) \rVert$  and \textbf{update magnitude} $\lVert u_s \rVert$, totaling five decomposition features. Since both classes are stratified along duplication in the training data, we do not consider it as a feature. For each step, we use the features derived from only that step's checkpoint. This way, our insights could benefit realistic settings, where predictions need to be made without access to the model's history or future.

\begin{figure}[b]
    \centering
    \vspace{-10pt}
    \begin{subfigure}{0.26\linewidth}
        \includegraphics[width=1\linewidth]{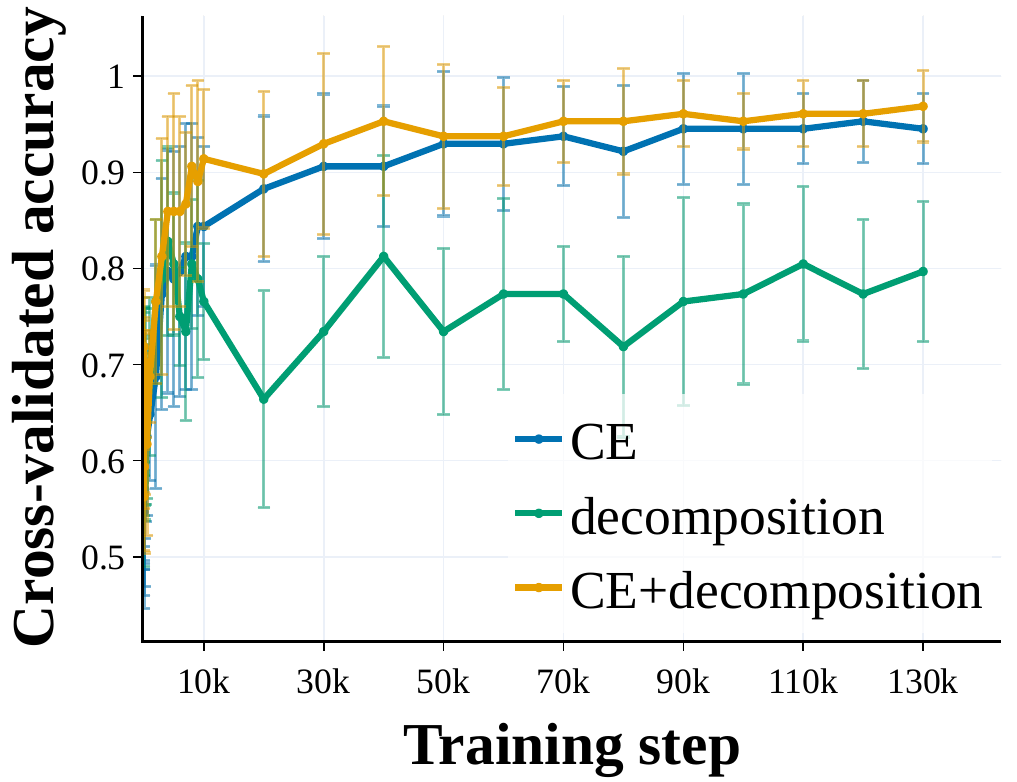}
        \caption{Accuracy on held out}
        \label{fig:pythia-1b-pred-acc}
    \end{subfigure}
    \begin{subfigure}{0.18\linewidth}
        \includegraphics[trim=0 0 240 25,clip,width=1\linewidth]{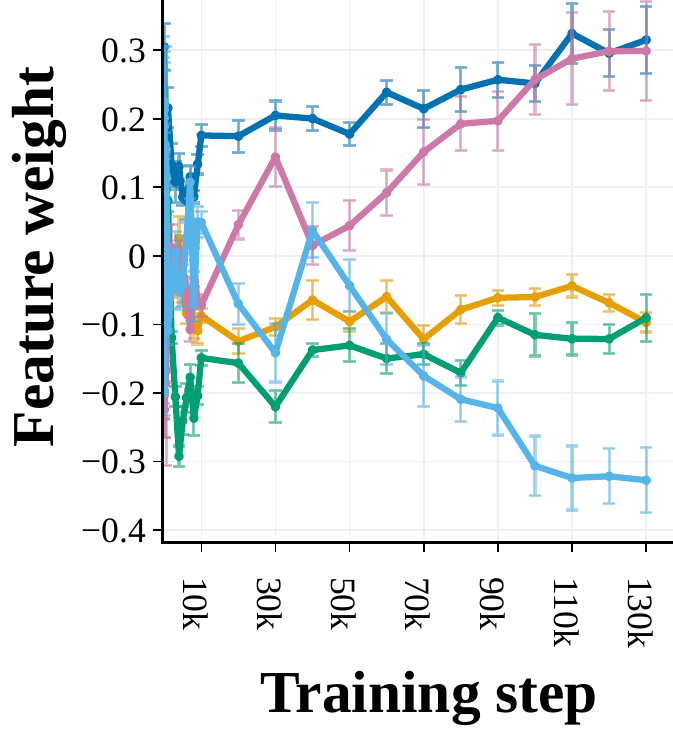}
    \caption{Weights decomp}
    \label{fig:pythia-1b-pred-weight-decomp-only}
    \end{subfigure}
    \begin{subfigure}{0.35\linewidth}
    \includegraphics[trim=0 0 240 25,clip,width=0.515\linewidth]{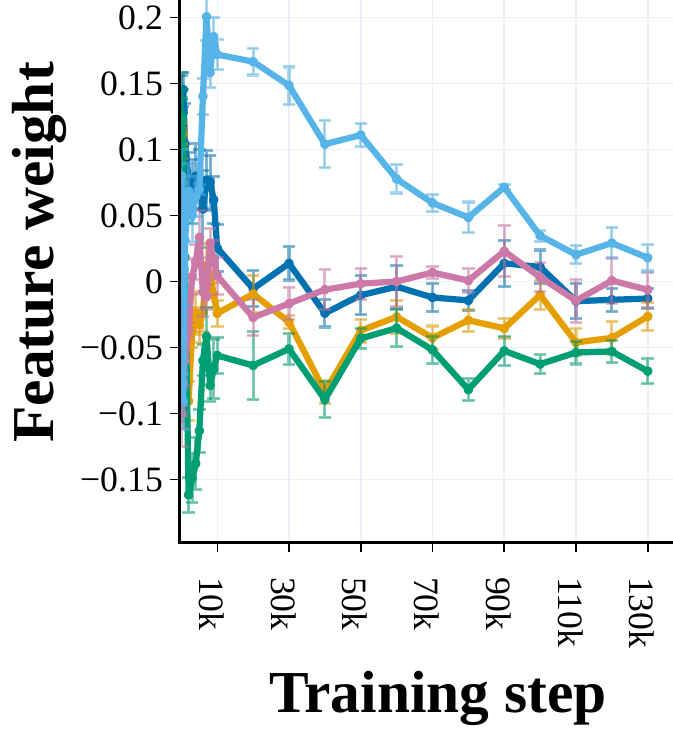}
    \includegraphics[trim=335 60 0 60,clip,width=0.4\linewidth]{figures/prediction/1b/single_loss_decomposition_features_without_loss_weights.pdf}
    \caption{Weights CE+decomp}
    \label{fig:pythia-1b-pred-weight-ce-decomp}
    \end{subfigure}
    \caption{\textbf{Memorization classification.} We plot the mean accuracy on held-out set and mean weights of the decomposition features of the decomposition and CE+decomposition. Others in App. Fig.~\ref{fig:pythia-pred-combined}.}
    \label{fig:pythia-1b-pred}
\end{figure}

\textbf{Memorization is predictable early in training.} We find that very early on in training, all linear models perform above chance. As shown in Fig.~\ref{fig:pythia-1b-pred-acc}, up until step 256 all feature combinations perform roughly similarly at around $67$ to $69\%$ accuracy. After that, their performance trajectories split with the CE+decomposition feature scoring the highest at $90.6\%$ accuracy at step $10k$, while cross-entropy only and decomposition only follow at $84.4\%$ and $75.0\%$ respectively. The combined model keeps outperforming for the remaining checkpoints, peaking at $96.1\%$ accuracy by step $40$k. This confirms our observations above: By that point more than half of memorization learning is already completed, i.e., many examples are already 32-extractable and the mean cross-entropies have separated by almost two standard deviations (see Fig.~\ref{fig:pythia-1b-dynamics}). Besides, this prediction at scale has been reported at such later training stages by \citet{bidermanEmergentPredictableMemorization2023}. %

\textbf{Besides cross-entropy, self and train alignment are the strongest predictors of memorization.} To investigate which features are the most predictive of memorization, we plot the feature weights in Fig.~\ref{fig:pythia-1b-pred-weight-decomp-only}. Underlining our assessment above, self alignment is the highest positive predictor over almost all checkpoints, while alignment with the original training influences is the most negative. We also look at the decomposition feature weights of the combined model in Fig.~\ref{fig:pythia-1b-pred-weight-ce-decomp}. Interestingly, update magnitude and self alignment do not seem to add information beyond cross entropy as their weight is low over most models. Train and weight decay alignment also show negative predictive influence, while especially test sensitivity seems like the most informative component in early training, indicating that alignment with and sensitivity to forgetting are the factors not explained by CE alone.

\textbf{For larger models, cross-entropy is less predictive of memorization.} We perform the same naive classification across model scales (see App. Fig.~\ref{fig:pythia-pred-combined}) and find that cross entropy is a weaker early predictor of memorization as model size increases. Considering that the average cross entropies of the control sequences decreases as model size increases (see App. Fig. \ref{fig:pythia-all-dynamics}), this is consistent with our expectations. Interestingly, adding our decomposition features recovers a lot of the early predictability. For Pythia-$6.9$B, the CE+decomposition model still achieves $88.3\%$ accuracy at step $10$k, while the CE only model is at $77.3\%$. There, the CE model never catches up with the combined feature model, reaching comparable accuracy only in the final checkpoint.

\subsection{Localizing memorization in Pythia}

\begin{figure}[hb]
    \centering
    \begin{subfigure}{0.205\linewidth}
        \includegraphics[trim=0 0 255 0,clip,width=1\linewidth]{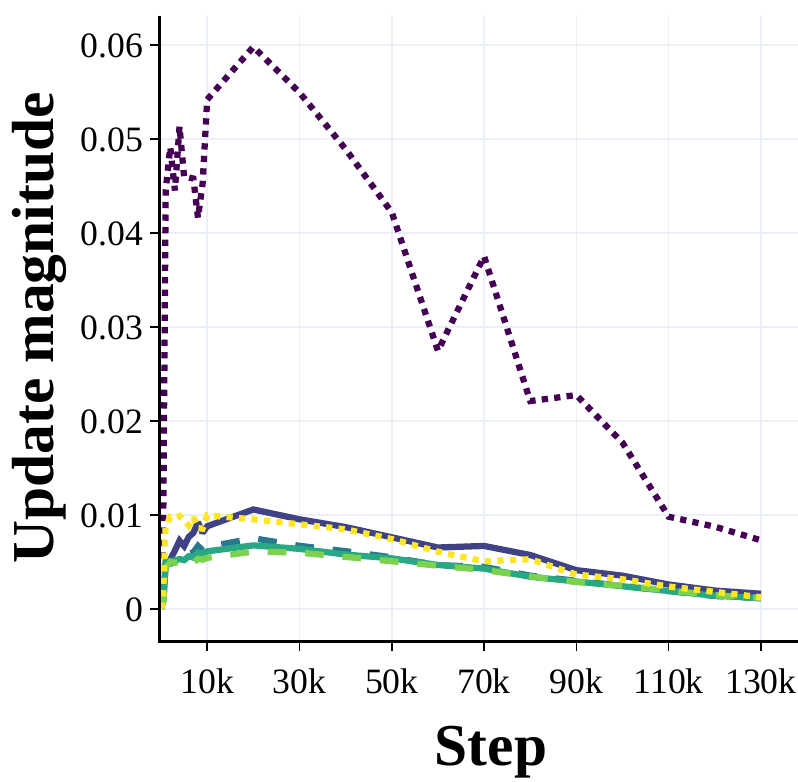}
        \caption{Upd. magn. recited}
    \end{subfigure}
    \begin{subfigure}{0.205\linewidth}
        \includegraphics[trim=0 0 255 0,clip,width=1\linewidth]{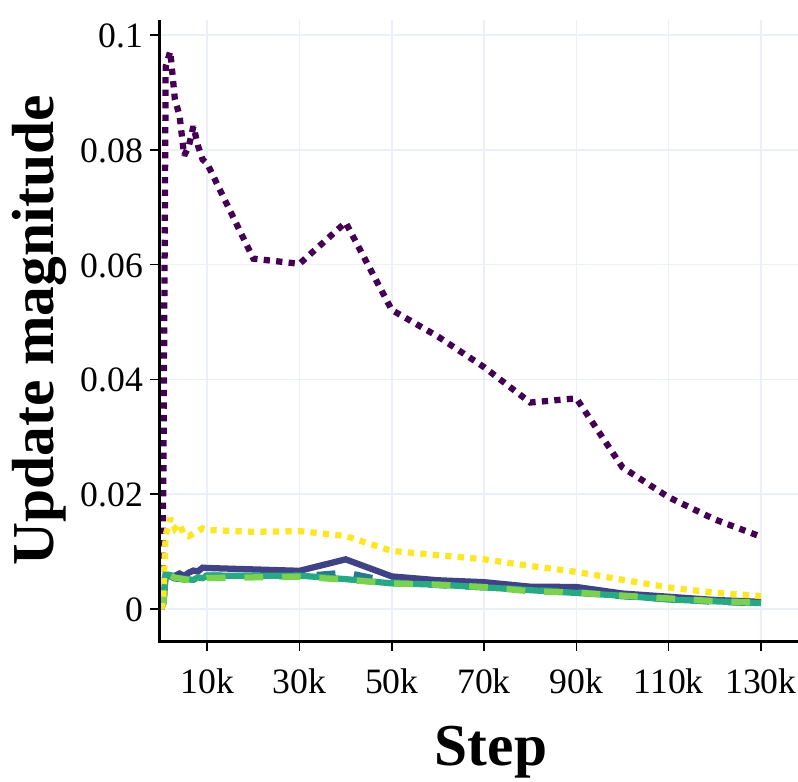}
        \caption{Upd. magn. recoll.}
    \end{subfigure}
    \begin{subfigure}{0.205\linewidth}
        \includegraphics[trim=0 0 255 0,clip,width=1\linewidth]{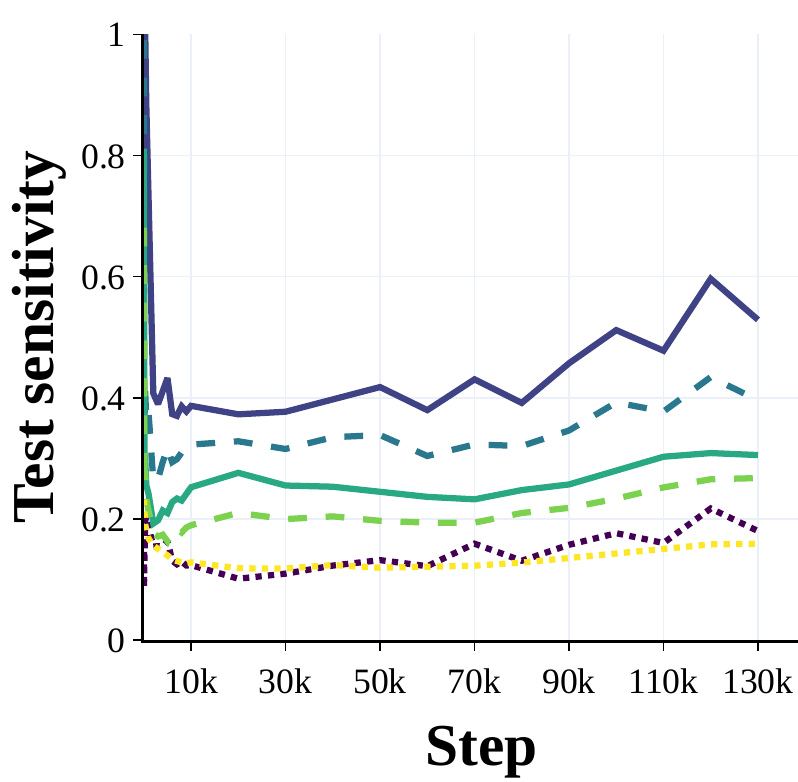}
        \caption{Test sens. recited}
    \end{subfigure}
    \begin{subfigure}{0.205\linewidth}
        \includegraphics[trim=0 0 255 0,clip,width=1\linewidth]{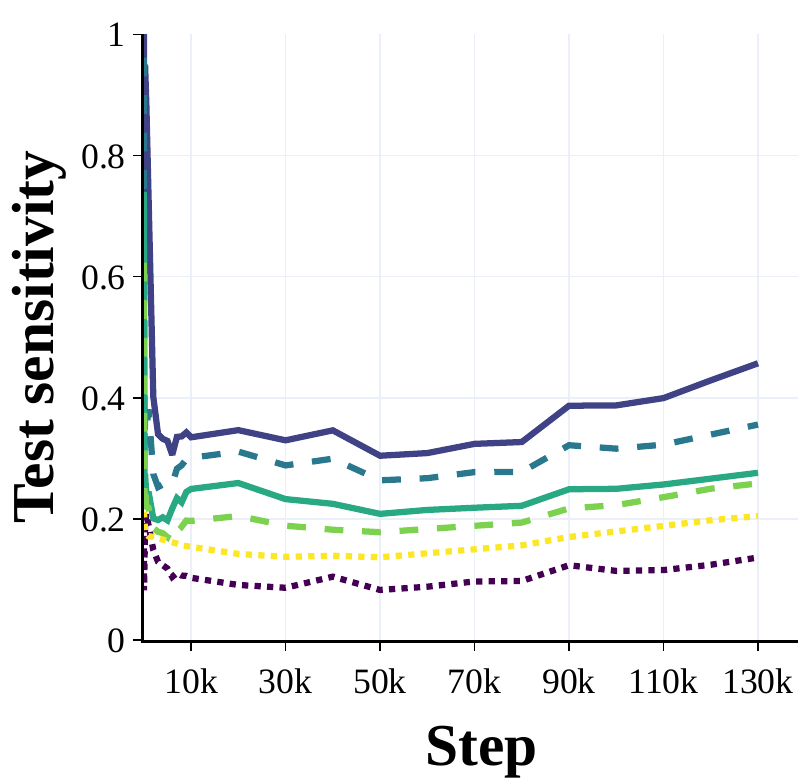}
        \caption{Test sens. recoll.}
    \end{subfigure}
    \begin{subfigure}{0.15\linewidth}
        \includegraphics[trim=400 20 0 40,clip,width=1\linewidth]{figures/decomp/four_classes/layer_regions_coarse/test_norm/test=memorized_recited/1b/train=single_sample_self_mean.pdf}
    \end{subfigure}
    \caption{\textbf{Pythia-1B update magnitude and test sensitivity by layer depth.} Others in App. Fig.~\ref{fig:pythia-all-norms-layer}.}
    \label{fig:pythia-1b-norms-layer}
\end{figure}

\begin{figure}[hb]
    \vspace{-5pt}
    \centering
    \begin{subfigure}{0.19\linewidth}
        \includegraphics[trim=0 0 255 0,clip,width=1\linewidth]{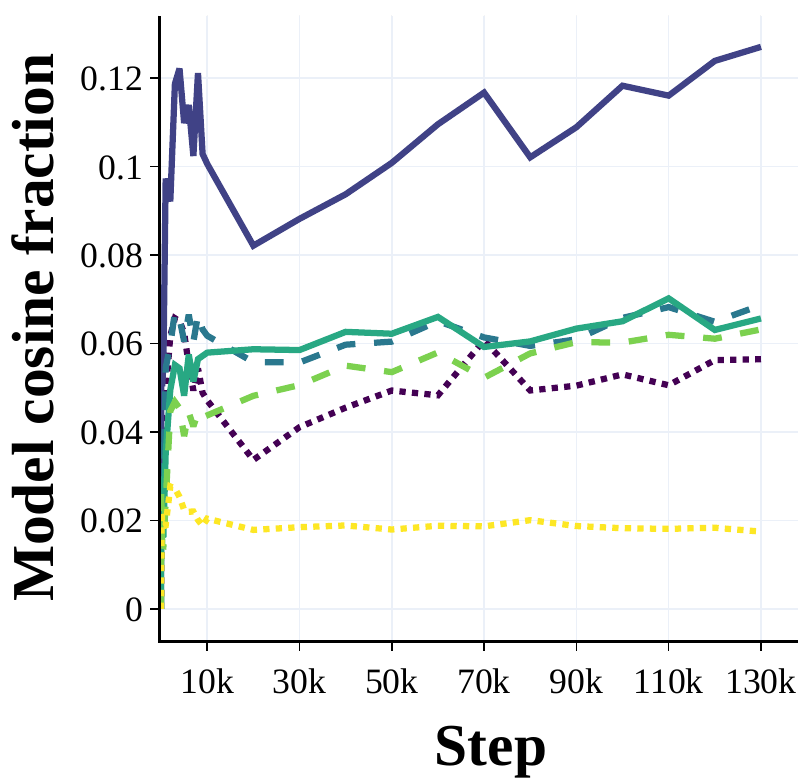}
        \caption{Recited$\rightarrow$recited}
    \end{subfigure}
    \begin{subfigure}{0.19\linewidth}
        \includegraphics[trim=0 0 255 0,clip,width=1\linewidth]{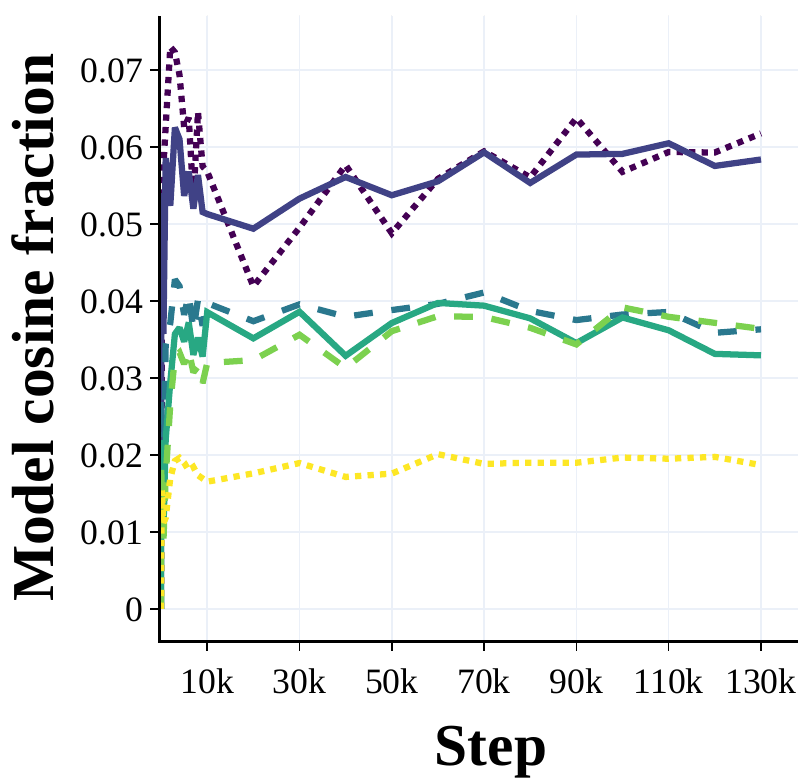}
        \caption{recoll.$\rightarrow$recoll.}
    \end{subfigure}
    \begin{subfigure}{0.115\linewidth}
        \includegraphics[trim=400 0 0 0,clip,width=1\linewidth]{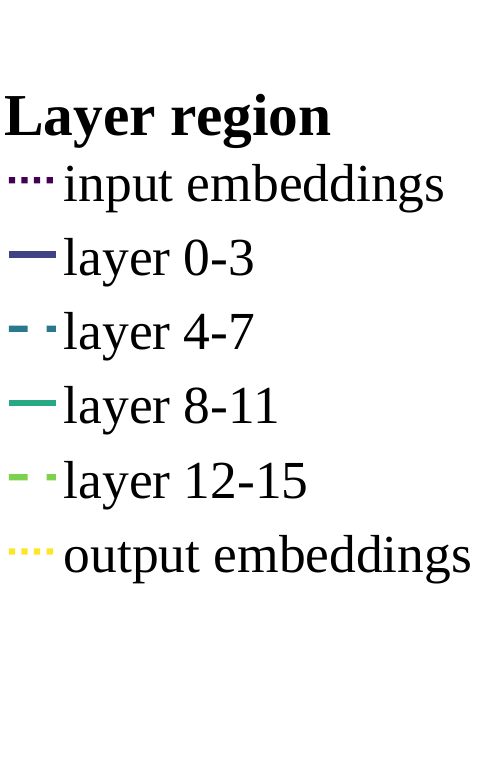}
    \end{subfigure}
    \begin{subfigure}{0.19\linewidth}
        \includegraphics[trim=0 0 190 0,clip,width=1\linewidth]{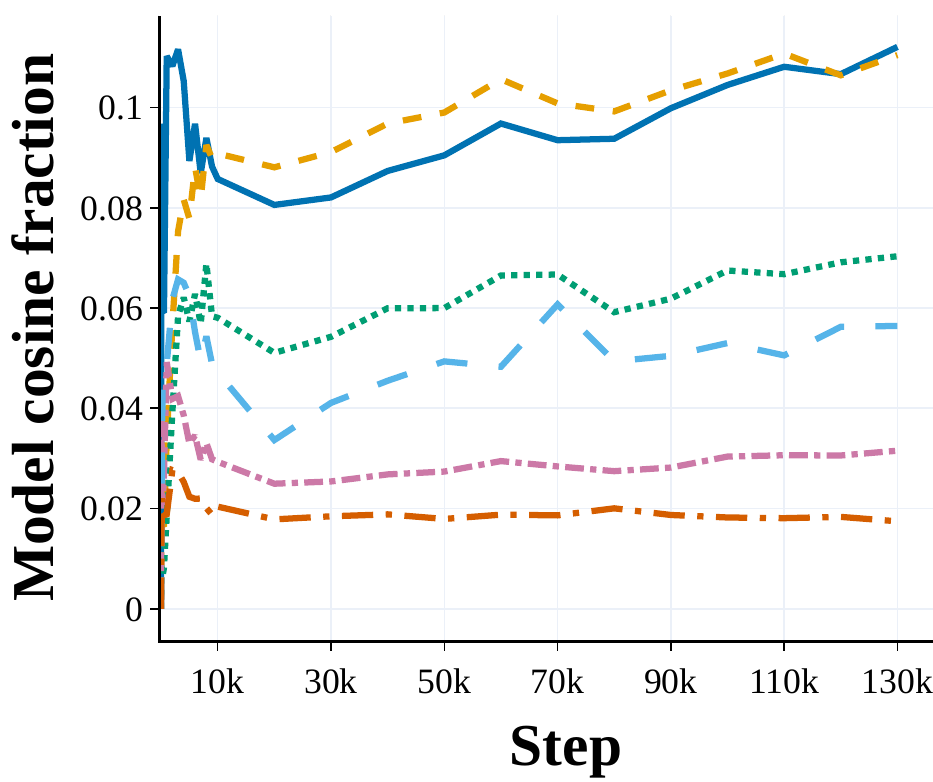}
        \caption{Recited$\rightarrow$recited}
    \end{subfigure}
    \begin{subfigure}{0.19\linewidth}
        \includegraphics[trim=0 0 190 0,clip,width=1\linewidth]{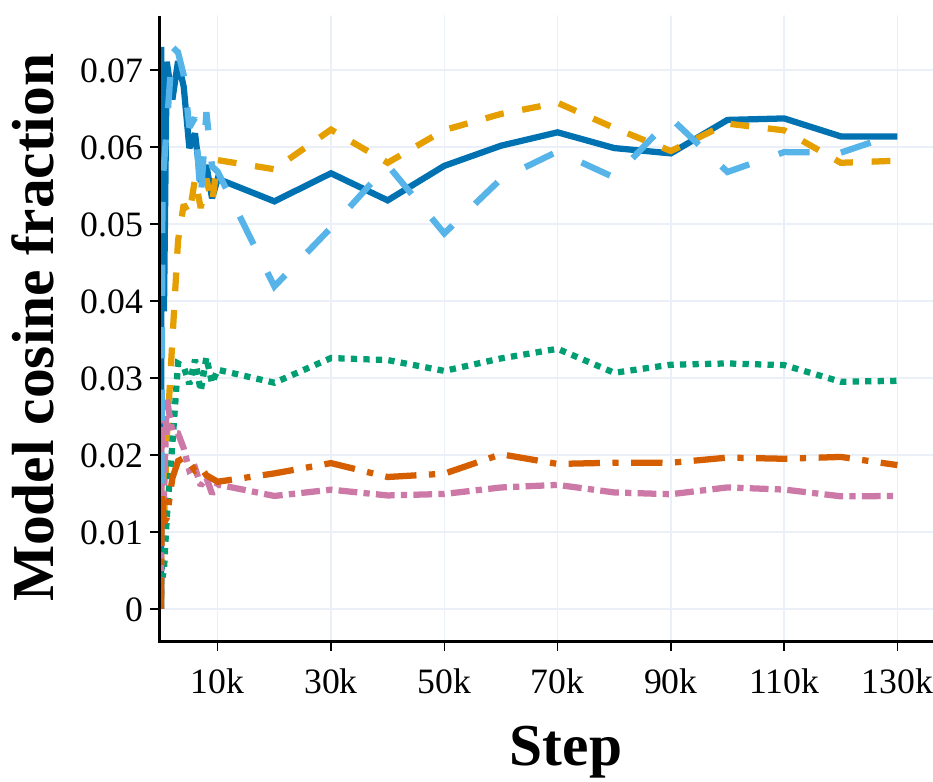}
        \caption{recoll.$\rightarrow$recoll.}
    \end{subfigure}
    \begin{subfigure}{0.085\linewidth}
        \includegraphics[trim=470 0 0 0,clip,width=1\linewidth]{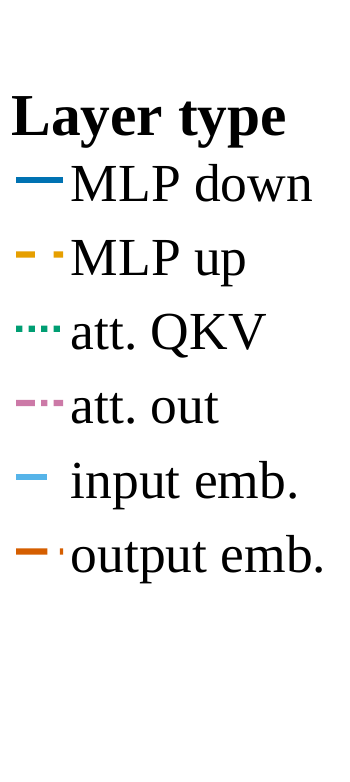}
    \end{subfigure}
    
    \caption{\textbf{Pythia-1B alignment by layer depth and type.} Other models, control in App. Fig.~\ref{fig:pythia-all-self-align-layer},~\ref{fig:pythia-all-self-align-layer-type}.}
    \label{fig:pythia-1b-self-inf-layer-type}
    \label{fig:pythia-1b-self-align-layer}
\end{figure}

\begin{figure}[ht]
    \centering
    \begin{subfigure}{0.2\linewidth}
        \includegraphics[trim=0 0 250 0,clip,width=1\linewidth]{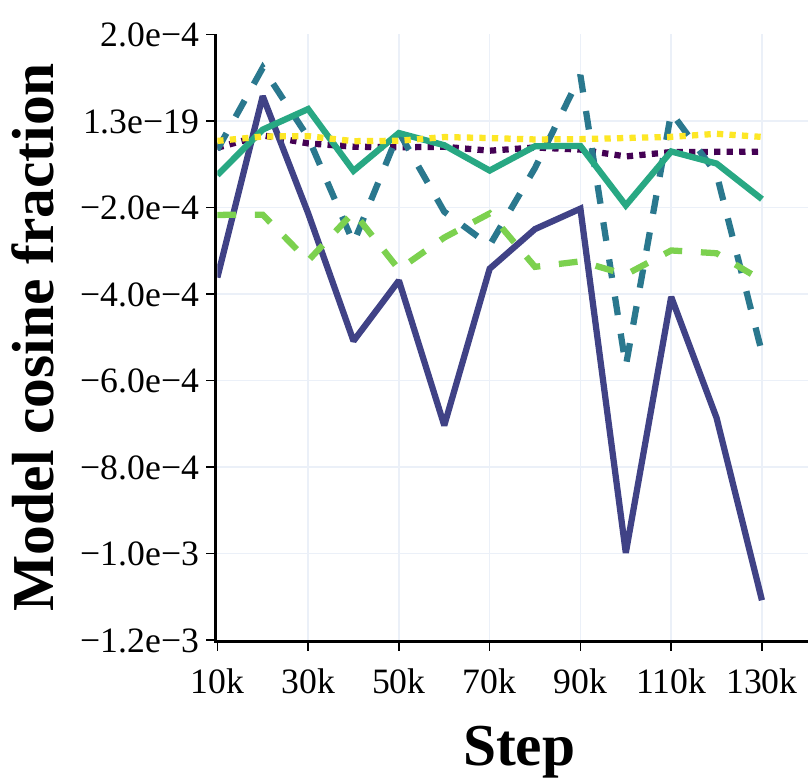}
        \caption{train$\rightarrow$recited}
    \end{subfigure}
    \begin{subfigure}{0.2\linewidth}
        \includegraphics[trim=0 0 250 0,clip,width=1\linewidth]{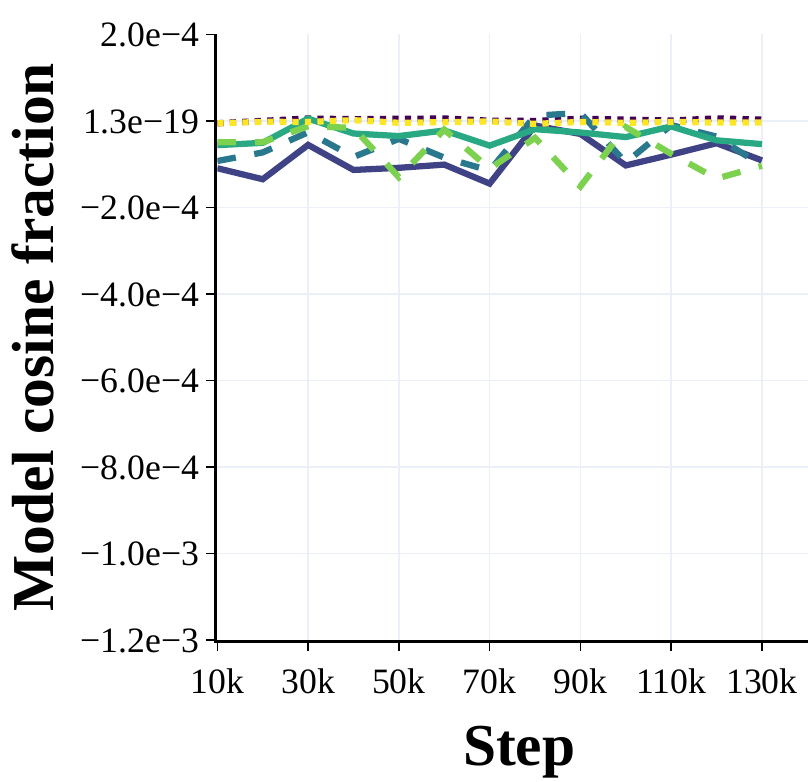}
        \caption{train$\rightarrow$recoll.}
    \end{subfigure}
    \begin{subfigure}{0.2\linewidth}
        \includegraphics[trim=0 0 250 0,clip,width=1\linewidth]{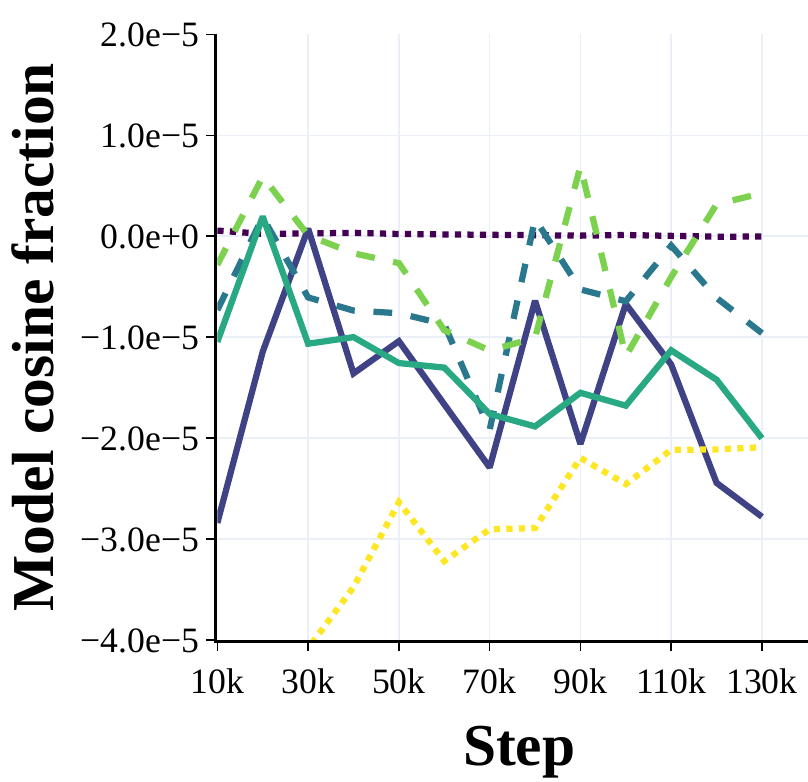}
        \caption{w. decay$\rightarrow$recited}
    \end{subfigure}
    \begin{subfigure}{0.2\linewidth}
        \includegraphics[trim=0 0 250 0,clip,width=1\linewidth]{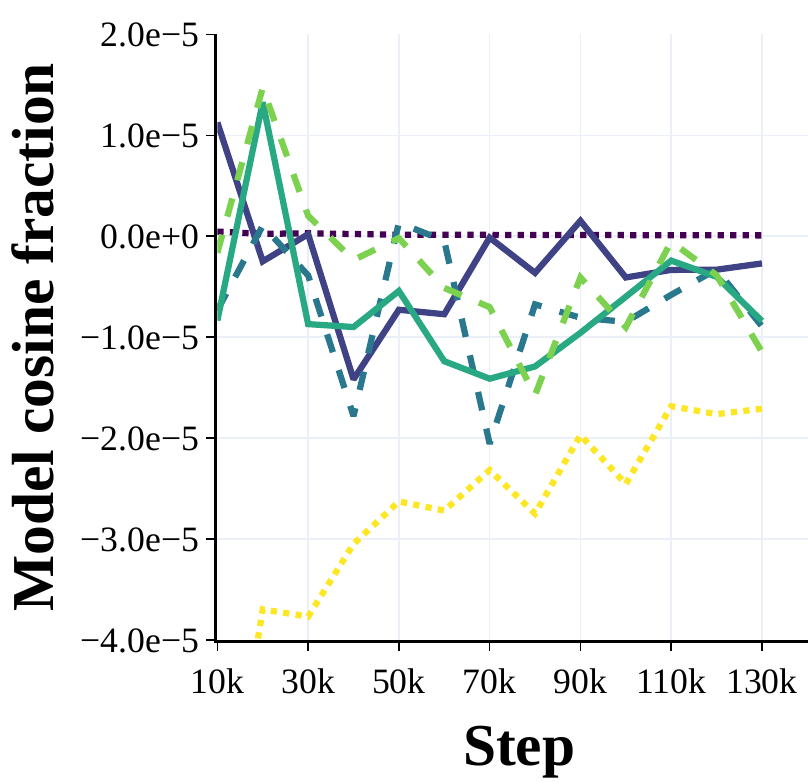}
        \caption{w. decay$\rightarrow$recoll.}
    \end{subfigure}
    \begin{subfigure}{0.15\linewidth}
        \includegraphics[trim=395 20 0 40,clip,width=1\linewidth]{figures/decomp/phases/from_10000/layer_regions_coarse/model_cosine_component/test=memorized_recollected/1b/train=regularization_mean.pdf}
    \end{subfigure}
    \caption{\textbf{Pythia-1B train/weight decay alignment by model region.} Other models in App. Fig.~\ref{fig:pythia-all-train-reg-align-layer}}
    \label{fig:pythia-1b-train-align-layer}
\end{figure}

We decompose our influences to the level of the model layers to see where the above learning influences stem from and to find the parameters implementing memorization. Our decomposition reveals that memorization influences are spread across the model, while peaking distinctly in some regions which we describe in the following. Our findings agree with arguments by which verbatim memorization is a distributed mechanism across the model \citep{maini-localization-2023,dankers-2026-memorisation}, and add further explanation to the conflicting findings in previous work regarding localization, beyond variation in memorization settings and methodology.

\textbf{Memorization learning is largely located in lower layers.} As we show in Fig.~\ref{fig:pythia-1b-norms-layer}, our decomposition reveals that the memorized instances are most sensitive to model changes in the lower layers of the model throughout all of training. Simultaneously, self alignment as shown in Fig.~\ref{fig:pythia-1b-self-align-layer} is the highest in the same layers, showing that a large fraction of memorization learning is located there. 
On the level of the update magnitude, Fig.~\ref{fig:pythia-1b-norms-layer} shows that training introduces the biggest changes to the input embeddings, while the other layer groups follow behind at similar levels. %

\textbf{Memorization learning focuses on the MLP layers.} Turning to Fig.~\ref{fig:pythia-1b-self-inf-layer-type}, where we plot self alignment grouped by layer type, we find that both forms of memorization self align the most in the MLP layers beginning early in training, while for both control groups it is roughly half of that throughout training. As we saw before, input embedding alignment is the most relevant source of alignment for both control groups and on par with MLP alignment in the recollected examples. Alignment in the attention QKV projections is also much higher in the memorized examples than the control groups.

\textbf{Recited memories degrade in lower layers.} In Fig.~\ref{fig:pythia-1b-train-align-layer}, we plot the alignment across layer regions with the original training data's parameter updates. We find that the effects from the training data onto the recited instances align in the lower model layers $0$-$3$, much more than for recollected examples and especially during late training. We find the same for alignment with weight decay, which we plot in Fig.~\ref{fig:pythia-1b-train-align-layer}. There, alignment of recited instances is high throughout training, while alignment of recollected examples tends to zero in the second half of training. Combined with the fact that test sensitivity is the highest in these layers, this explains why recited instances depend on a high level of duplication during training: As training progresses, their memories written to the parameters of the lower model layers are subject to constant forgetting due to the influences of other data and regularization, thus requiring the continued restoration of these memories. %
On the other hand, recollected sequences are subject to such degrading influences to a much smaller extent. The mechanisms for recall that their gradients write onto lower model layers are more robust against other training influences and thus require only a low level of verbatim duplication to remain effective.

\subsection{Causal memorization ablations}

\begin{wrapfigure}{!t}{0.5\textwidth}
    \centering
    \begin{subfigure}{0.49\linewidth}
        \includegraphics[trim=0 0 0 0,clip,width=1\linewidth]{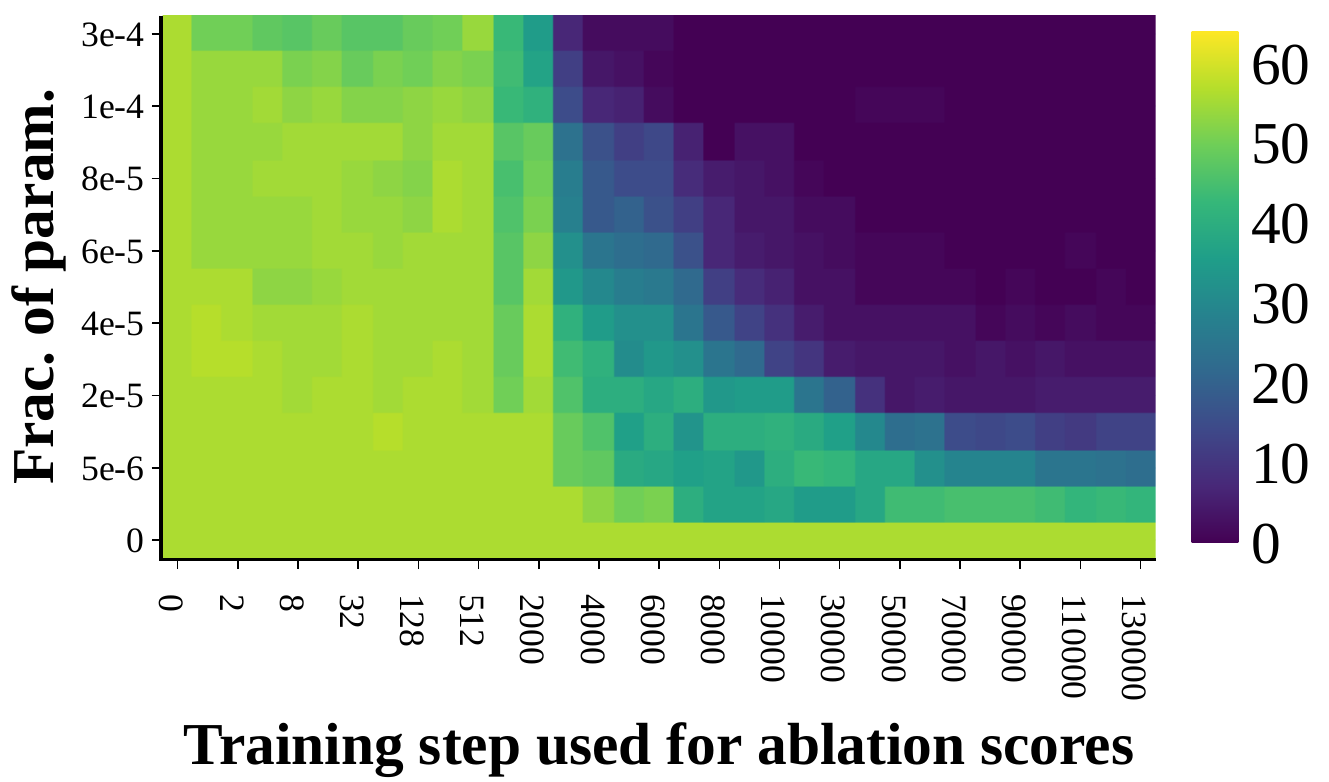}
    \end{subfigure}
    \begin{subfigure}{0.49\linewidth}
        \includegraphics[trim=0 0 0 0,clip,width=1\linewidth]{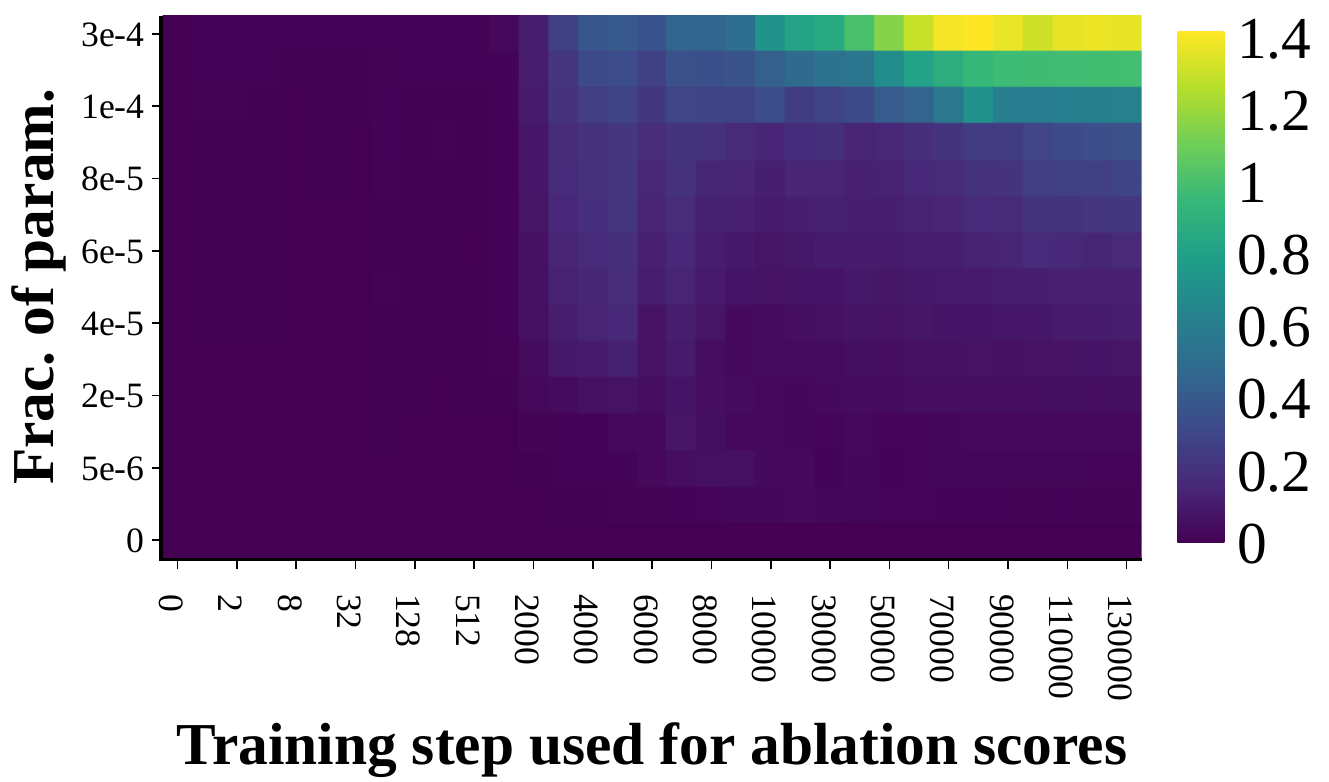}
    \end{subfigure}
    \caption{\textbf{Pythia-1B memorization ablation.} Left: Number of 32-extractable sequences in the memorized sample, right: CE loss on the control sample. Other models in App. Fig.~\ref{fig:pythia-ablation-app}.}
    \label{fig:pythia-1b-ablation}
    \vspace{-10pt}
\end{wrapfigure}

To validate our insights from the loss decomposition, we perform causal ablations on the parameter level to observe changes in the model memorization behavior or training dynamics. Strongly negative self-influences $\mathcal{I}_s^{x} (x;\theta)$ indicate that a parameter has contributed strongly to the loss decreasing on the predictions $x$ due to memorization. %
We thus identify a top-$K\%$ set $\mathcal{T}_K$ of parameters $\theta$ by the negative magnitude of their memorization influences $\mathcal{I}_s^{x} (x;\theta)$.
We intervene by setting the identified parameters to zero and observing the 32-extractability of the sequences we decompose. We also tried other interventions which performed comparably, in particular replacing the parameters with the mean value of their group.

\textbf{Ablation in the embeddings is not effective.} Recall that a large fraction of memorization updates stems from the input embeddings. Consequently, we find that directly using the influence scores to rank parameters is biased heavily towards them (see App. Fig.~\ref{fig:pythia-ablation-app-params-embs}). As we scale up the fraction of ablated parameters, model behavior in terms of CE on the control groups breaks down in sync with memorization. This indicates that the memorization influences in the embeddings, especially update magnitude, are an artifact of training rather than a causal building block of memorization. We thus restrict our following interventions to internal parameters.

\textbf{Our influences enable sparse and early memorization ablations.} We plot 32-extractability of our sampled sequences across different fractions of ablated parameters $K$ and using the accumulated influence scores from different points in training in Fig.~\ref{fig:pythia-1b-ablation}. Strikingly, as early as step $2000$, the information from our decomposition robustly ablates memorization behavior across our samples, with 32-extractability approaching zero when ablating the top-0.0002 parameters of Pythia-$1$B. Later in training, these interventions become focused, as complete ablation succeeds up to a sparsity of top-$0.00002$ intervened. At the same time, we observe cross entropy on the control samples to check whether our interventions simply break general model behavior and indeed, only the largest ablations break model behavior, while changes in the smaller interventions are modest below $0.1$. We plot the results of the other models up to $2.8$B parameters in App. Fig.~\ref{fig:pythia-ablation-app} which show that our findings generalize across models, i.e., for each model size there is a number of parameters which both effectively ablates memorization and leaves control cross-entropy intact. We consider this a very interesting result, since both our localization, as well as the intervention on parameters, are quite naively building on our decomposition. It causally underlines that our decomposition identifies the locations of the memories in the model successfully and that actionable post-hoc fixes are possible.

\textbf{Memorization in smaller models is ablated by a single attention head.} Interestingly, most of the ablated parameters are located in the attention layers (see App. Fig.~\ref{fig:pythia-ablation-app-params}) and we find that recited instances tend to align with other recited instances in the lower attention layers (see App.~\ref{app:attention_ablation}). Motivated by this, we intervene on the attention heads in these layers. Interestingly, we find that, up to $1$B parameters, memorization can be ablated by intervening on a single attention head using our naive approach, a result which echoes \citet{stoehrLocalizingParagraphMemorization2024} who find similar results for fine-tuning an attention head in a small GPT-Neo model. Details in App. Sec.~\ref{app:attention_ablation}.

\section{Discussion \& Conclusion}

Our work presents novel perspectives on the training dynamics of memorization in LLMs. Here we review our results along the time and model scale dimensions.

\textbf{Memorization follows three training phases.} Across models, the loss of both the memorized and control instances flattens around the $5$k step mark. This co-occurs with the saturation of the model's performance in BLiMP, which marks an important phase transition in language model generalization \citep{chen2024sudden}. Most memorization follows a similar trajectory, with more than $80\%$ of the tokens already greedily decodeable at the same point. Decomposing their loss trajectory reveals that on the model- and layer-level almost all the trends of the influences start to stabilize around that point. %
This development seems to be completed around the $20$k mark. Further, our layer-level results indicate that in the first $10$k steps, the training dynamics introduce a shared mechanism for memorization in an attention head in the first transformer layer. This aligns with other reports of early heuristics for context processing, for instance context copying \citep{körner2026copyfirsttranslatelater}. Further, memorization predictability starts to become successful around step $10k$ and reaches its peak shortly after for most models. %
In sum, this suggests that most of the memorization mechanisms are formed in early training and that later memorization dynamics are largely refining and maintaining existing memories, rather than building new ones. Further, it suggests that memorization and generalization appear in parallel in the model. We argue the point of emergence of both, around step $5$k, to be the first phase transition of memorization. The second transition seems to happen somewhere around step $20$k, when most of our influences stabilize and after which predictability converges. %

\textbf{Memorization and model scale.} Previous work asks how memorization behaves as a function of model scale \citep{tirumala2022memorization}. %
As we report in Sec.~\ref{sec:train_dynamics}, as model scale increases, the training data is fit to a higher extent, resulting in the average CE loss approaching that of the memorized instances. Predicting memorization based on solely CE or perplexity is more difficult in these models, though we show features derived from our gradient decomposition to recover some of the early predictability observed in the smaller models. Further, as model scale increases, our decomposition shows that the constant forgetting effects in the lower parts of the model due to training and weight decay become weaker, thus explaining the findings by \citet{tirumala2022memorization}. For the largest models, weight decay even is slightly aligned with memorization on average. This could explain previous observations by which memorization scales with model size \citep{carlini2023quantifying,bidermanEmergentPredictableMemorization2023}.

\textbf{Limitations.} Our methodology omits the influences of the first moment term in AdamW and estimates the second moment. Both choices could bias our results. %
Besides, our memorization prediction should be seen as a validation of our explanations and a proof-of-concept rather than a direct competitor to \citet{bidermanEmergentPredictableMemorization2023} and \citet{prashanth2025recite}'s approaches over the entire training dataset. Computing our ExPLAIND features over the entire dataset is infeasible, though we think that gradient based proxies derived from parameter updates of the train loop are a promising possibility. Our interventions likewise validate the analysis but are post-hoc in nature: they do not establish how to prevent memorization during training, and their effects on robustness and other predictions remain unclear.
Finally, our findings concern verbatim memorization in the Pythia family up to $6.9$B parameters. Their generalizability to larger models, other architectures, and other forms of memorization, for instance in the vision domain or in a noisy label setting, remains open.

\subsubsection*{Acknowledgments}
We thank Felicia Körner and Martin Bär for their feedback on earlier versions of this paper. We acknowledge the support for BP through the ERC Consolidator Grant DIALECT 101043235.

\subsubsection*{AI use statement}
This study was conceived and written by its human authors, who take full responsibility for the claims presented and artifacts produced. We acknowledge the help of AI tools for implementing our experiments, though we checked all generated code for its correctness. Further, AI helped verifying and improving our math and proofs. In particular, the second moment estimator presented in Appendix \ref{app:second_moment} was developed with the help of AI. We incorporated a small amount feedback from an AI tool into our writing to improve clarity and grammar.

\subsubsection*{Reproducibility statement}

Our study focuses on the fully open sourced Pythia family, for which checkpoints, training data, and training loop are available on Github and Huggingface. To ensure reproducibility of our results, we will provide the full implementation of our method and experiments which allows to reproduce all our results and comes with all documentation necessary to run all the experiments described in this paper on Github upon publication. Besides, we did our best to document the parameters and design choices necessary for reproduction of our study in Appendix \ref{app:experiment_settings}.
\bibliography{iclr2027_conference,custom,MyLibrary}

\begin{thebibliography}{33}
\providecommand{\natexlab}[1]{#1}
\providecommand{\url}[1]{\texttt{#1}}
\expandafter\ifx\csname urlstyle\endcsname\relax
  \providecommand{\doi}[1]{doi: #1}\else
  \providecommand{\doi}{doi: \begingroup \urlstyle{rm}\Url}\fi

\bibitem[Arpit et~al.(2017)Arpit, Jastrzkebski, Ballas, Krueger, Bengio, Kanwal, Maharaj, Fischer, Courville, Bengio, and Lacoste-Julien]{pmlr-v70-arpit17a}
Devansh Arpit, Stanislaw Jastrzkebski, Nicolas Ballas, David Krueger, Emmanuel Bengio, Maxinder~S. Kanwal, Tegan Maharaj, Asja Fischer, Aaron Courville, Yoshua Bengio, and Simon Lacoste-Julien.
\newblock A closer look at memorization in deep networks.
\newblock In Doina Precup and Yee~Whye Teh (eds.), \emph{Proceedings of the 34th International Conference on Machine Learning}, volume~70 of \emph{Proceedings of Machine Learning Research}, pp.\  233--242. PMLR, 06--11 Aug 2017.
\newblock URL \url{https://proceedings.mlr.press/v70/arpit17a.html}.

\bibitem[Bell et~al.(2023)Bell, Geyer, Glickenstein, Fernandez, and Moore]{bellExactKernelEquivalence2023}
Brian~Wesley Bell, Michael Geyer, David Glickenstein, Amanda~S. Fernandez, and Juston Moore.
\newblock An {{Exact Kernel Equivalence}} for {{Finite Classification Models}}.
\newblock In \emph{Proceedings of 2nd {{Annual Workshop}} on {{Topology}}, {{Algebra}}, and {{Geometry}} in {{Machine Learning}} ({{TAG-ML}})}, pp.\  206--217. PMLR, September 2023.

\bibitem[Biderman et~al.(2023{\natexlab{a}})Biderman, Prashanth, Sutawika, Schoelkopf, Anthony, Purohit, and Raff]{bidermanEmergentPredictableMemorization2023}
Stella Biderman, USVSN~Sai Prashanth, Lintang Sutawika, Hailey Schoelkopf, Quentin~Gregory Anthony, Shivanshu Purohit, and Edward Raff.
\newblock Emergent and {{Predictable Memorization}} in {{Large Language Models}}.
\newblock In \emph{Thirty-Seventh {{Conference}} on {{Neural Information Processing Systems}}}, November 2023{\natexlab{a}}.

\bibitem[Biderman et~al.(2023{\natexlab{b}})Biderman, Schoelkopf, Anthony, Bradley, O'Brien, Hallahan, Khan, Purohit, Prashanth, Raff, Skowron, Sutawika, and Wal]{bidermanPythiaSuiteAnalyzing2023}
Stella Biderman, Hailey Schoelkopf, Quentin~Gregory Anthony, Herbie Bradley, Kyle O'Brien, Eric Hallahan, Mohammad~Aflah Khan, Shivanshu Purohit, Usvsn~Sai Prashanth, Edward Raff, Aviya Skowron, Lintang Sutawika, and Oskar Van~Der Wal.
\newblock Pythia: {{A Suite}} for {{Analyzing Large Language Models Across Training}} and {{Scaling}}.
\newblock In \emph{Proceedings of the 40th {{International Conference}} on {{Machine Learning}}}, pp.\  2397--2430. PMLR, July 2023{\natexlab{b}}.

\bibitem[Biderman et~al.(2026)Biderman, Khan, Mireshghallah, Arnett, Barez, and Saphra]{biderman2026position}
Stella Biderman, Mohammad~Aflah Khan, Niloofar Mireshghallah, Catherine Arnett, Fazl Barez, and Naomi Saphra.
\newblock Position: Don't just ''fix it in post'': A science of {AI} must study learning dynamics.
\newblock In \emph{Forty-third International Conference on Machine Learning Position Paper Track}, 2026.
\newblock URL \url{https://openreview.net/forum?id=VbHh05xQ7y}.

\bibitem[Carlini et~al.(2023)Carlini, Ippolito, Jagielski, Lee, Tramer, and Zhang]{carlini2023quantifying}
Nicholas Carlini, Daphne Ippolito, Matthew Jagielski, Katherine Lee, Florian Tramer, and Chiyuan Zhang.
\newblock Quantifying memorization across neural language models.
\newblock In \emph{The Eleventh International Conference on Learning Representations}, 2023.
\newblock URL \url{https://openreview.net/forum?id=TatRHT_1cK}.

\bibitem[Chang et~al.(2024)Chang, Thomason, and Jia]{changLocalizationMethodsActually2024}
Ting-Yun Chang, Jesse Thomason, and Robin Jia.
\newblock Do {{Localization Methods Actually Localize Memorized Data}} in {{LLMs}}? {{A Tale}} of {{Two Benchmarks}}.
\newblock In Kevin Duh, Helena Gomez, and Steven Bethard (eds.), \emph{Proceedings of the 2024 {{Conference}} of the {{North American Chapter}} of the {{Association}} for {{Computational Linguistics}}: {{Human Language Technologies}} ({{Volume}} 1: {{Long Papers}})}, pp.\  3190--3211, Mexico City, Mexico, June 2024. Association for Computational Linguistics.
\newblock \doi{10.18653/v1/2024.naacl-long.176}.

\bibitem[Chatterjee(2018)]{pmlr-v80-chatterjee18a}
Satrajit Chatterjee.
\newblock Learning and memorization.
\newblock In Jennifer Dy and Andreas Krause (eds.), \emph{Proceedings of the 35th International Conference on Machine Learning}, volume~80 of \emph{Proceedings of Machine Learning Research}, pp.\  755--763. PMLR, 10--15 Jul 2018.
\newblock URL \url{https://proceedings.mlr.press/v80/chatterjee18a.html}.

\bibitem[Chen et~al.(2024)Chen, Shwartz-Ziv, Cho, Leavitt, and Saphra]{chen2024sudden}
Angelica Chen, Ravid Shwartz-Ziv, Kyunghyun Cho, Matthew~L Leavitt, and Naomi Saphra.
\newblock Sudden drops in the loss: Syntax acquisition, phase transitions, and simplicity bias in {MLM}s.
\newblock In \emph{The Twelfth International Conference on Learning Representations}, 2024.
\newblock URL \url{https://openreview.net/forum?id=MO5PiKHELW}.

\bibitem[Dai et~al.(2022)Dai, Dong, Hao, Sui, Chang, and Wei]{daiKnowledgeNeuronsPretrained2022}
Damai Dai, Li~Dong, Yaru Hao, Zhifang Sui, Baobao Chang, and Furu Wei.
\newblock Knowledge {{Neurons}} in {{Pretrained Transformers}}.
\newblock In Smaranda Muresan, Preslav Nakov, and Aline Villavicencio (eds.), \emph{Proceedings of the 60th {{Annual Meeting}} of the {{Association}} for {{Computational Linguistics}} ({{Volume}} 1: {{Long Papers}})}, pp.\  8493--8502, Dublin, Ireland, May 2022. Association for Computational Linguistics.
\newblock \doi{10.18653/v1/2022.acl-long.581}.

\bibitem[Dankers(2026)]{dankers-2026-memorisation}
Verna Dankers.
\newblock Memorisation meets compositionality in natural language processing.
\newblock In Yanai Elazar, Allyson Ettinger, Nora Kassner, and Sebastian Ruder (eds.), \emph{Proceedings of The Big Picture v2: Crafting a Research Narrative}, pp.\  144--159, San Diego, CA, USA, July 2026. Association for Computational Linguistics.
\newblock ISBN 979-8-89176-416-3.
\newblock \doi{10.18653/v1/2026.bigpicture-main.12}.
\newblock URL \url{https://aclanthology.org/2026.bigpicture-main.12/}.

\bibitem[Dankers \& Titov(2024)Dankers and Titov]{dankersGeneralisationFirstMemorisation2024}
Verna Dankers and Ivan Titov.
\newblock Generalisation {{First}}, {{Memorisation Second}}? {{Memorisation Localisation}} for {{Natural Language Classification Tasks}}.
\newblock In Lun-Wei Ku, Andre Martins, and Vivek Srikumar (eds.), \emph{Findings of the {{Association}} for {{Computational Linguistics}}: {{ACL}} 2024}, pp.\  14348--14366, Bangkok, Thailand, August 2024. Association for Computational Linguistics.
\newblock \doi{10.18653/v1/2024.findings-acl.852}.

\bibitem[Du et~al.(2025)Du, Mondorf, Casola, Yao, Litschko, and Plank]{du-etal-2025-reason}
Yupei Du, Philipp Mondorf, Silvia Casola, Yuekun Yao, Robert Litschko, and Barbara Plank.
\newblock Reason to rote: Rethinking memorization in reasoning.
\newblock In Christos Christodoulopoulos, Tanmoy Chakraborty, Carolyn Rose, and Violet Peng (eds.), \emph{Proceedings of the 2025 Conference on Empirical Methods in Natural Language Processing}, pp.\  8659--8679, Suzhou, China, November 2025. Association for Computational Linguistics.
\newblock ISBN 979-8-89176-332-6.
\newblock \doi{10.18653/v1/2025.emnlp-main.437}.
\newblock URL \url{https://aclanthology.org/2025.emnlp-main.437/}.

\bibitem[Eichin et~al.(2026)Eichin, Du, Mondorf, Matveev, Plank, and Hedderich]{eichin2026explaind}
Florian Eichin, Yupei Du, Philipp Mondorf, Maria Matveev, Barbara Plank, and Michael~A. Hedderich.
\newblock Ex{PLAIND}: Unifying model, data, and training attribution to study model behavior.
\newblock In \emph{Forty-third International Conference on Machine Learning}, 2026.
\newblock URL \url{https://openreview.net/forum?id=7G6x9QTaN4}.

\bibitem[Feldman(2020)]{feldman-2020-memorization}
Vitaly Feldman.
\newblock Does learning require memorization? a short tale about a long tail.
\newblock In \emph{Proceedings of the 52nd Annual ACM SIGACT Symposium on Theory of Computing}, STOC 2020, pp.\  954–959, New York, NY, USA, 2020. Association for Computing Machinery.
\newblock ISBN 9781450369794.
\newblock \doi{10.1145/3357713.3384290}.
\newblock URL \url{https://doi.org/10.1145/3357713.3384290}.

\bibitem[Feldman \& Zhang(2020)Feldman and Zhang]{feldmanWhatNeuralNetworks2020}
Vitaly Feldman and Chiyuan Zhang.
\newblock What {{Neural Networks Memorize}} and {{Why}}: {{Discovering}} the {{Long Tail}} via {{Influence Estimation}}.
\newblock volume~33, pp.\  2881--2891. Curran Associates, Inc., 2020.

\bibitem[Geva et~al.(2021)Geva, Schuster, Berant, and Levy]{gevaTransformerFeedForwardLayers2021}
Mor Geva, Roei Schuster, Jonathan Berant, and Omer Levy.
\newblock Transformer {{Feed-Forward Layers Are Key-Value Memories}}.
\newblock In Marie-Francine Moens, Xuanjing Huang, Lucia Specia, and Scott Wen-tau Yih (eds.), \emph{Proceedings of the 2021 {{Conference}} on {{Empirical Methods}} in {{Natural Language Processing}}}, pp.\  5484--5495, Online and Punta Cana, Dominican Republic, November 2021. Association for Computational Linguistics.
\newblock \doi{10.18653/v1/2021.emnlp-main.446}.

\bibitem[Jia et~al.(2025)Jia, Wallace, Huang, Pimentel, Maini, Dankers, Wei, and Lesci]{l2m2-ws-2025-1}
Robin Jia, Eric Wallace, Yangsibo Huang, Tiago Pimentel, Pratyush Maini, Verna Dankers, Johnny Wei, and Pietro Lesci (eds.).
\newblock \emph{Proceedings of the First Workshop on Large Language Model Memorization (L2M2)}, Vienna, Austria, August 2025. Association for Computational Linguistics.
\newblock ISBN 979-8-89176-278-7.
\newblock \doi{10.18653/v1/2025.l2m2-1.0}.
\newblock URL \url{https://aclanthology.org/2025.l2m2-1.0/}.

\bibitem[Kangaslahti et~al.(2025)Kangaslahti, Rosenfeld, and Saphra]{kangaslahtiHiddenBreakthroughsLanguage2025}
Sara Kangaslahti, Elan Rosenfeld, and Naomi Saphra.
\newblock Hidden {{Breakthroughs}} in {{Language Model Training}}, June 2025.

\bibitem[Koh \& Liang(2017)Koh and Liang]{kohUnderstandingBlackboxPredictions2017}
Pang~Wei Koh and Percy Liang.
\newblock Understanding {{Black-box Predictions}} via {{Influence Functions}}.
\newblock In \emph{Proceedings of the 34th {{International Conference}} on {{Machine Learning}}}, pp.\  1885--1894. PMLR, July 2017.

\bibitem[Körner et~al.(2026)Körner, Matveev, Eichin, Kutyniok, Plank, and Hedderich]{körner2026copyfirsttranslatelater}
Felicia Körner, Maria Matveev, Florian Eichin, Gitta Kutyniok, Barbara Plank, and Michael~A. Hedderich.
\newblock Copy first, translate later: Interpreting translation dynamics in multilingual pretraining, 2026.
\newblock URL \url{https://arxiv.org/abs/2604.17633}.

\bibitem[Lan et~al.(2019)Lan, Liu, Zhou, and Yosinski]{lanLCALossChange2019}
Janice Lan, Rosanne Liu, Hattie Zhou, and Jason Yosinski.
\newblock {{LCA}}: {{Loss Change Allocation}} for {{Neural Network Training}}.
\newblock In \emph{Advances in {{Neural Information Processing Systems}}}, volume~32. Curran Associates, Inc., 2019.

\bibitem[Lee et~al.(2022)Lee, Ippolito, Nystrom, Zhang, Eck, Callison-Burch, and Carlini]{lee-etal-2022-deduplicating}
Katherine Lee, Daphne Ippolito, Andrew Nystrom, Chiyuan Zhang, Douglas Eck, Chris Callison-Burch, and Nicholas Carlini.
\newblock Deduplicating training data makes language models better.
\newblock In Smaranda Muresan, Preslav Nakov, and Aline Villavicencio (eds.), \emph{Proceedings of the 60th Annual Meeting of the Association for Computational Linguistics (Volume 1: Long Papers)}, pp.\  8424--8445, Dublin, Ireland, May 2022. Association for Computational Linguistics.
\newblock \doi{10.18653/v1/2022.acl-long.577}.
\newblock URL \url{https://aclanthology.org/2022.acl-long.577/}.

\bibitem[Maini et~al.(2023)Maini, Mozer, Sedghi, Lipton, Kolter, and Zhang]{maini-localization-2023}
Pratyush Maini, Michael~C. Mozer, Hanie Sedghi, Zachary~C. Lipton, J.~Zico Kolter, and Chiyuan Zhang.
\newblock Can neural network memorization be localized?
\newblock In \emph{Proceedings of the 40th International Conference on Machine Learning}, ICML'23. JMLR.org, 2023.

\bibitem[Meng et~al.(2022)Meng, Bau, Andonian, and Belinkov]{mengLocatingEditingFactual2022}
Kevin Meng, David Bau, Alex~J. Andonian, and Yonatan Belinkov.
\newblock Locating and {{Editing Factual Associations}} in {{GPT}}.
\newblock In \emph{Advances in {{Neural Information Processing Systems}}}, October 2022.

\bibitem[Mireshghallah et~al.(2022)Mireshghallah, Uniyal, Wang, Evans, and Berg-Kirkpatrick]{mireshghallah-etal-2022-empirical}
Fatemehsadat Mireshghallah, Archit Uniyal, Tianhao Wang, David Evans, and Taylor Berg-Kirkpatrick.
\newblock An empirical analysis of memorization in fine-tuned autoregressive language models.
\newblock In Yoav Goldberg, Zornitsa Kozareva, and Yue Zhang (eds.), \emph{Proceedings of the 2022 Conference on Empirical Methods in Natural Language Processing}, pp.\  1816--1826, Abu Dhabi, United Arab Emirates, December 2022. Association for Computational Linguistics.
\newblock \doi{10.18653/v1/2022.emnlp-main.119}.
\newblock URL \url{https://aclanthology.org/2022.emnlp-main.119/}.

\bibitem[Ortu et~al.(2024)Ortu, Jin, Doimo, Sachan, Cazzaniga, and Sch{\"o}lkopf]{ortuCompetitionMechanismsTracing2024}
Francesco Ortu, Zhijing Jin, Diego Doimo, Mrinmaya Sachan, Alberto Cazzaniga, and Bernhard Sch{\"o}lkopf.
\newblock Competition of {{Mechanisms}}: {{Tracing How Language Models Handle Facts}} and {{Counterfactuals}}.
\newblock In Lun-Wei Ku, Andre Martins, and Vivek Srikumar (eds.), \emph{Proceedings of the 62nd {{Annual Meeting}} of the {{Association}} for {{Computational Linguistics}} ({{Volume}} 1: {{Long Papers}})}, pp.\  8420--8436, Bangkok, Thailand, August 2024. Association for Computational Linguistics.
\newblock \doi{10.18653/v1/2024.acl-long.458}.

\bibitem[Prashanth et~al.(2025)Prashanth, Deng, O'Brien, V, Khan, Borkar, Choquette-Choo, Fuehne, Biderman, Ke, Lee, and Saphra]{prashanth2025recite}
USVSN~Sai Prashanth, Alvin Deng, Kyle O'Brien, Jyothir~S V, Mohammad~Aflah Khan, Jaydeep Borkar, Christopher~A. Choquette-Choo, Jacob~Ray Fuehne, Stella Biderman, Tracy Ke, Katherine Lee, and Naomi Saphra.
\newblock Recite, reconstruct, recollect: Memorization in {LM}s as a multifaceted phenomenon.
\newblock In \emph{The Thirteenth International Conference on Learning Representations}, 2025.
\newblock URL \url{https://openreview.net/forum?id=3E8YNv1HjU}.

\bibitem[Pruthi et~al.(2020)Pruthi, Liu, Kale, and Sundararajan]{pruthi2020tracin}
Garima Pruthi, Frederick Liu, Satyen Kale, and Mukund Sundararajan.
\newblock Estimating training data influence by tracing gradient descent.
\newblock In Hugo Larochelle, Marc'Aurelio Ranzato, Raia Hadsell, Maria{-}Florina Balcan, and Hsuan{-}Tien Lin (eds.), \emph{Advances in Neural Information Processing Systems 33: Annual Conference on Neural Information Processing Systems 2020, NeurIPS 2020, December 6-12, 2020, virtual}, 2020.
\newblock URL \url{https://proceedings.neurips.cc/paper/2020/hash/e6385d39ec9394f2f3a354d9d2b88eec-Abstract.html}.

\bibitem[Stoehr et~al.(2024)Stoehr, Gordon, Zhang, and Lewis]{stoehrLocalizingParagraphMemorization2024}
Niklas Stoehr, Mitchell Gordon, Chiyuan Zhang, and Owen Lewis.
\newblock Localizing {{Paragraph Memorization}} in {{Language Models}}, March 2024.

\bibitem[Tirumala et~al.(2022)Tirumala, Markosyan, Zettlemoyer, and Aghajanyan]{tirumala2022memorization}
Kushal Tirumala, Aram~H. Markosyan, Luke Zettlemoyer, and Armen Aghajanyan.
\newblock Memorization without overfitting: Analyzing the training dynamics of large language models.
\newblock In Alice~H. Oh, Alekh Agarwal, Danielle Belgrave, and Kyunghyun Cho (eds.), \emph{Advances in Neural Information Processing Systems}, 2022.
\newblock URL \url{https://openreview.net/forum?id=u3vEuRr08MT}.

\bibitem[Warstadt et~al.(2020)Warstadt, Parrish, Liu, Mohananey, Peng, Wang, and Bowman]{warstadt-etal-2020-blimp-benchmark}
Alex Warstadt, Alicia Parrish, Haokun Liu, Anhad Mohananey, Wei Peng, Sheng-Fu Wang, and Samuel~R. Bowman.
\newblock {BL}i{MP}: The benchmark of linguistic minimal pairs for {E}nglish.
\newblock \emph{Transactions of the Association for Computational Linguistics}, 8:\penalty0 377--392, 2020.
\newblock \doi{10.1162/tacl_a_00321}.
\newblock URL \url{https://aclanthology.org/2020.tacl-1.25/}.

\bibitem[Zheng \& Jiang(2022)Zheng and Jiang]{zhengEmpiricalStudyMemorization2022}
Xiaosen Zheng and Jing Jiang.
\newblock An {{Empirical Study}} of {{Memorization}} in {{NLP}}.
\newblock In Smaranda Muresan, Preslav Nakov, and Aline Villavicencio (eds.), \emph{Proceedings of the 60th {{Annual Meeting}} of the {{Association}} for {{Computational Linguistics}} ({{Volume}} 1: {{Long Papers}})}, pp.\  6265--6278, Dublin, Ireland, May 2022. Association for Computational Linguistics.
\newblock \doi{10.18653/v1/2022.acl-long.434}.

\end{thebibliography}
\bibliographystyle{iclr2027_conference}

\newpage
\appendix
\section*{Appendix}

\section{Proofs and additional details on methodology}
\label{app:methodology}

\subsection{Proof of Equation~\ref{eq:acc_decomp}}

We restate Theorem 3.1 and the output-function part of Corollary 3.3 of
\citet{eichin2026explaind}, then apply them to the sequence loss. We collect
the contributions of earlier gradients into the incoming first moment and
retain one contribution per training sequence. In the restatement, we use
$S$ for the number of training steps, reserving $N$ for the fixed sequence-loss
normalizer.

\begin{theorem}[ExPLAIND decomposition under AdamW, \citealt{eichin2026explaind}]
\label{cor:epk_loss_eq}
Let $f_\theta:\mathcal X\to\mathcal Y$ be a model with parameters
$\theta\in\mathbb R^D$, where $\mathcal X\subseteq\mathbb R^I$ and
$\mathcal Y\subseteq\mathbb R^O$. Let
$\mathcal D=\{(x_1,y_1),\ldots,(x_M,y_M)\}$ be a dataset and
$L:\mathcal Y\times\mathcal Y\to\mathbb R_{\geq0}$ a per-sample loss.
Assume that $f_\theta(x)$ is continuously differentiable in $\theta$ and
$L(z,y)$ in $z$ along the parameter segments considered below. At training
step $i$, define
$\mathcal L_i(\theta):=\sum_{k\in B_i}w_{i,k}L(f_\theta(x_k),y_k)$,
where $B_i\subseteq\{1,\ldots,M\}$ and the weights $w_{i,k}\in\mathbb R$
are fixed when taking the gradient.

Suppose that $\theta_S$ is obtained from $\theta_0$ by $S$ AdamW steps with
learning rates $\alpha_s>0$, weight decay $\lambda\geq0$,
$\beta_1,\beta_2\in[0,1)$, and $\epsilon>0$. With $m_0=v_0=0$, the updates are
\begin{equation}
\label{eq:app_adamw}
\begin{aligned}
g_s&:=\nabla_\theta\mathcal L_s(\theta_s),\\
m_{s+1}&:=\beta_1m_s+(1-\beta_1)g_s,\\
v_{s+1}&:=\beta_2v_s+(1-\beta_2)g_s^2,
&\hat v_{s+1}&:=\frac{v_{s+1}}{1-\beta_2^{s+1}},\\
\theta_{s+1}&:=\theta_s-
\frac{\alpha_s}{1-\beta_1^{s+1}}
\frac{m_{s+1}}{\sqrt{\hat v_{s+1}}+\epsilon}
-\alpha_s\lambda\theta_s.
\end{aligned}
\end{equation}
Then, for every $x\in\mathcal X$,
\begin{equation}
\label{eq:app_output_decomp}
\begin{aligned}
    f_{\theta_S}(x)=f_{\theta_0}(x)
    &-\sum_{k=1}^M\sum_{s=0}^{S-1}
      \phi_s^{\mathrm{test}}(x)\,\phi_s^{\mathrm{train}}(x_k)-\sum_{s=0}^{S-1}\phi_s^{\mathrm{test}}(x)\,\mathbf r_s,
\end{aligned}
\end{equation}
where
\begin{equation}
\label{eq:app_feature_maps}
\begin{aligned}
\phi_s^{\mathrm{test}}(x)
&:=\int_0^1\nabla_\theta f_{\theta_s(t)}(x)\,dt
  \in\mathbb R^{O\times D},\\
\phi_s^{\mathrm{train}}(x_k)
&:=\sum_{i=0}^s\alpha_{s,i}\,\mathbf1_{k\in B_i}\,w_{i,k}
\frac{\nabla_\theta L(f_{\theta_i}(x_k),y_k)}
     {\sqrt{\hat v_{s+1}}+\epsilon}
  \in\mathbb R^D,\\
\mathbf r_s&:=\alpha_s\lambda\theta_s,\\
\theta_s(t)&:=\theta_s+t(\theta_{s+1}-\theta_s),\\
\alpha_{s,i}&:=
\frac{\alpha_s(1-\beta_1)\beta_1^{s-i}}{1-\beta_1^{s+1}}.
\end{aligned}
\end{equation}
Here $\nabla_\theta f$ denotes the output Jacobian, while gradients of scalar
losses are column vectors. Powers, square roots, and division of vectors are
coordinatewise.
\end{theorem}

\begin{corollary}[ExPLAIND decomposition for functions of the outputs, \citealt{eichin2026explaind}]
\label{cor:app_output_functions}
Under the assumptions of Theorem~\ref{cor:epk_loss_eq}, let
$h:\mathcal Y\to\mathbb R^Q$ be continuously differentiable on the output
paths, and define $\widetilde f_\theta:=h\circ f_\theta$.
Equation~\ref{eq:app_output_decomp} also holds for $\widetilde f$ after
replacing the test feature map by
\begin{equation}
\label{eq:app_composed_feature_map}
\widetilde\phi_s^{\mathrm{test}}(x)
:=\int_0^1\nabla_\theta\bigl(h(f_{\theta_s(t)}(x))\bigr)\,dt.
\end{equation}
The training feature maps, regularization terms, and optimizer trajectory
remain unchanged. For a scalar output, the integrand in
Eq.~\ref{eq:app_composed_feature_map} is written as a row vector.
\end{corollary}

Combining the theorem and corollary, we now specialize to the sequence loss.
Fix $N\in\mathbb N_{>0}$ independently of sequence length. For a sequence
$x=(x_1,\ldots,x_{|x|})$, set
\begin{equation}
\label{eq:app_sequence_loss}
\ell(\theta;x):=\frac1N\sum_{i=1}^{|x|}
L\bigl(f_\theta(x_{<i}),x_i\bigr).
\end{equation}
For each fixed $x$, apply Corollary~\ref{cor:app_output_functions} to the model's stacked next-token outputs with $h_x(a):=N^{-1}\sum_{i=1}^{|x|}L(a_i,x_i)$. The test feature map is therefore
\begin{equation}
\label{eq:app_loss_feature_map}
    \widetilde\phi_s^{\mathrm{test}}(x)
    =\left(\int_0^1\nabla_\theta\ell(\theta_s(t);x)\,dt\right)^\top
    =\Phi_s(x)^\top.
\end{equation}

For training, the theorem's per-sample loss is the same sequence loss.
Writing $B_s$ for the batch of training sequences, we sum these losses,
so that the theorem's weights are $w_{s,k}=1$ and
\begin{equation}
\label{eq:app_batch_gradient}
    g_s =\sum_{z\in B_s}\nabla_\theta\ell(\theta_s;z)
    =\frac1N\sum_{z\in B_s}\sum_{i=1}^{|z|}\nabla_\theta L\bigl(f_{\theta_s}(z_{<i}),z_i\bigr).
\end{equation}

Redistributing the terms of Eq.~\ref{eq:app_feature_maps}, separating the
current step from earlier steps, gives
\begin{equation}
\label{eq:app_collect_history}
\begin{aligned}
\sum_{k=1}^M\phi_s^{\mathrm{train}}(x_k)
&=\sum_{i=0}^s\alpha_{s,i}
  \frac{g_i}{\sqrt{\hat v_{s+1}}+\epsilon}\\
&=\frac{\alpha_s}{1-\beta_1^{s+1}}
  \frac{(1-\beta_1)g_s+\beta_1m_s}
       {\sqrt{\hat v_{s+1}}+\epsilon}\\
&=\sum_{z\in B_s}u_s^z+u_s^{\mathrm{fm}},
\end{aligned}
\end{equation}
where
\begin{equation}
\label{eq:app_sequence_updates}
\begin{aligned}
u_s^z&:=\frac{\alpha_s(1-\beta_1)}{1-\beta_1^{s+1}}
       \frac{\nabla_\theta\ell(\theta_s;z)}
            {\sqrt{\hat v_{s+1}}+\epsilon},\\
u_s^{\mathrm{fm}}&:=\frac{\alpha_s\beta_1}{1-\beta_1^{s+1}}
       \frac{m_s}{\sqrt{\hat v_{s+1}}+\epsilon},\\
u_s^{\mathrm{reg}}&:=\mathbf r_s=\alpha_s\lambda\theta_s.
\end{aligned}
\end{equation}
Here $m_s$ is the uncorrected first moment entering step $s$, and $s$ retains
the optimizer's original step count. Substituting
Eqs.~\ref{eq:app_loss_feature_map} and~\ref{eq:app_collect_history} into the
loss version of Eq.~\ref{eq:app_output_decomp} with a single update step gives
\begin{equation}
\label{eq:app_sequence_decomp}
\begin{aligned}
\ell(\theta_{s+1};x)-\ell(\theta_s;x)
&=-\Phi_s(x)^\top
  \left(\sum_{z\in B_s}u_s^z+u_s^{\mathrm{fm}}+u_s^{\mathrm{reg}}\right)\\
&=\sum_{z\in B_s}\mathcal I_s^z(x)
  +\mathcal I_s^{\mathrm{fm}}(x)+\mathcal I_s^{\mathrm{reg}}(x),
\end{aligned}
\end{equation}
where $\mathcal I_s^\rho(x):=-\Phi_s(x)^\top u_s^\rho$.
This yields Eq.~\ref{eq:acc_decomp} with sequence-indexed data contributions.

\noindent\textit{Batch reduction.}
The specialization above assumes the gradient in
Eq.~\ref{eq:app_batch_gradient}. If the implementation additionally averages
over training sequences, multiply each $u_s^z$ by $|B_s|^{-1}$ and form the
moment states using that batch-mean gradient. This is separate from the
fixed token-loss normalization. 

\noindent\textit{Batch gradient clipping.}
If the accumulated batch gradient is norm-clipped before the AdamW moment
updates, let $c_s$ be its realized scalar clipping factor, so that the
gradient supplied to AdamW is
$c_s\sum_{z\in B_s}\nabla_\theta\ell(\theta_s;z)$ under the sum convention.
The proof applies with weights $w_{s,k}=c_s$, treating $c_s$ as fixed
when attributing the realized update. Accordingly, multiply each $u_s^z$
in Eq.~\ref{eq:app_sequence_updates} by the same $c_s$.

\subsection{Estimating the second-moment state of Pythia}
\label{app:second_moment}

Since intermediate Pythia checkpoints are released without optimizer state, we reconstruct the Adam second moment $v$ empirically. With $\beta_2 = 0.95$, the exponential moving average $v_t = \beta_2 v_{t-1} + (1-\beta_2) g_t^2$ assigns weight $(1-\beta_2)\beta_2^{k}$ to the squared gradient $k$ steps in the past, an effective horizon of $(1-\beta_2)^{-1} = 20$ steps. Assuming the gradient distribution is approximately stationary over this window, $v$ at a checkpoint is well approximated by the expected squared per-batch gradient at that checkpoint, $\mathbb{E}[g_B^2]$, where $g_B$ is the per-coordinate gradient of a training batch of $B$ i.i.d.\ sequences. 

Writing per-sequence gradients as $g_i = \mu + \varepsilon_i$ with full-data gradient $\mu$ and i.i.d.\ zero-mean noise $\varepsilon_i$ of variance $\sigma^2$, the batch gradient satisfies $\mathbb{E}[g_B^2] = \mu^2 + \sigma^2/B$. Estimating this target from batches of $b \ll B$ sequences is therefore biased: $\mathbb{E}[g_b^2] = \mu^2 + \sigma^2/b$, which exceeds the target by $(1/b - 1/B)\,\sigma^2$, worst case up to a factor $B/b$ in the noise-dominated regime $\mu^2 \ll \sigma^2/B$, and this bias does not vanish with more samples.

Instead, we compute $N$ per-sequence gradients $g_1,\dots,g_N$ (clipped as in training) on held-out Pile data disjoint from the batches whose contributions we decompose, and form the per-coordinate sample mean $\bar{g}$ and sample variance $s^2$. Since $\mathbb{E}[\bar{g}^2] = \mu^2 + \sigma^2/N$ and $\mathbb{E}[s^2] = \sigma^2$,
the plug-in estimator
\begin{equation}
    \hat{v} \;=\; \max\!\left(\bar{g}^{\,2} - \tfrac{s^2}{N},\, 0\right)
    \;+\; \frac{s^2}{B}
\end{equation}
satisfies $\mathbb{E}[\hat{v}] = \mu^2 + \sigma^2/B$ exactly, apart from the
truncation at zero, which introduces a small positive bias only where
$\mu^2$ is statistically indistinguishable from zero; there the retained term
$s^2/B$ dominates the target in any case. We use $B = 1024$, the Pythia
training batch size.

To choose $N$, note that in the noise-dominated regime the variance of a single squared gradient is $\operatorname{Var}(\varepsilon_i^2) = (\kappa - 1)\,\sigma^4$, where $\kappa$ is the noise kurtosis ($\kappa = 3$ for Gaussian noise), so the relative standard error of $\hat{v}$ scales as $\sqrt{(\kappa-1)/N}$. Adam applies $\hat{v}$ through $\sqrt{\hat{v}}$ in the update denominator, which halves the relative error to first order. With $N = 256$ this yields a relative error in the update of roughly $\tfrac{1}{2}\sqrt{2/256} \approx 4\%$ under Gaussian noise, and remains below $10\%$ even for heavy-tailed gradients with $\kappa \le 13$. Both figures are small compared to the intrinsic sampling fluctuation of the quantity being replaced: the true $20$-step EMA is itself an average of around $20$ noisy squared gradients and thus carries relative fluctuations of order $\sqrt{2/20} \approx 30\%$. Finally, because $\hat{v}$ is estimated on data independent of the update batch, the reconstructed update remains linear in the gradients being attributed.

\section{Additional experiment settings}
\label{app:experiment_settings}

\subsection{Models}
\label{app:model}
We use \texttt{EleutherAI/pythia-<size>-deduped} GPT NeoX checkpoints loaded from huggingface and the corresponding tokenizer. Our decomposition implementation is a fork of the original ExPLAIND repository, extended with the Pythia reconstruction, sample level decomposition, prediction, and ablation code used here. Models are loaded through \texttt{FixedLossGPTNeoXForCausalLM} with Flash Attention 2, and dropout is disabled. Computation uses \texttt{float32} by default and \texttt{bfloat16} for the 6.9B model to reduce memory use. We restricted each decomposition to one H200 GPU; we did not decompose the 12B model because of the computational cost. We reconstruct AdamW with Pythia's learning rate schedule, betas $(0.9,0.95)$, epsilon $10^{-8}$, weight decay $0.01$, gradient clipping with norm limit a $1$, and zero the first moments. Second moments are estimated from the 256 sequences above. The path integral in the test feature map of ExPLAIND uses $100$ integration steps following \citet{eichin2026explaind}'s recommendation.

\subsection{Data}
\label{app:data}

For each selected deduplicated Pythia model, we sample the first 64 Pythia tokenizer tokens of 32 examples in each of four classes. The classes are defined by 32-extractability in as precomputed by \citet{bidermanEmergentPredictableMemorization2023} (code at github repo \texttt{EleutherAI/pythia-memorized-evals}) and by the occurrence count of the 32-token continuation in the deduplicated Pile, taken from huggingface dataset \texttt{usvsnsp/deduped-num-duplicates}: \emph{memorized\_recited} (extractable, count $>5$), \emph{memorized\_recollected} (extractable, count $\leq5$), \emph{control\_duplicated} (not extractable, count $>5$), and \emph{control} (not extractable, count $\leq5$), following \citet{prashanth2025recite}. Controls come from the tail of the preshuffled deduplicated Pile mmap and exclude memorized indices. The 128 prediction examples pass a text filter that rejects excessive capitalization, selected code characters, more than five digits, and repeated substrings. The reconstruction uses 1,024 sequences of 2,048 tokens for the original update and a disjoint set of 256 sequences to estimate the second moment; these sequences also come from the Pile tail but are not subject to the text filter.

\subsection{ExPLAIND decomposition and features}
Our implementation is a fork of the original ExPLAIND repository, extended with the Pythia reconstruction, sample level decomposition, prediction, and ablation code used here. We evaluate checkpoints at steps $0,1,2,4,8,16,32,64,128,256,512$, every $1,000$ steps from $1,000$ through $9,000$, and every $10,000$ steps from $10,000$ through $130,000$.  We decompose all 128 sequences separately and retain training sequences as separate, accumulated group. Sequence loss follows Pythia's update convention, namely the token loss sum divided by 2,048.

\subsection{Memorization prediction}
For the prediction, at each checkpoint we compute five features for every prediction sequence: cosine alignment with the original training update, cosine alignment with the matching sequence update, the prediction feature norm, the matching update feature norm, and alignment with the weight decay contribution. Dot products and squared norms are summed across parameters. Prediction uses the features of one checkpoint at a time; accumulated scores are used only for the descriptive decomposition plots and ablations. The plots report class means and standard deviations for both the memorized versus control grouping and the four classes. We also evaluate BLiMP at every checkpoint by comparing the summed sentence log probabilities in each minimal pair.

We fit ridge linear models with an unpenalized intercept and coefficient $10^{-8}$ using six stratified folds generated with seed 42. Feature standardization is fitted only on the training portion of each fold. Binary predictions threshold the scalar output at $0.5$. We evaluate decomposition features, CE, and decomposition+CE, using one checkpoint at a time. We report accuracy from predictions on the excluded folds and retain the scores, predictions, labels, fold assignments, and fitted weights.

\subsection{Causal ablation}
For each checkpoint, we sum parameter influences from the 64 memorized sequences over that checkpoint and all earlier checkpoints. We select the most negative fractions specified by the ablation script and replace those parameters with zero. Normalization parameters are always excluded. We report separate scopes that exclude embeddings or include them; each fraction is relative to the eligible parameters in its scope. Random masks with the same number of parameters provide a baseline. After intervention, we measure teacher forced greedy accuracy following a 32-token prefix, mean token loss, and the number of sequences for which all 32 continuation tokens are decoded greedily, both for the binary groups and for each of the four classes. We additionally measure control CE and the fraction of greedily generated control tokens changed by the intervention.

\section{Additional results}
\label{app:results}

\subsection{Training dynamics and gradient norms across scales}
\label{app:additional_dynamics}
\begin{figure}[h]
    \centering
    Pythia-160m \\
    \begin{subfigure}{0.245\linewidth}
        \includegraphics[width=1\linewidth]{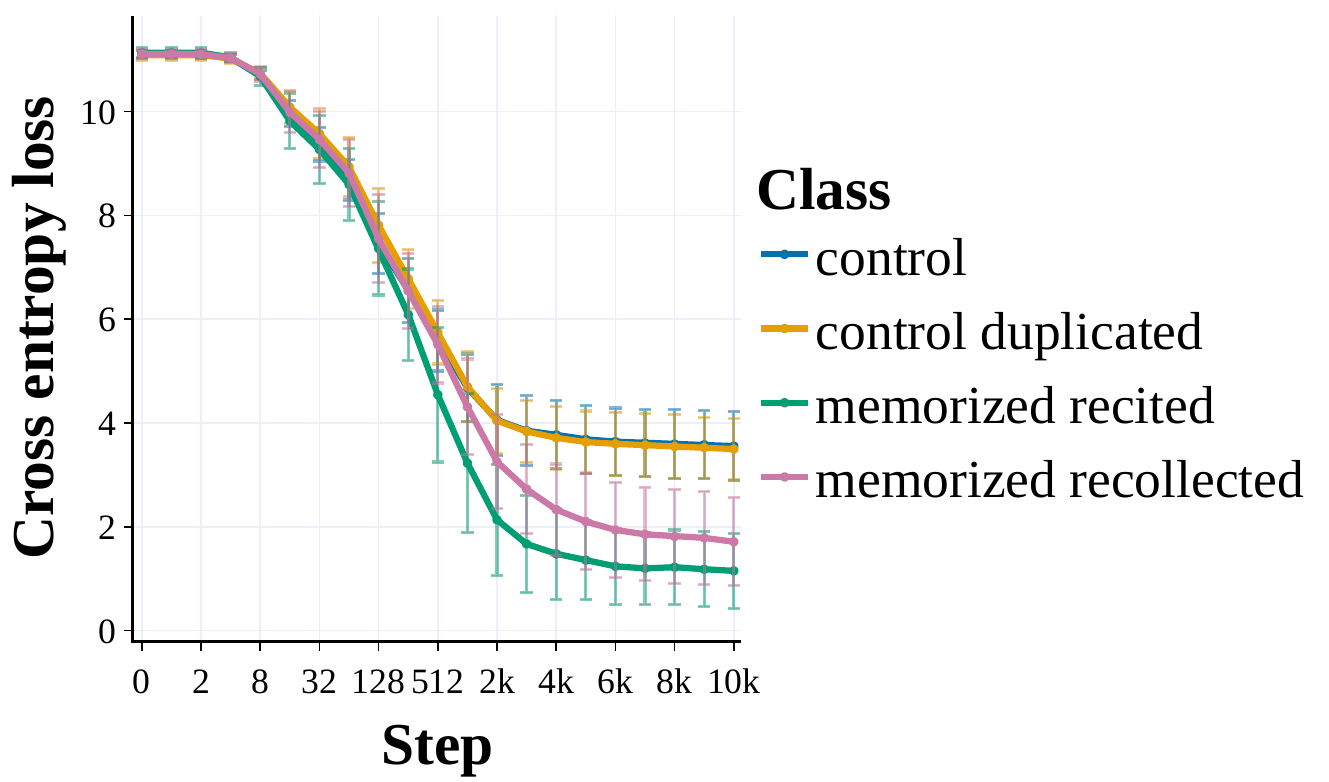}
    \end{subfigure}
    \begin{subfigure}{0.245\linewidth}
        \includegraphics[width=1\linewidth]{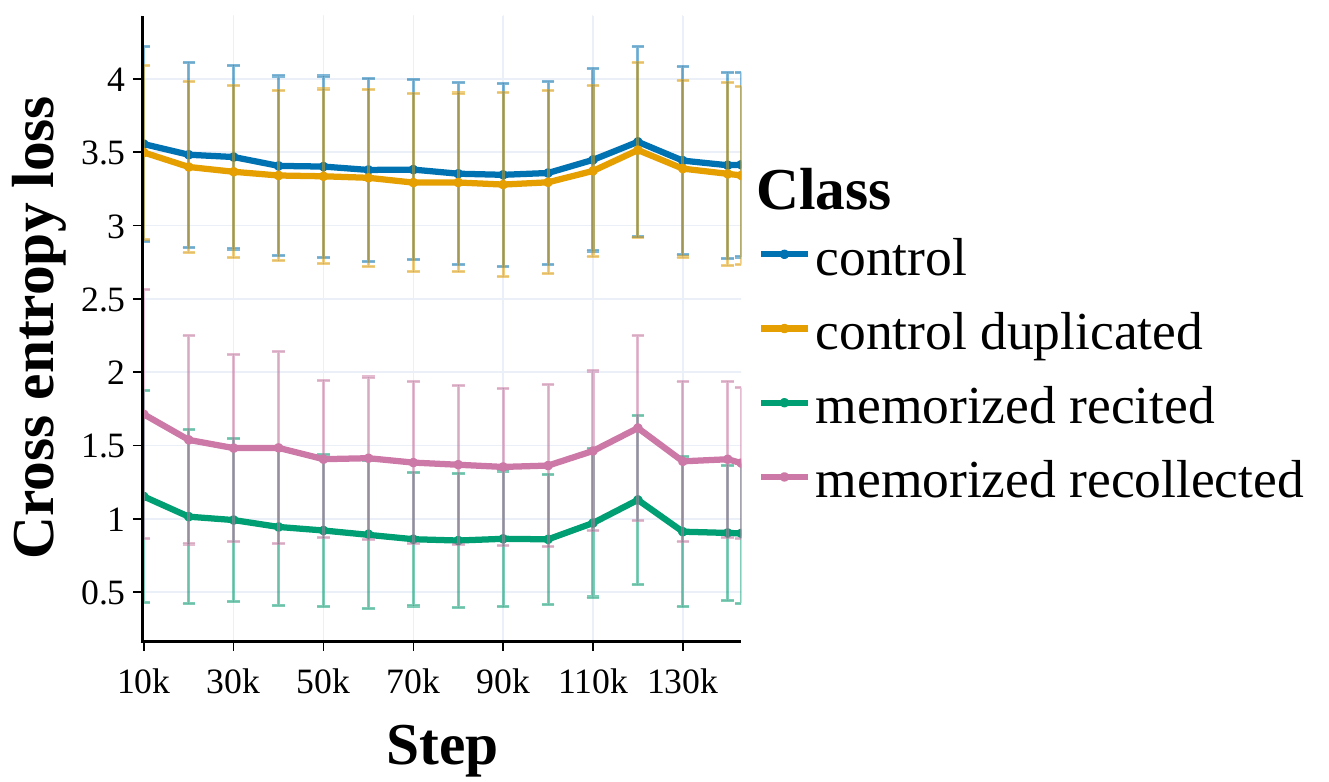}
    \end{subfigure}
    \begin{subfigure}{0.245\linewidth}
        \includegraphics[width=1\linewidth]{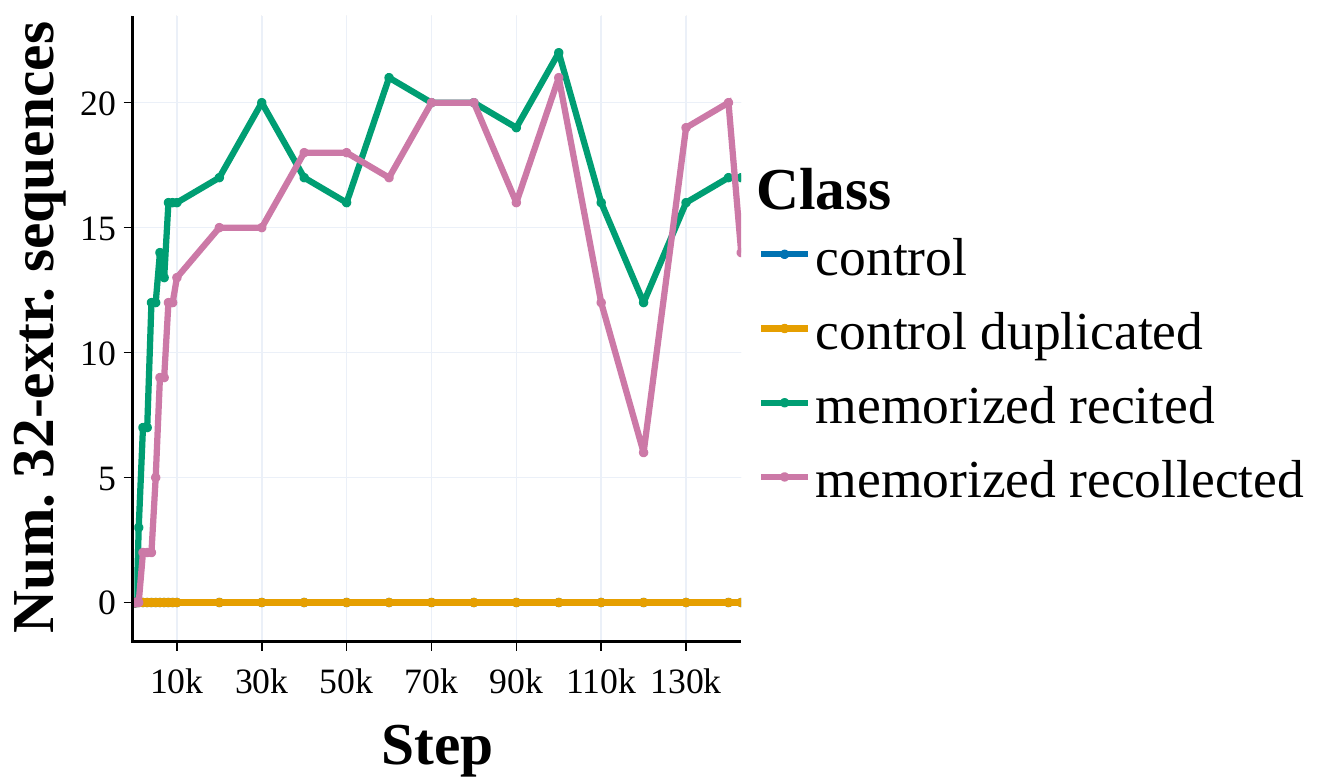}
    \end{subfigure}
    \begin{subfigure}{0.245\linewidth}
        \includegraphics[width=1\linewidth]{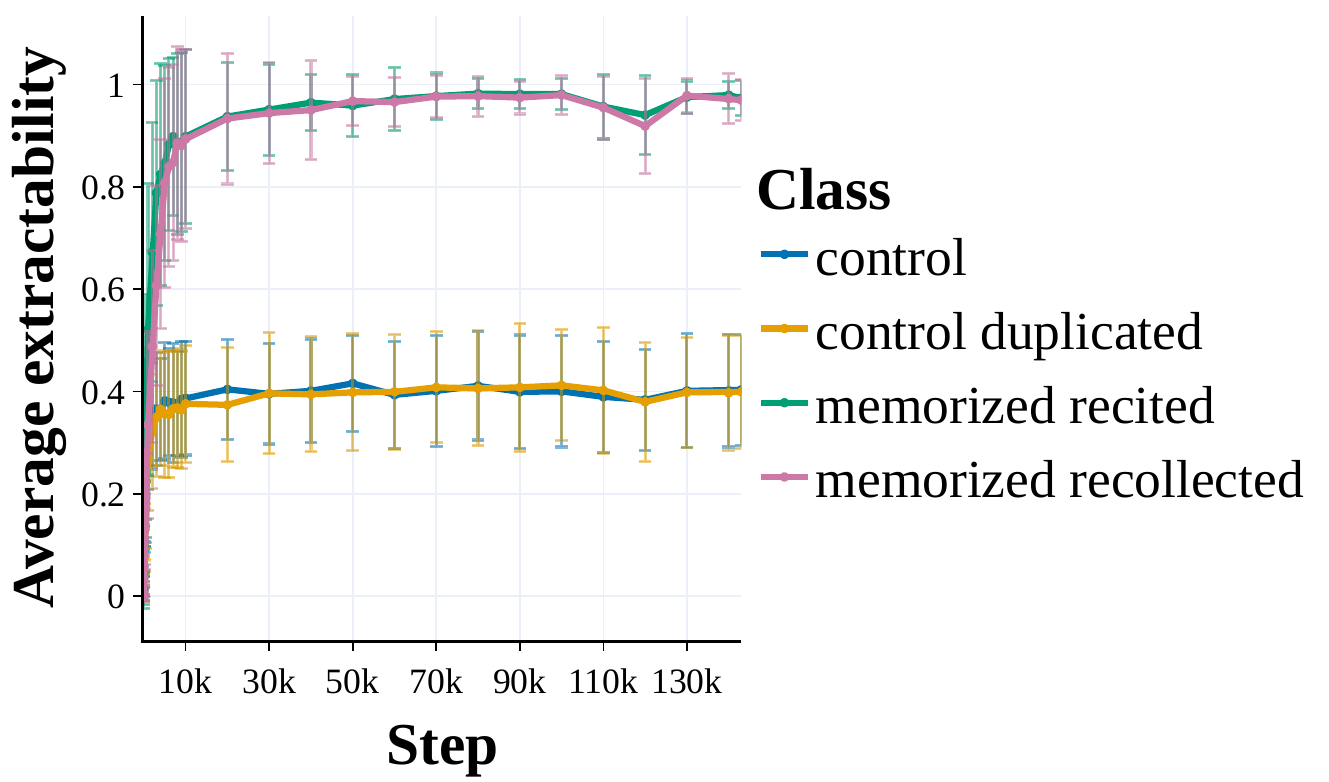}
    \end{subfigure}
    Pythia-410m \\
    \begin{subfigure}{0.245\linewidth}
        \includegraphics[width=1\linewidth]{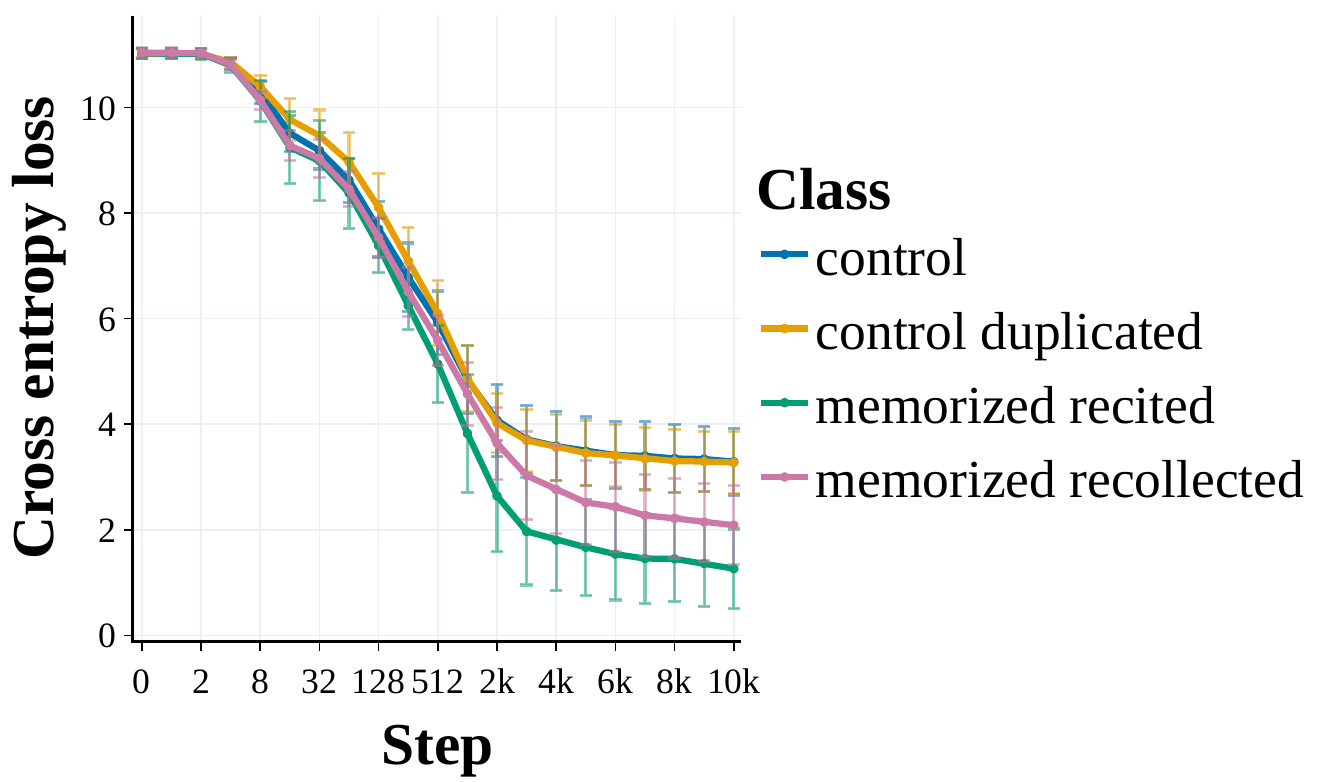}
    \end{subfigure}
    \begin{subfigure}{0.245\linewidth}
        \includegraphics[width=1\linewidth]{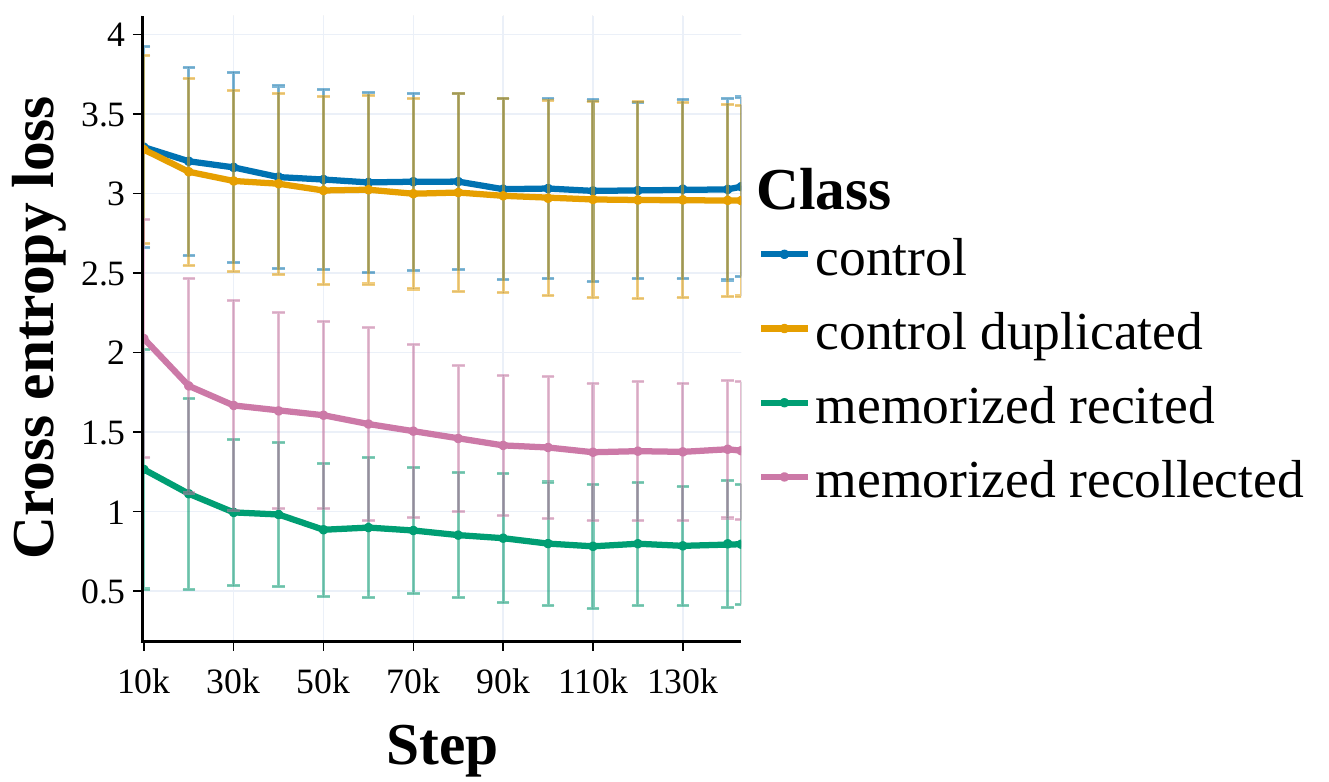}
    \end{subfigure}
    \begin{subfigure}{0.245\linewidth}
        \includegraphics[width=1\linewidth]{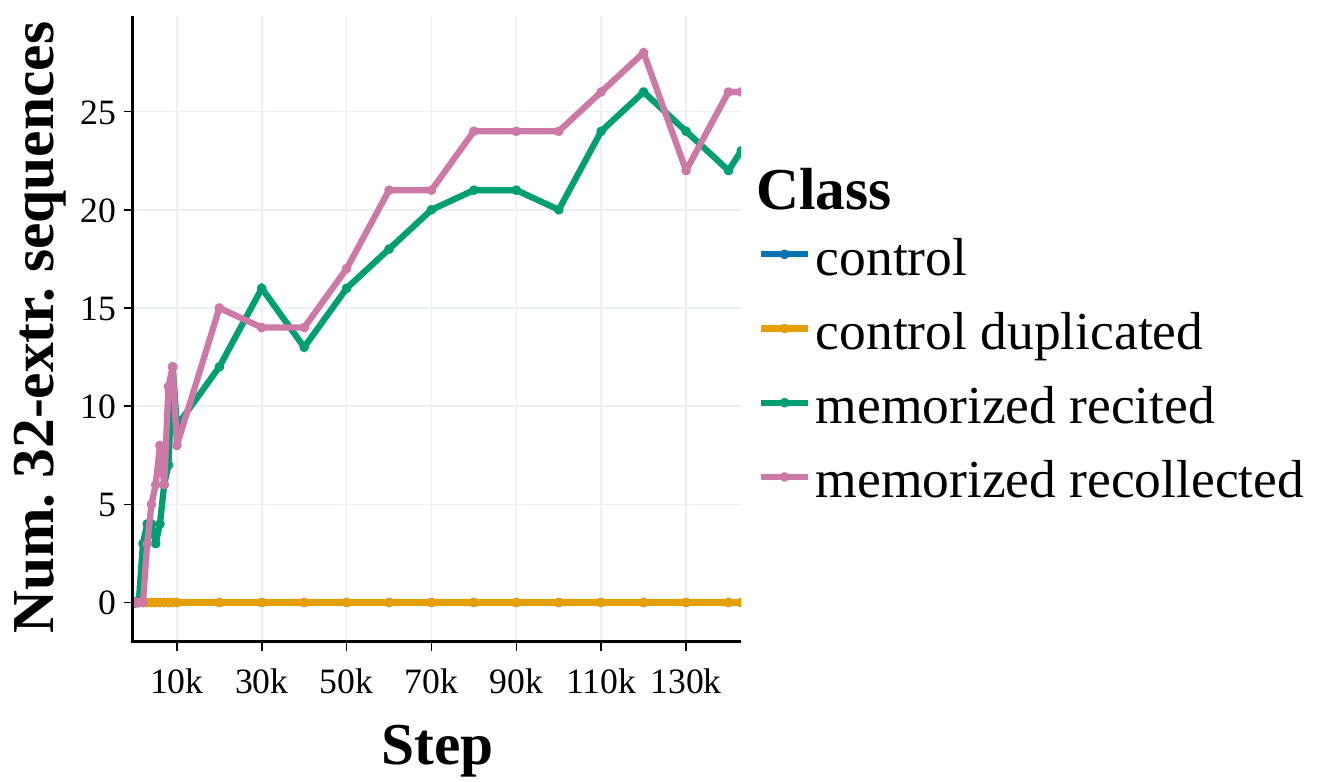}
    \end{subfigure}
    \begin{subfigure}{0.245\linewidth}
        \includegraphics[width=1\linewidth]{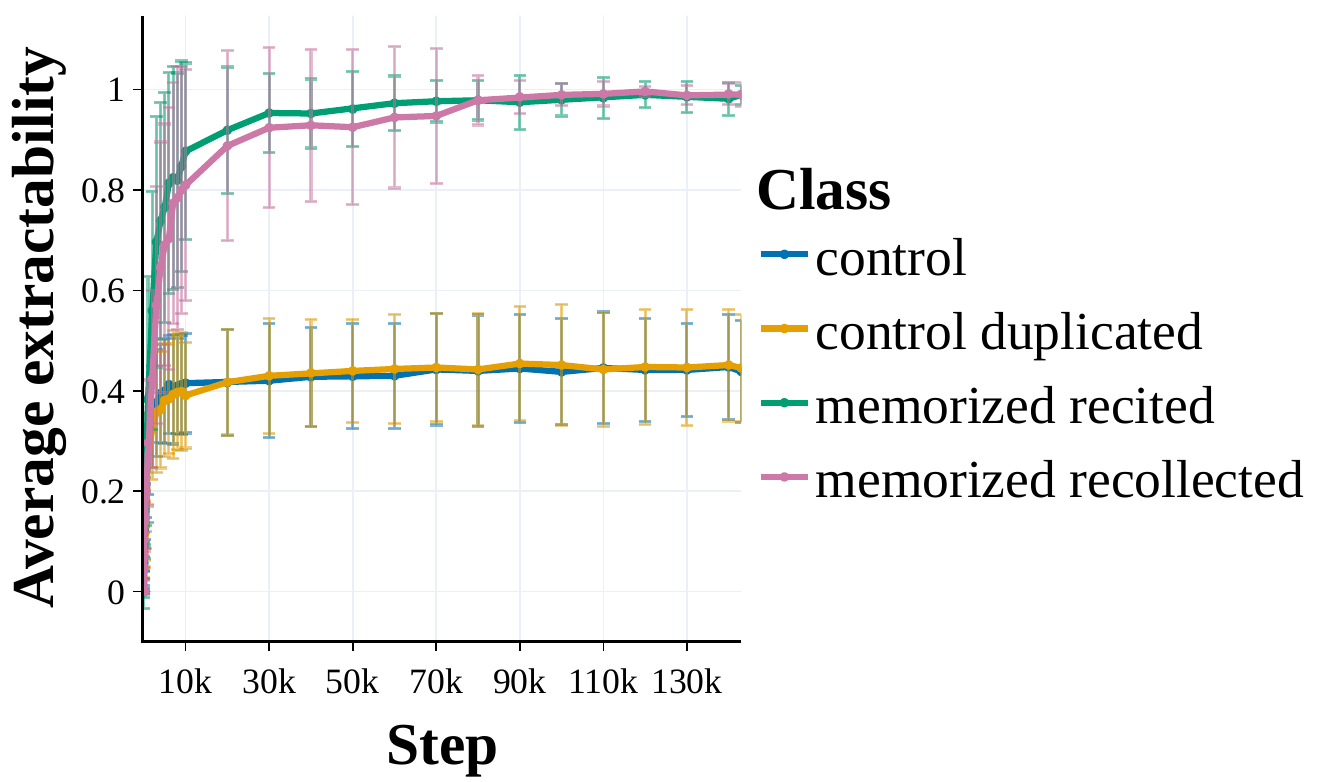}
    \end{subfigure}
    Pythia-1B \\
    \begin{subfigure}{0.245\linewidth}
        \includegraphics[width=1\linewidth]{figures/checkpoint_evals/phases/through_10000/accumulated/four_classes/1b/loss.pdf}
    \end{subfigure}
    \begin{subfigure}{0.245\linewidth}
        \includegraphics[width=1\linewidth]{figures/checkpoint_evals/phases/from_10000/accumulated/four_classes/1b/loss.pdf}
    \end{subfigure}
    \begin{subfigure}{0.245\linewidth}
        \includegraphics[width=1\linewidth]{figures/checkpoint_evals/accumulated/four_classes/1b/is_32_extractable.pdf}
    \end{subfigure}
    \begin{subfigure}{0.245\linewidth}
        \includegraphics[width=1\linewidth]{figures/checkpoint_evals/accumulated/four_classes/1b/extractability.pdf}
    \end{subfigure}
    Pythia-1.4b \\
    \begin{subfigure}{0.245\linewidth}
        \includegraphics[width=1\linewidth]{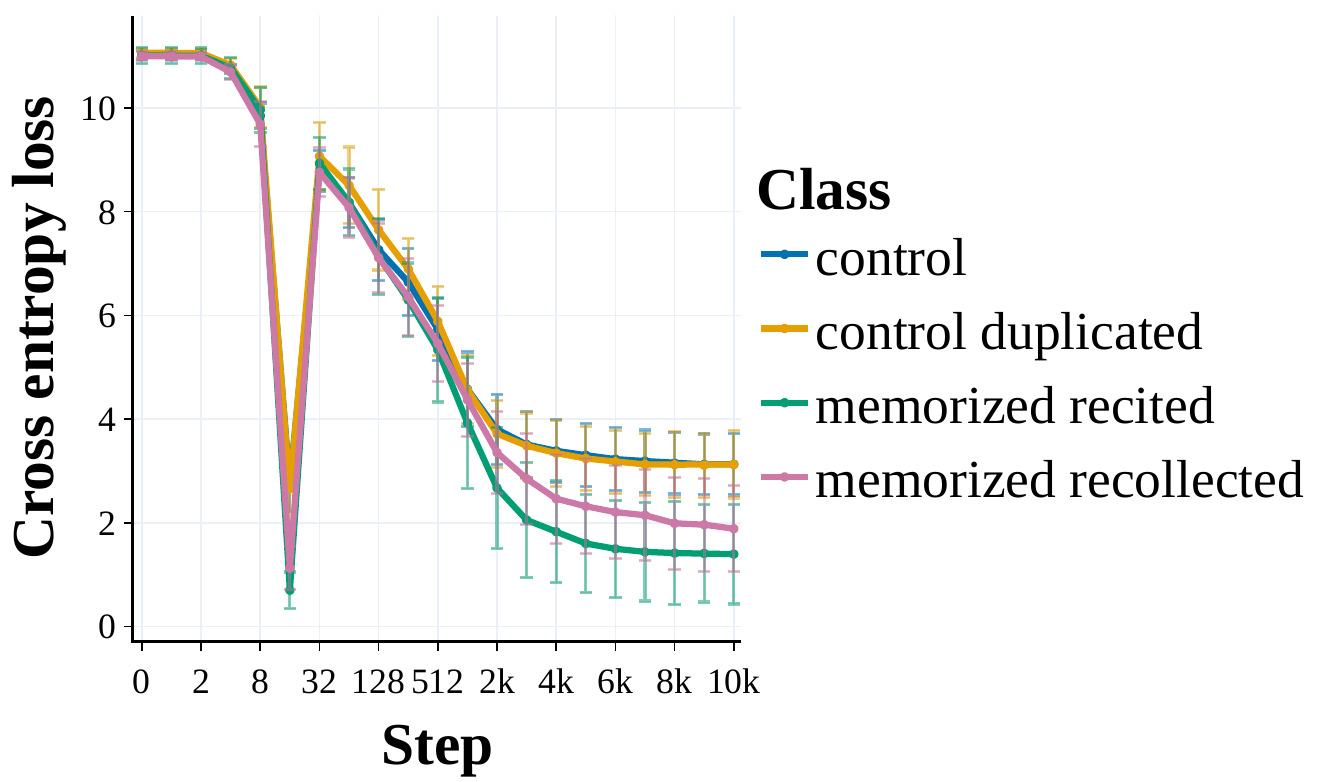}
    \end{subfigure}
    \begin{subfigure}{0.245\linewidth}
        \includegraphics[width=1\linewidth]{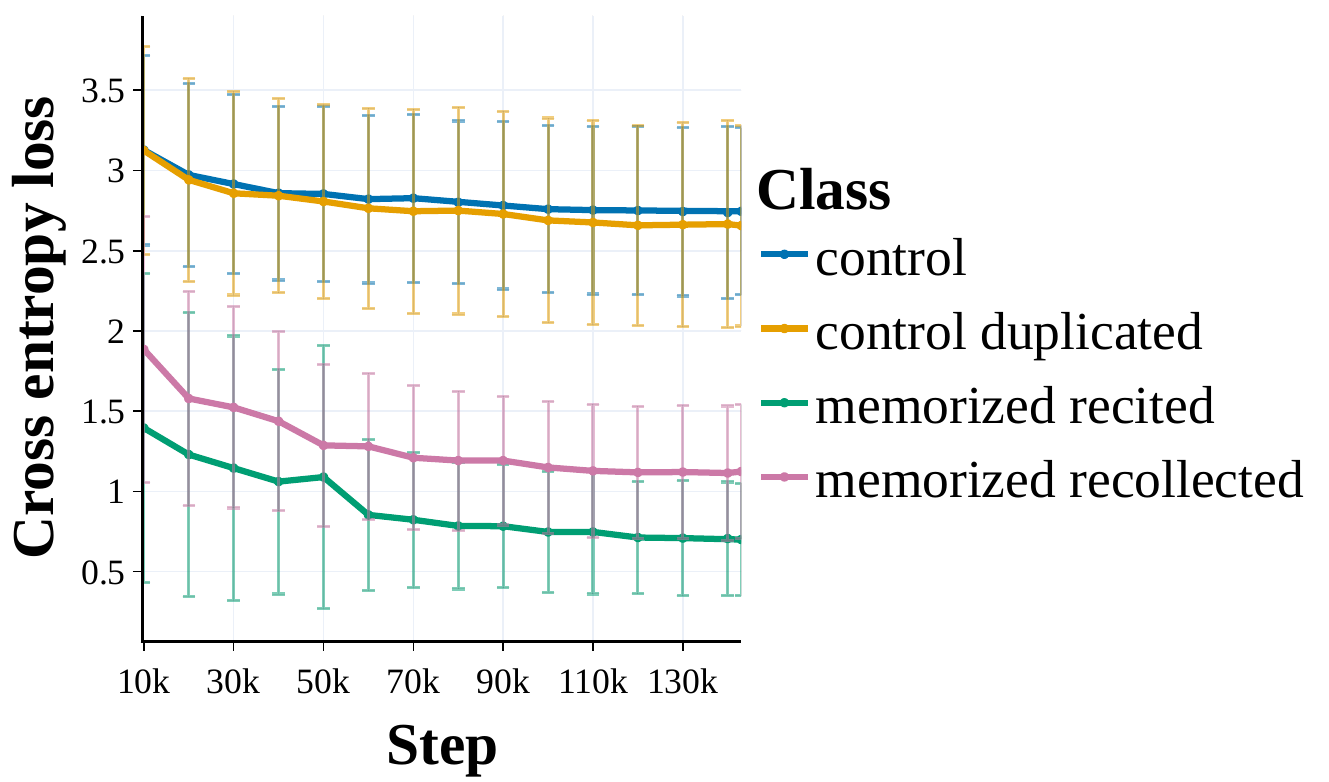}
    \end{subfigure}
    \begin{subfigure}{0.245\linewidth}
        \includegraphics[width=1\linewidth]{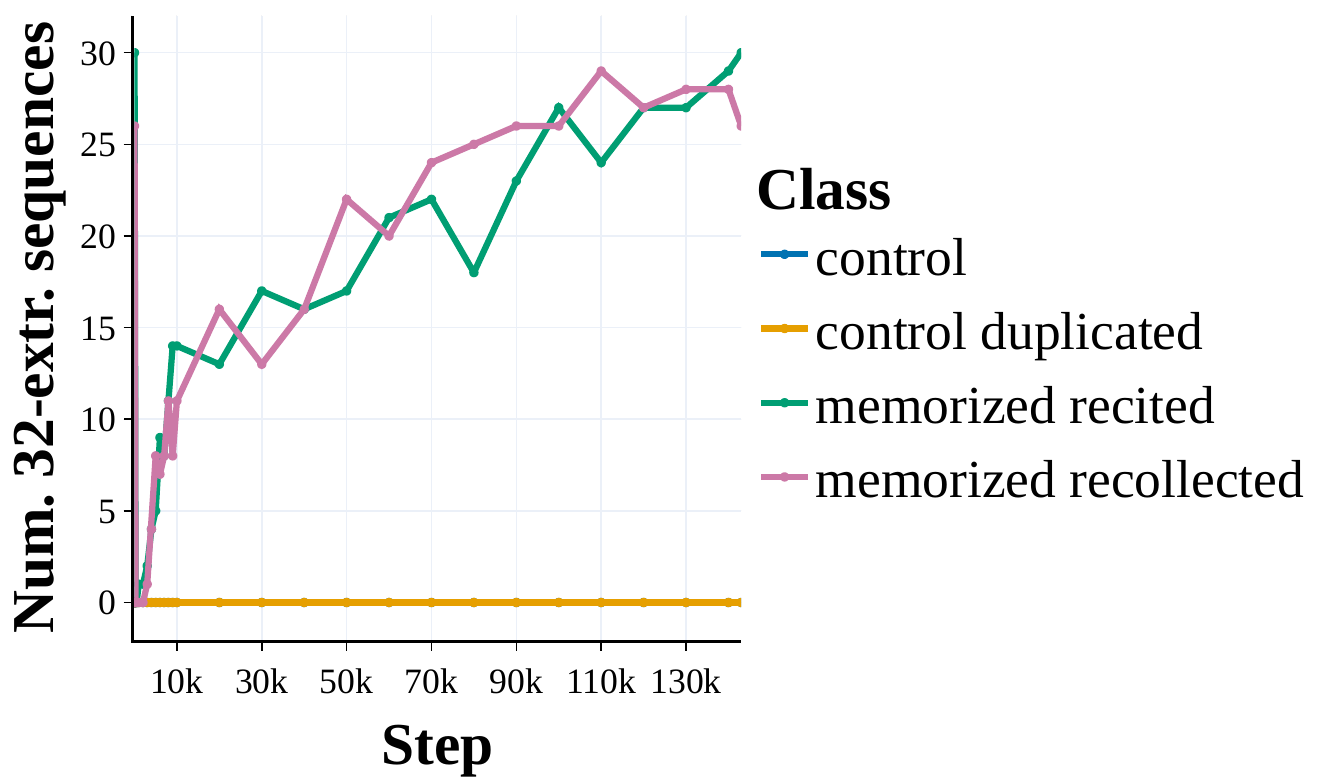}
    \end{subfigure}
    \begin{subfigure}{0.245\linewidth}
        \includegraphics[width=1\linewidth]{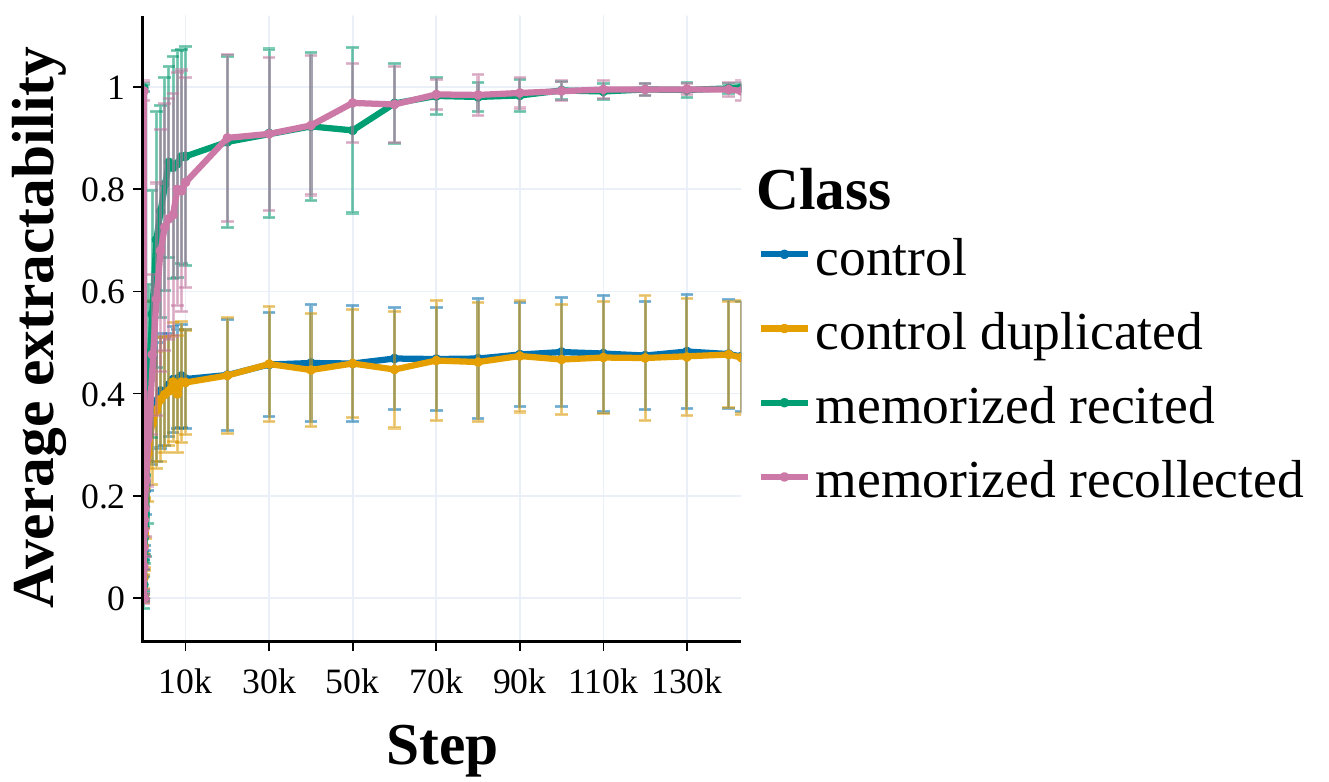}
    \end{subfigure}
    Pythia-2.8b \\
    \begin{subfigure}{0.245\linewidth}
        \includegraphics[width=1\linewidth]{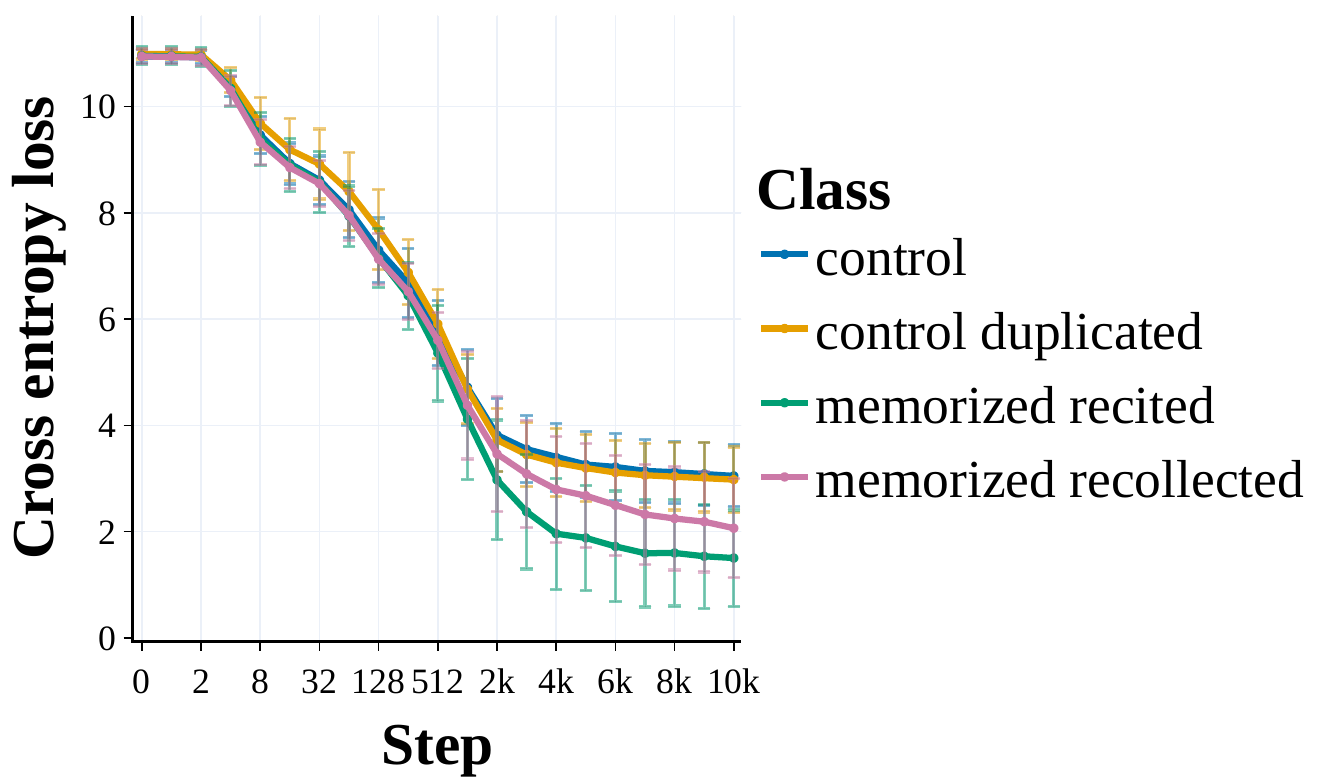}
    \end{subfigure}
    \begin{subfigure}{0.245\linewidth}
        \includegraphics[width=1\linewidth]{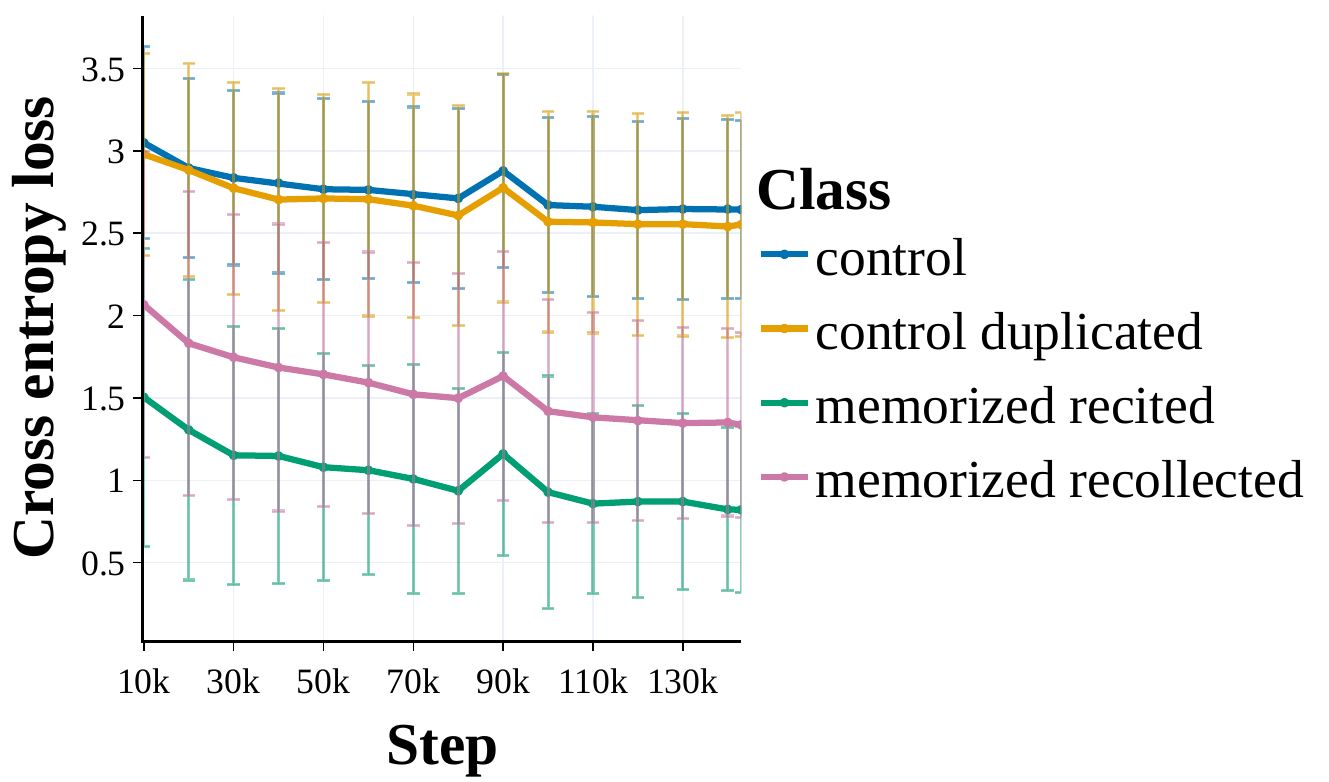}
    \end{subfigure}
    \begin{subfigure}{0.245\linewidth}
        \includegraphics[width=1\linewidth]{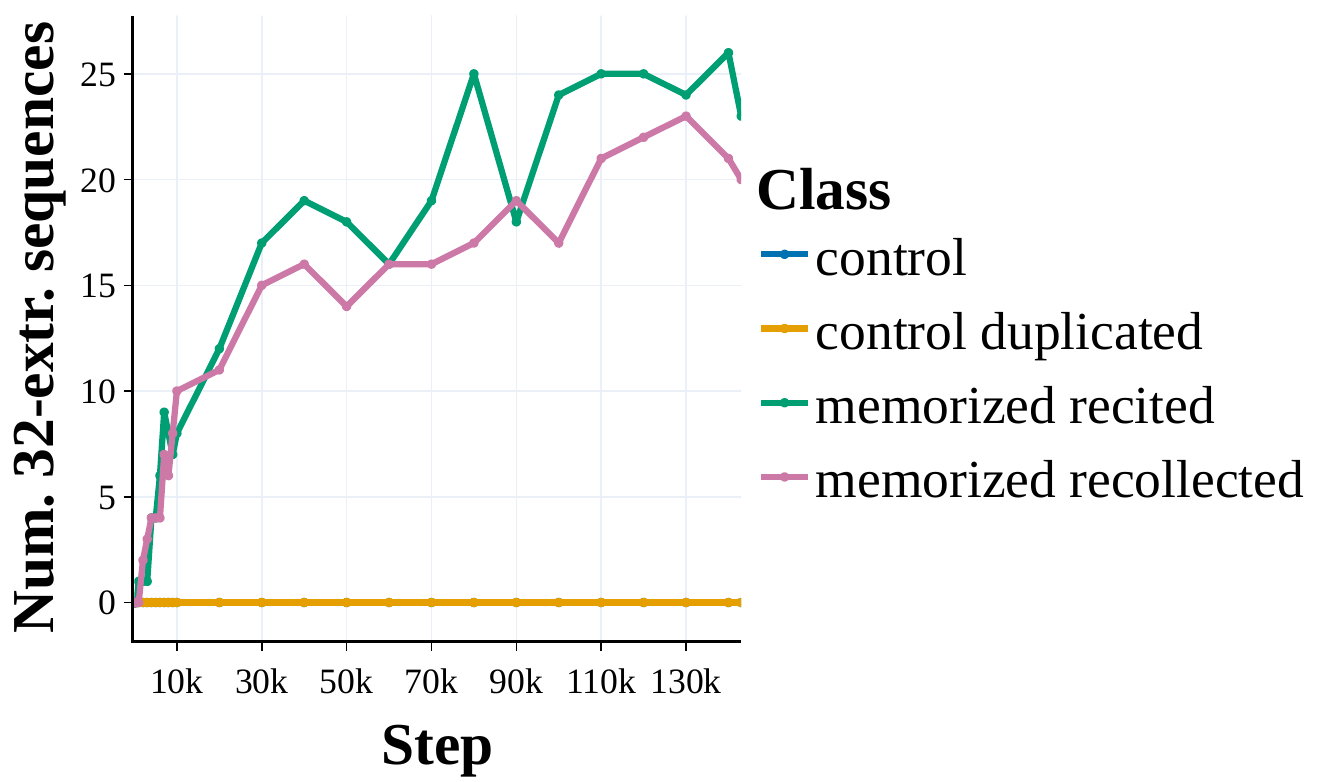}
    \end{subfigure}
    \begin{subfigure}{0.245\linewidth}
        \includegraphics[width=1\linewidth]{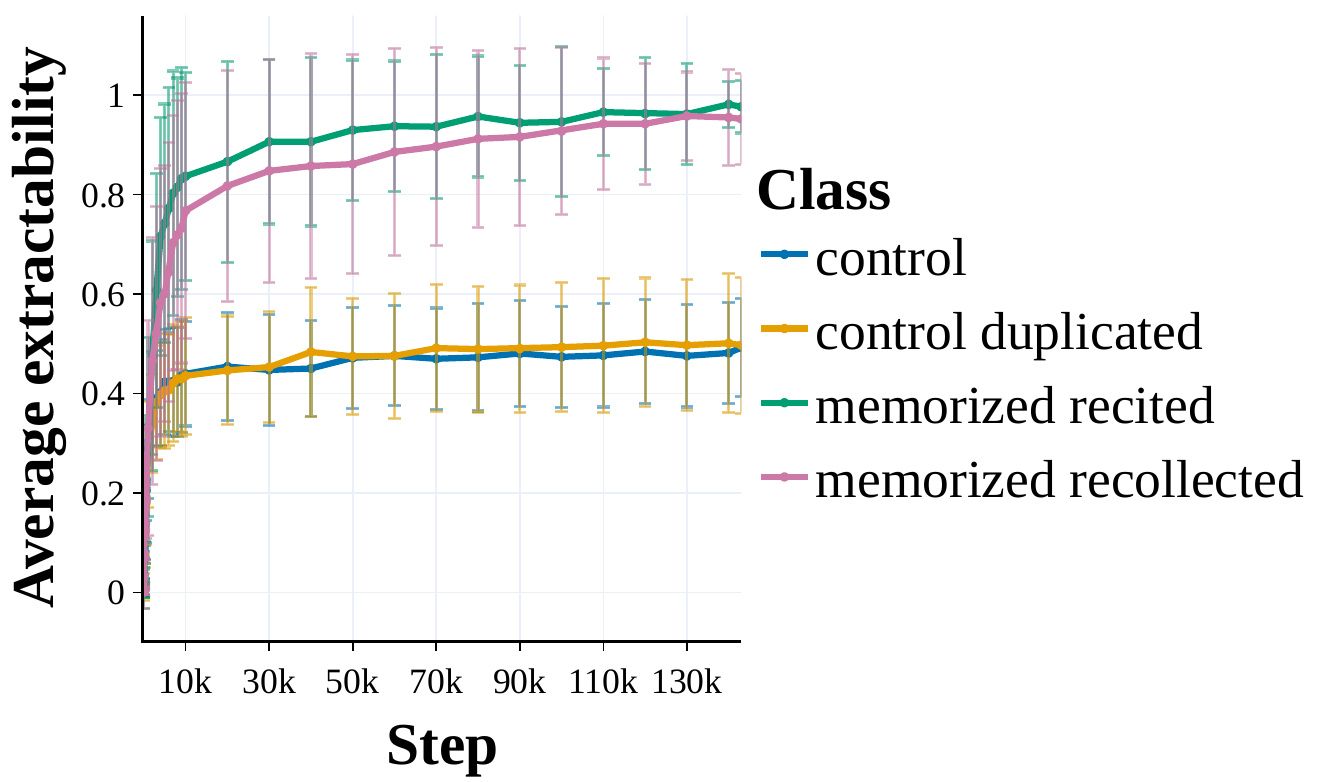}
    \end{subfigure}
    Pythia-6.9B \\
    \begin{subfigure}{0.245\linewidth}
        \includegraphics[width=1\linewidth]{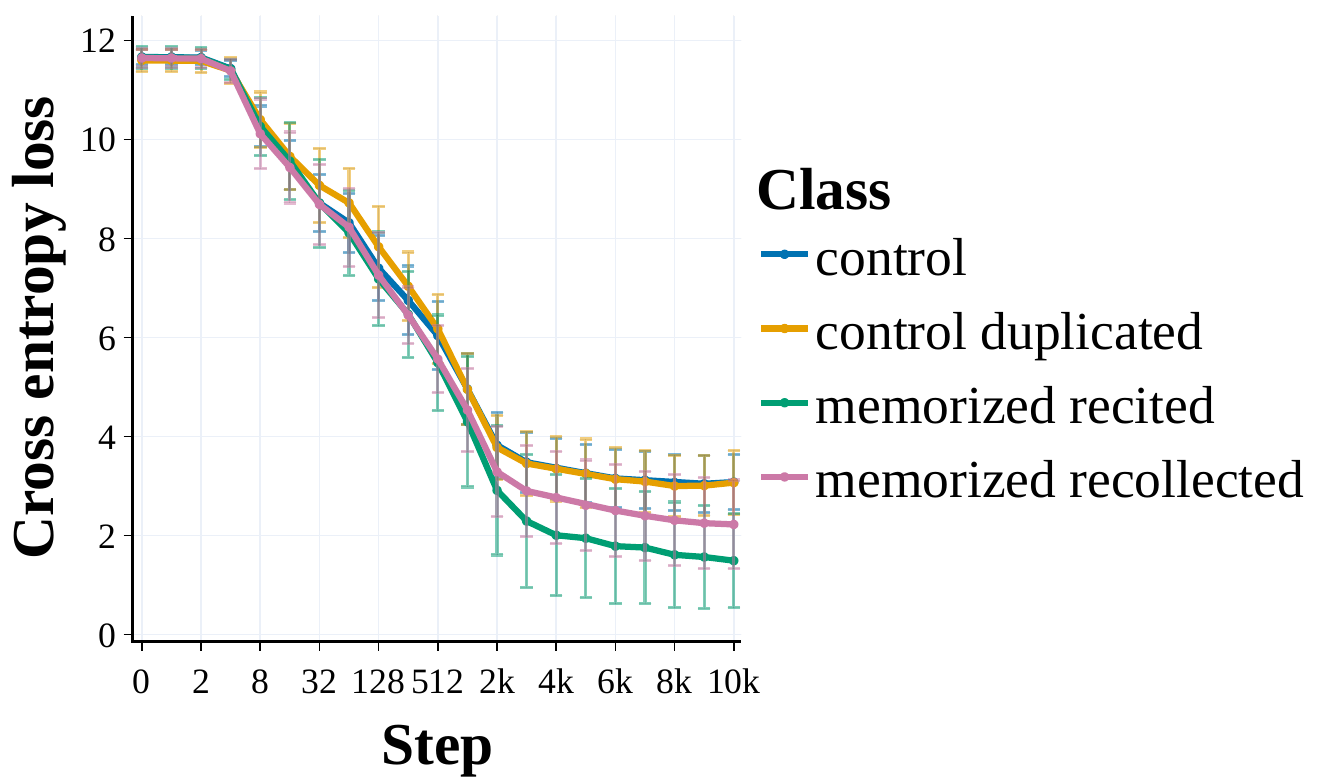}
    \end{subfigure}
    \begin{subfigure}{0.245\linewidth}
        \includegraphics[width=1\linewidth]{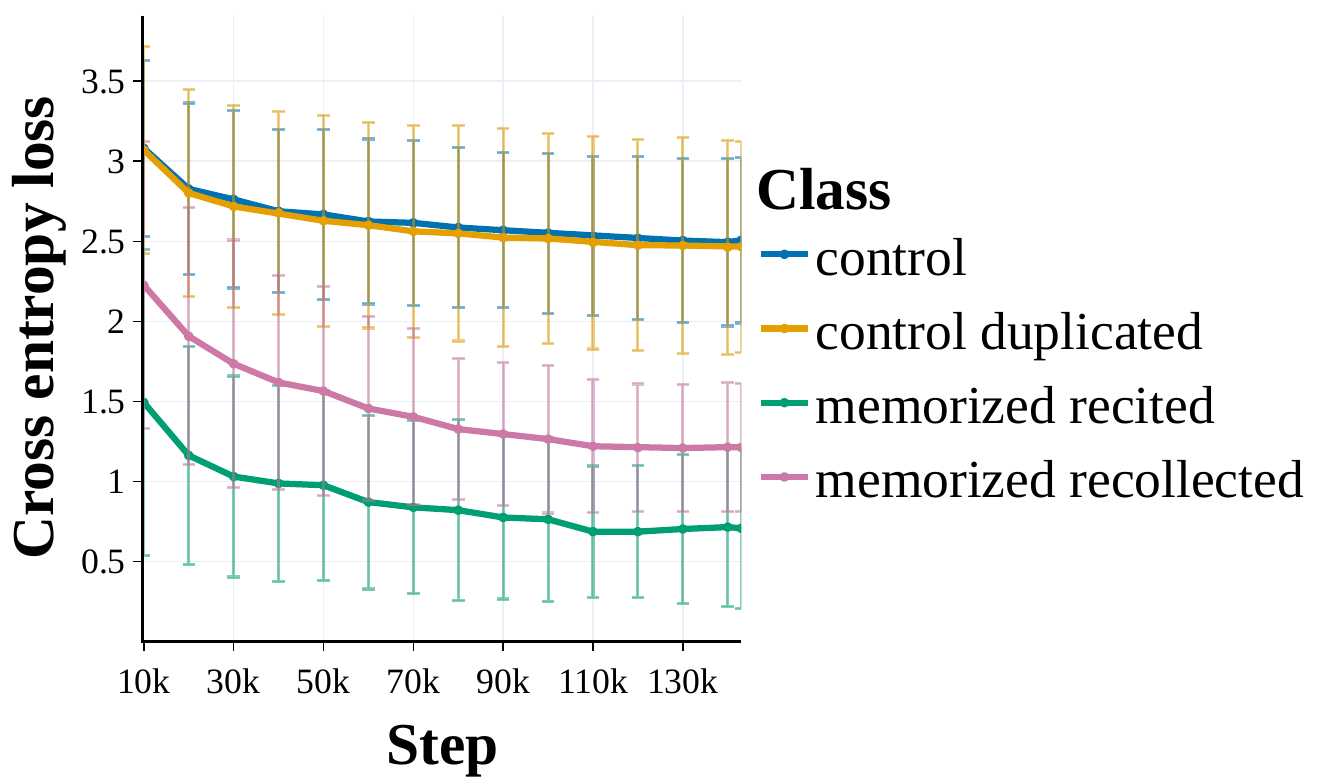}
    \end{subfigure}
    \begin{subfigure}{0.245\linewidth}
        \includegraphics[width=1\linewidth]{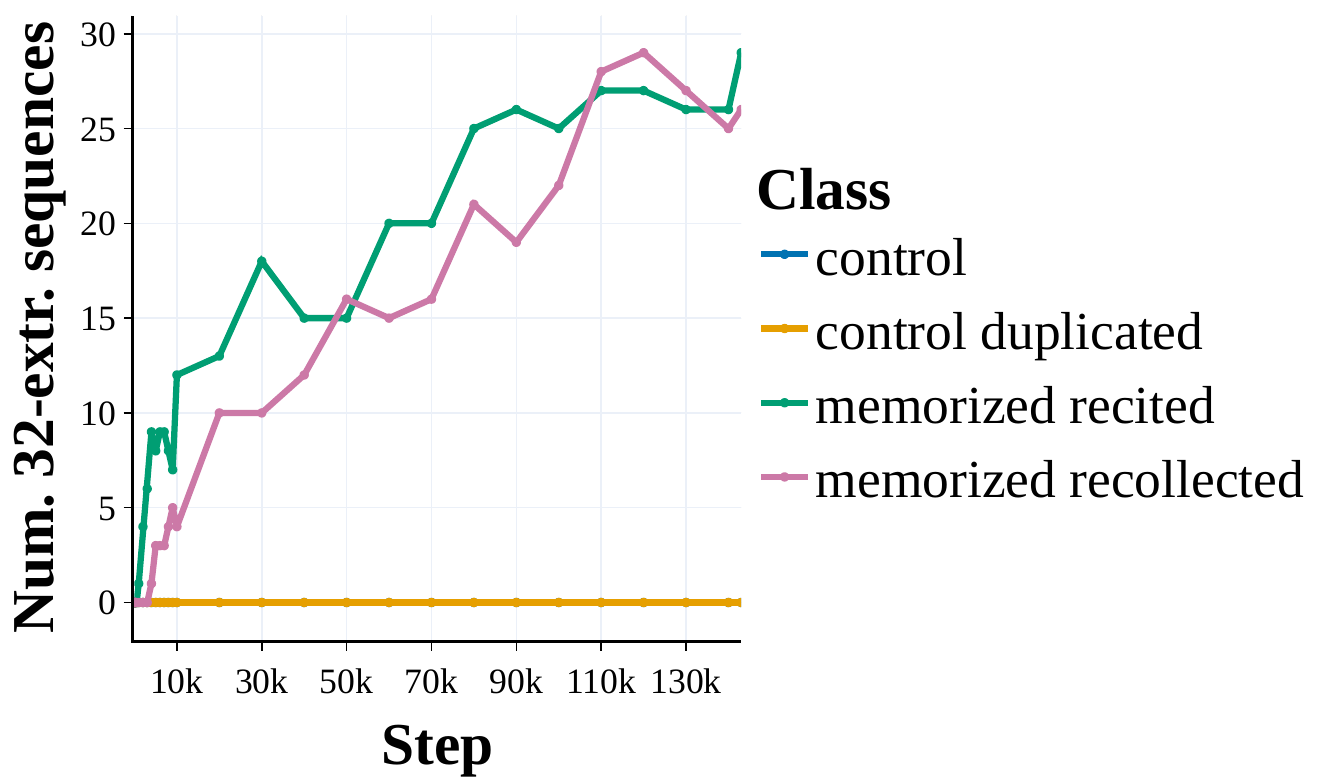}
    \end{subfigure}
    \begin{subfigure}{0.245\linewidth}
        \includegraphics[width=1\linewidth]{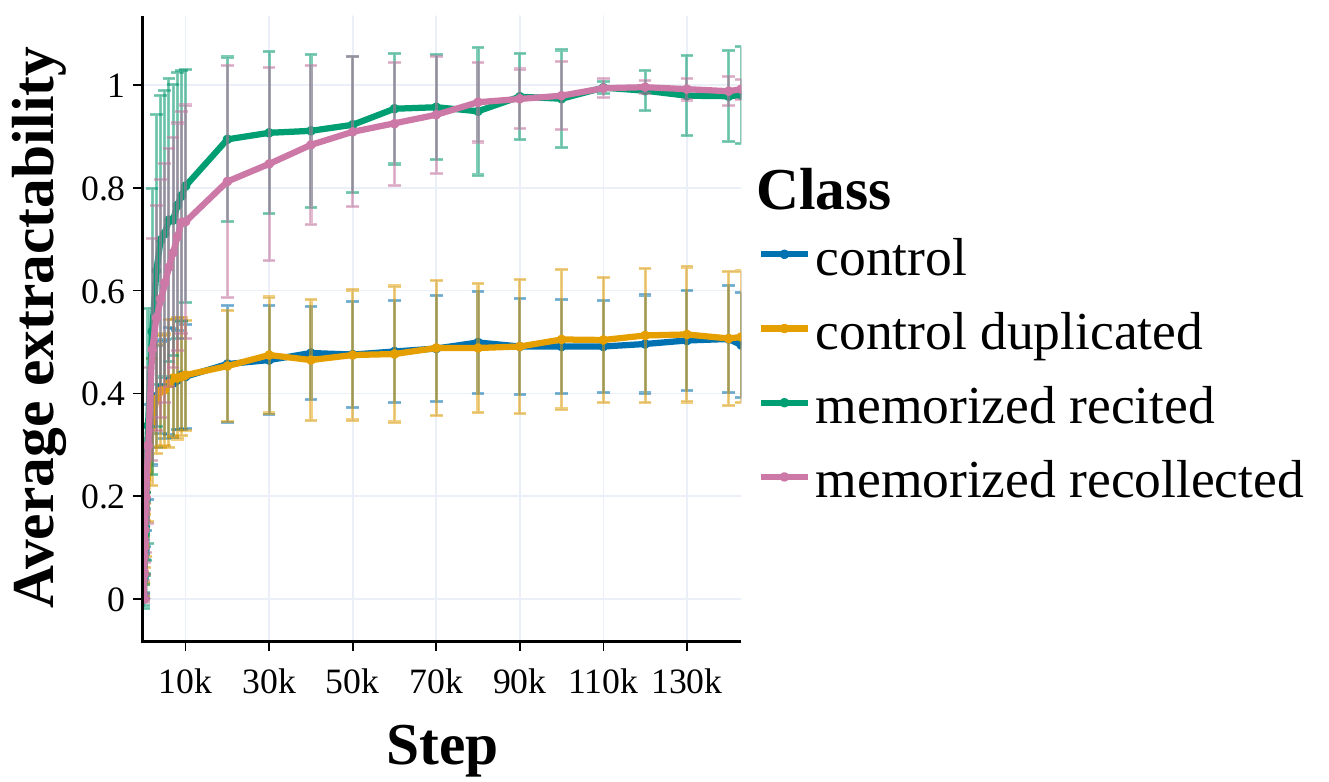}
    \end{subfigure}
    \caption{\textbf{Pythia training dynamics across scales.} We plot the loss over our sampled sequences and 32-extractability over time.}
    \label{fig:pythia-all-dynamics}
\end{figure}

\begin{figure}
    \centering
    \begin{subfigure}{0.3\linewidth}
        \includegraphics[width=1\linewidth]{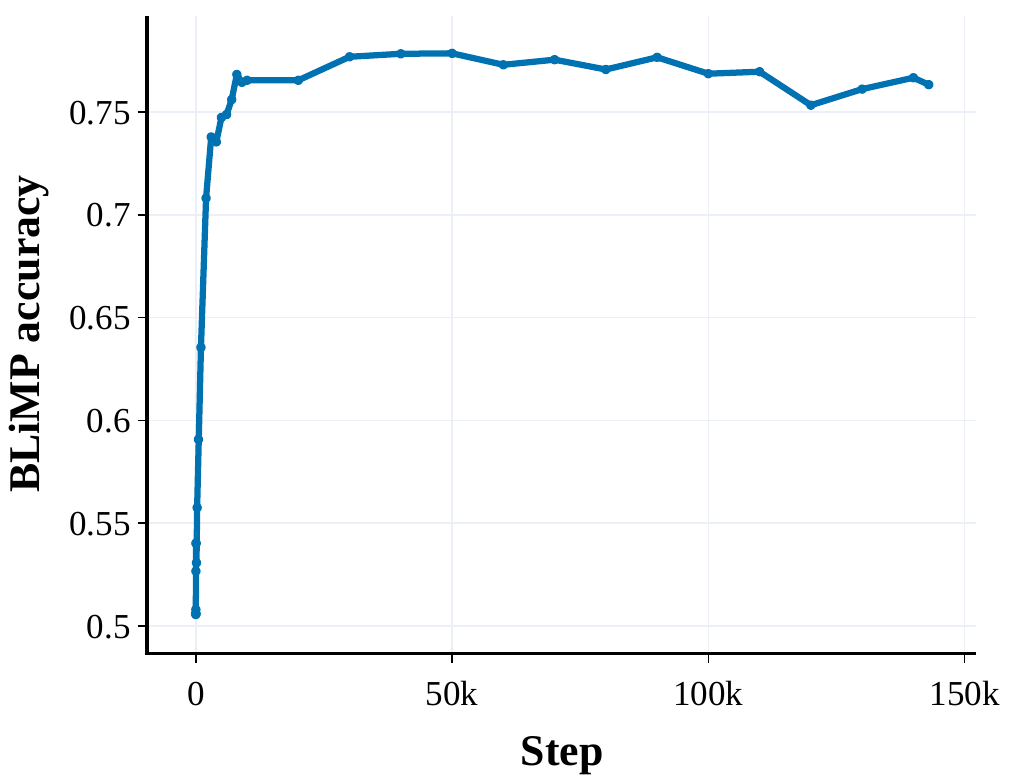}
        \caption{Pythia-160m}
    \end{subfigure}
    \begin{subfigure}{0.3\linewidth}
        \includegraphics[width=1\linewidth]{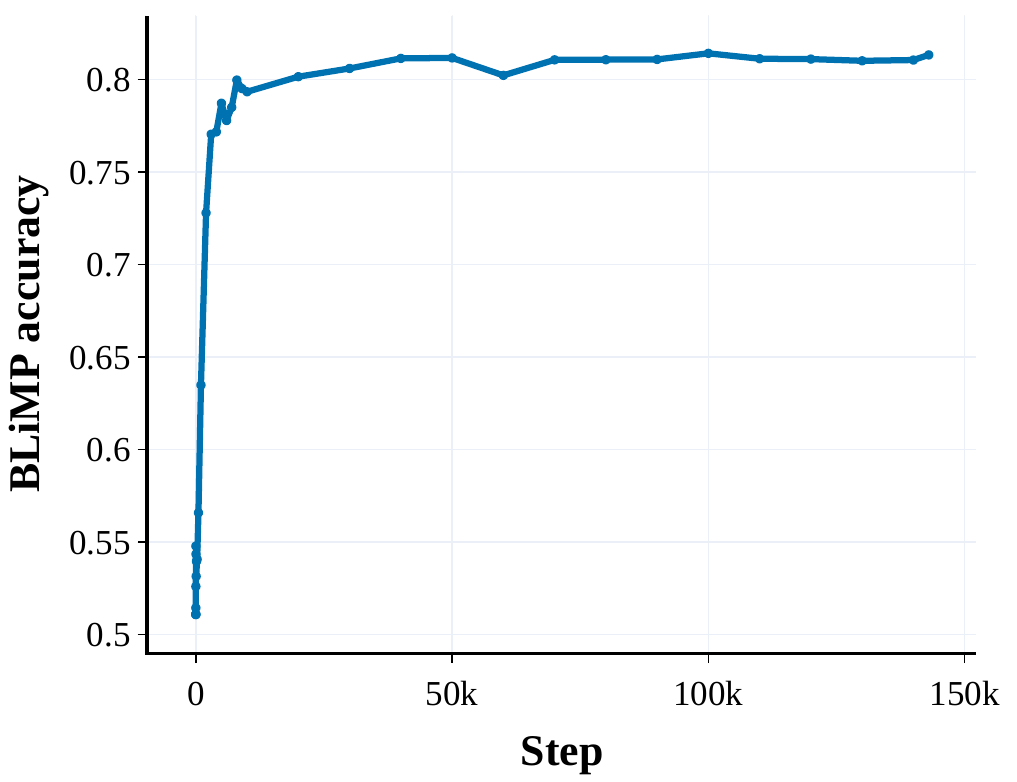}
        \caption{Pythia-410m}
    \end{subfigure}
    \begin{subfigure}{0.3\linewidth}
        \includegraphics[width=1\linewidth]{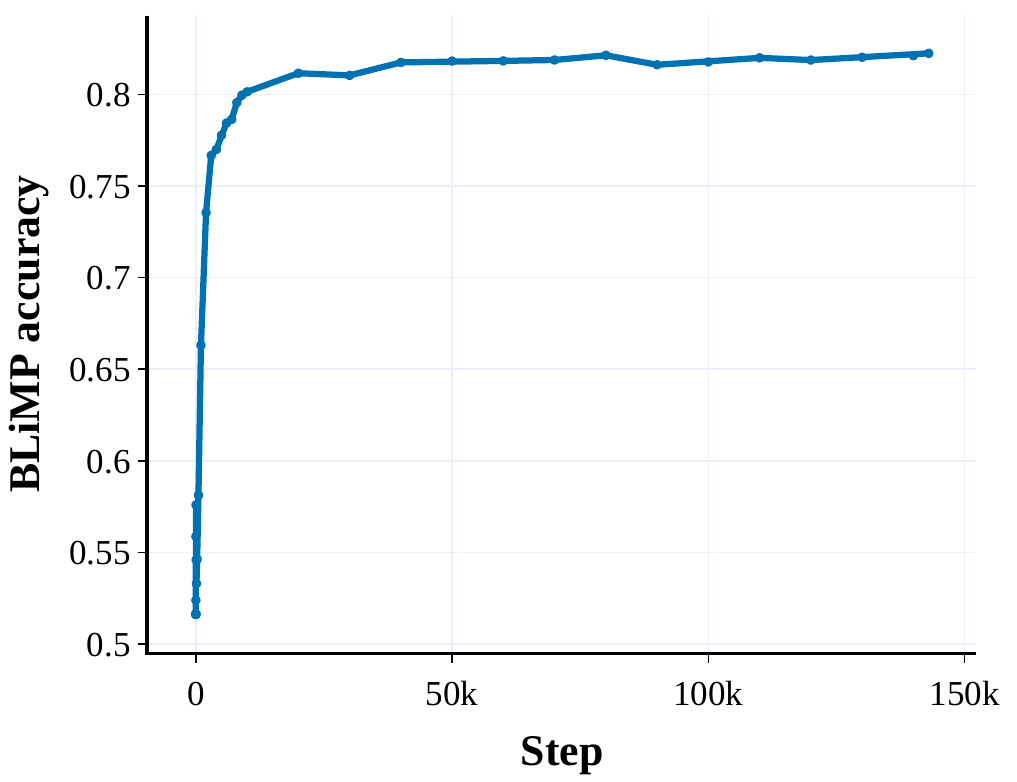}
        \caption{Pythia-1b}
    \end{subfigure}
    \begin{subfigure}{0.3\linewidth}
        \includegraphics[width=1\linewidth]{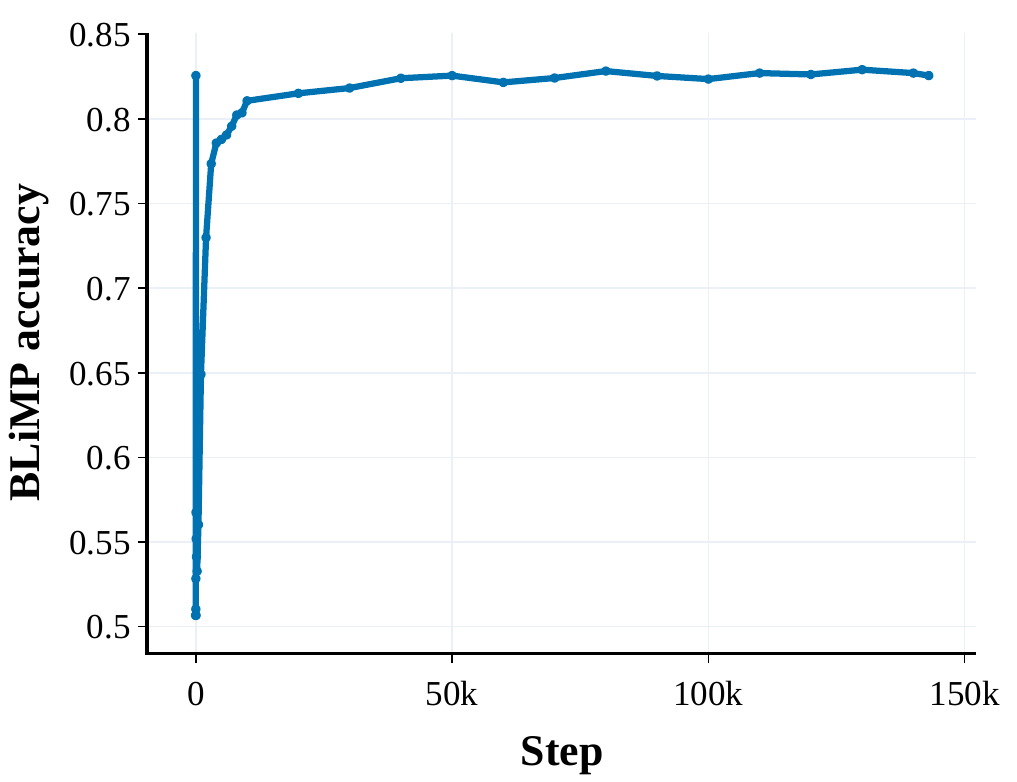}
        \caption{Pythia-1.4b}
    \end{subfigure}
    \begin{subfigure}{0.3\linewidth}
        \includegraphics[width=1\linewidth]{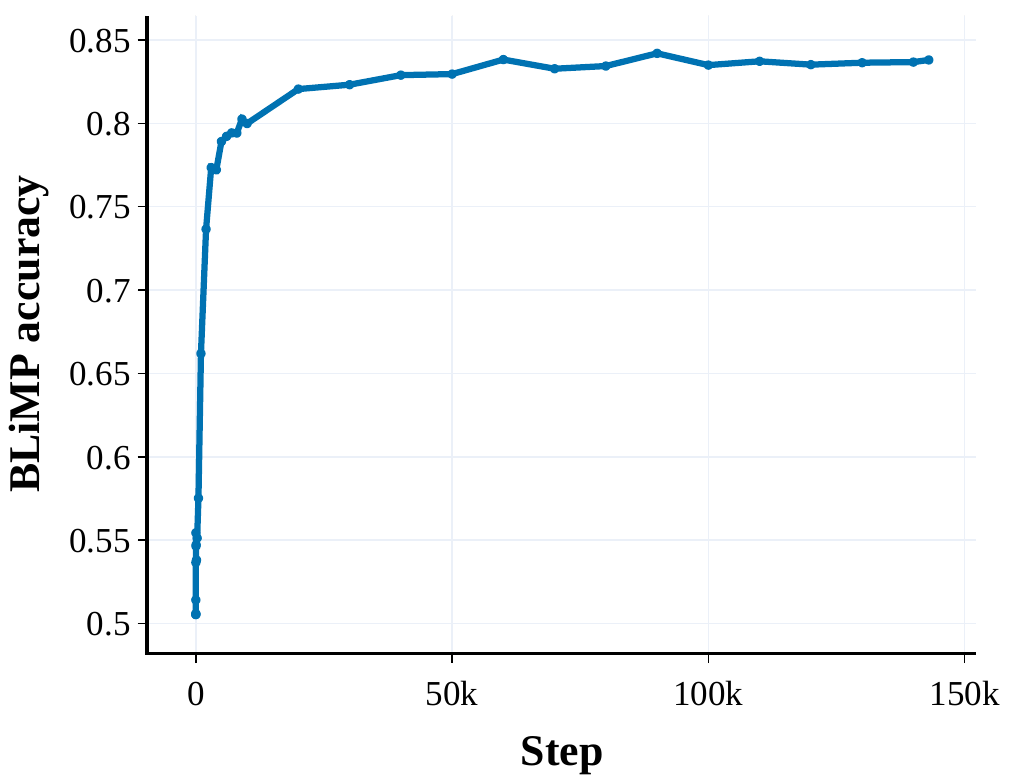}
        \caption{Pythia-2.8b}
    \end{subfigure}
    \begin{subfigure}{0.3\linewidth}
        \includegraphics[width=1\linewidth]{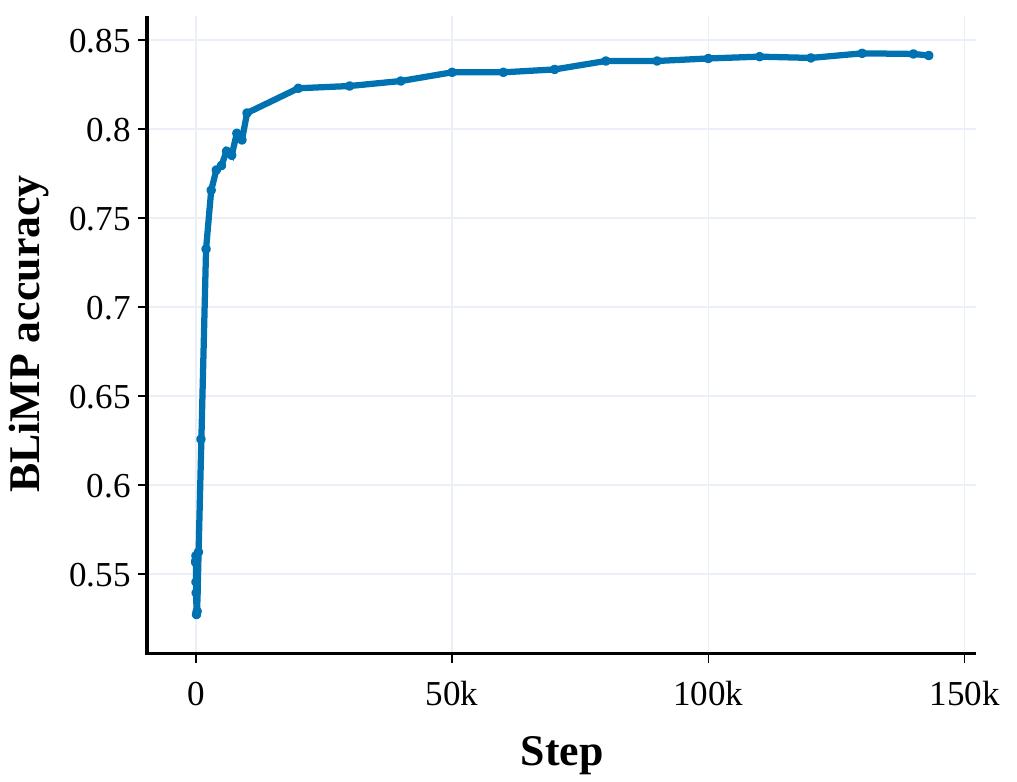}
        \caption{Pythia-6.9b}
    \end{subfigure}
    \caption{\textbf{Pythia BLiMP accuracy across scales.}}
    \label{fig:blimp}
\end{figure}

\begin{figure}
    \centering
\begin{tabular}{c}
    \textbf{Pythia-160m} \\
    \begin{subfigure}{0.215\linewidth}
        \includegraphics[trim=0 0 242 0,clip,width=1\linewidth]{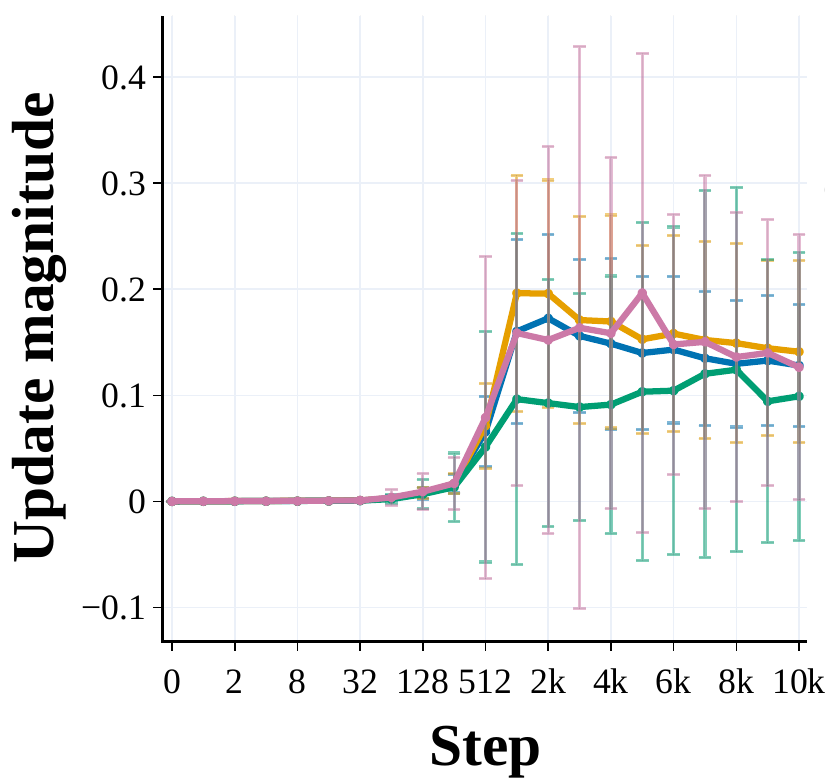}
    \end{subfigure}
    \begin{subfigure}{0.215\linewidth}
        \includegraphics[trim=0 0 250 0,clip,width=1\linewidth]{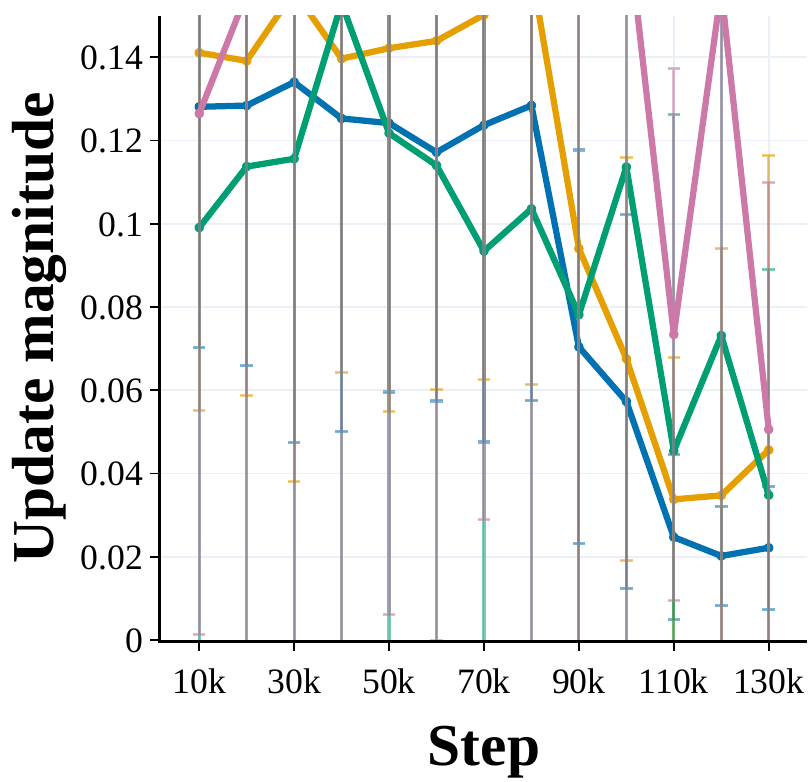}
    \end{subfigure}
    \begin{subfigure}{0.215\linewidth}
        \includegraphics[trim=0 0 242 0,clip,width=1\linewidth]{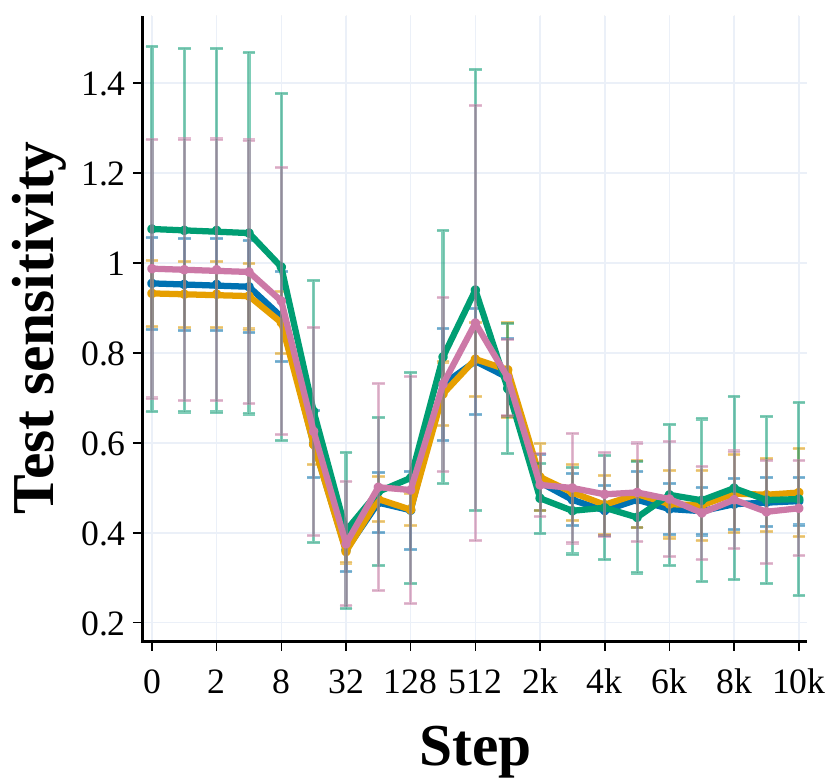}
    \end{subfigure}
    \begin{subfigure}{0.215\linewidth}
        \includegraphics[trim=0 0 250 0,clip,width=1\linewidth]{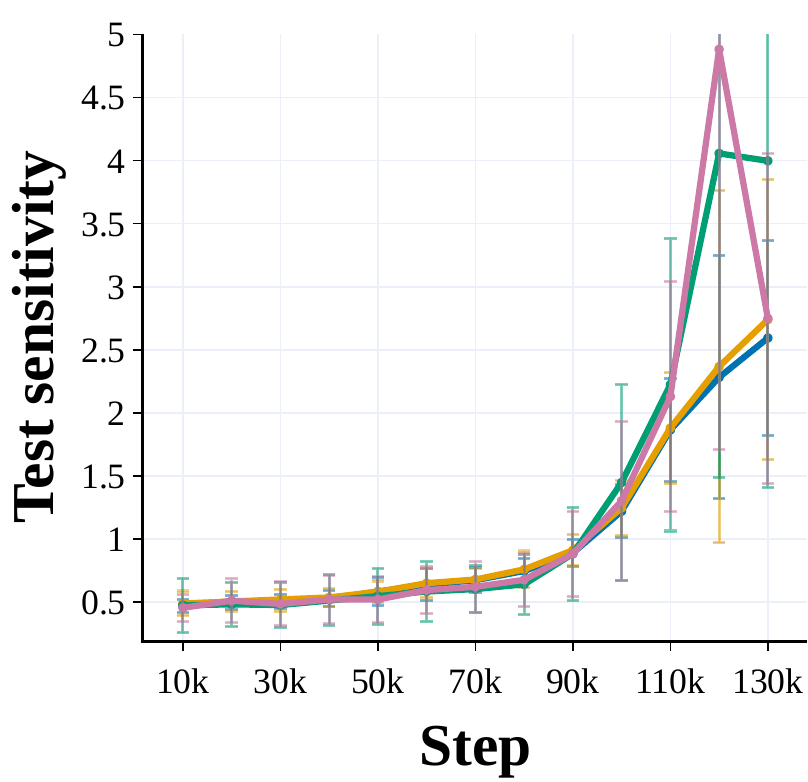}
    \end{subfigure}
    \begin{subfigure}{0.1\linewidth}
        \includegraphics[trim=425 0 0 0,clip,width=1\linewidth]{figures/decomp/phases/from_10000/feature_norms/160m/test_norm.pdf}
    \end{subfigure} \\
    \textbf{Pythia-410m} \\
    \begin{subfigure}{0.215\linewidth}
        \includegraphics[trim=0 0 242 0,clip,width=1\linewidth]{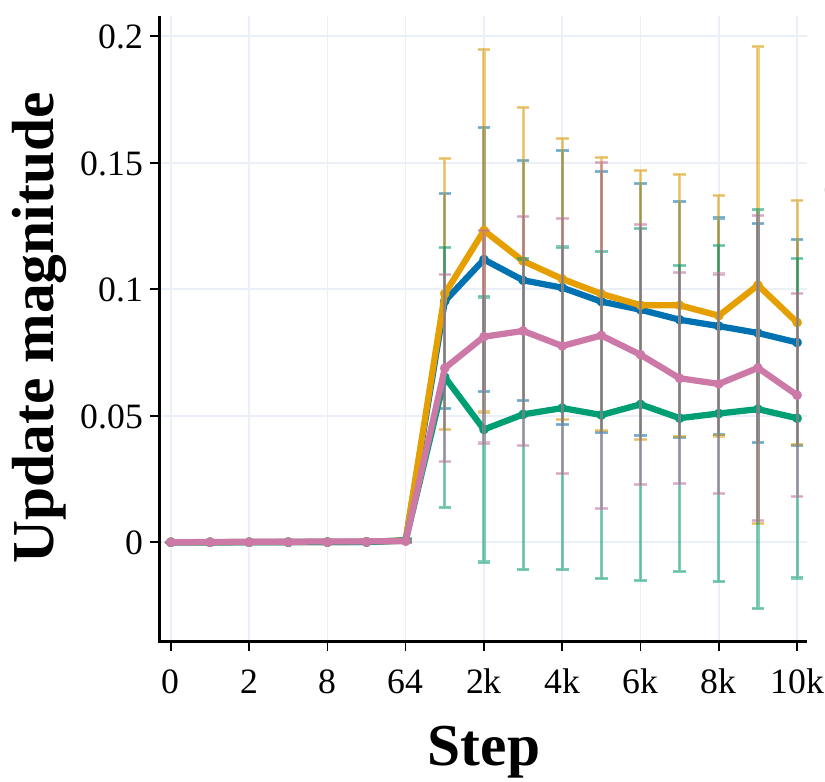}
    \end{subfigure}
    \begin{subfigure}{0.215\linewidth}
        \includegraphics[trim=0 0 250 0,clip,width=1\linewidth]{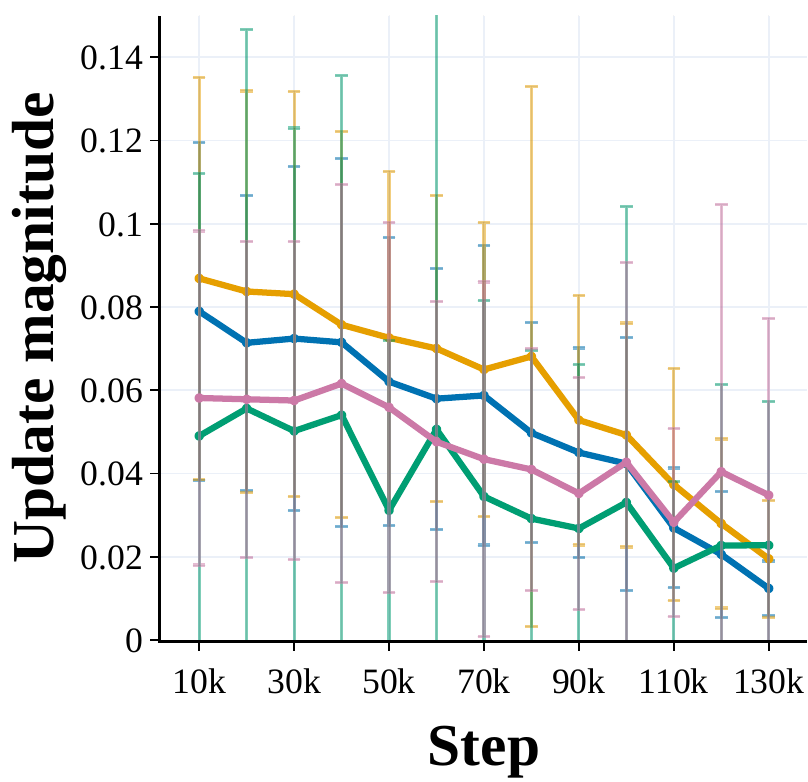}
    \end{subfigure}
    \begin{subfigure}{0.215\linewidth}
        \includegraphics[trim=0 0 242 0,clip,width=1\linewidth]{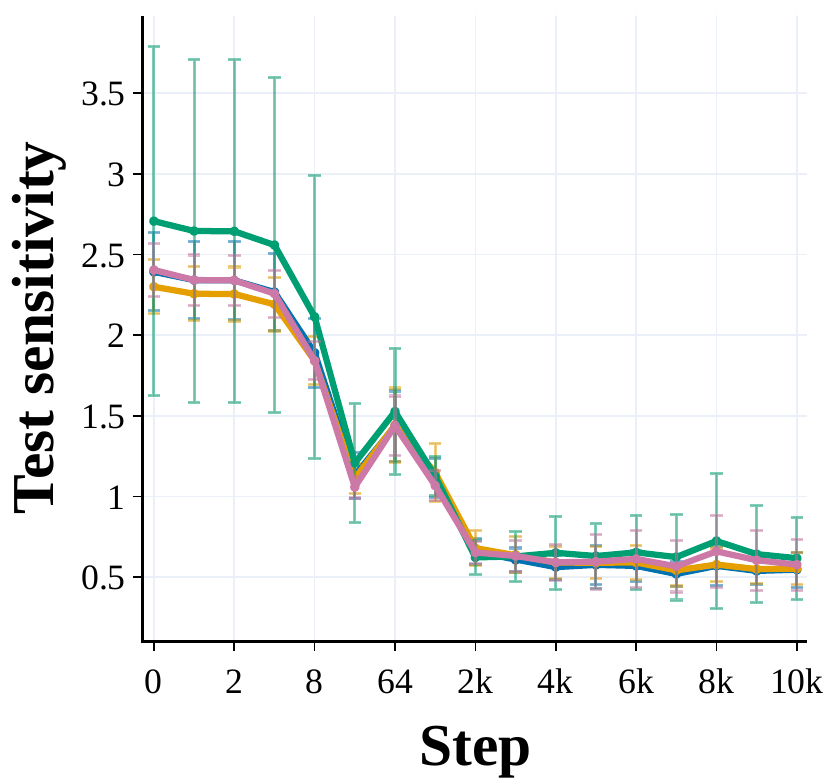}
    \end{subfigure}
    \begin{subfigure}{0.215\linewidth}
        \includegraphics[trim=0 0 250 0,clip,width=1\linewidth]{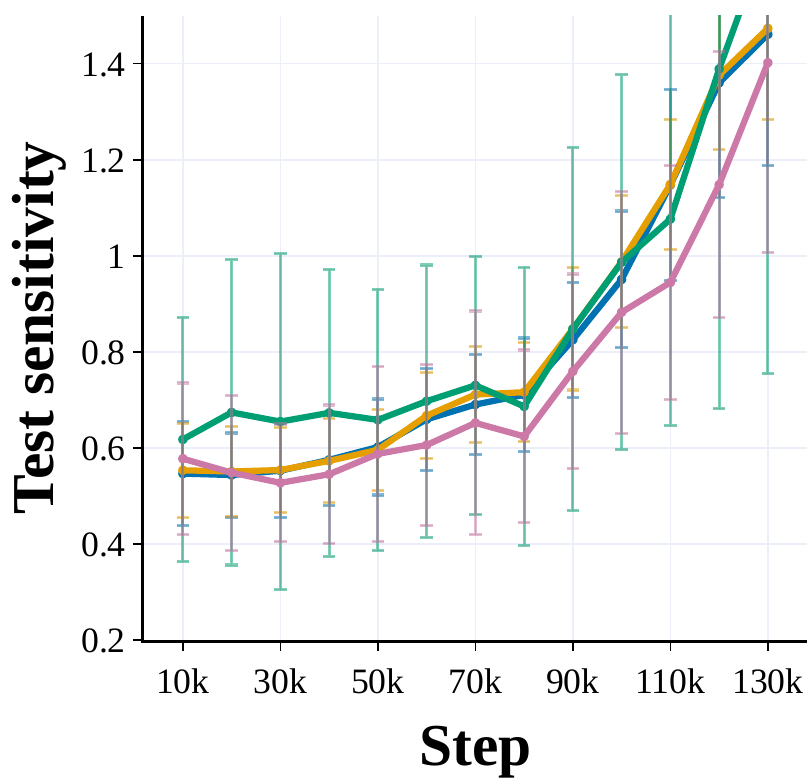}
    \end{subfigure}
    \begin{subfigure}{0.1\linewidth}
        \includegraphics[trim=425 0 0 0,clip,width=1\linewidth]{figures/decomp/phases/from_10000/feature_norms/410m/test_norm.pdf}
    \end{subfigure} \\
\end{tabular}
    \caption{\textbf{Pythia update magnitudes and test sensitivities across scales.} We plot each at early and later pretraining.}
    \label{fig:pythia-all-norms}
\end{figure}

\begin{figure}
    \centering
    \begin{tabular}{c}
    \textbf{Pythia-160m} \\
        \begin{subfigure}{0.215\linewidth}
            \includegraphics[trim=0 0 180 0,clip,width=1\linewidth]{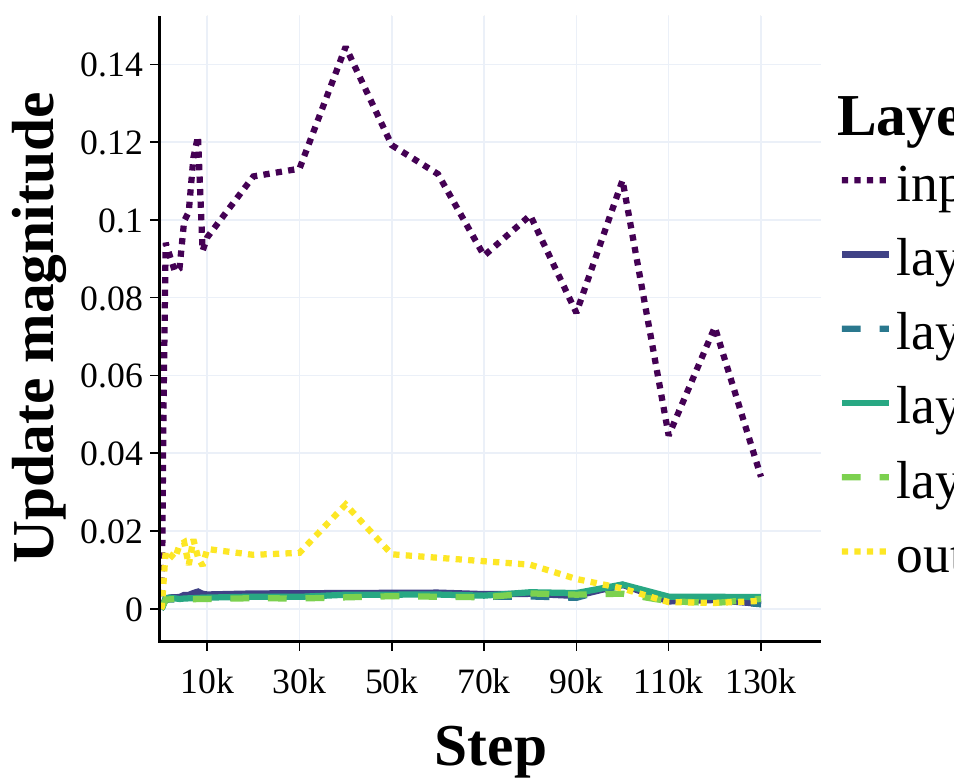}
        \end{subfigure}
        \begin{subfigure}{0.215\linewidth}
            \includegraphics[trim=0 0 180 0,clip,width=1\linewidth]{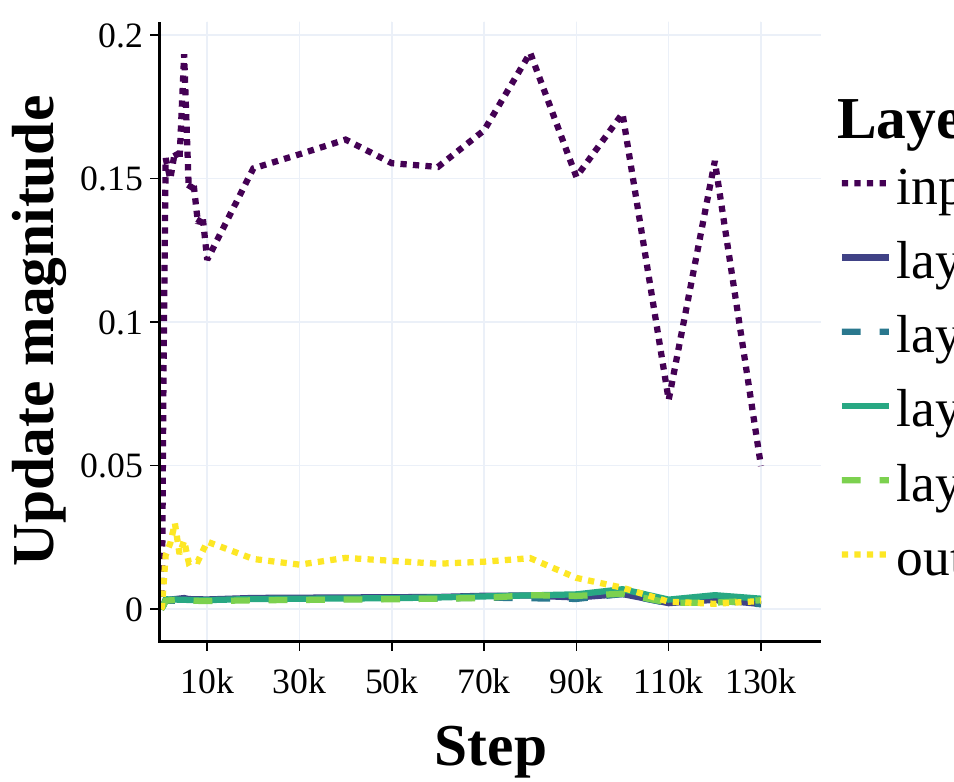}
        \end{subfigure}
        \begin{subfigure}{0.215\linewidth}
            \includegraphics[trim=0 0 180 0,clip,width=1\linewidth]{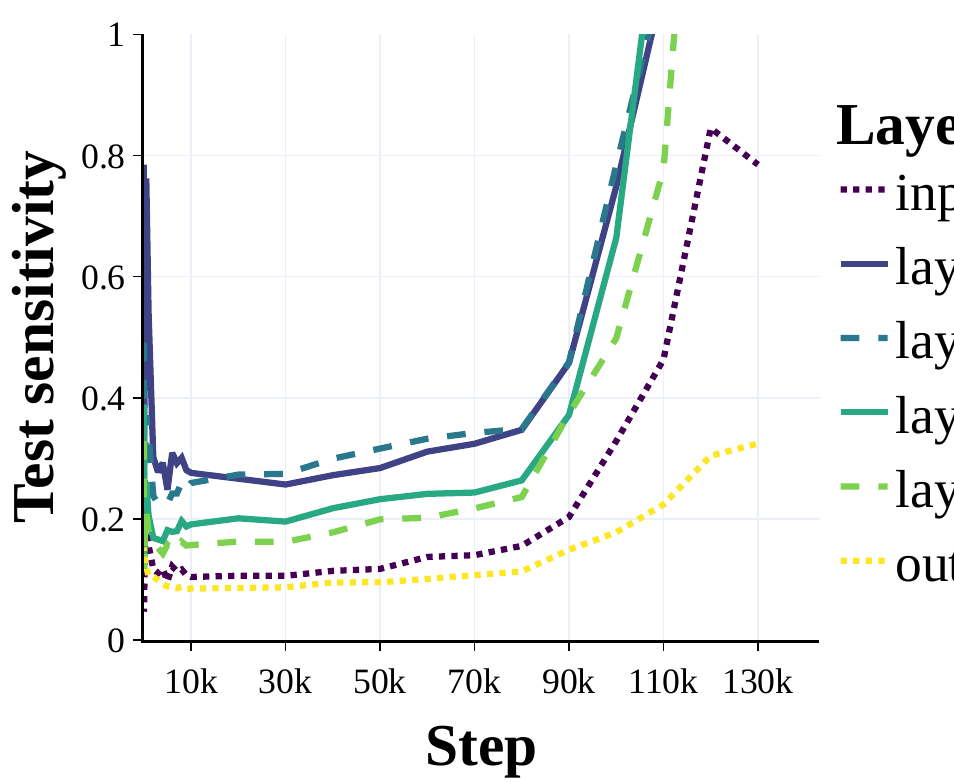}
        \end{subfigure}
        \begin{subfigure}{0.215\linewidth}
            \includegraphics[trim=0 0 180 0,clip,width=1\linewidth]{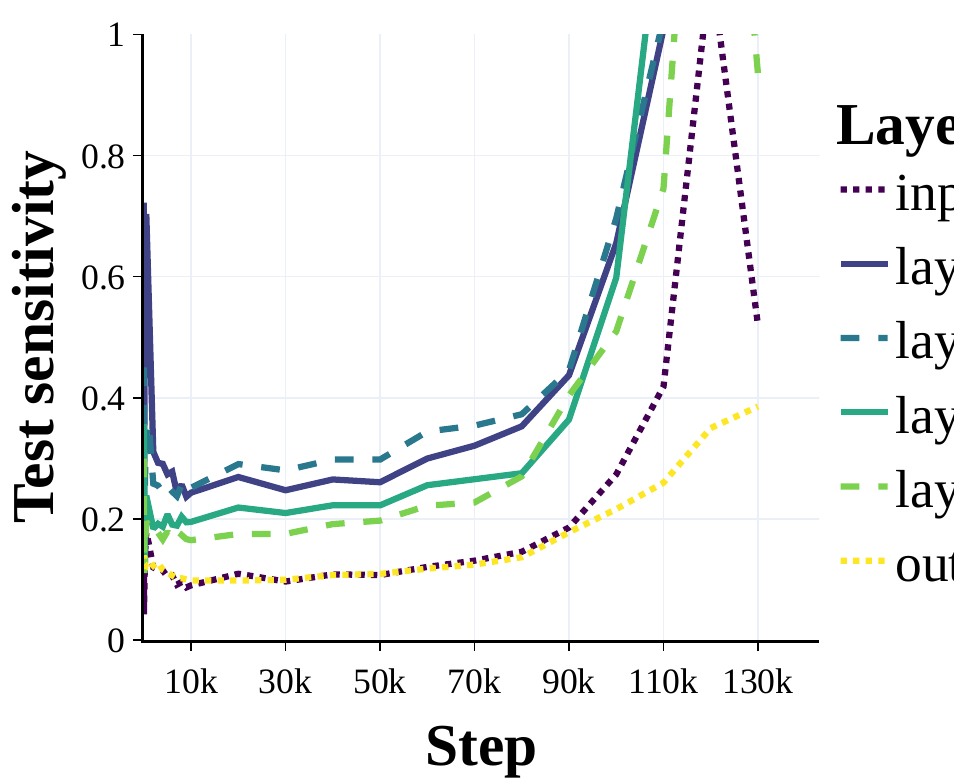}
        \end{subfigure}
        \begin{subfigure}{0.1\linewidth}
            \includegraphics[trim=425 0 0 0,clip,width=1\linewidth]{figures/decomp/four_classes/layer_regions_coarse/test_norm/test=memorized_recited/160m/train=single_sample_self_mean.pdf}
        \end{subfigure} \\
        \textbf{Pythia-1B} \\
\end{tabular}
    \caption{\textbf{Pythia update magnitudes and test sensitivity by model region across scales.}}
    \label{fig:pythia-all-norms-layer}
\end{figure}

Figure~\ref{fig:pythia-all-dynamics} reports loss and extractability for the memorized and control groups across Pythia models from $160$M to $6.9$B parameters, while Fig.~\ref{fig:blimp} reports BLiMP accuracy over training. Figures~\ref{fig:pythia-all-norms} and~\ref{fig:pythia-all-norms-layer} show update magnitudes and test sensitivities over training and by model region, respectively.

\begin{figure}
    \centering
    \begin{tabular}{c}
        \textbf{Pythia-160m} \\
        \includegraphics[trim=0 0 250 0,clip,width=0.215\linewidth]{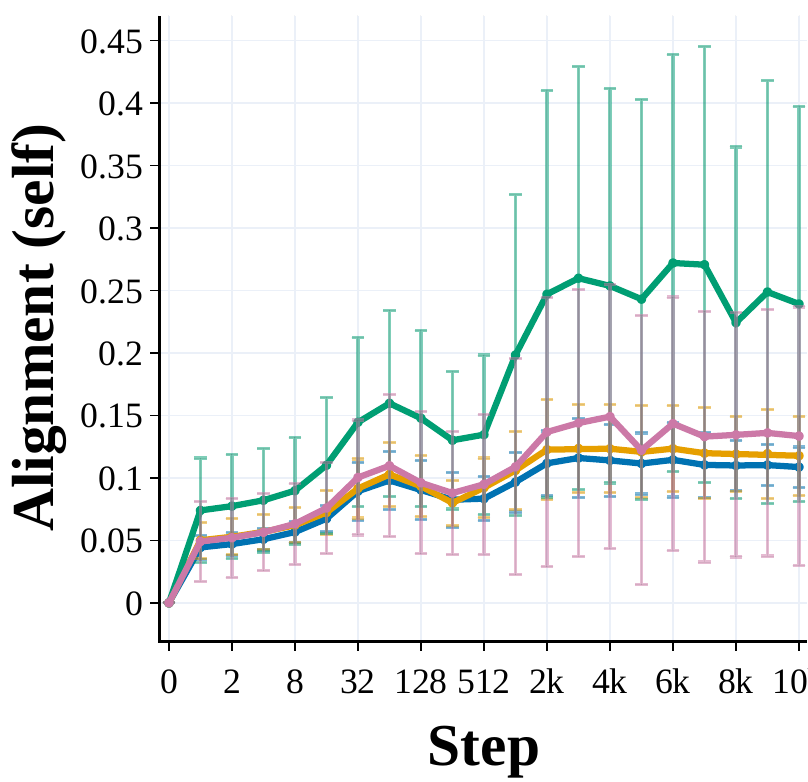}
        \includegraphics[trim=0 0 250 0,clip,width=0.215\linewidth]{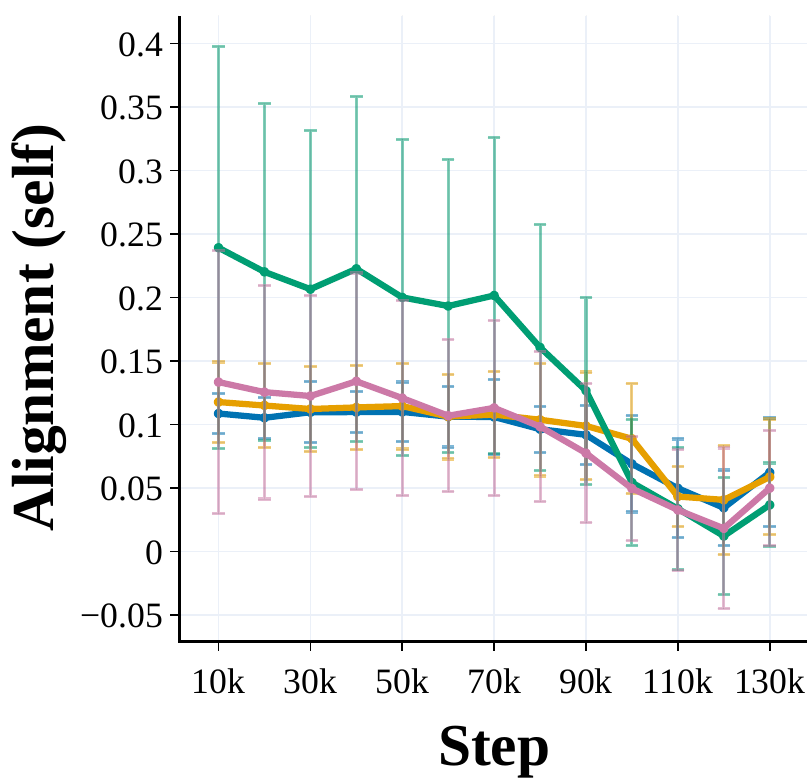}
        \includegraphics[trim=0 0 250 0,clip,width=0.215\linewidth]{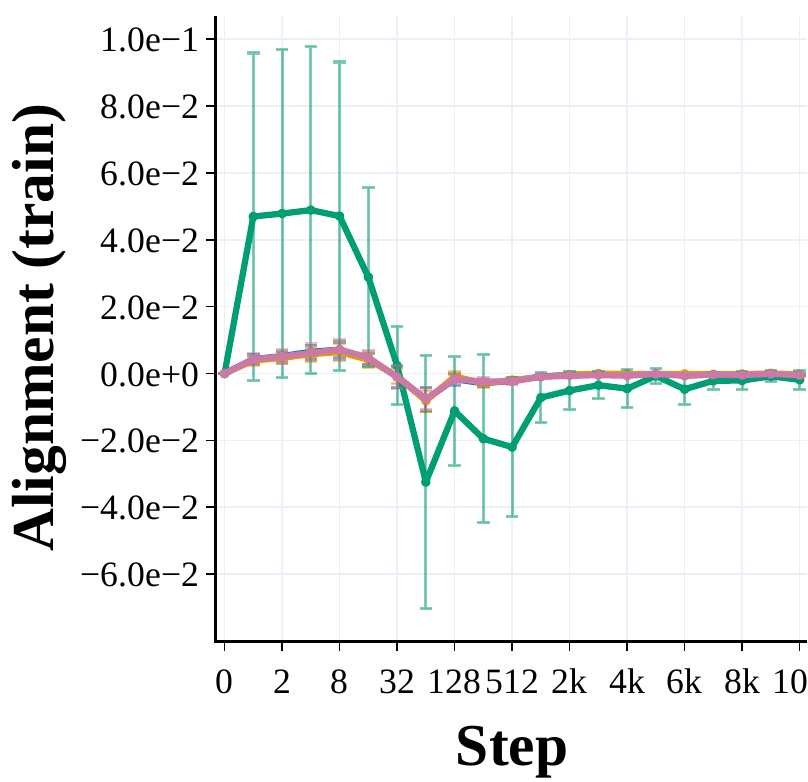}
        \includegraphics[trim=0 0 250 0,clip,width=0.215\linewidth]{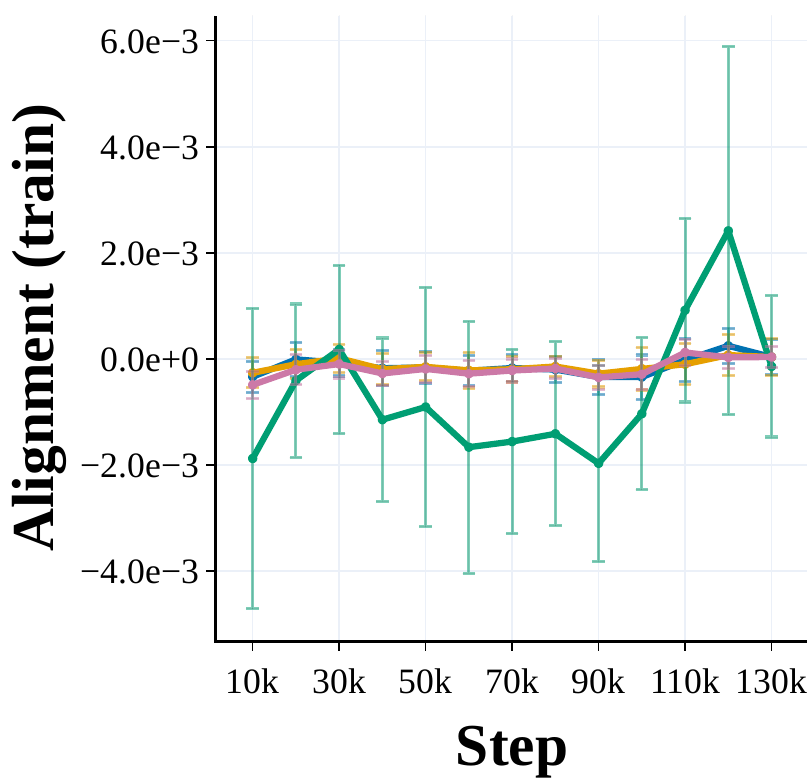}
        \includegraphics[trim=390 0 0 0,clip,width=0.09\linewidth]{figures/decomp/phases/from_10000/cosines/160m/train_original.pdf} \\
        
        \textbf{Pythia-410m} \\
        \includegraphics[trim=0 0 250 0,clip,width=0.215\linewidth]{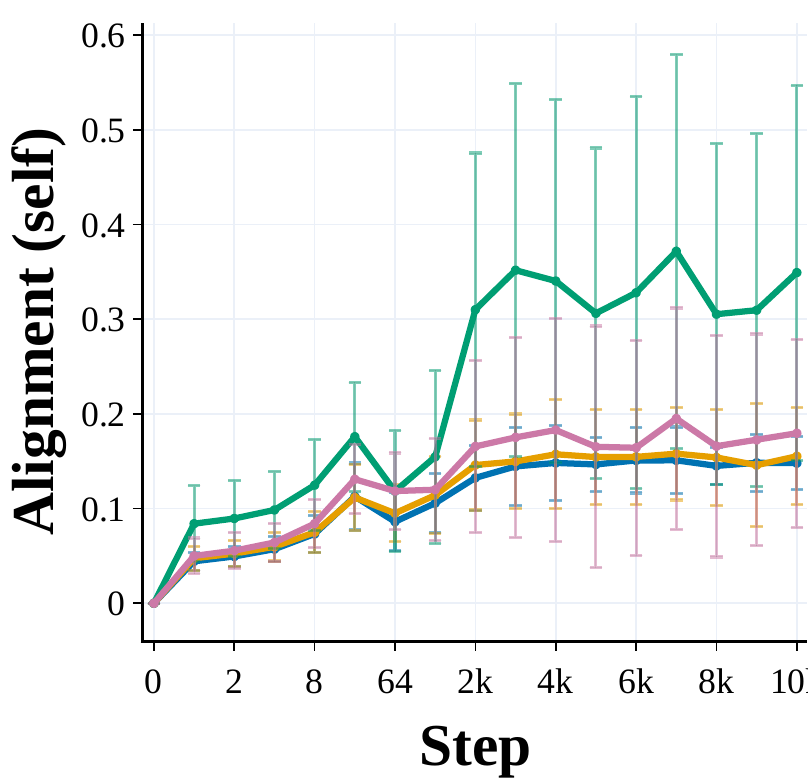}
        \includegraphics[trim=0 0 250 0,clip,width=0.215\linewidth]{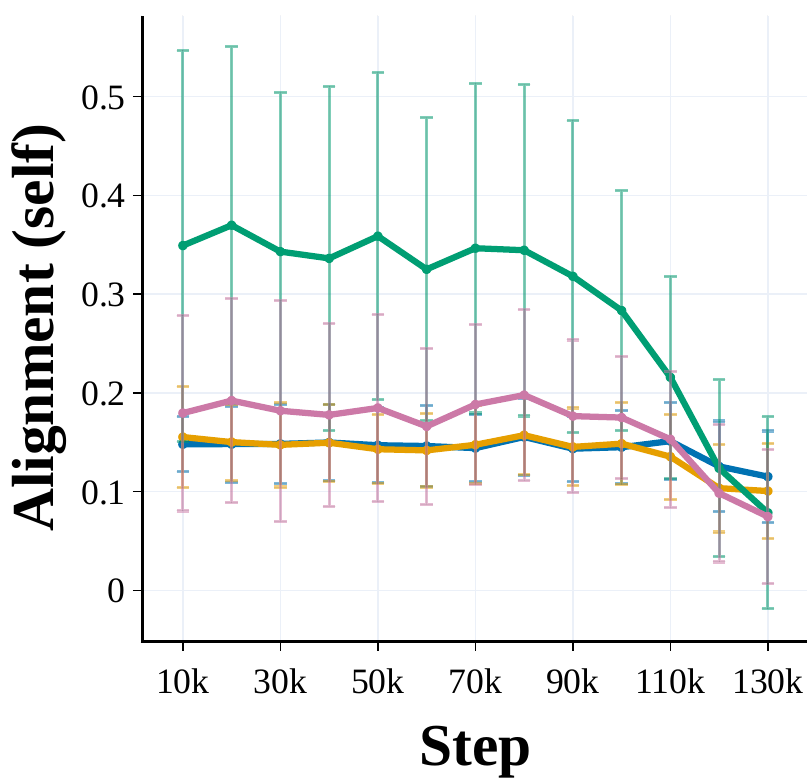}
        \includegraphics[trim=0 0 250 0,clip,width=0.215\linewidth]{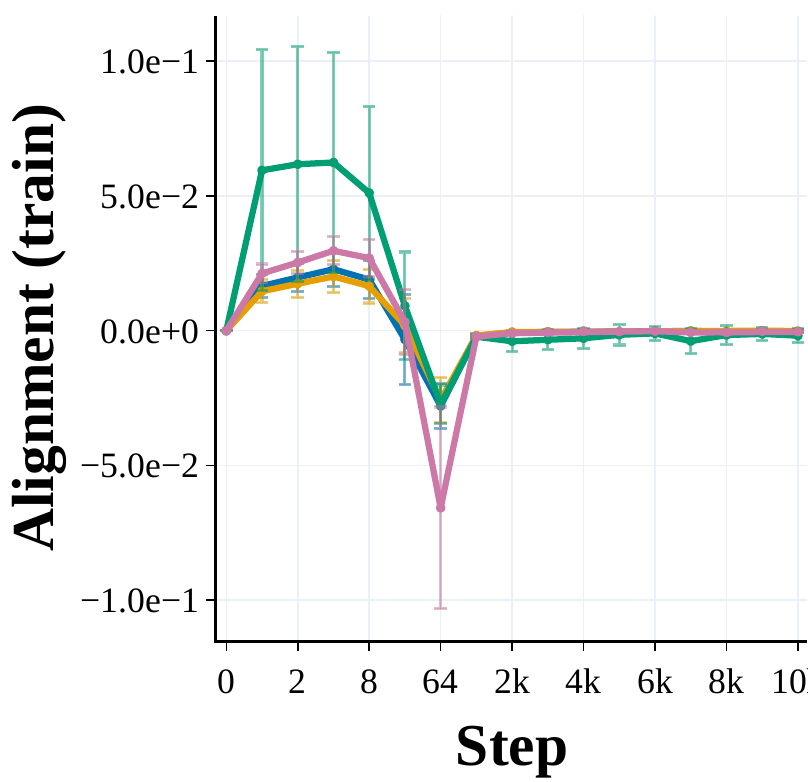}
        \includegraphics[trim=0 0 250 0,clip,width=0.215\linewidth]{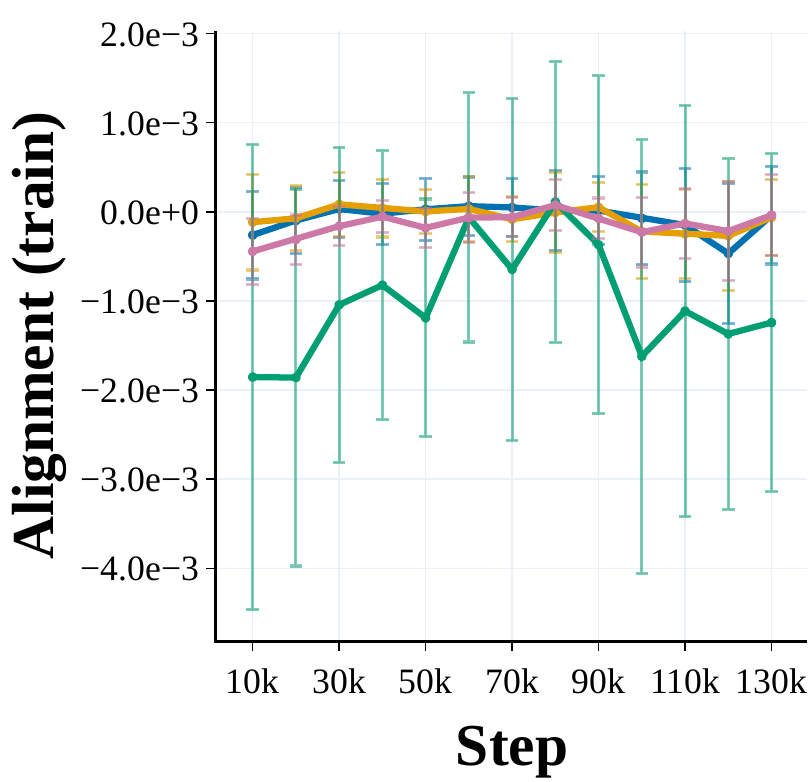}
        \includegraphics[trim=390 0 0 0,clip,width=0.09\linewidth]{figures/decomp/phases/from_10000/cosines/410m/train_original.pdf} \\

        \textbf{Pythia-1b (model shown in paper)} \\
\end{tabular}
    
    \caption{\textbf{Pythia self alignment across scales.} We plot early and later training stages.}
    \label{fig:pythia-alignment-combined-early-and-late}
\end{figure}

\begin{figure}
    \centering
    \begin{tabular}{c}
        \textbf{Pythia-160m} \\
        \includegraphics[trim=0 0 250 0,clip,width=0.215\linewidth]{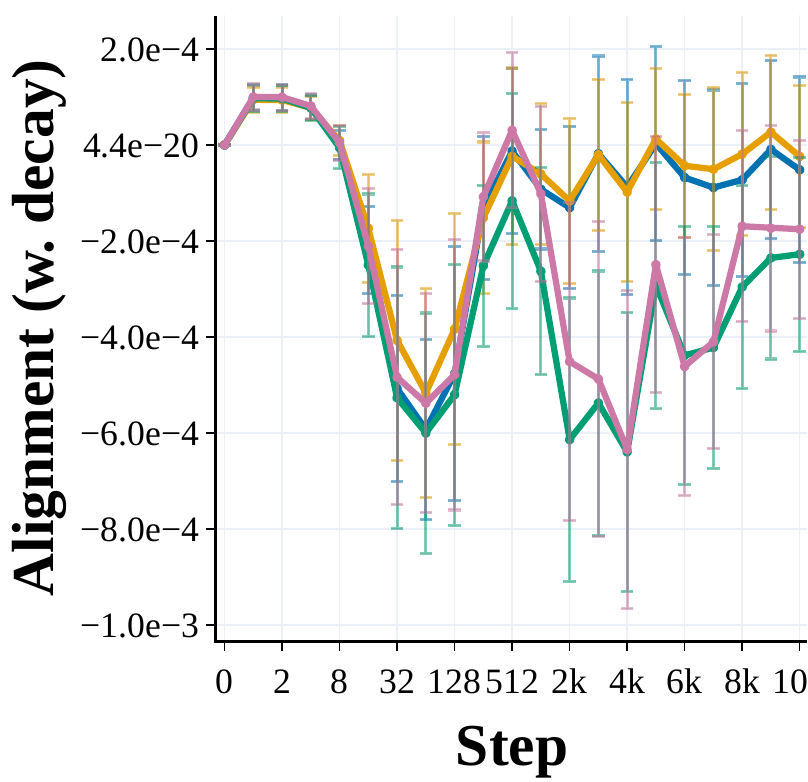}
        \includegraphics[trim=0 0 250 0,clip,width=0.215\linewidth]{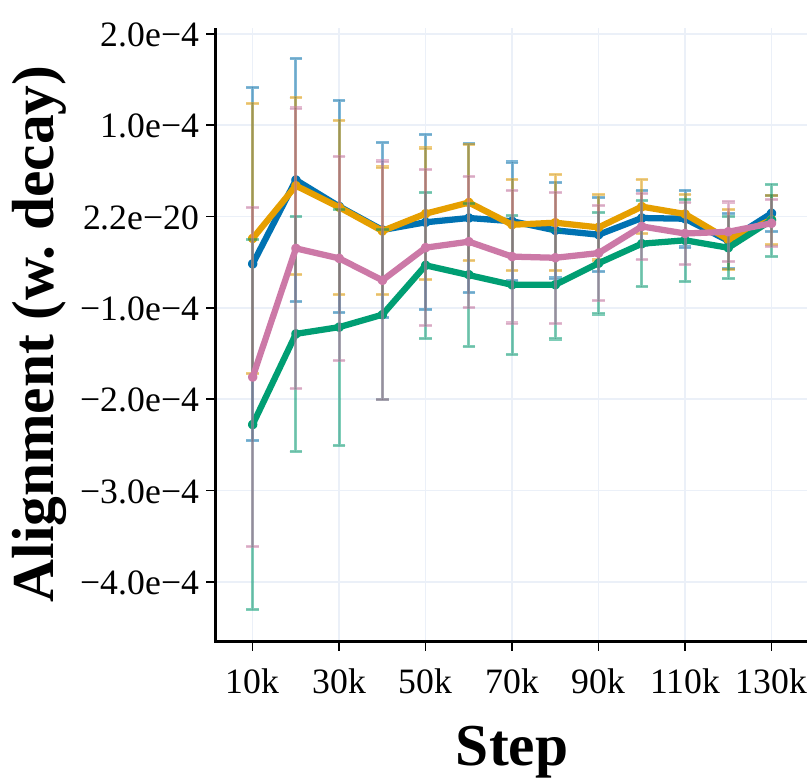}
        \includegraphics[trim=390 0 0 0,clip,width=0.09\linewidth]{figures/decomp/phases/from_10000/cosines/160m/regularization.pdf} \\
        
        \textbf{Pythia-410m} \\
        \includegraphics[trim=0 0 250 0,clip,width=0.215\linewidth]{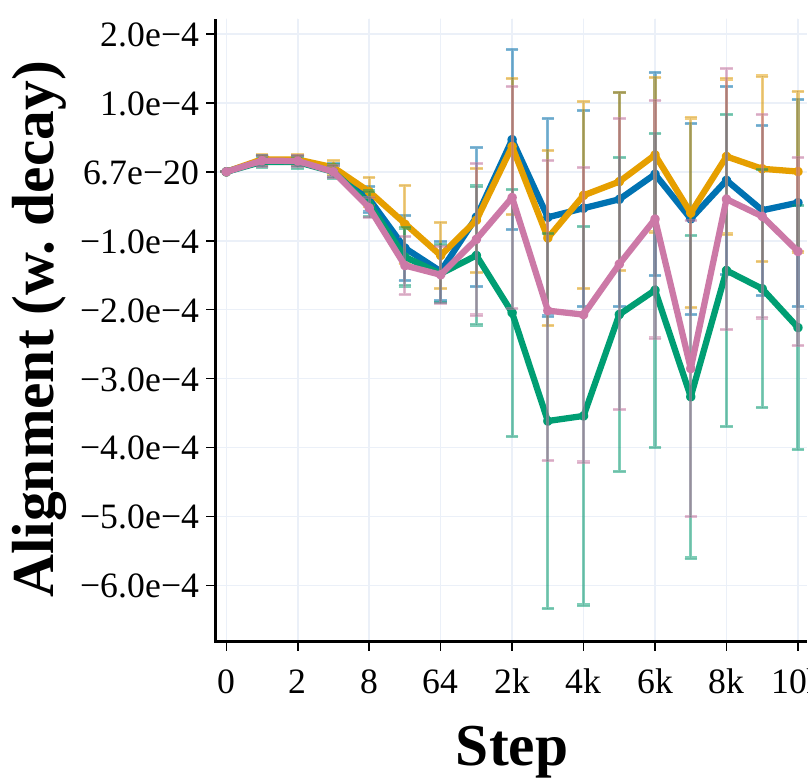}
        \includegraphics[trim=0 0 250 0,clip,width=0.215\linewidth]{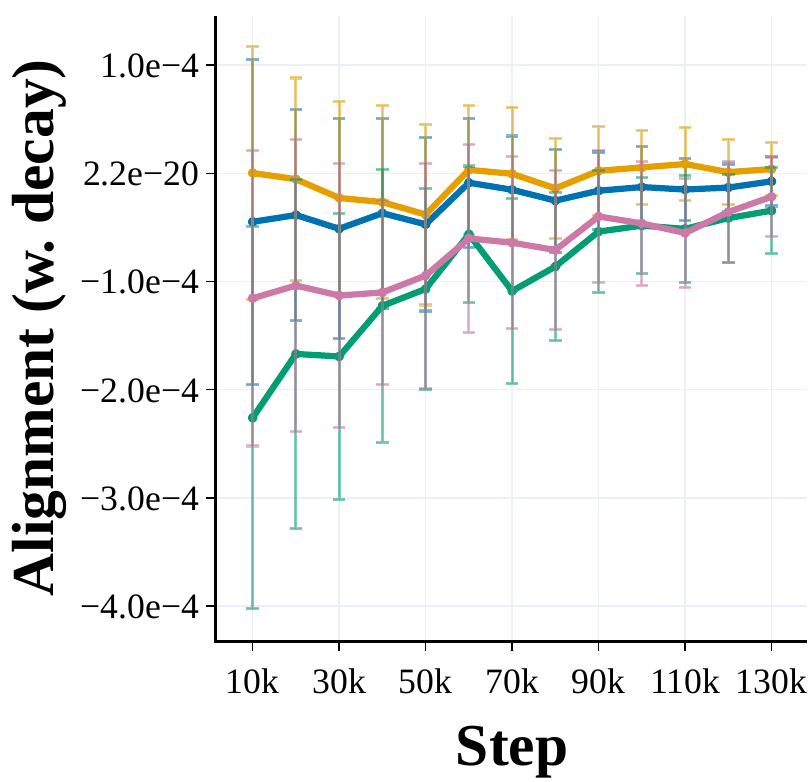}
        \includegraphics[trim=390 0 0 0,clip,width=0.09\linewidth]{figures/decomp/phases/from_10000/cosines/410m/regularization.pdf} \\

        \textbf{Pythia-1b (model shown in paper)} \\
        \includegraphics[trim=0 0 250 0,clip,width=0.215\linewidth]{figures/decomp/phases/through_10000/cosines/1b/regularization.pdf}
        \includegraphics[trim=0 0 250 0,clip,width=0.215\linewidth]{figures/decomp/phases/from_10000/cosines/1b/regularization.pdf}
        \includegraphics[trim=390 0 0 0,clip,width=0.09\linewidth]{figures/decomp/phases/from_10000/cosines/1b/regularization.pdf} \\

        \textbf{Pythia-1.4b} \\
        \includegraphics[trim=0 0 250 0,clip,width=0.215\linewidth]{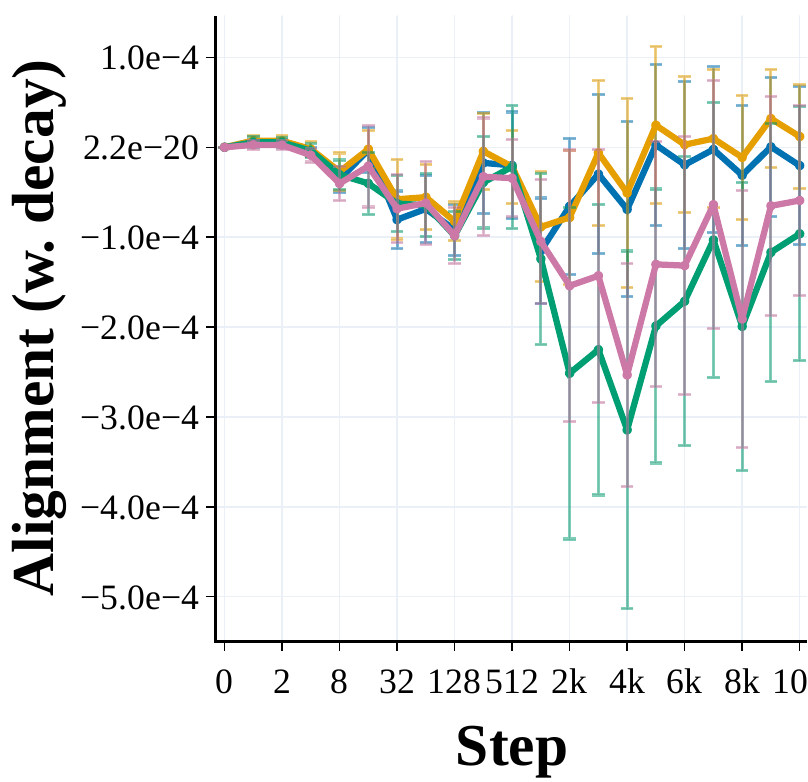}
        \includegraphics[trim=0 0 250 0,clip,width=0.215\linewidth]{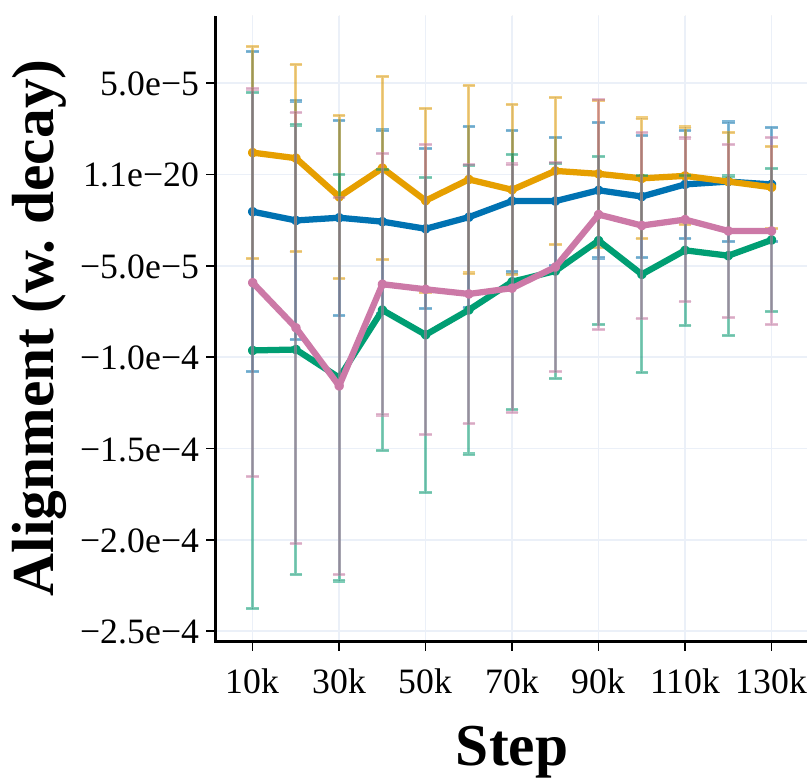}
        \includegraphics[trim=390 0 0 0,clip,width=0.09\linewidth]{figures/decomp/phases/from_10000/cosines/1.4b/regularization.pdf} \\
\end{tabular}
    
    \caption{\textbf{Pythia weight decay alignment across scales.} We plot the alignment of the instances we decompose with the parameter update introduced by weight decay, early and late in training.}
    \label{fig:pythia-alignment-regularization-combined}
\end{figure}

\begin{figure}
    \centering
    \begin{tabular}{c}
    \textbf{Pythia-160m} \\
    \begin{subfigure}{0.215\linewidth}
        \includegraphics[trim=0 0 180 0,clip,width=1\linewidth]{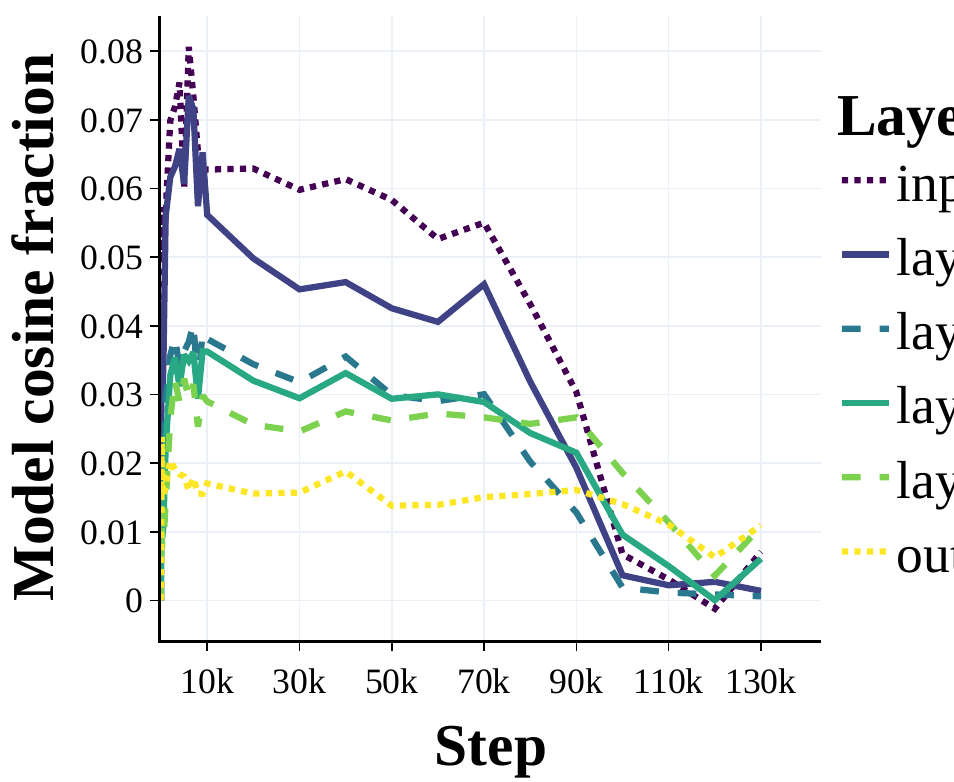}
    \end{subfigure}
    \begin{subfigure}{0.215\linewidth}
        \includegraphics[trim=0 0 180 0,clip,width=1\linewidth]{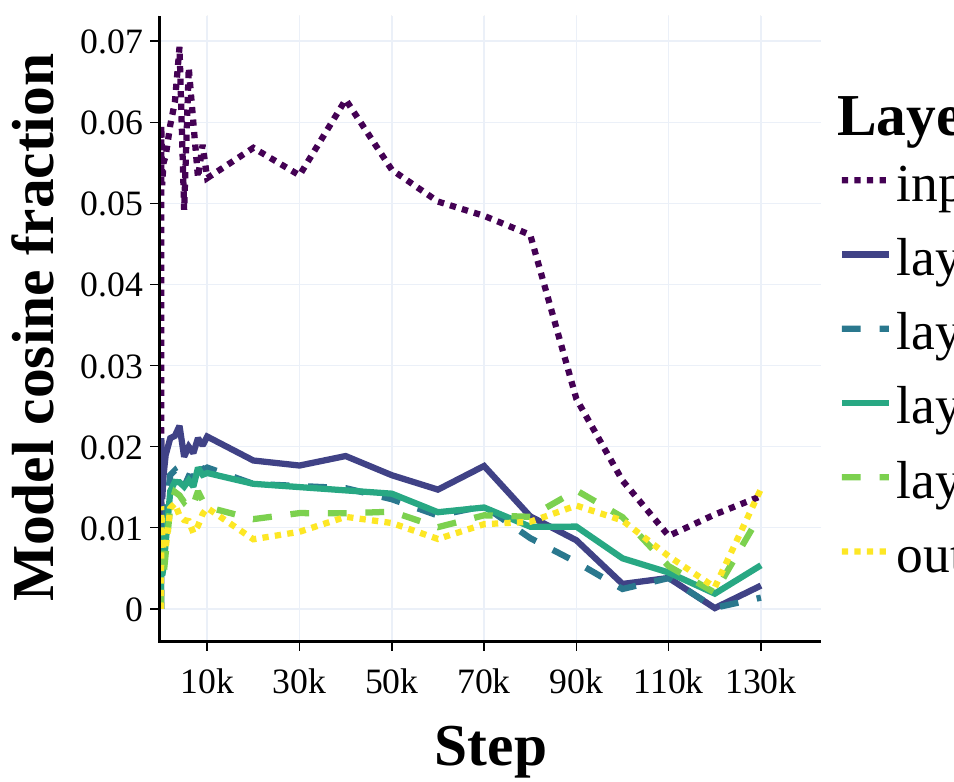}
    \end{subfigure}
    \begin{subfigure}{0.215\linewidth}
        \includegraphics[trim=0 0 180 0,clip,width=1\linewidth]{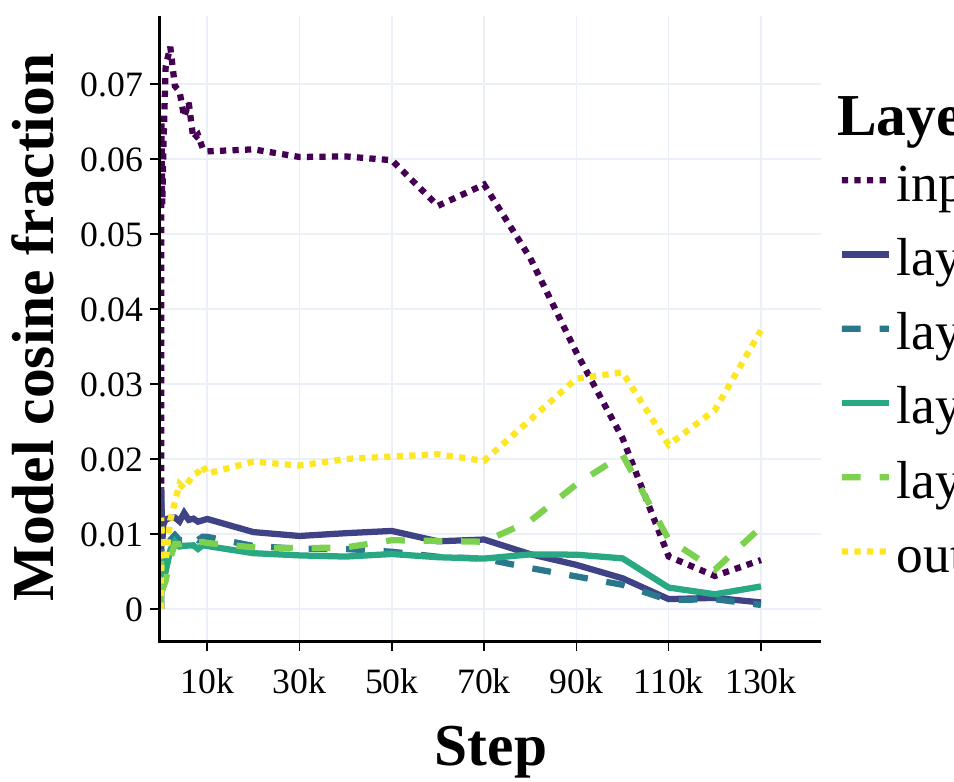}
    \end{subfigure}
    \begin{subfigure}{0.215\linewidth}
        \includegraphics[trim=0 0 180 0,clip,width=1\linewidth]{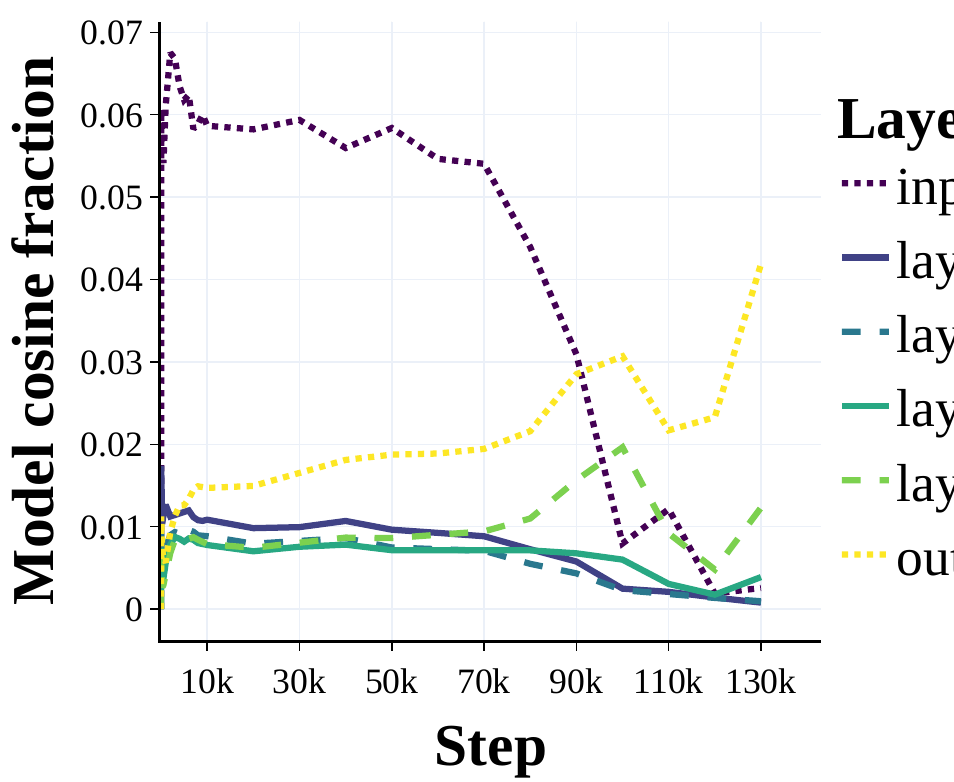}
    \end{subfigure}
    \begin{subfigure}{0.1\linewidth}
        \includegraphics[trim=425 0 0 0,clip,width=1\linewidth]{figures/decomp/four_classes/layer_regions_coarse/model_cosine_component/test=control/160m/train=single_sample_self_mean.pdf}
    \end{subfigure} \\
    \textbf{Pythia-410m} \\
\end{tabular}
    \caption{\textbf{Pythia self alignment by model region across scales.}}
    \label{fig:pythia-all-self-align-layer}
\end{figure}

\begin{figure}
    \centering
    \begin{tabular}{c}
    \textbf{Pythia-160m} \\
    \begin{subfigure}{0.215\linewidth}
        \includegraphics[trim=0 0 220 0,clip,width=1\linewidth]{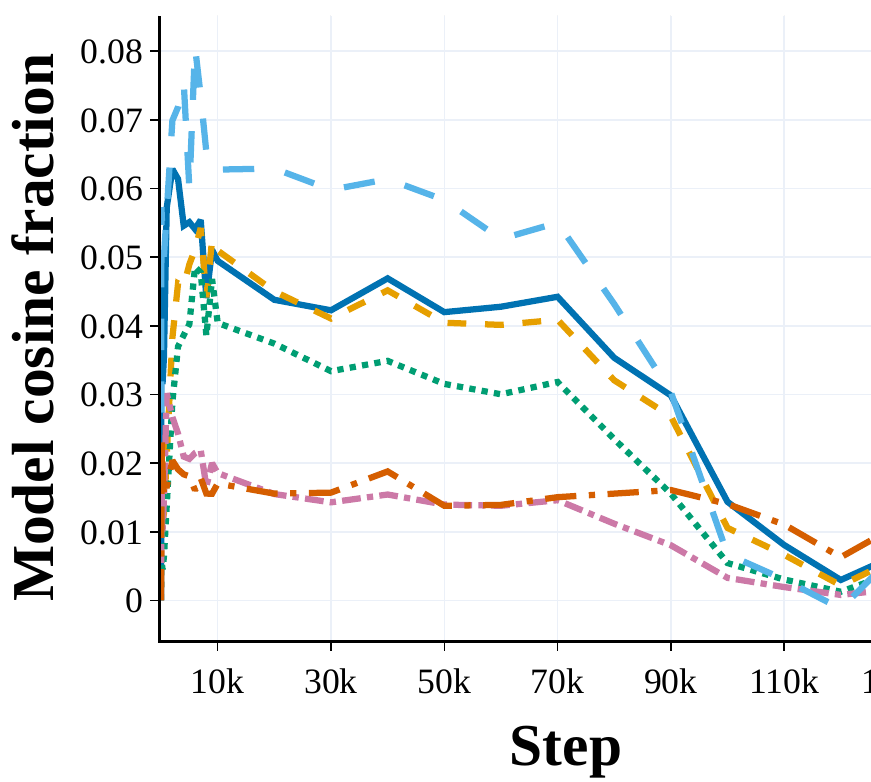}
    \end{subfigure}
    \begin{subfigure}{0.215\linewidth}
        \includegraphics[trim=0 0 220 0,clip,width=1\linewidth]{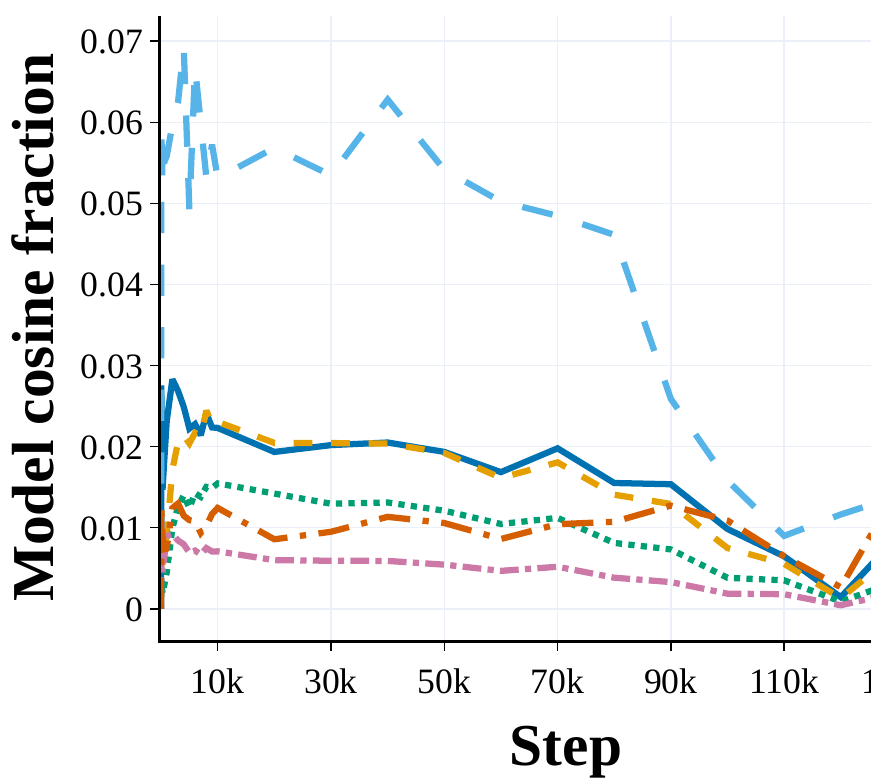}
    \end{subfigure}
    \begin{subfigure}{0.215\linewidth}
        \includegraphics[trim=0 0 220 0,clip,width=1\linewidth]{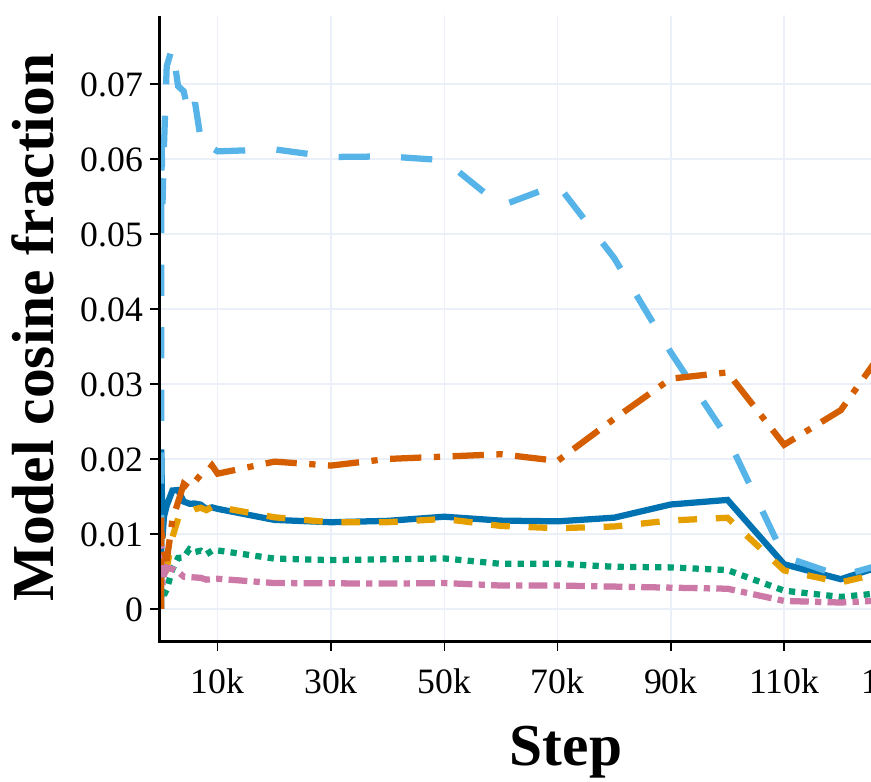}
    \end{subfigure}
    \begin{subfigure}{0.215\linewidth}
        \includegraphics[trim=0 0 220 0,clip,width=1\linewidth]{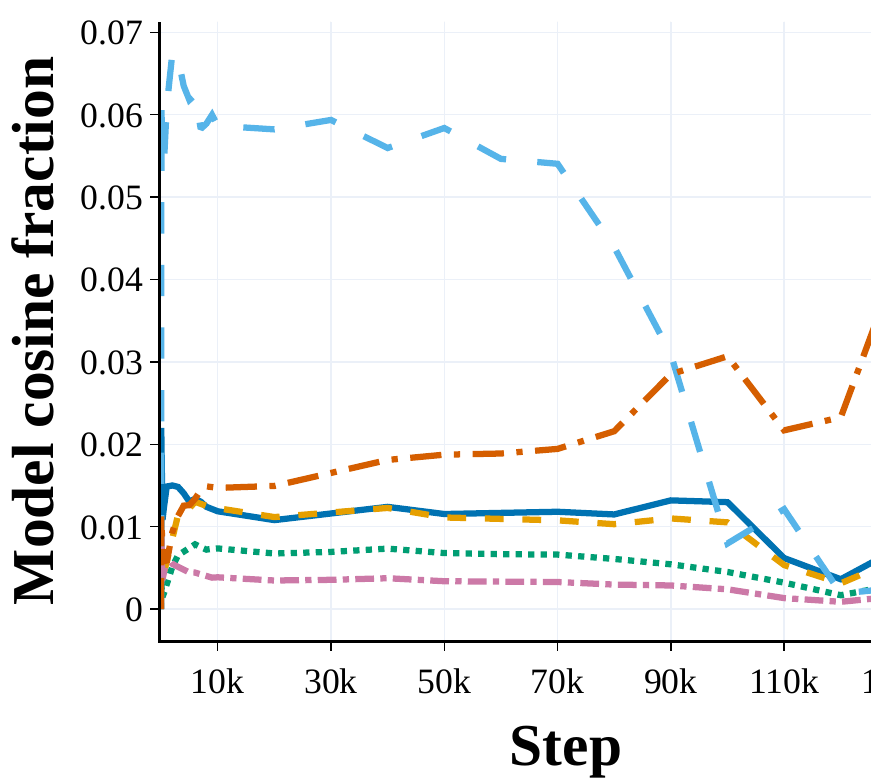}
    \end{subfigure}
    \begin{subfigure}{0.1\linewidth}
        \includegraphics[trim=380 0 0 0,clip,width=1\linewidth]{figures/decomp/four_classes/layer_types/model_cosine_component/test=control/160m/train=single_sample_self_mean.pdf}
    \end{subfigure} \\
    \textbf{Pythia-410m} \\
\end{tabular}
    \caption{\textbf{Pythia self alignment by layer type across scales.}}
    \label{fig:pythia-all-self-align-layer-type}
\end{figure}

\begin{figure}
    \centering
    \begin{tabular}{c}
        \textbf{Pythia-160m} \vspace{-8pt}\\
        \begin{subfigure}{0.2\linewidth}
            \includegraphics[trim=0 0 250 0,clip,width=1\linewidth]{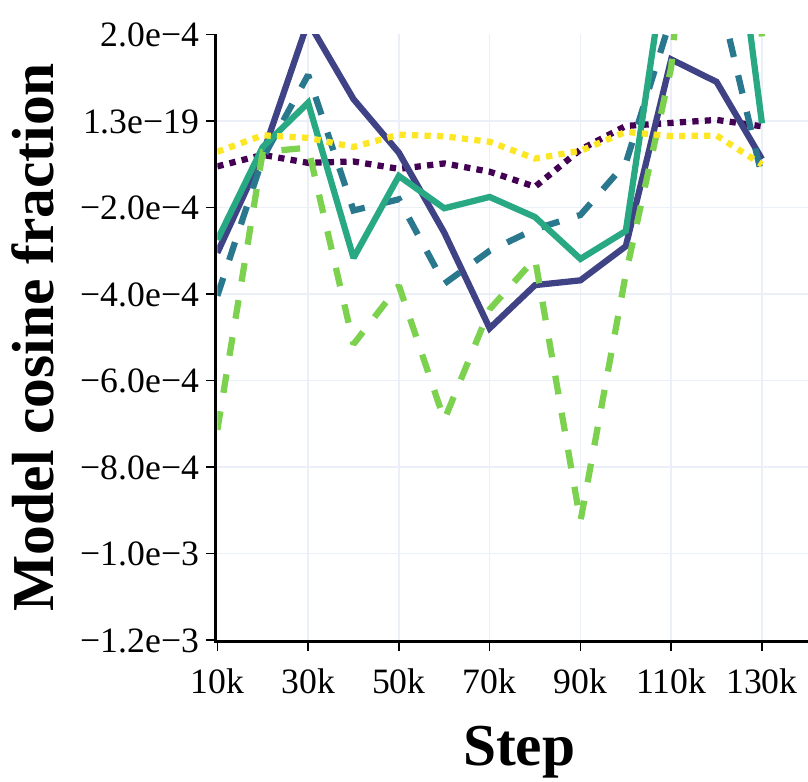}
        \end{subfigure}
        \begin{subfigure}{0.2\linewidth}
            \includegraphics[trim=0 0 250 0,clip,width=1\linewidth]{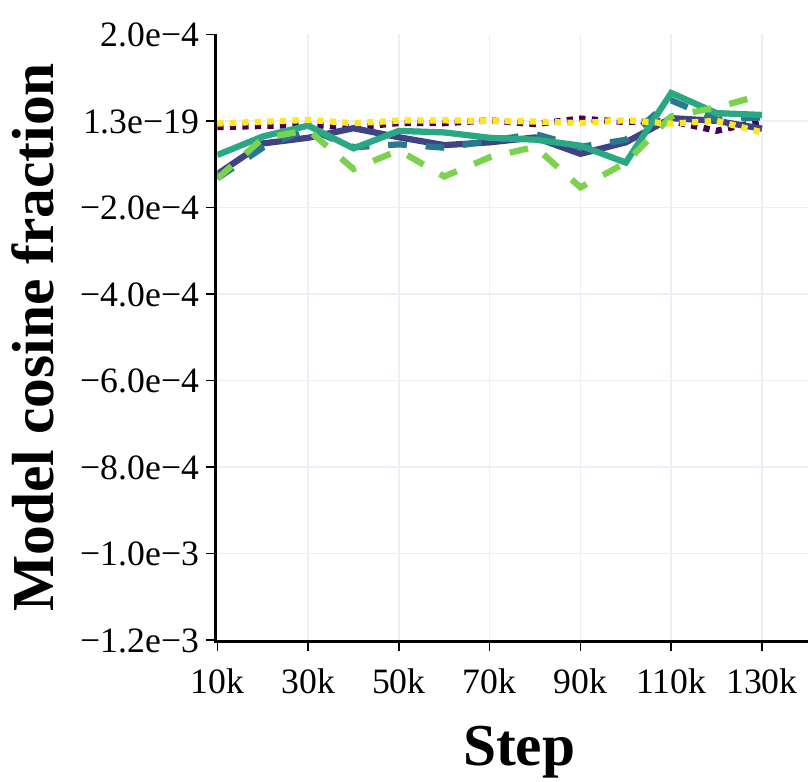}
        \end{subfigure}
        \begin{subfigure}{0.2\linewidth}
            \includegraphics[trim=0 0 250 0,clip,width=1\linewidth]{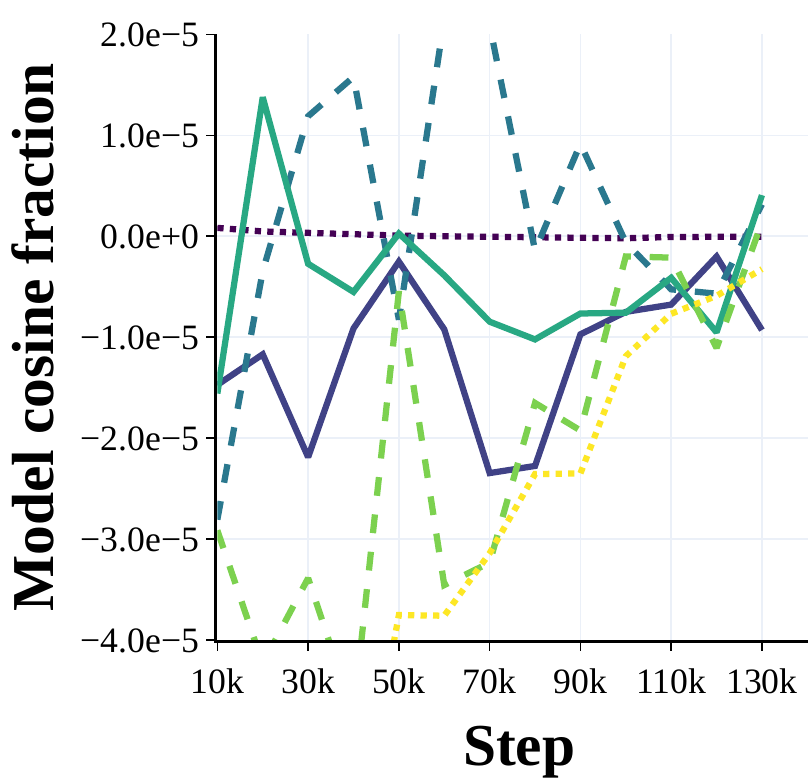}
        \end{subfigure}
        \begin{subfigure}{0.2\linewidth}
            \includegraphics[trim=0 0 250 0,clip,width=1\linewidth]{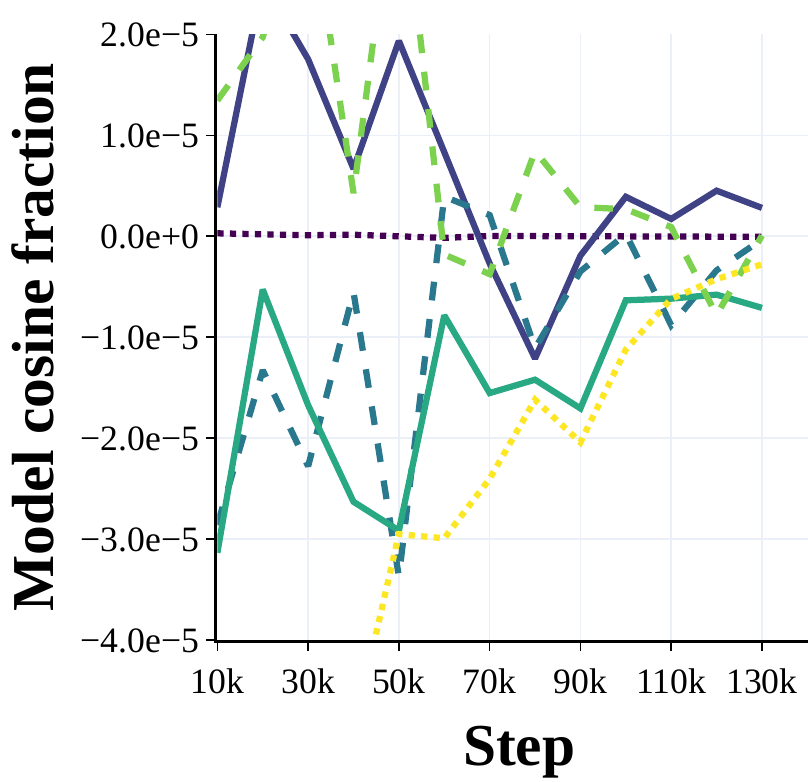}
        \end{subfigure}
        \begin{subfigure}{0.15\linewidth}
            \includegraphics[trim=395 0 0 0,clip,width=1\linewidth]{figures/decomp/phases/from_10000/layer_regions_coarse/model_cosine_component/test=memorized_recollected/160m/train=regularization_mean.pdf}
        \end{subfigure} \\
        \textbf{Pythia-410m} \vspace{-8pt}\\
\end{tabular}
    \caption{\textbf{Pythia train and regularization alignment by model region across scales.}}
    \label{fig:pythia-all-train-reg-align-layer}
\end{figure}

Figures~\ref{fig:pythia-alignment-combined-early-and-late} and~\ref{fig:pythia-alignment-regularization-combined} compare alignment with updates from a sequence's own occurrences, the original training data, and weight decay. Memorized sequences exhibit strong self-alignment, while recited sequences show more negative alignment with other training data and weight decay than recollected sequences. Contributions to self-alignment are broken down by model region in Fig.~\ref{fig:pythia-all-self-align-layer} and layer type in Fig.~\ref{fig:pythia-all-self-align-layer-type}, highlighting the role of lower layers. Figure~\ref{fig:pythia-all-train-reg-align-layer} provides the corresponding layer-wise decomposition for training data and weight decay alignment. Alignment between other injected sequences across scales is shown in Fig.~\ref{fig:align-all-other}.

\subsection{Prediction}
\label{app:additional_prediction}
\begin{figure}
    \centering
    \begin{tabular}{c}
    \textbf{Pythia-160m}\vspace{-5pt}\\
    \begin{subfigure}{0.27\linewidth}
        \includegraphics[width=1\linewidth]{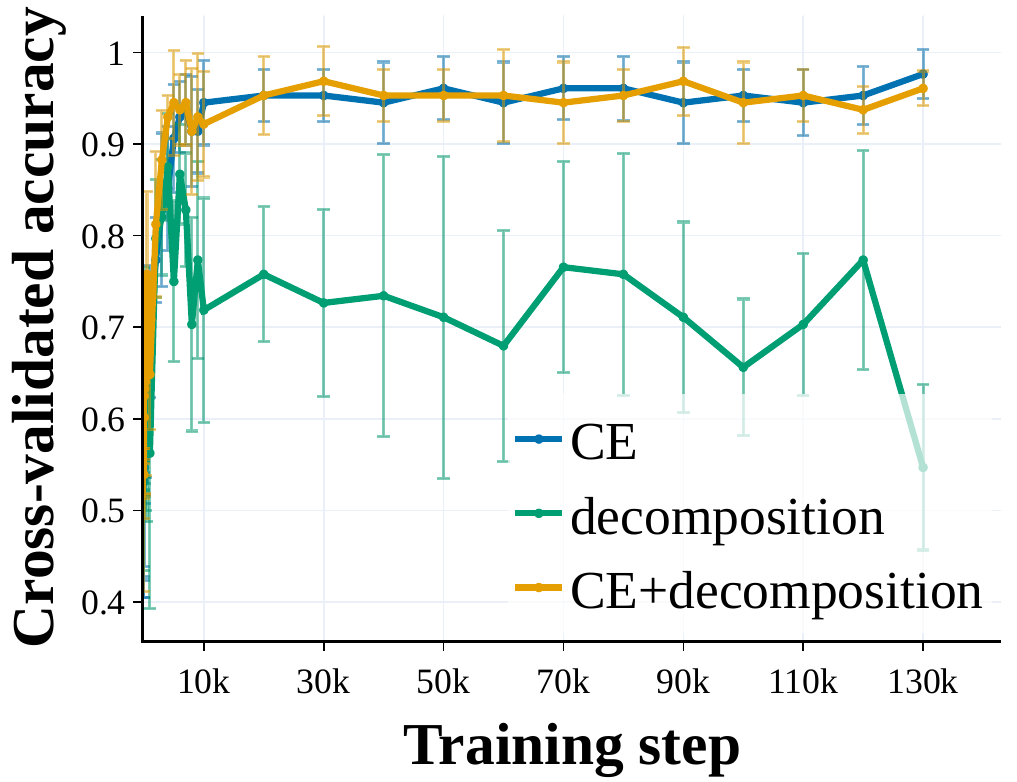}
    \end{subfigure}
    \begin{subfigure}{0.19\linewidth}
        \includegraphics[trim=0 0 240 0,clip,width=1\linewidth]{figures/prediction/1.4b/single_decomposition_only_weights.pdf}
    \end{subfigure}
    \begin{subfigure}{0.35\linewidth}
    \includegraphics[trim=0 0 240 0,clip,width=0.5\linewidth]{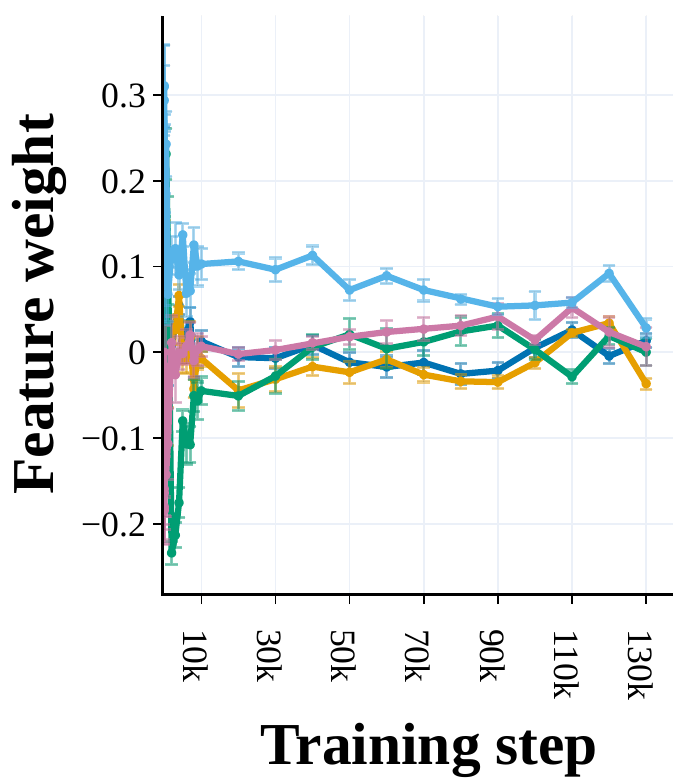}
    \includegraphics[trim=335 0 0 0,clip,width=0.4\linewidth]{figures/prediction/160m/single_loss_decomposition_features_without_loss_weights.pdf}
    \end{subfigure} \\
    \textbf{Pythia-410m}\vspace{-5pt}\\
    \begin{subfigure}{0.27\linewidth}
        \includegraphics[width=1\linewidth]{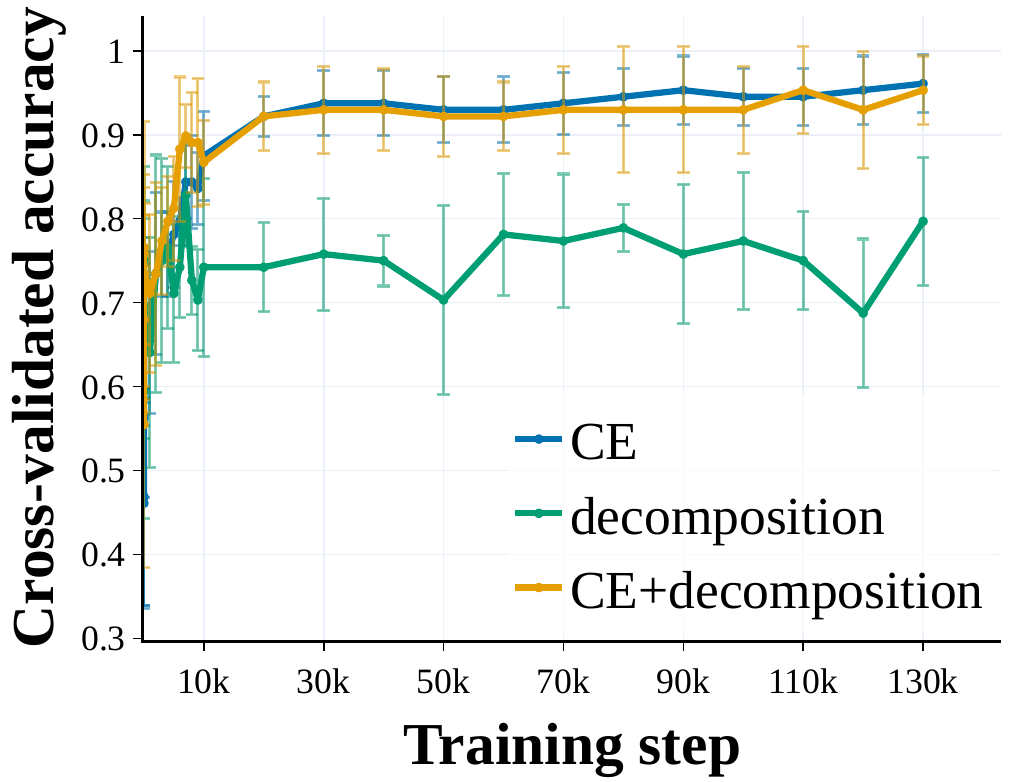}
    \end{subfigure}
    \begin{subfigure}{0.19\linewidth}
        \includegraphics[trim=0 0 240 0,clip,width=1\linewidth]{figures/prediction/1.4b/single_decomposition_only_weights.pdf}
    \end{subfigure}
    \begin{subfigure}{0.35\linewidth}
    \includegraphics[trim=0 0 240 0,clip,width=0.5\linewidth]{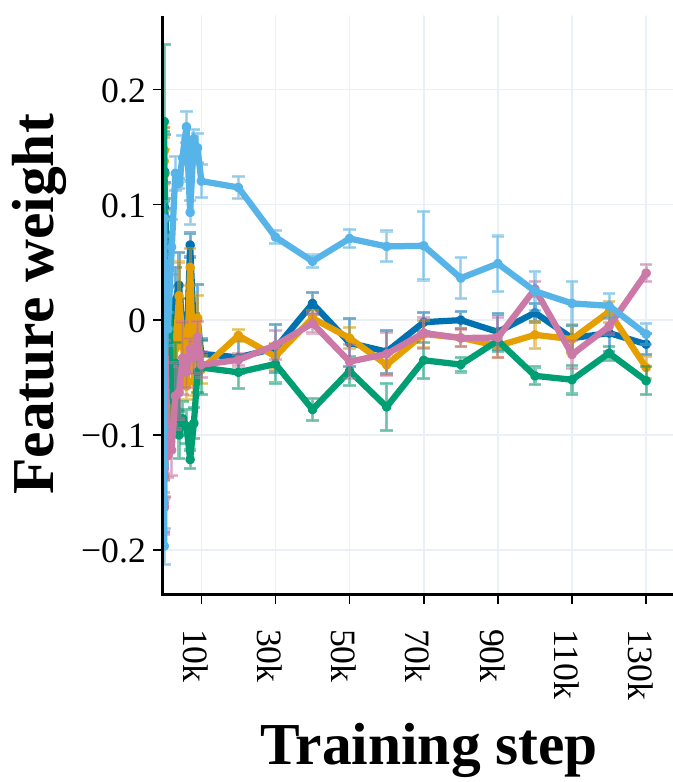}
    \includegraphics[trim=335 0 0 0,clip,width=0.4\linewidth]{figures/prediction/410m/single_loss_decomposition_features_without_loss_weights.pdf}
    \end{subfigure} \\
    \textbf{Pythia-1B}\vspace{-5pt}\\
\end{tabular}
    \caption{\textbf{Memorization vs. not memorization classification across model scales.} We also plot the weights of the decomposition features for the decomposition only and CE+decomposition model.}
    \label{fig:pythia-pred-combined}
\end{figure}

Figure~\ref{fig:pythia-pred-combined} reports held-out memorization classification accuracy and the coefficients of linear predictors using decomposition features, alone or with cross-entropy. These features improve prediction over the cross-entropy baseline, particularly in larger models and early in training.

\subsection{Causal parameter interventions}
\label{app:additional_interventions}
\begin{figure}
    \centering
    \begin{tabular}{c}
    \textbf{Pythia-160m} \\
    \begin{subfigure}{0.33\linewidth}
        \includegraphics[trim=0 0 0 0,clip,width=1\linewidth]{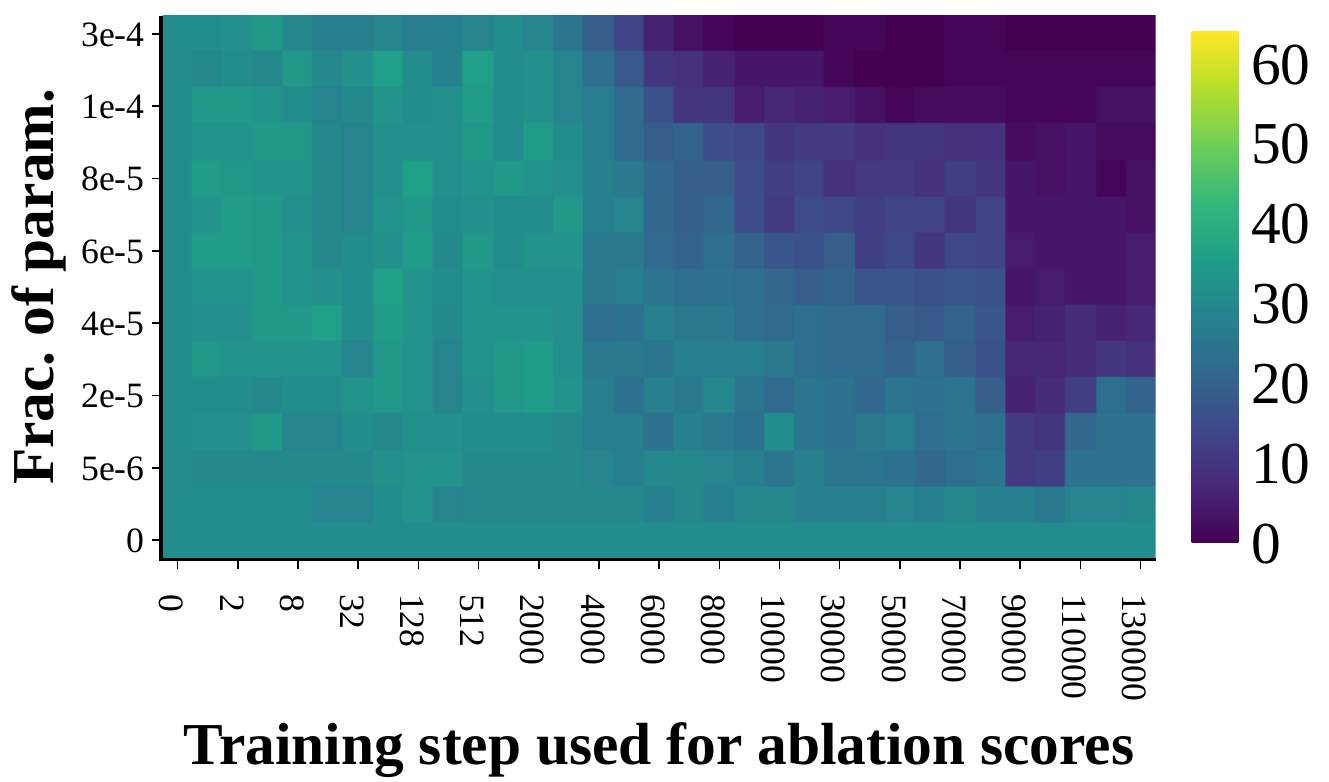}
    \end{subfigure}
    \begin{subfigure}{0.33\linewidth}
        \includegraphics[trim=0 0 0 0,clip,width=1\linewidth]{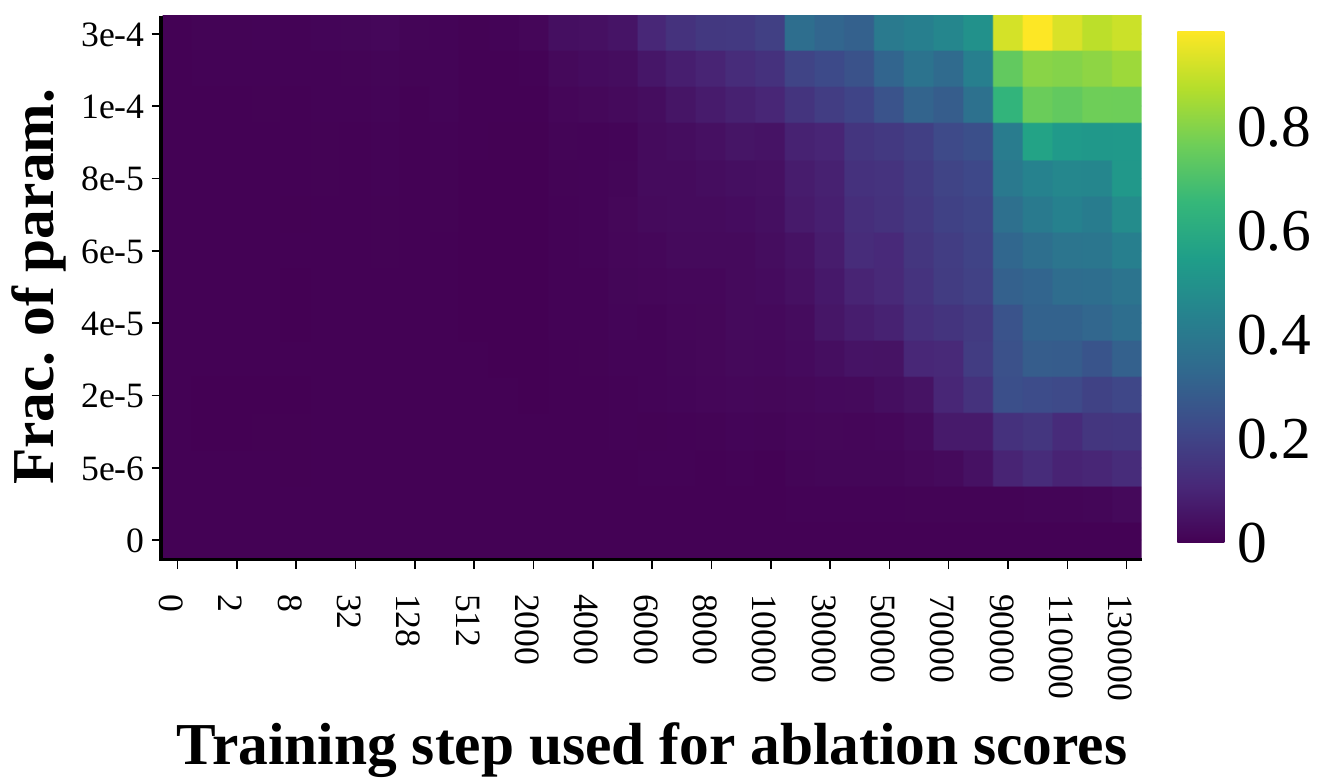}
    \end{subfigure} \\
    \textbf{Pythia-410m} \\
    \begin{subfigure}{0.33\linewidth}
        \includegraphics[trim=0 0 0 0,clip,width=1\linewidth]{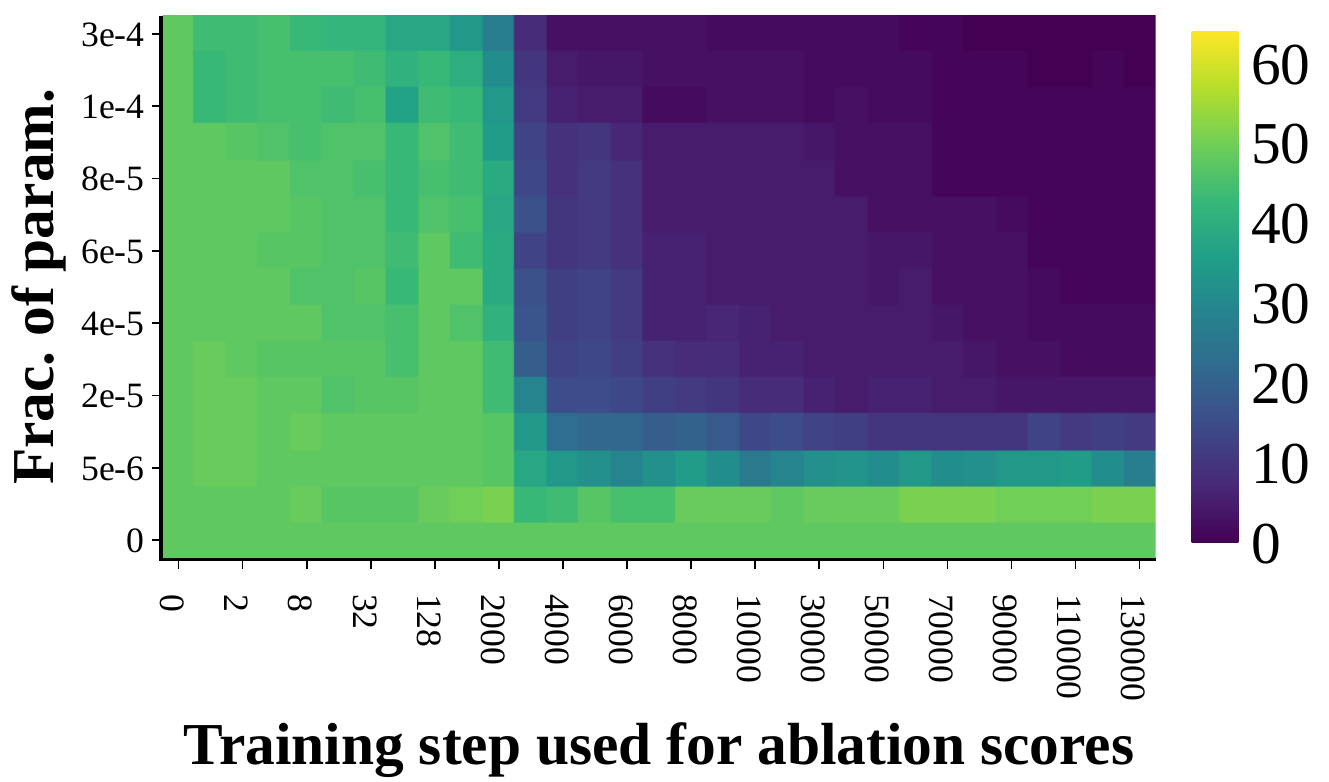}
    \end{subfigure}
    \begin{subfigure}{0.33\linewidth}
        \includegraphics[trim=0 0 0 0,clip,width=1\linewidth]{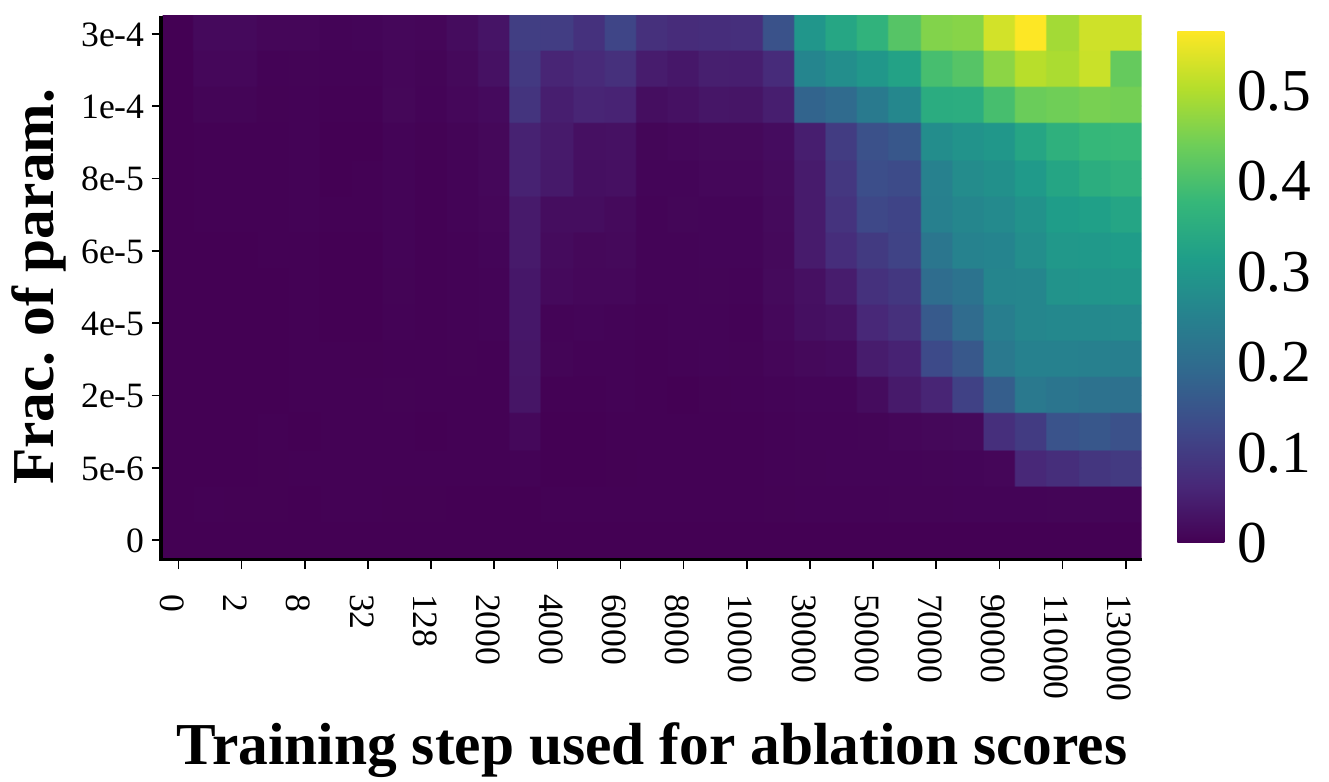}
    \end{subfigure} \\
    \textbf{Pythia-1B} \\
    \begin{subfigure}{0.33\linewidth}
        \includegraphics[trim=0 0 0 0,clip,width=1\linewidth]{figures/causal_ablations/410m/pythia_ablation_grid_32_extractable_all_no_embeddings_zero.pdf}
    \end{subfigure}
    \begin{subfigure}{0.33\linewidth}
        \includegraphics[trim=0 0 0 0,clip,width=1\linewidth]{figures/causal_ablations/1b/pythia_ablation_grid_control_loss_change_no_embeddings_zero.pdf}
    \end{subfigure} \\
    \textbf{Pythia-1.4B} \\
    \begin{subfigure}{0.33\linewidth}
        \includegraphics[trim=0 0 0 0,clip,width=1\linewidth]{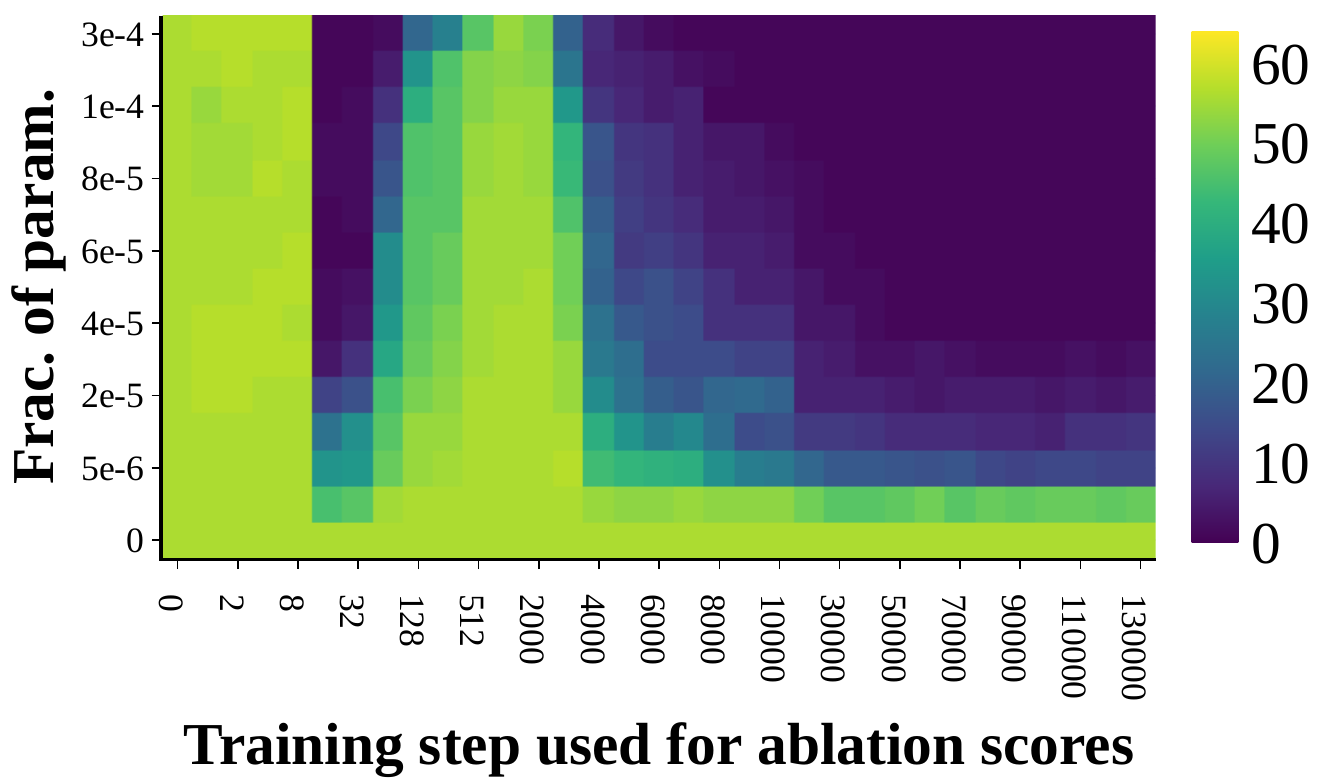}
    \end{subfigure}
    \begin{subfigure}{0.33\linewidth}
        \includegraphics[trim=0 0 0 0,clip,width=1\linewidth]{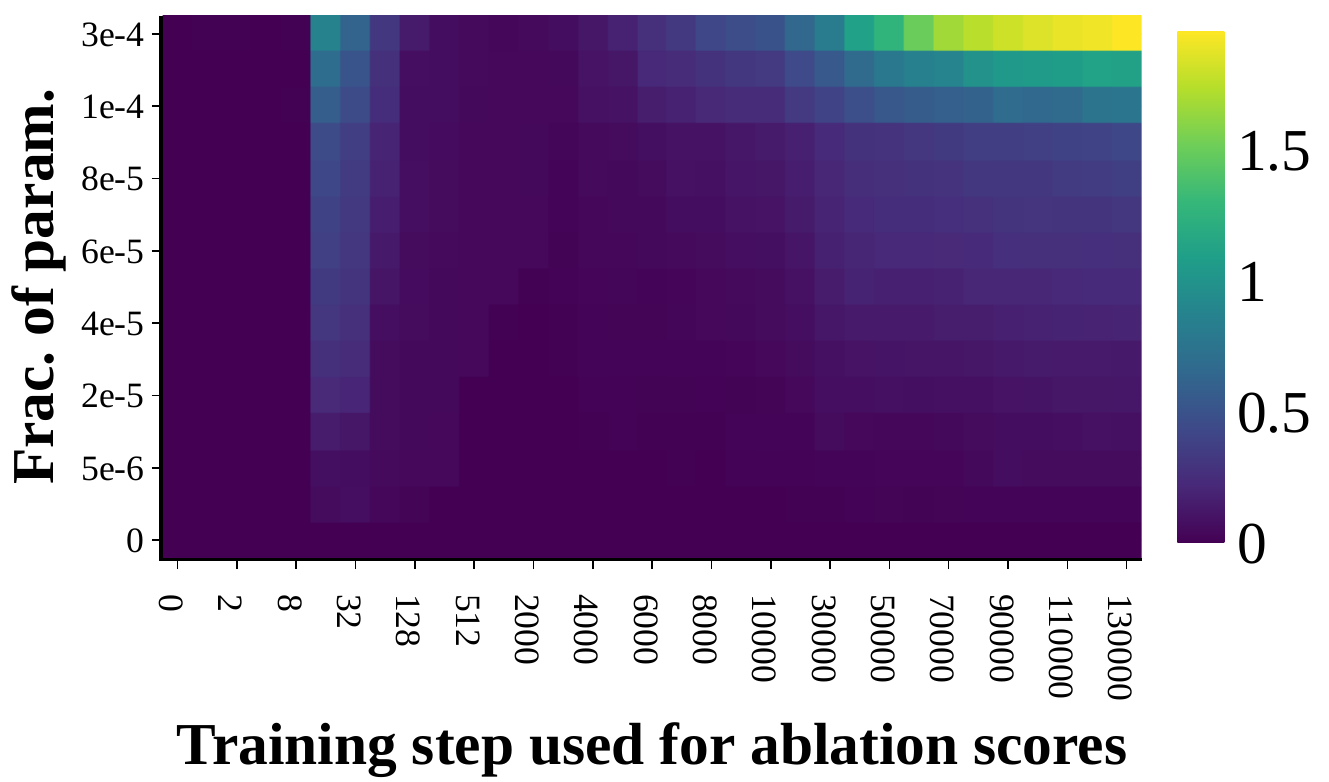}
    \end{subfigure} \\
\end{tabular}
    \caption{\textbf{Pythia memorization ablations across scales.} Left: Number of 32-extractable sequences in the memorized sample, right: CE loss on the control sample.}
    \label{fig:pythia-ablation-app}
\end{figure}

\begin{figure}
    \centering
    \begin{tabular}{c}
    \textbf{Pythia-160m} \\
    \begin{subfigure}{0.4\linewidth}
        \includegraphics[trim=0 0 0 0,clip,width=1\linewidth]{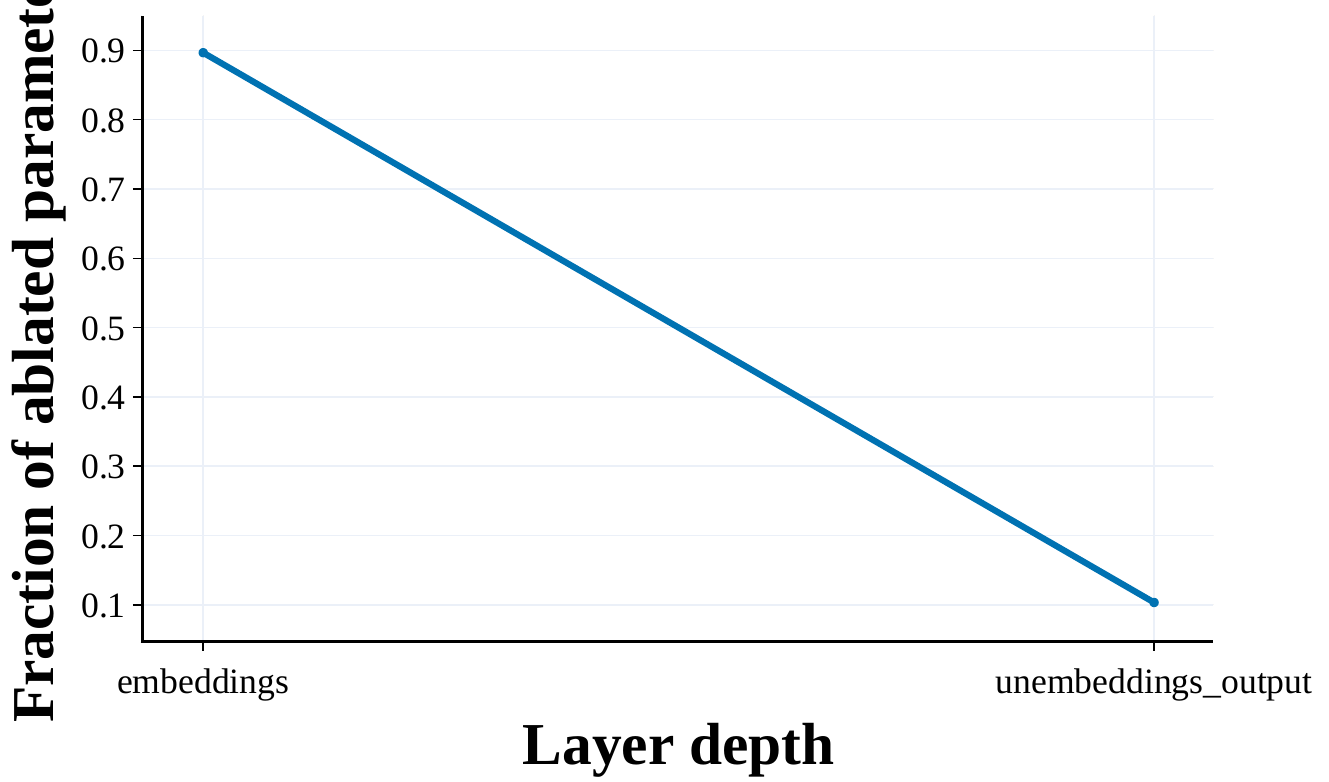}
    \end{subfigure}
    \begin{subfigure}{0.4\linewidth}
        \includegraphics[trim=0 0 0 0,clip,width=1\linewidth]{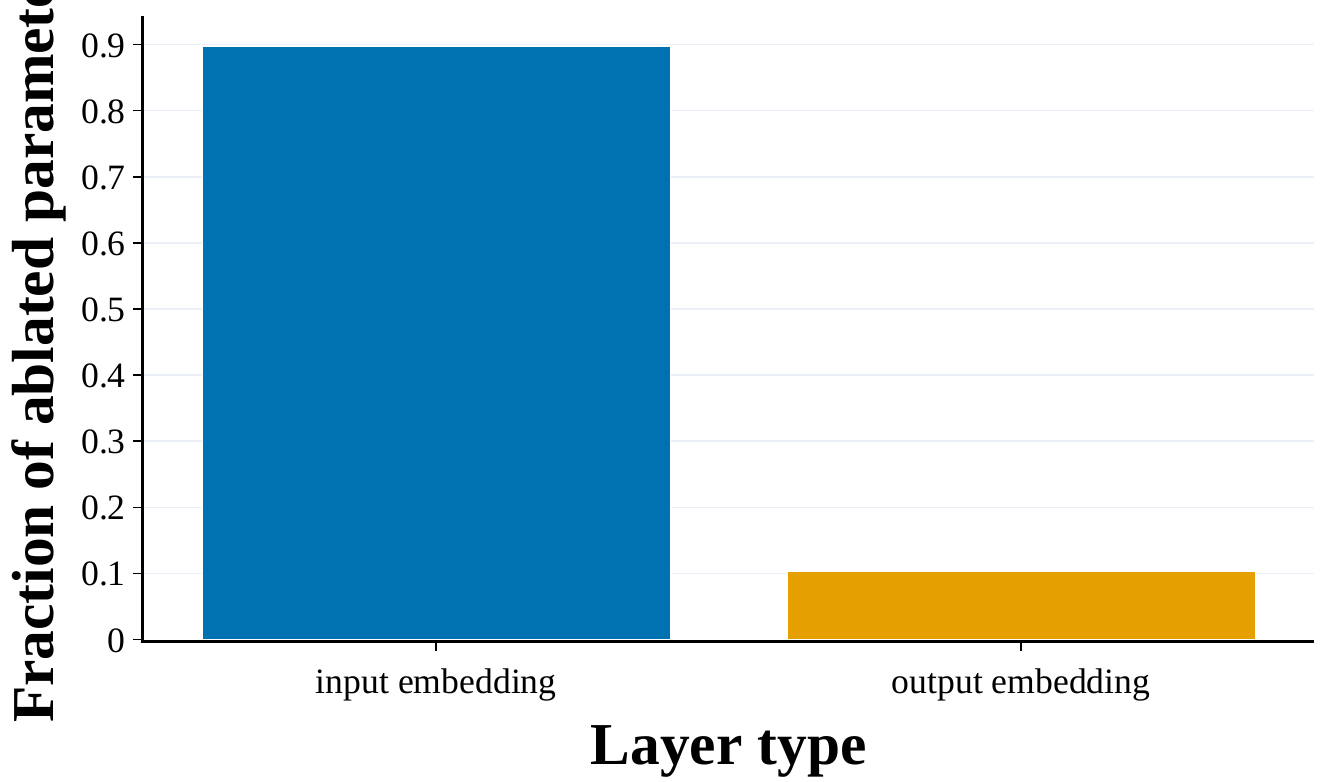}
    \end{subfigure} \\
    \textbf{Pythia-410m} \\
    \begin{subfigure}{0.4\linewidth}
        \includegraphics[trim=0 0 0 0,clip,width=1\linewidth]{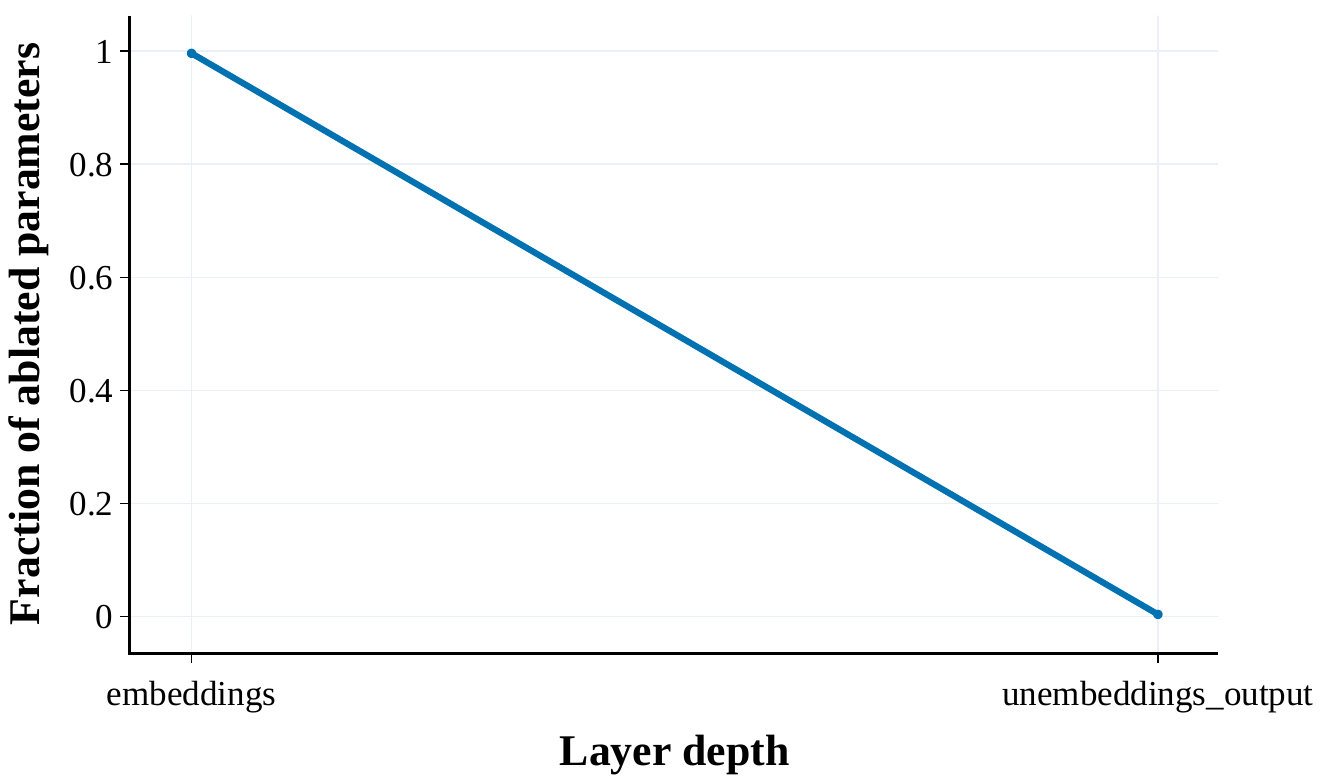}
    \end{subfigure}
    \begin{subfigure}{0.4\linewidth}
        \includegraphics[trim=0 0 0 0,clip,width=1\linewidth]{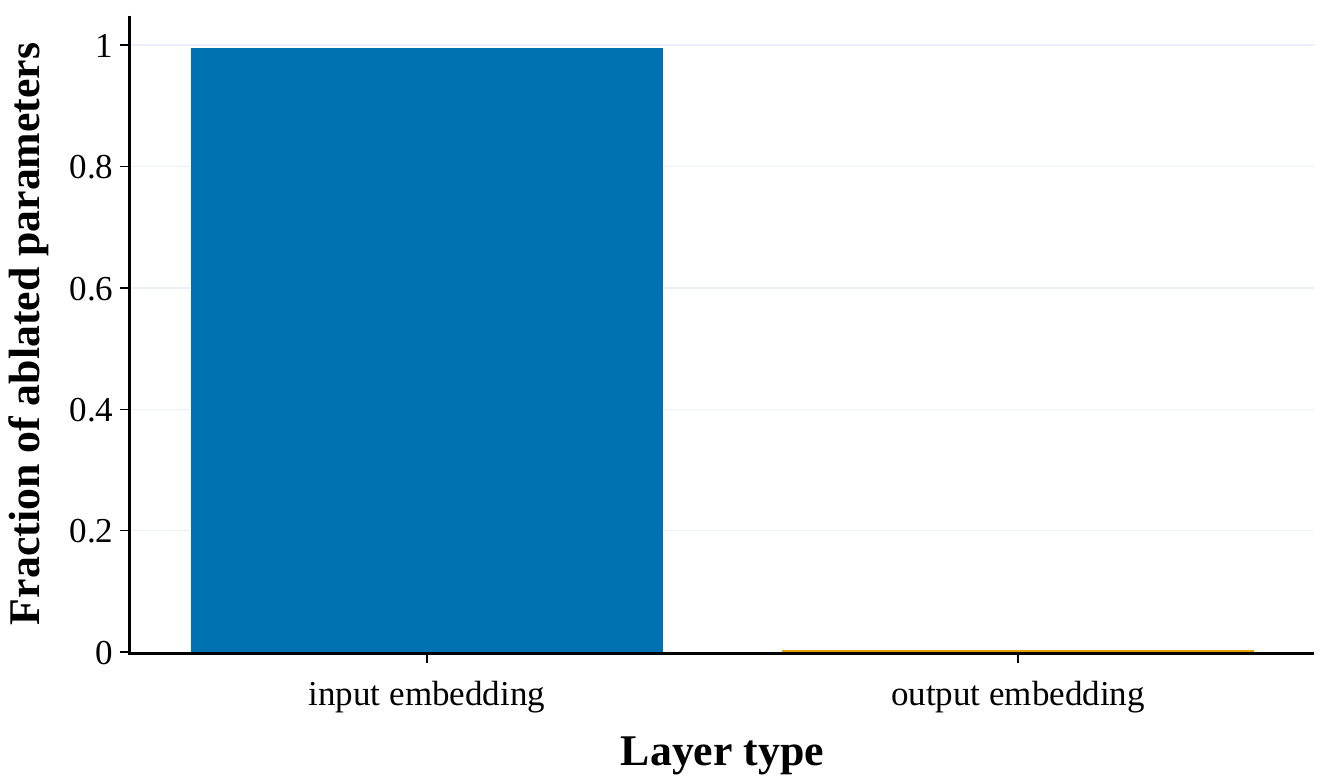}
    \end{subfigure} \\
    \textbf{Pythia-1B} \\
    \begin{subfigure}{0.4\linewidth}
        \includegraphics[trim=0 0 0 0,clip,width=1\linewidth]{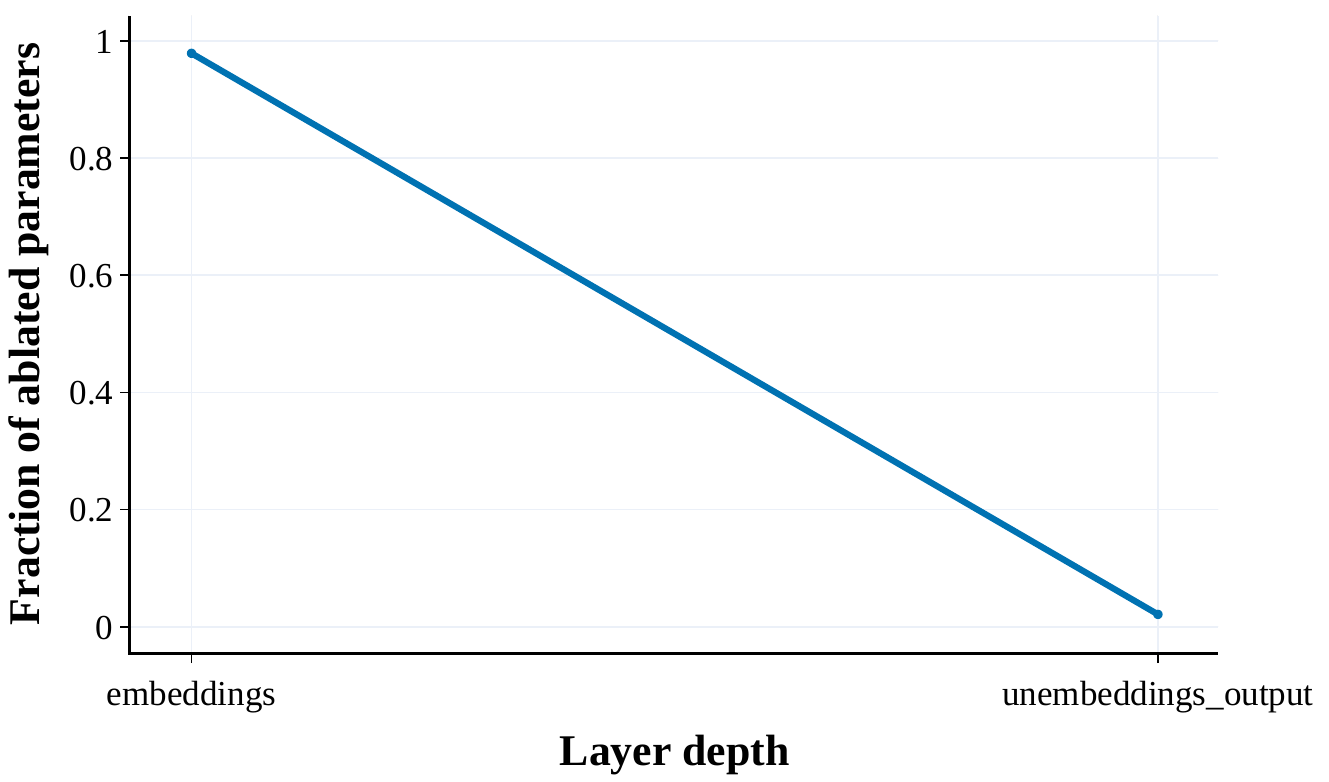}
    \end{subfigure}
    \begin{subfigure}{0.4\linewidth}
        \includegraphics[trim=0 0 0 0,clip,width=1\linewidth]{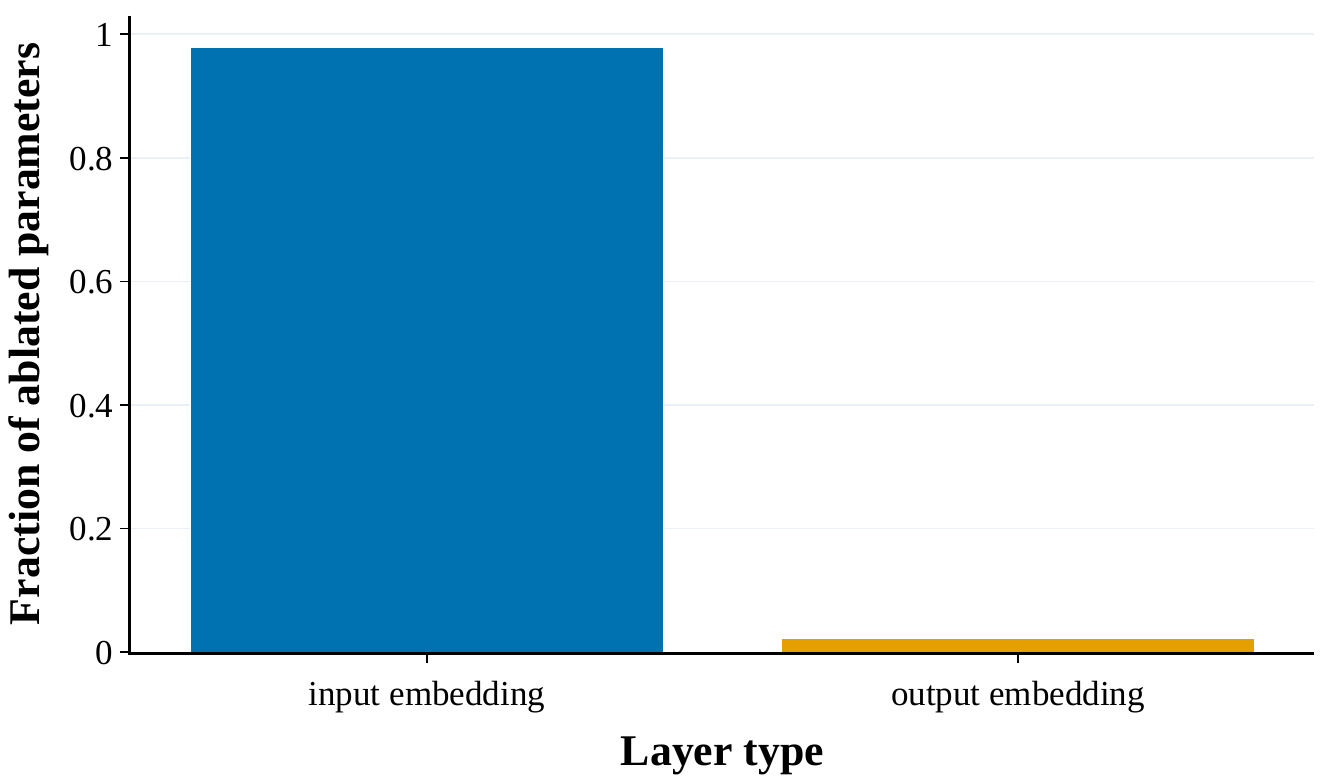}
    \end{subfigure} \\
    \textbf{Pythia-1.4B} \\
\end{tabular}
    \caption{\textbf{Pythia memorization ablations affected parameters, embeddings included.} Ablation at step $10,000$ with $0.00004$ of all model parameters ablated.}
    \label{fig:pythia-ablation-app-params-embs}
\end{figure}

\begin{figure}
    \centering
    \begin{tabular}{c}
    \textbf{Pythia-160m} \\
    \begin{subfigure}{0.4\linewidth}
        \includegraphics[trim=0 0 0 0,clip,width=1\linewidth]{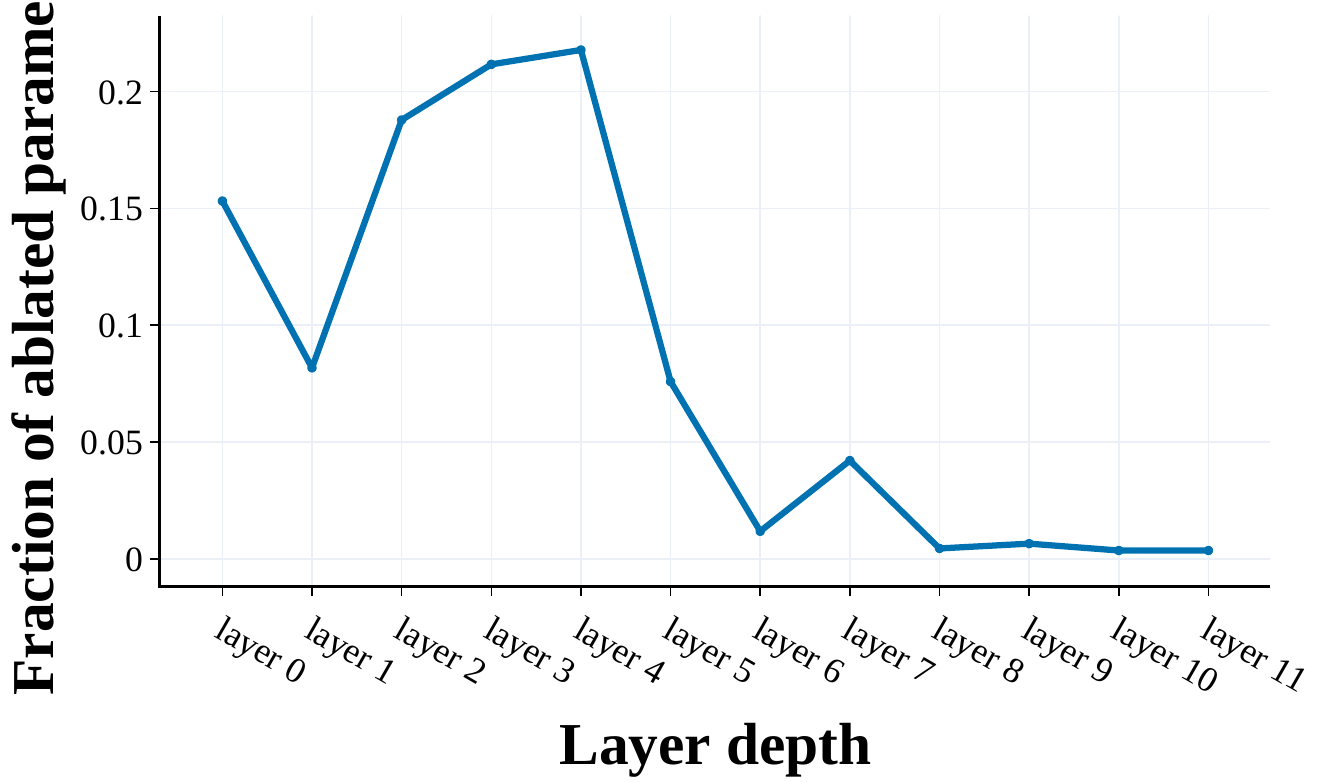}
    \end{subfigure}
    \begin{subfigure}{0.4\linewidth}
        \includegraphics[trim=0 0 0 0,clip,width=1\linewidth]{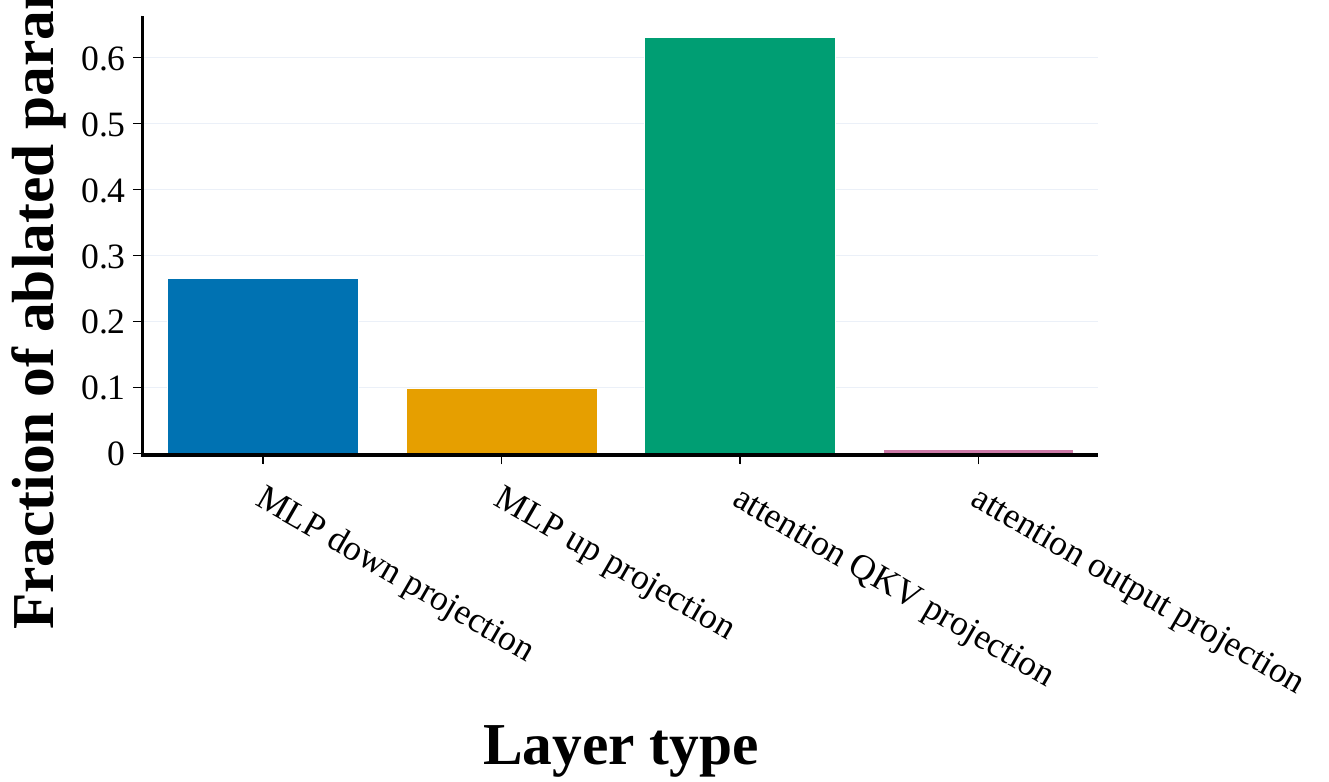}
    \end{subfigure} \\
    \textbf{Pythia-410m} \\
    \begin{subfigure}{0.4\linewidth}
        \includegraphics[trim=0 0 0 0,clip,width=1\linewidth]{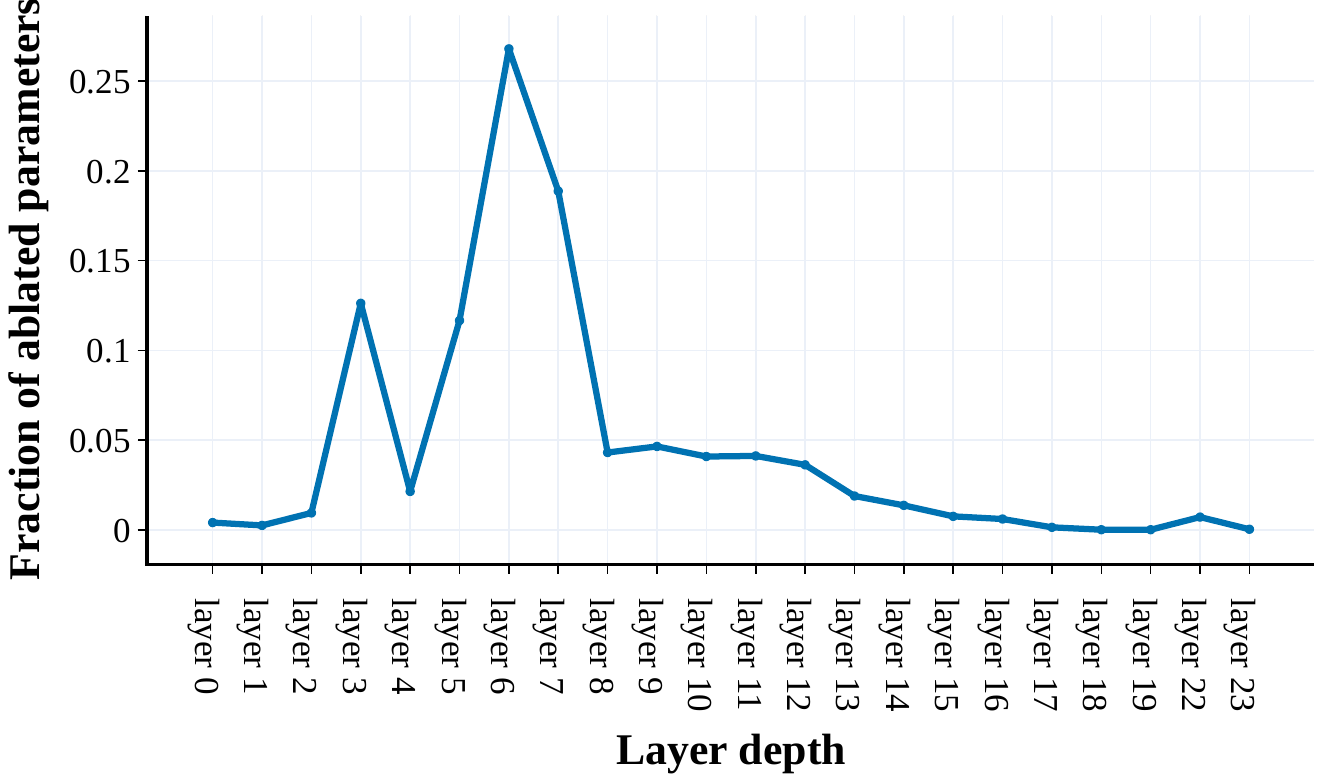}
    \end{subfigure}
    \begin{subfigure}{0.4\linewidth}
        \includegraphics[trim=0 0 0 0,clip,width=1\linewidth]{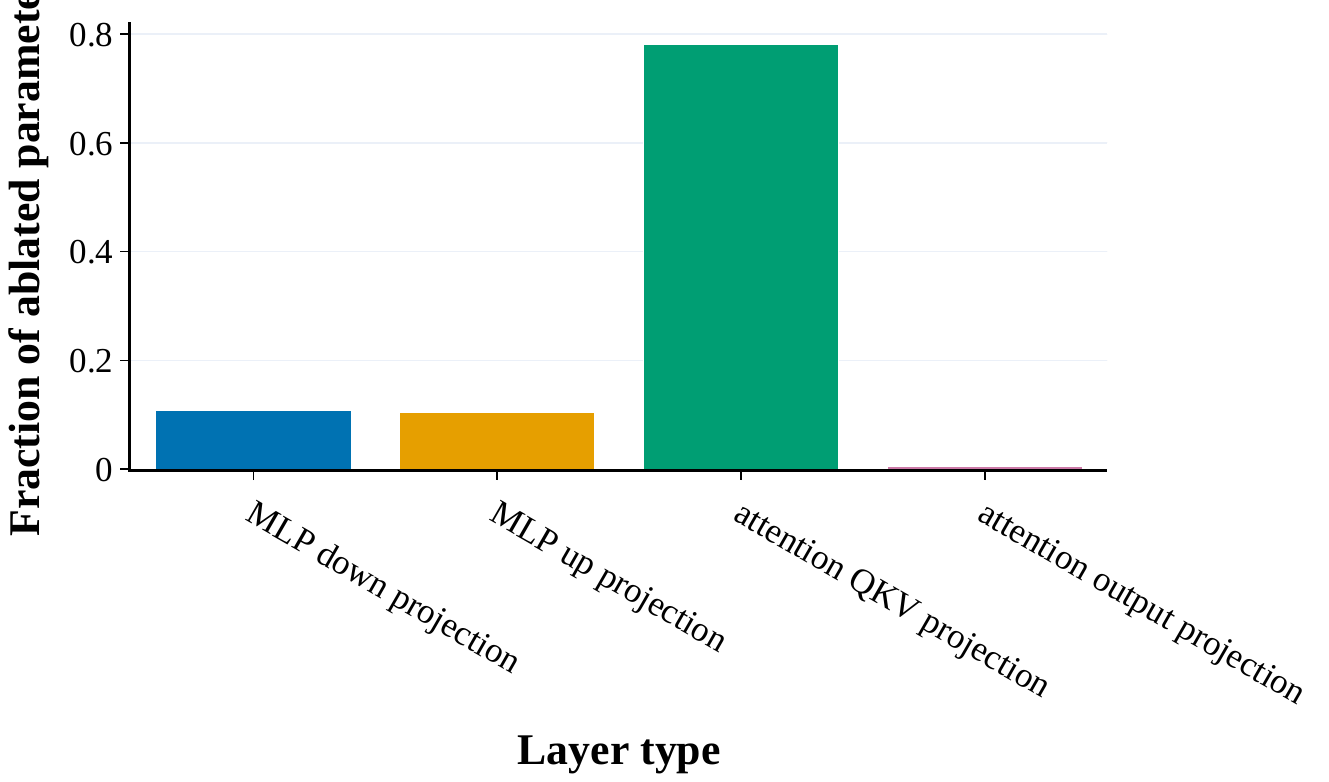}
    \end{subfigure} \\
    \textbf{Pythia-1B} \\
    \begin{subfigure}{0.4\linewidth}
        \includegraphics[trim=0 0 0 0,clip,width=1\linewidth]{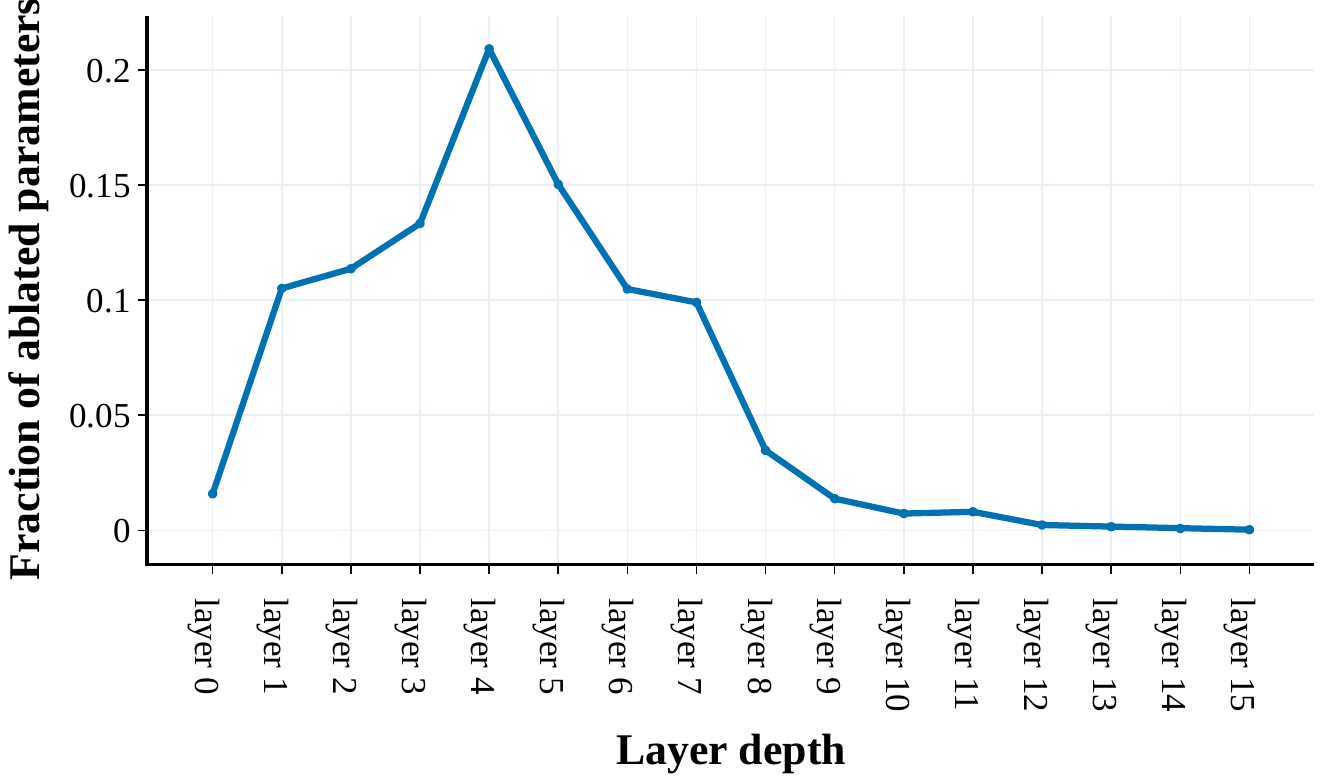}
    \end{subfigure}
    \begin{subfigure}{0.4\linewidth}
        \includegraphics[trim=0 0 0 0,clip,width=1\linewidth]{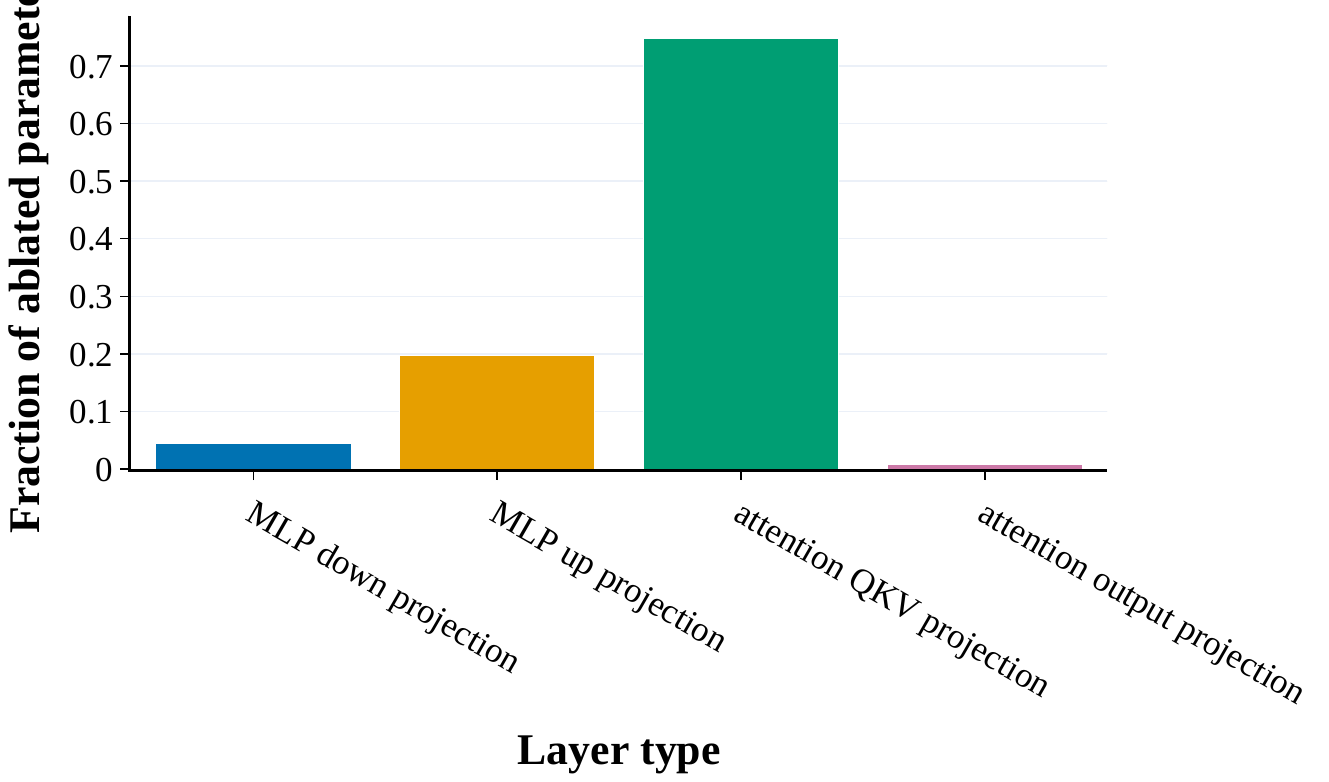}
    \end{subfigure} \\
    \textbf{Pythia-1.4B} \\
    \begin{subfigure}{0.4\linewidth}
        \includegraphics[trim=0 0 0 0,clip,width=1\linewidth]{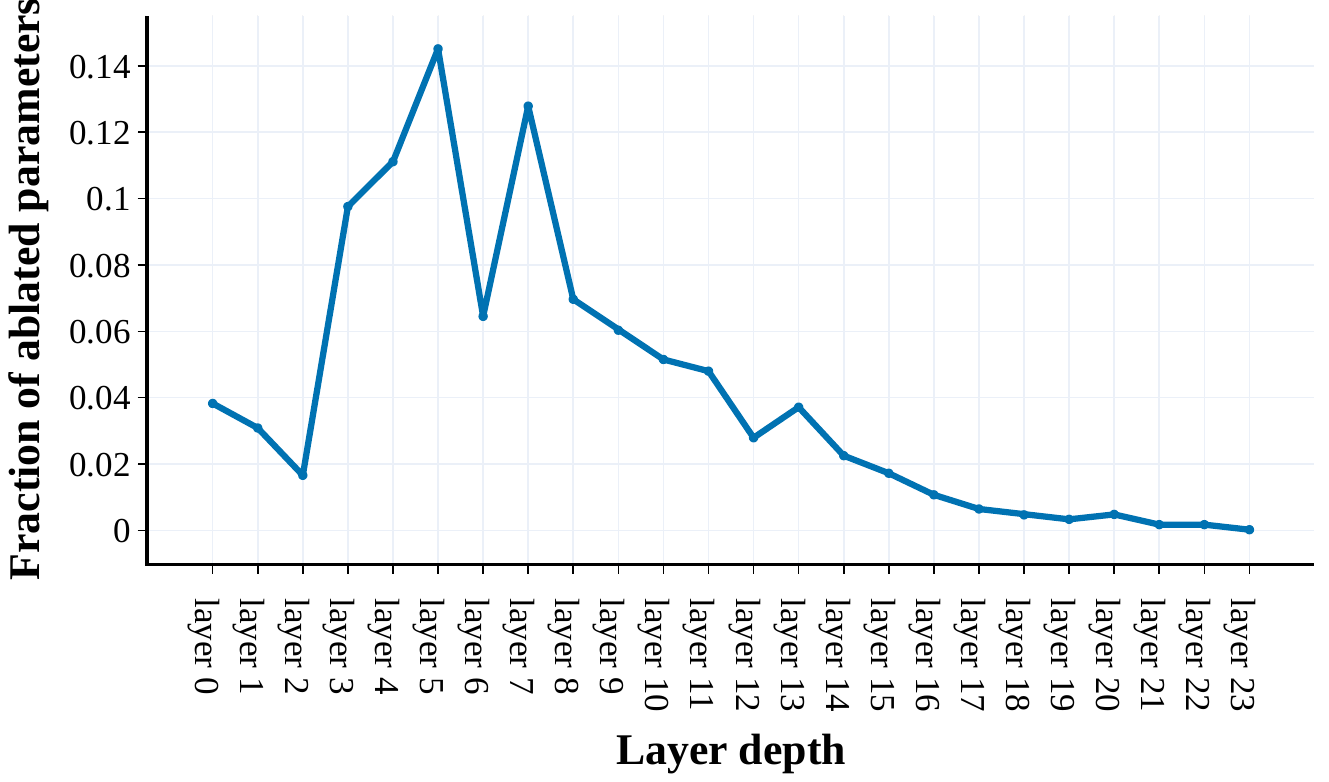}
    \end{subfigure}
    \begin{subfigure}{0.4\linewidth}
        \includegraphics[trim=0 0 0 0,clip,width=1\linewidth]{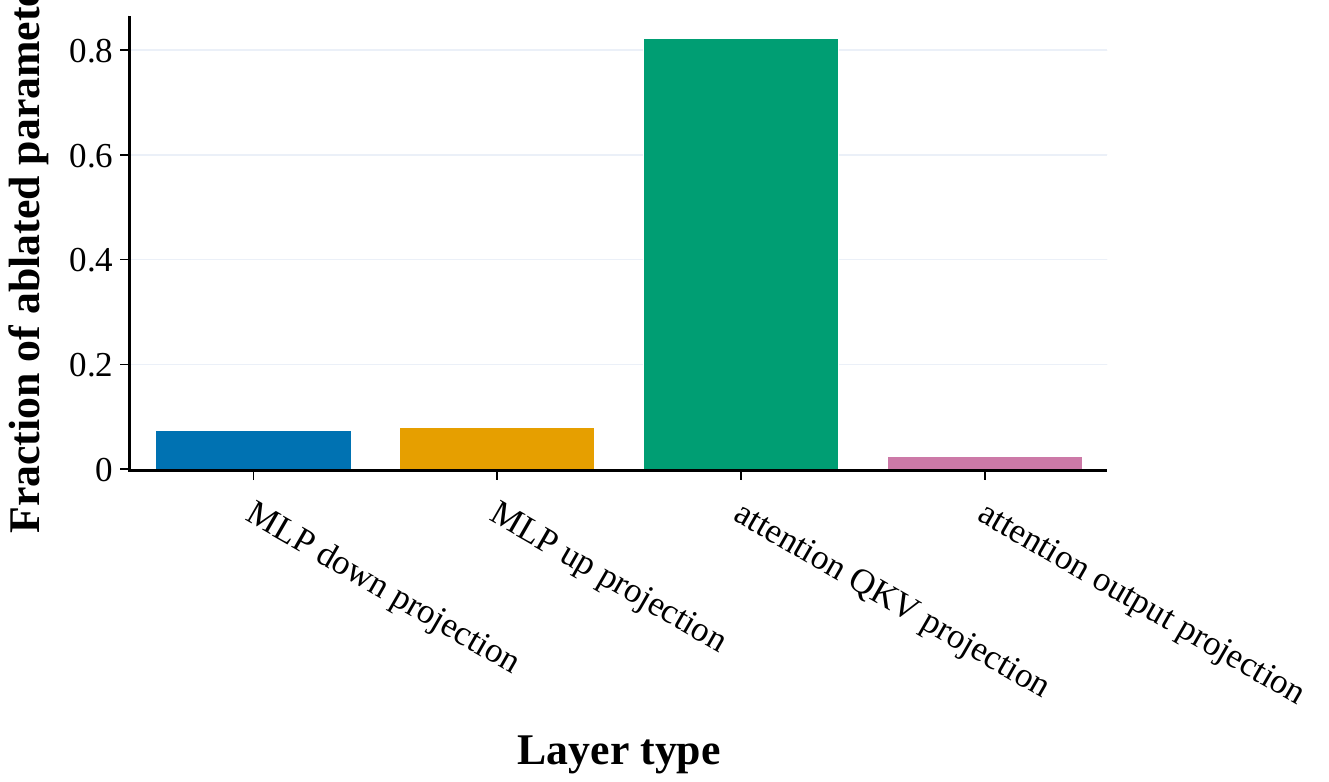}
    \end{subfigure} \\
\end{tabular}
    \caption{\textbf{Pythia memorization ablations affected parameters, embeddings excluded.} Ablation at step $10,000$ with $0.00004$ of all model parameters ablated.}
    \label{fig:pythia-ablation-app-params}
\end{figure}

Figure~\ref{fig:pythia-ablation-app} reports the effects of parameter ablation on $32$-extractability and control loss across model scales. Figures~\ref{fig:pythia-ablation-app-params-embs} and~\ref{fig:pythia-ablation-app-params} show the selected parameters by depth and layer type with embeddings included and excluded, respectively, for the $10,000$-step setting and an ablated fraction of $4\times10^{-5}$. The following section examines attention-head interventions in Figs.~\ref{fig:attention-ablation} and~\ref{fig:attention-ablation2}.

\subsection{A Shared memorization mechanism in lower layer attention heads}
\label{app:attention_ablation}
\begin{figure}
    \centering
    \begin{subfigure}{0.29\linewidth}
        \includegraphics[trim=0 0 0 0,clip,width=1.0\linewidth]{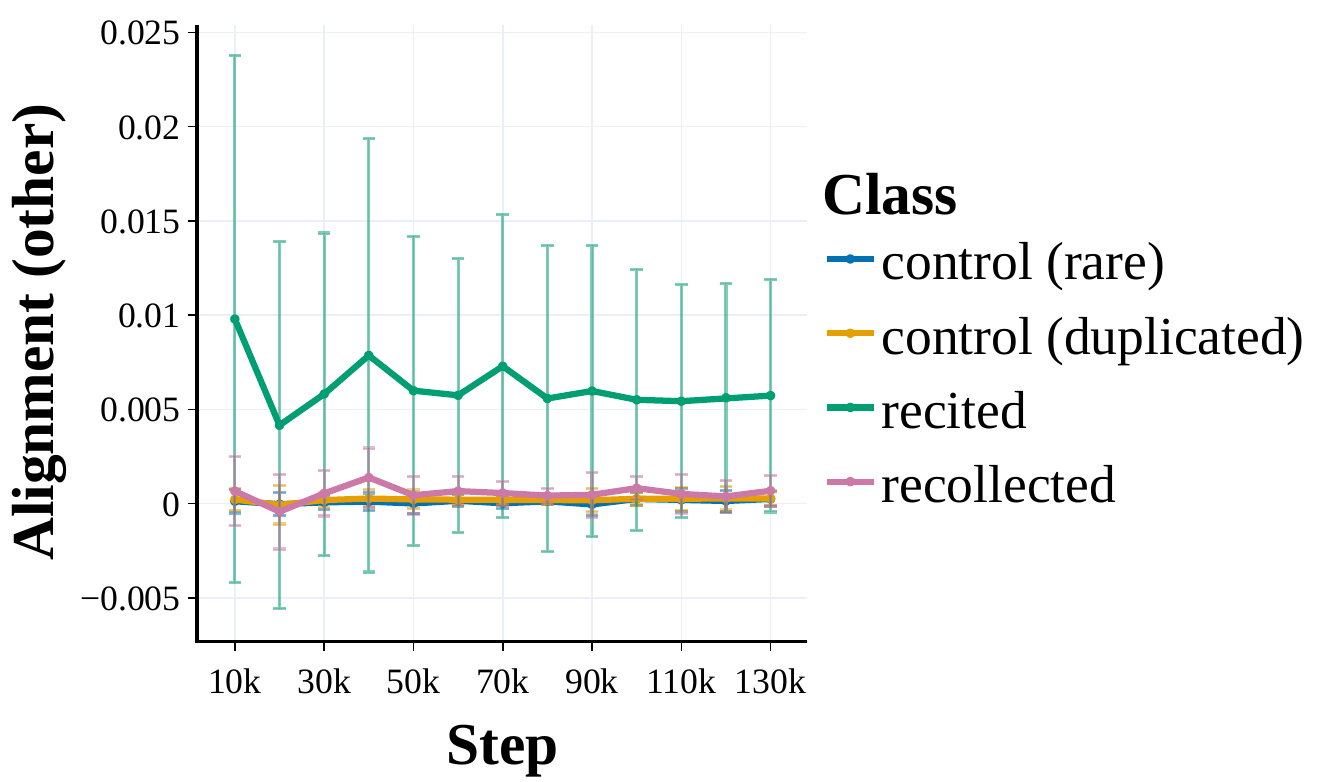}
    \caption{Alignment other}
    \end{subfigure}
    \begin{subfigure}{0.29\linewidth}
        \includegraphics[width=1.0\linewidth]{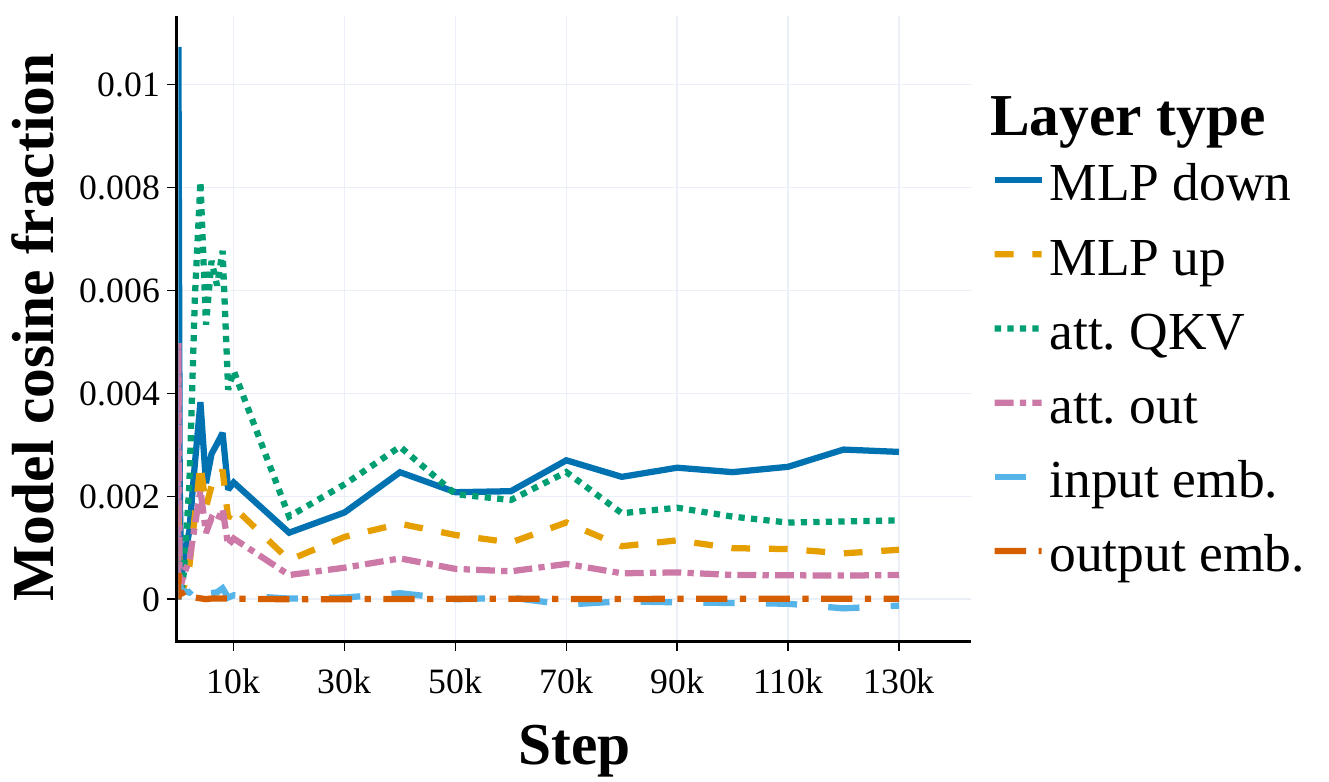}
        \caption{Decomp. layer type}
    \end{subfigure}
    \begin{subfigure}{0.29\linewidth}
        \includegraphics[width=1.0\linewidth]{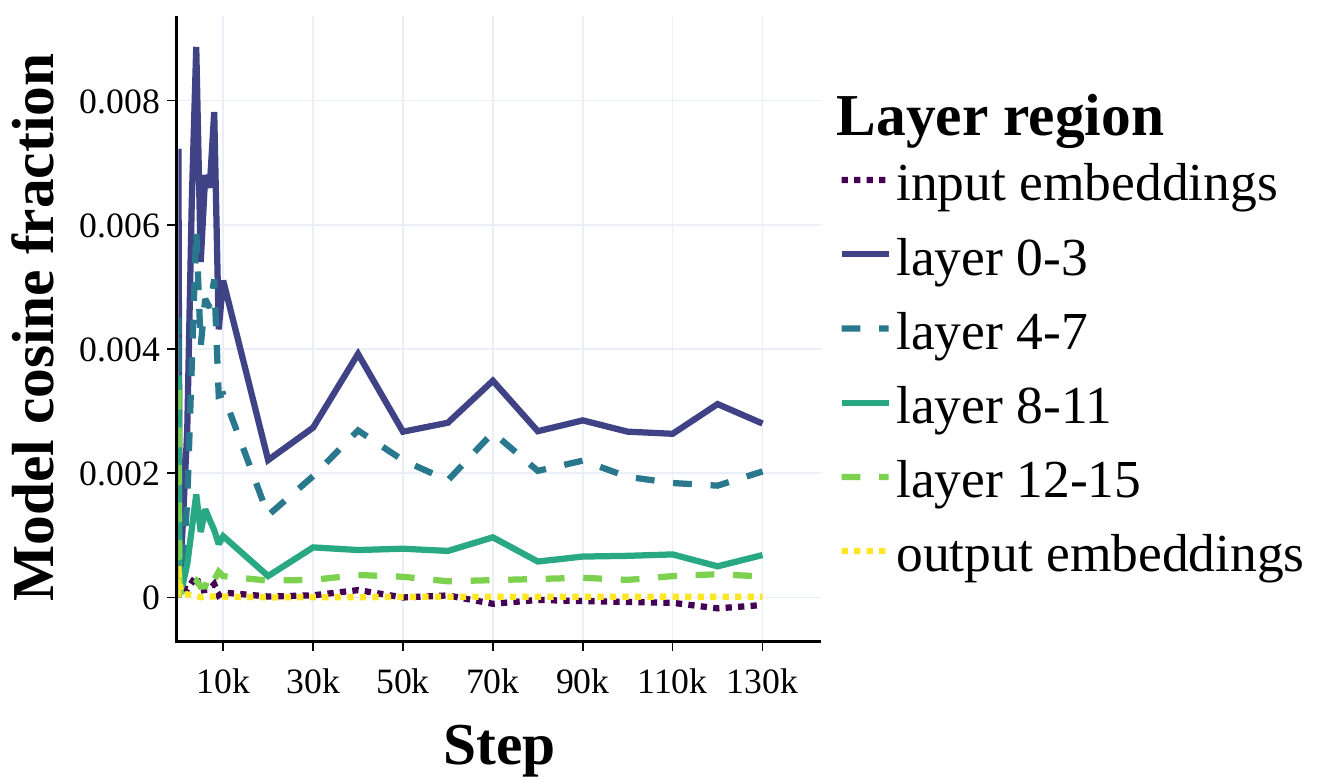}
        \caption{Decomp. layer depth}
    \end{subfigure}
    \caption{\textbf{Alignment with other injected instances.} Other models in App. Fig.~\ref{fig:align-all-other}}
    \label{fig:pythia-1b-alignment-other}
\end{figure}

We also track the alignment of our decomposed sequences with each other and plot the result in Fig.~\ref{fig:pythia-1b-alignment-other}. Interestingly, recited instances tend to have a high level of alignment with each other throughout all of training. This trend is also followed by the other instances in early training until around step $64$. This alignment suggests that memorization learning in these instances tends towards a shared mechanism. To understand where in the model such a mechanism might be located, we turn to our layer-wise alignment decomposition and find that until around step $20$k, the alignment largely stems from attention heads in the lower layers. 

\begin{figure}
    \centering
    \begin{tabular}{c}
    \textbf{Pythia-160m} \\
    \begin{subfigure}{0.3\linewidth}
    \includegraphics[width=1\linewidth]{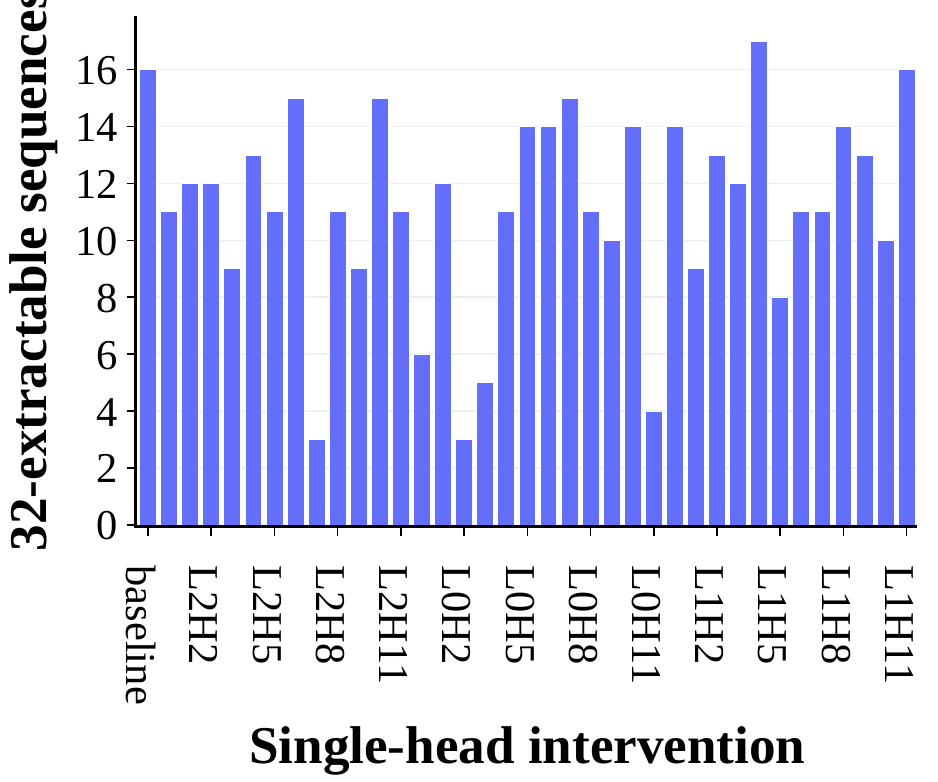}
    \caption{32-extr. recited}
    \end{subfigure}
    \begin{subfigure}{0.3\linewidth}
    \includegraphics[width=1\linewidth]{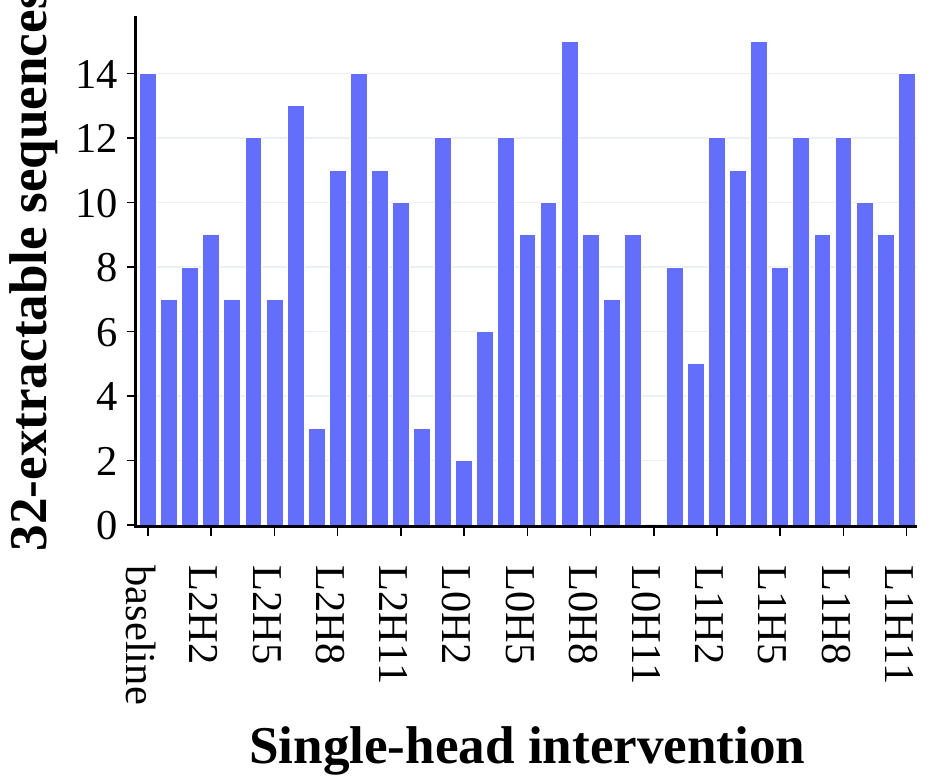}
    \caption{32-extr. recollected}
    \end{subfigure}
    \begin{subfigure}{0.3\linewidth}
    \includegraphics[width=1\linewidth]{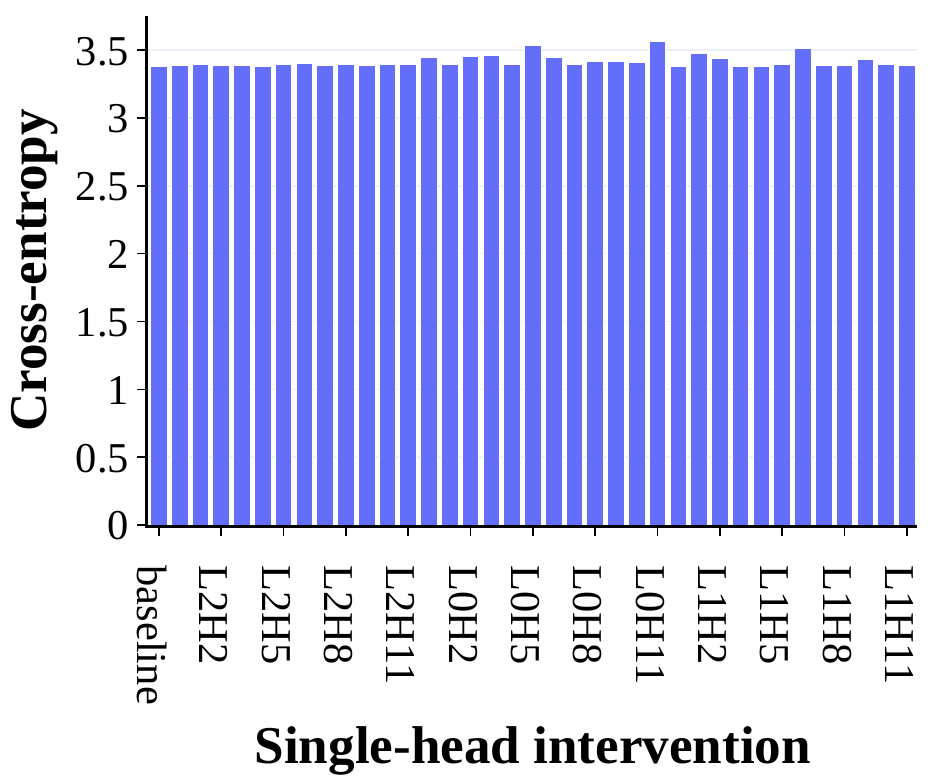}
    \caption{CE on control}
    \end{subfigure} \\
    \textbf{Pythia-410m} \\
    \begin{subfigure}{0.3\linewidth}
    \includegraphics[width=1\linewidth]{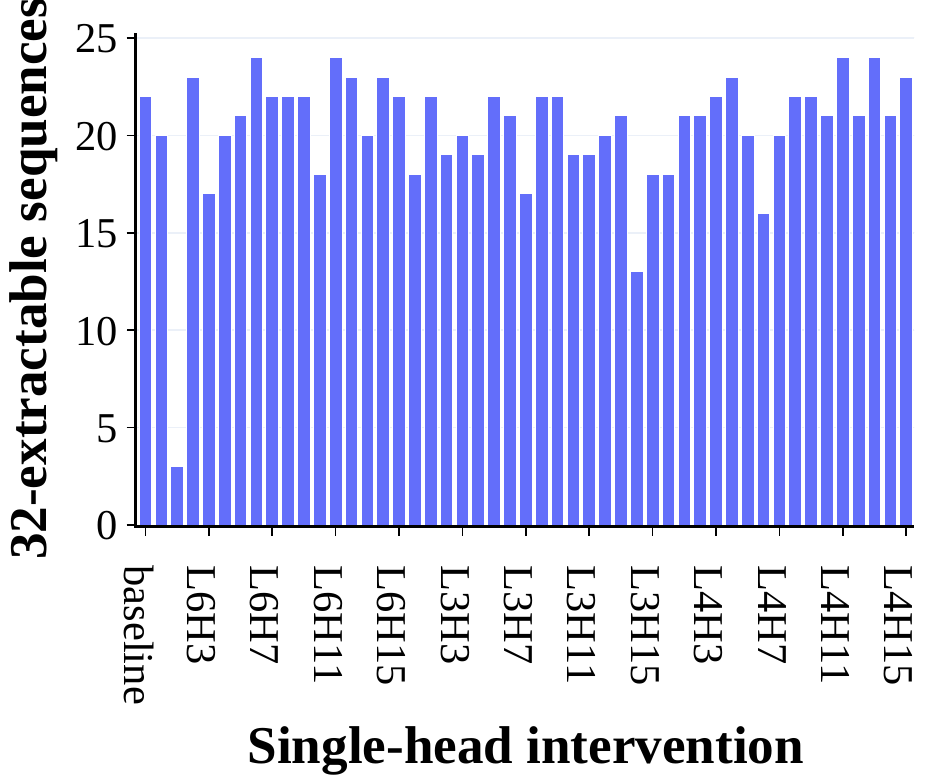}
    \caption{32-extr. recited}
    \end{subfigure}
    \begin{subfigure}{0.3\linewidth}
    \includegraphics[width=1\linewidth]{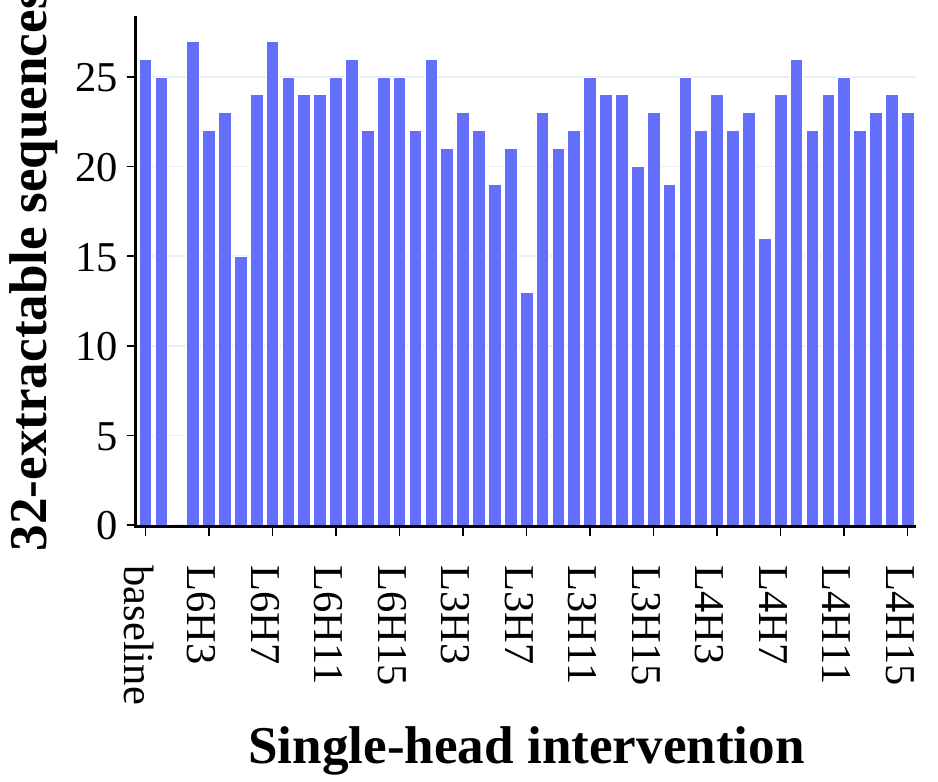}
    \caption{32-extr. recollected}
    \end{subfigure}
    \begin{subfigure}{0.3\linewidth}
    \includegraphics[width=1\linewidth]{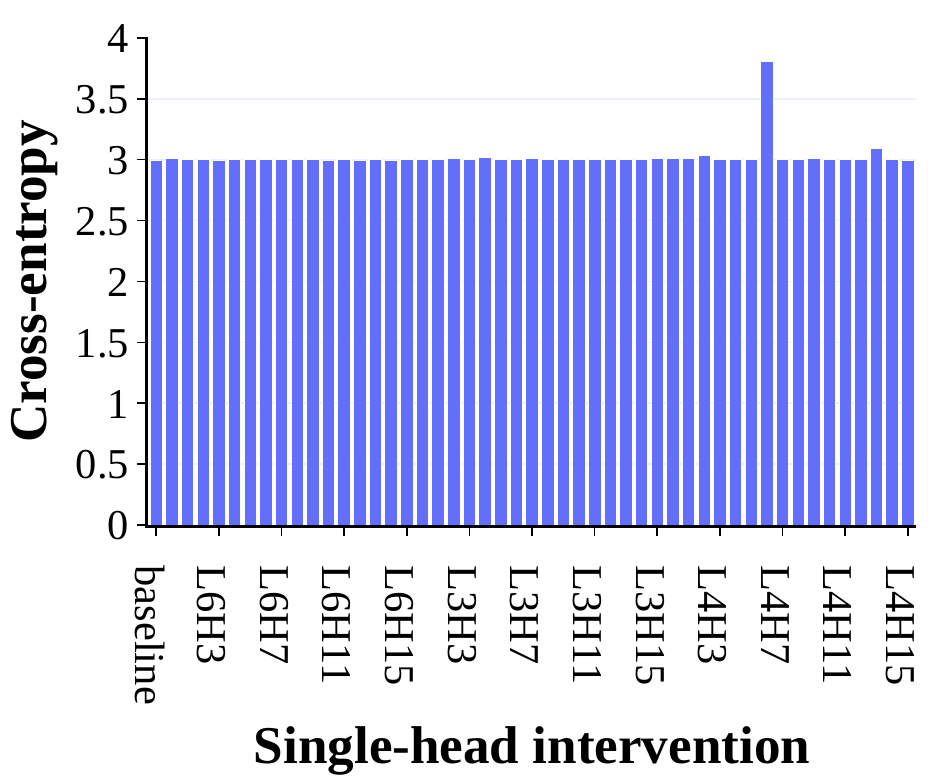}
    \caption{CE on control}
    \end{subfigure} \\
    \textbf{Pythia-1B} \\
    \begin{subfigure}{0.3\linewidth}
    \includegraphics[width=1\linewidth]{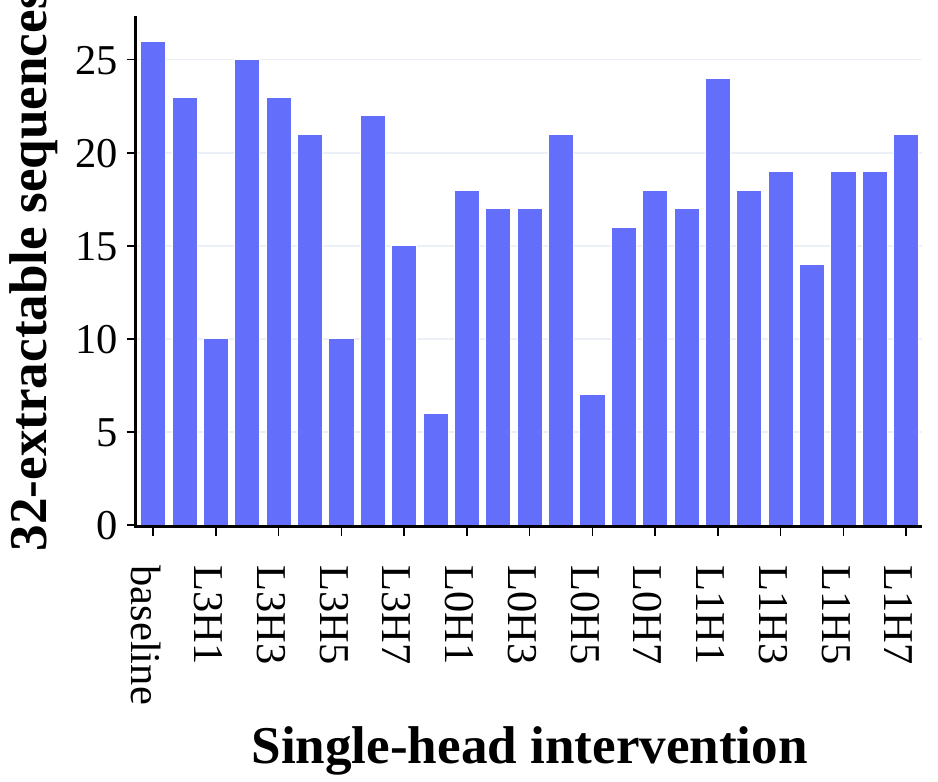}
    \caption{32-extr. recited}
    \end{subfigure}
    \begin{subfigure}{0.3\linewidth}
    \includegraphics[width=1\linewidth]{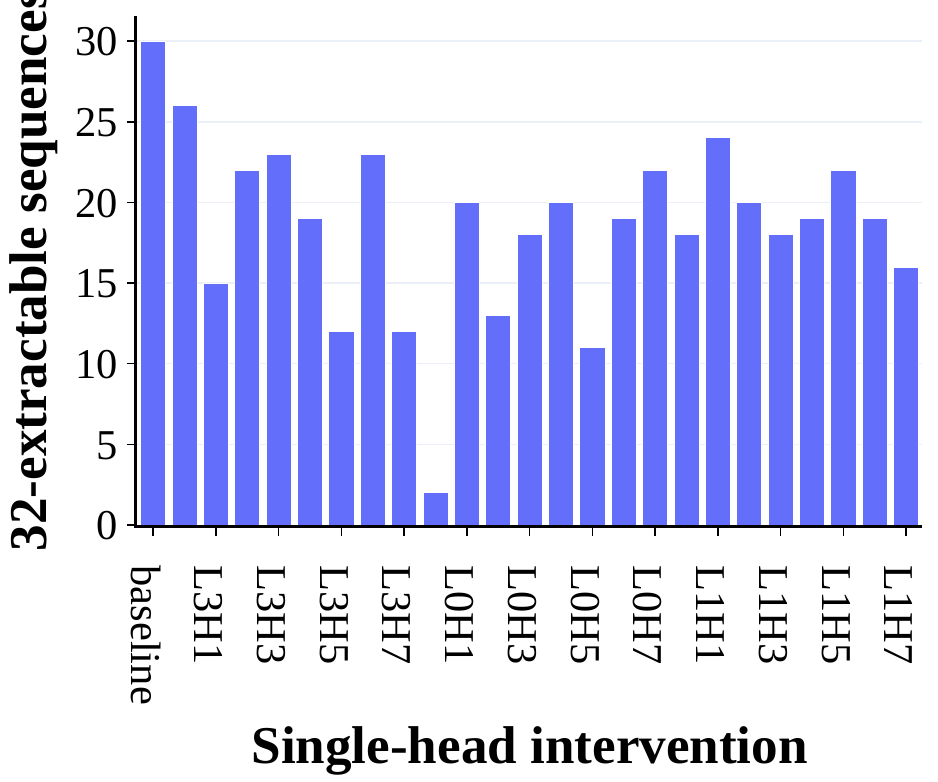}
    \caption{32-extr. recollected}
    \end{subfigure}
    \begin{subfigure}{0.3\linewidth}
    \includegraphics[width=1\linewidth]{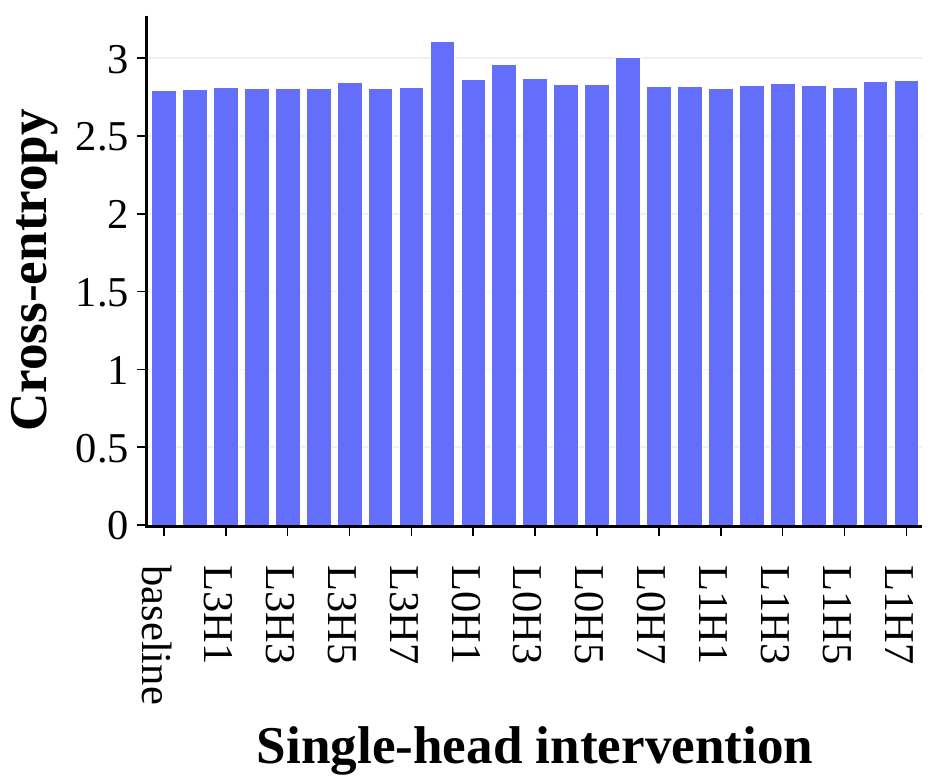}
    \caption{CE on control}
    \end{subfigure} \\
    \end{tabular}
    \caption{\textbf{Attention head ablation in Pythia (1/2).} We ablate each attention head, one by one, and plot number of 32-extractable memorized sequences and CE on control examples in intervened model. }
    \label{fig:attention-ablation}
\end{figure}

\textbf{Memorization in smaller models is ablated by a single attention head.} To further investigate the mechanism that could be implemented there, we turn to our parameter interventions. Due to our implementation, we do not have access to the decomposition on the level of single attention heads, but only entire attention layers. Therefore, one by one, we zero-intervene on the parameters of all the attention heads in the TOP-3 attention layers as ordered by their layer alignment. We observe 32-extractability of the memorized instances and cross-entropy of the control instances and plot the result in App. Fig.~\ref{fig:attention-ablation}. While most attention heads tend to have little influence on the memorization behavior of the memorized sequences we study, we find that ablating attention head $0$ in layer $0$ to have a big effect on memorization reducing the amount of 32-extractable sequences to $5$ and $1$ for the $64$ recited and recollected sequences respectively. At the same time, control cross-entropy increases by around $0.25$, leaving large parts of the model intact.

\begin{figure}
    \centering
    \begin{tabular}{c}
    \textbf{Pythia-160m} \\
    \begin{subfigure}{0.3\linewidth}
        \includegraphics[trim=0 0 0 0,clip,width=1.0\linewidth]{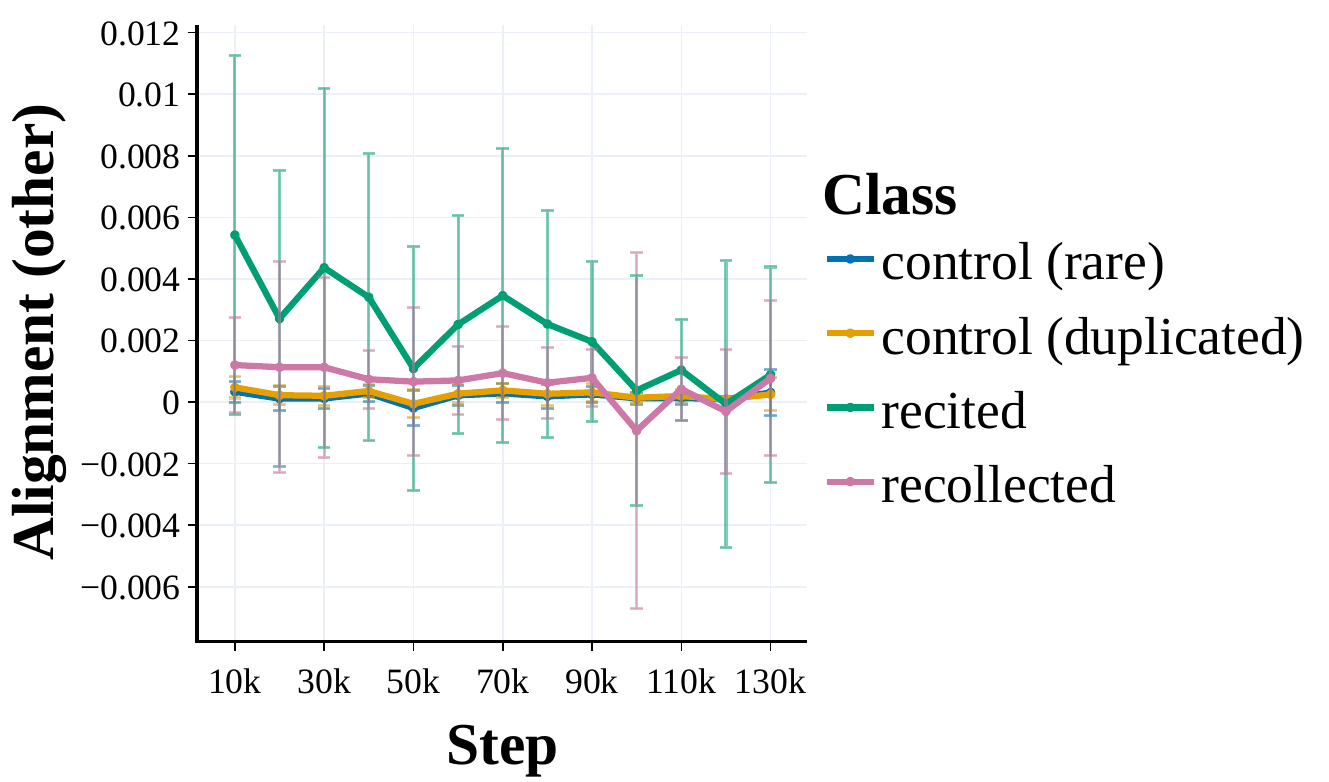}
    \end{subfigure}
    \begin{subfigure}{0.3\linewidth}
        \includegraphics[width=1.0\linewidth]{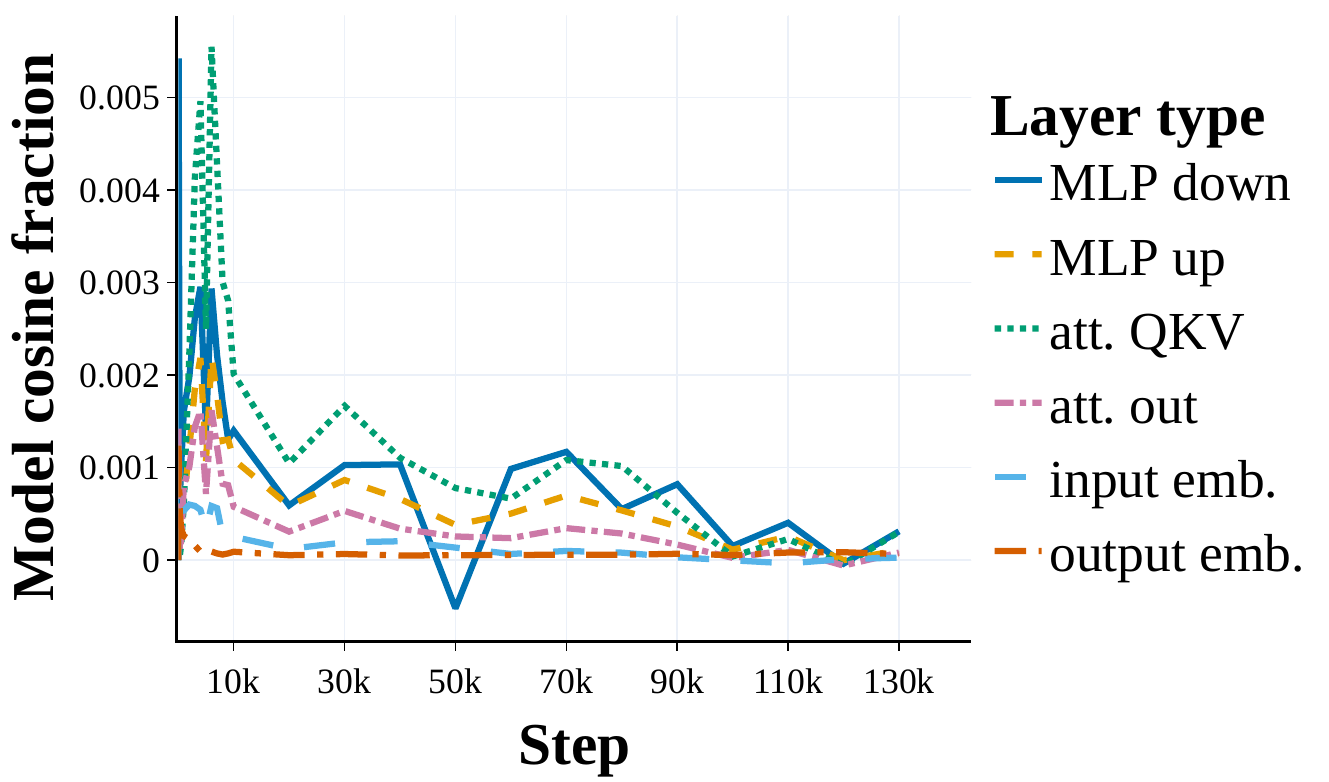}
    \end{subfigure}
    \begin{subfigure}{0.3\linewidth}
        \includegraphics[width=1.0\linewidth]{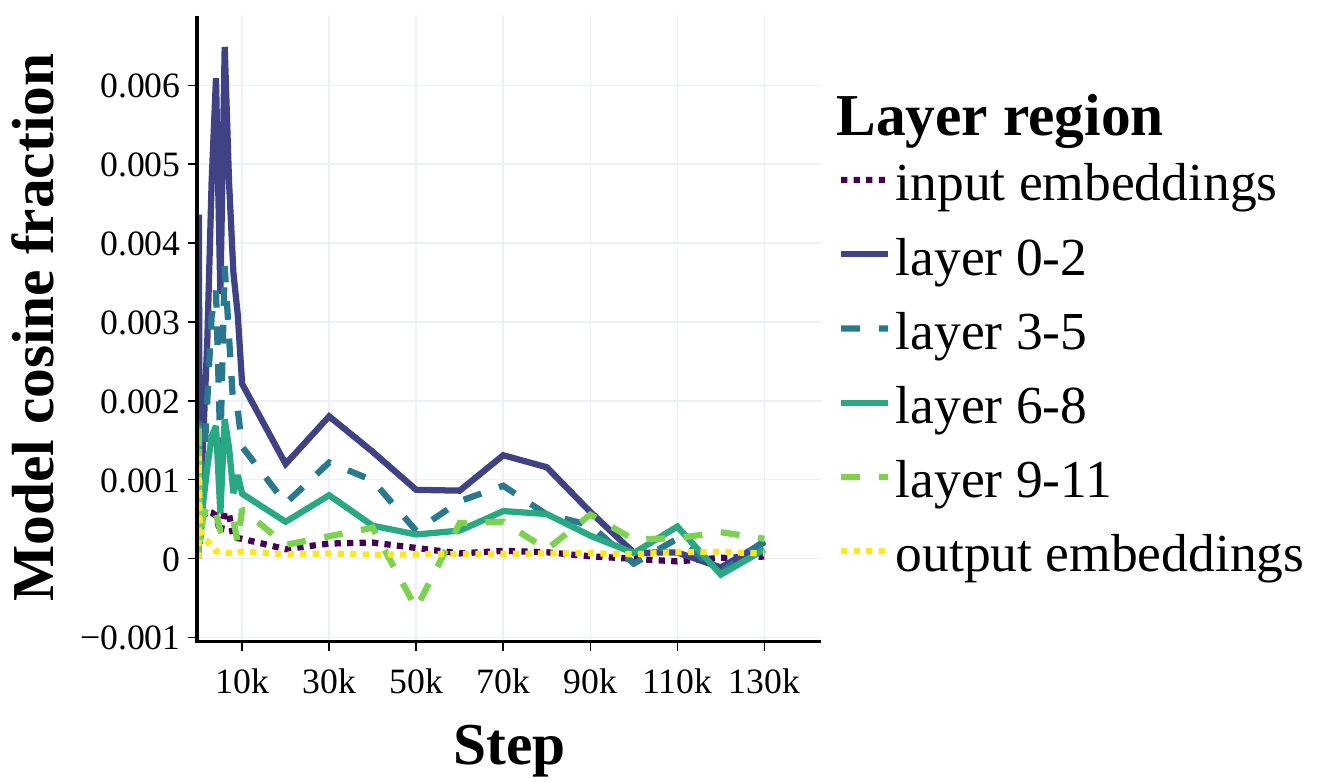}
    \end{subfigure} \\
    \textbf{Pythia-410m} \\
    \begin{subfigure}{0.3\linewidth}
        \includegraphics[trim=0 0 0 0,clip,width=1.0\linewidth]{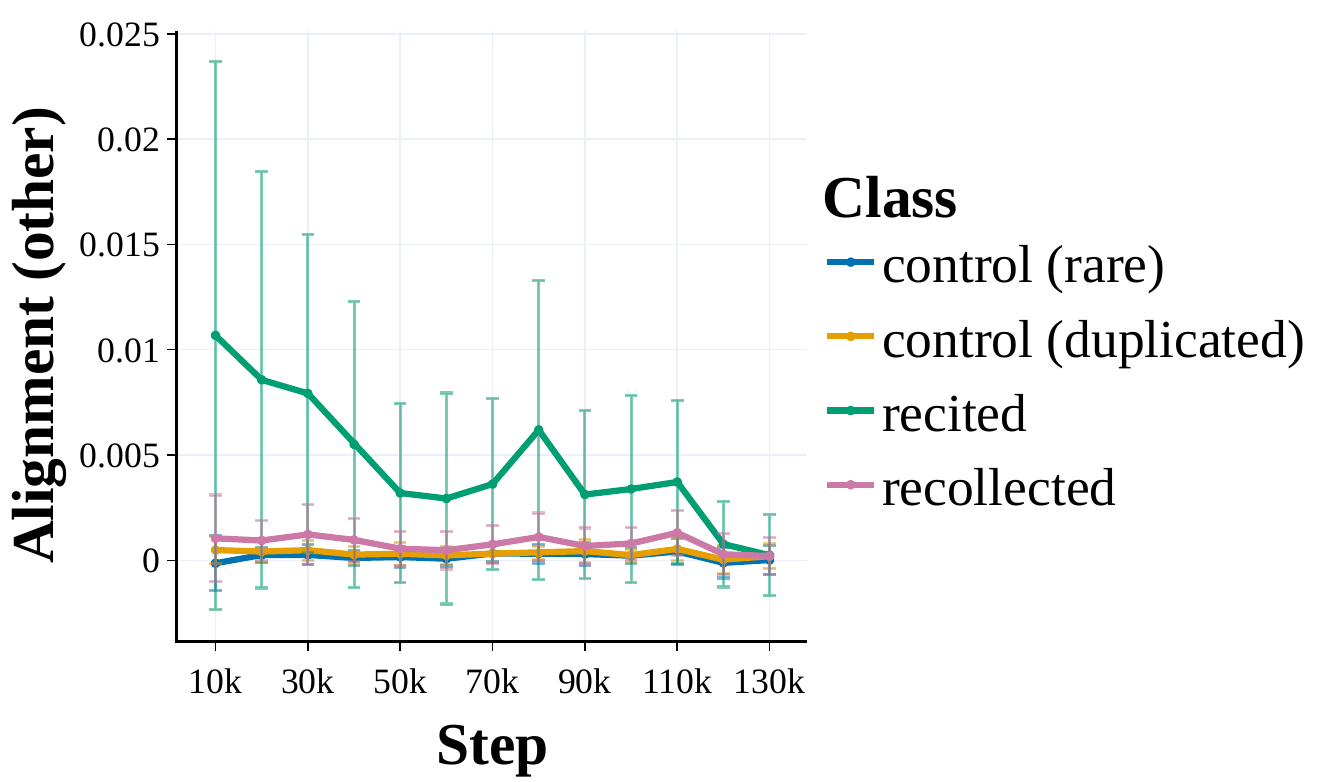}
    \end{subfigure}
    \begin{subfigure}{0.3\linewidth}
        \includegraphics[width=1.0\linewidth]{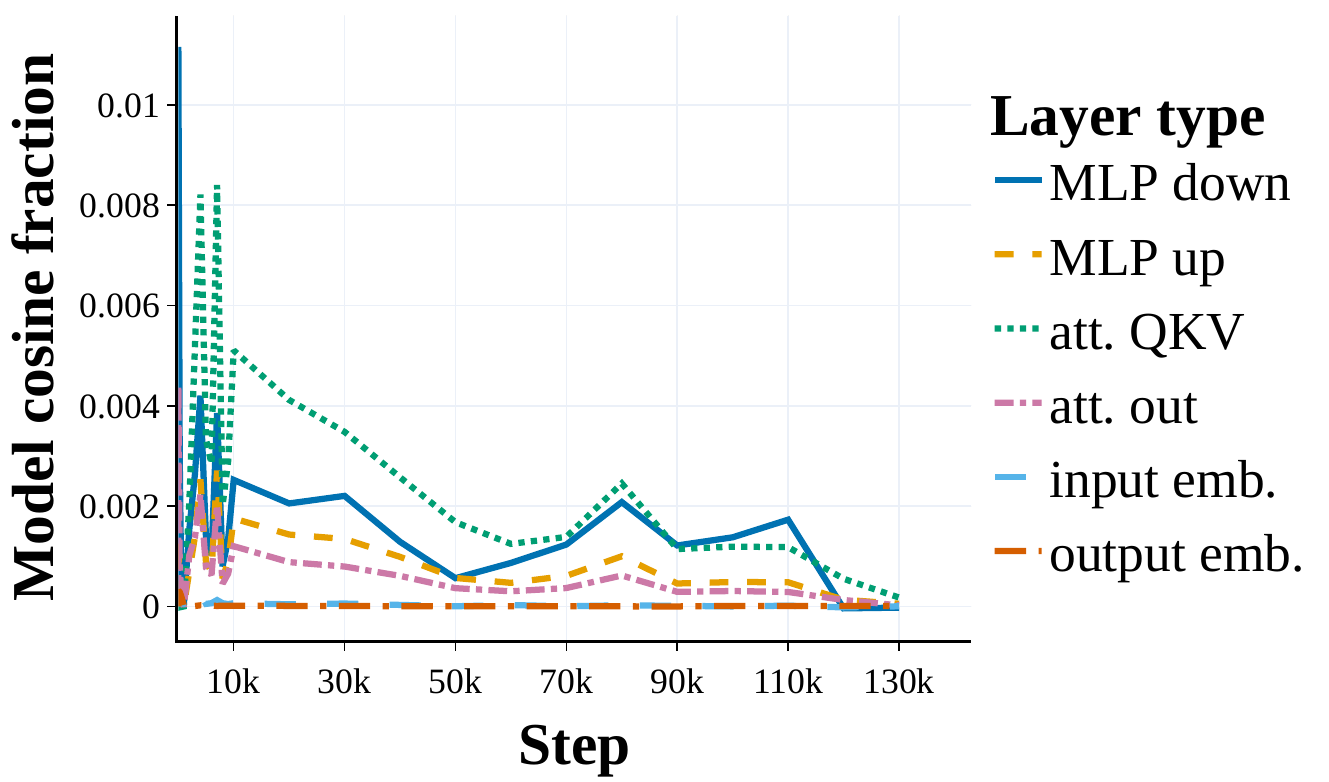}
    \end{subfigure}
    \begin{subfigure}{0.3\linewidth}
        \includegraphics[width=1.0\linewidth]{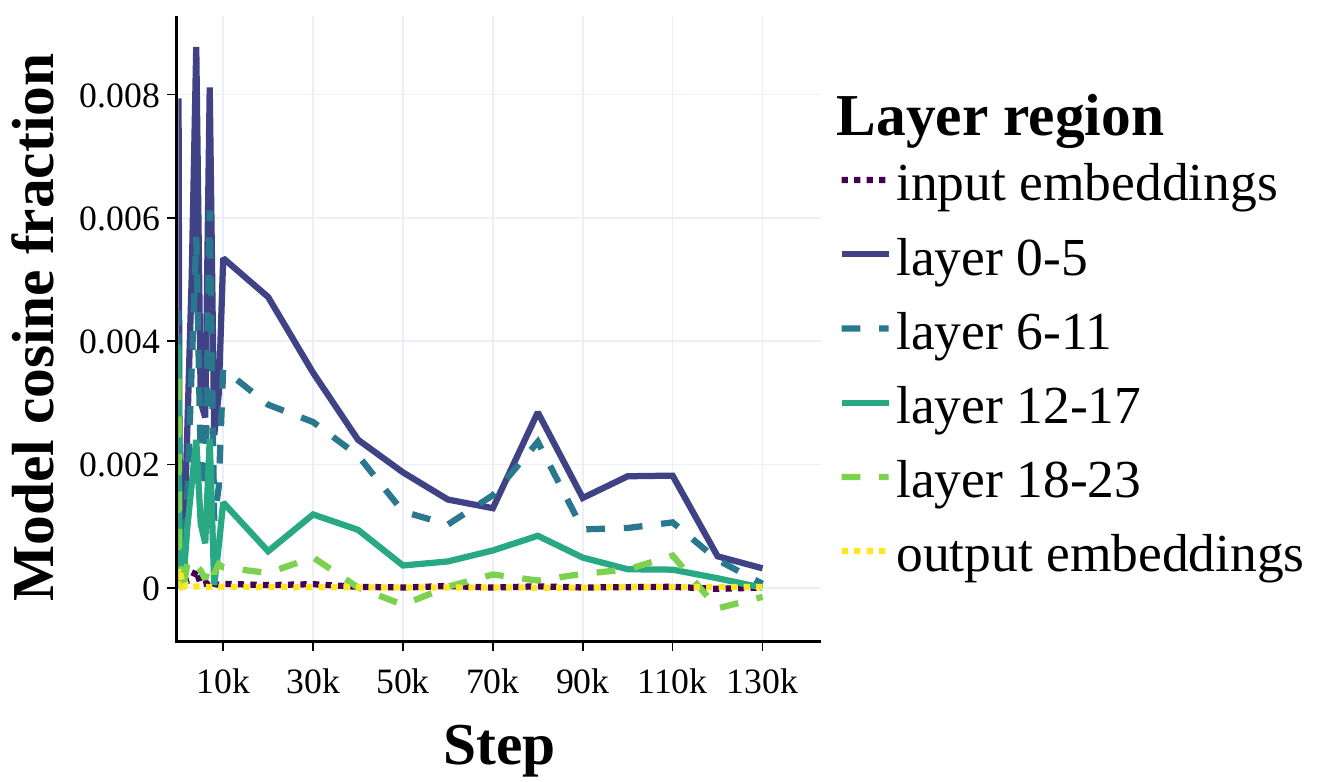}
    \end{subfigure} \\
    \textbf{Pythia-1B} \\
\end{tabular}
    \caption{\textbf{Alignment with other injected instances across scales.}}
    \label{fig:align-all-other}
\end{figure}

\begin{figure}
    \centering
    \begin{tabular}{c}
    \textbf{Pythia-1.4B} \\
    \begin{subfigure}{0.3\linewidth}
    \includegraphics[width=1\linewidth]{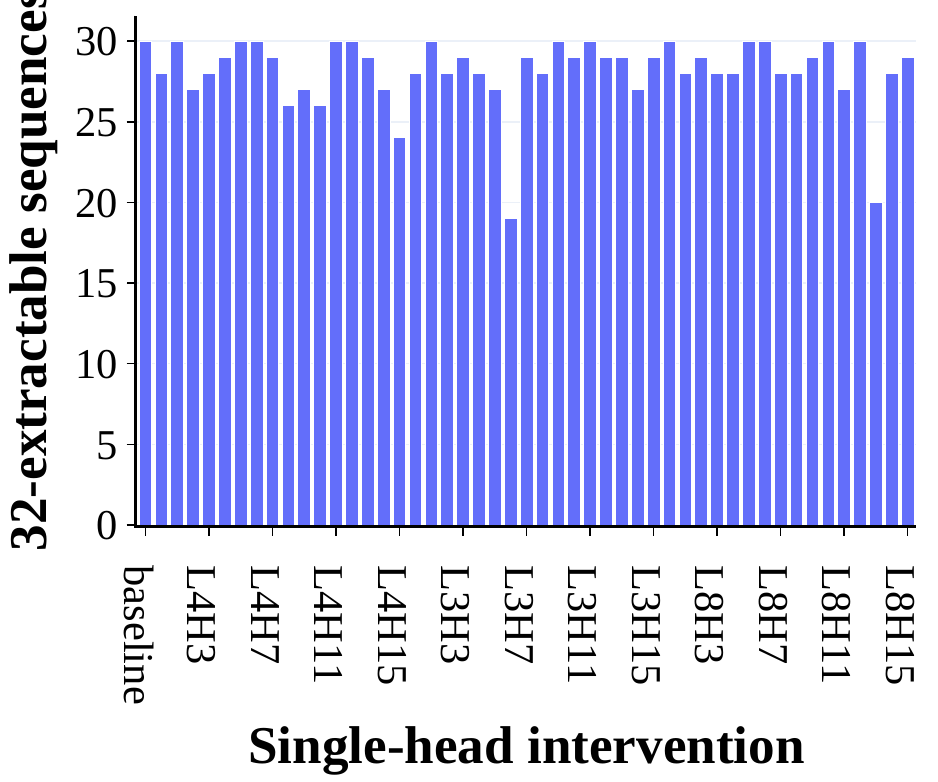}
    \caption{32-extr. recited}
    \end{subfigure}
    \begin{subfigure}{0.3\linewidth}
    \includegraphics[width=1\linewidth]{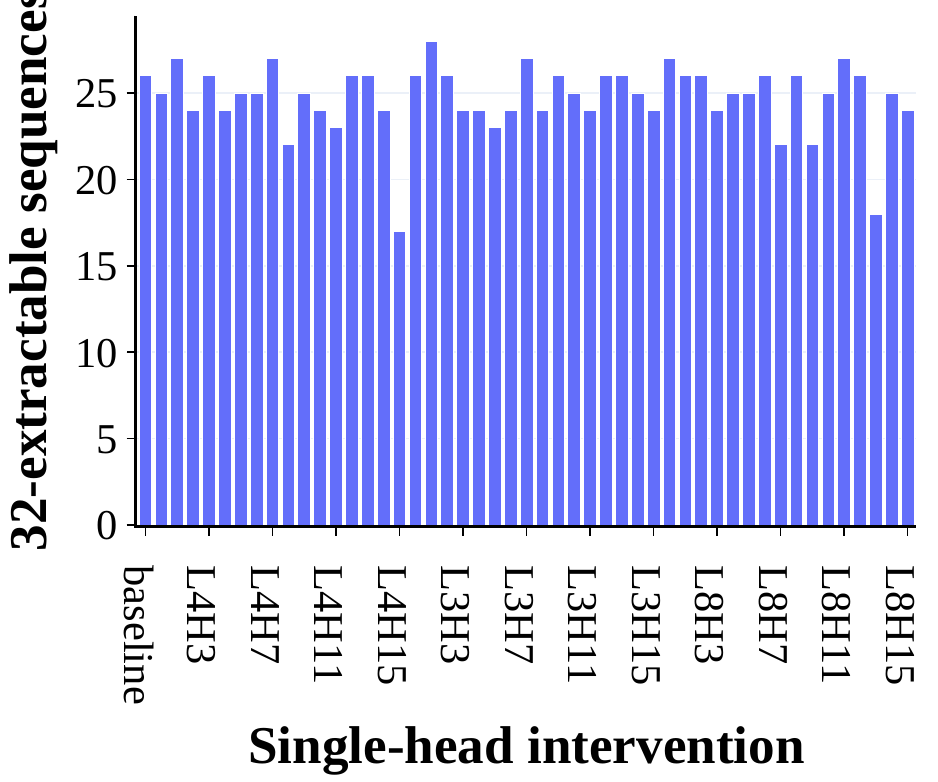}
    \caption{32-extr. recollected}
    \end{subfigure}
    \begin{subfigure}{0.3\linewidth}
    \includegraphics[width=1\linewidth]{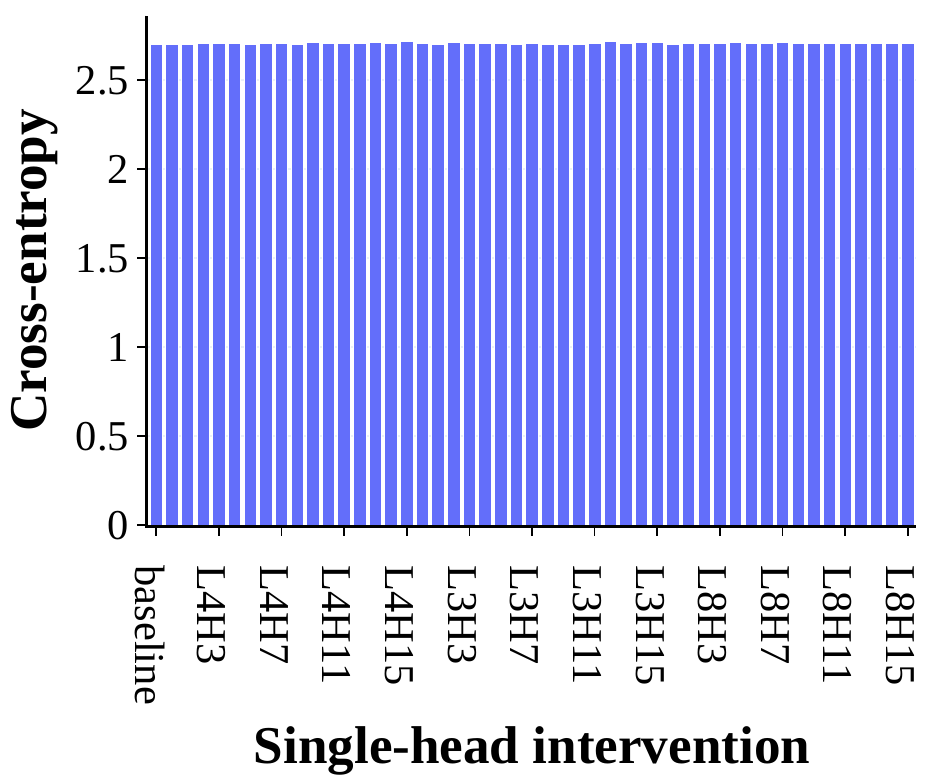}
    \caption{CE on control}
    \end{subfigure} \\
    \textbf{Pythia-2.8B} \\
    \begin{subfigure}{0.3\linewidth}
    \includegraphics[width=1\linewidth]{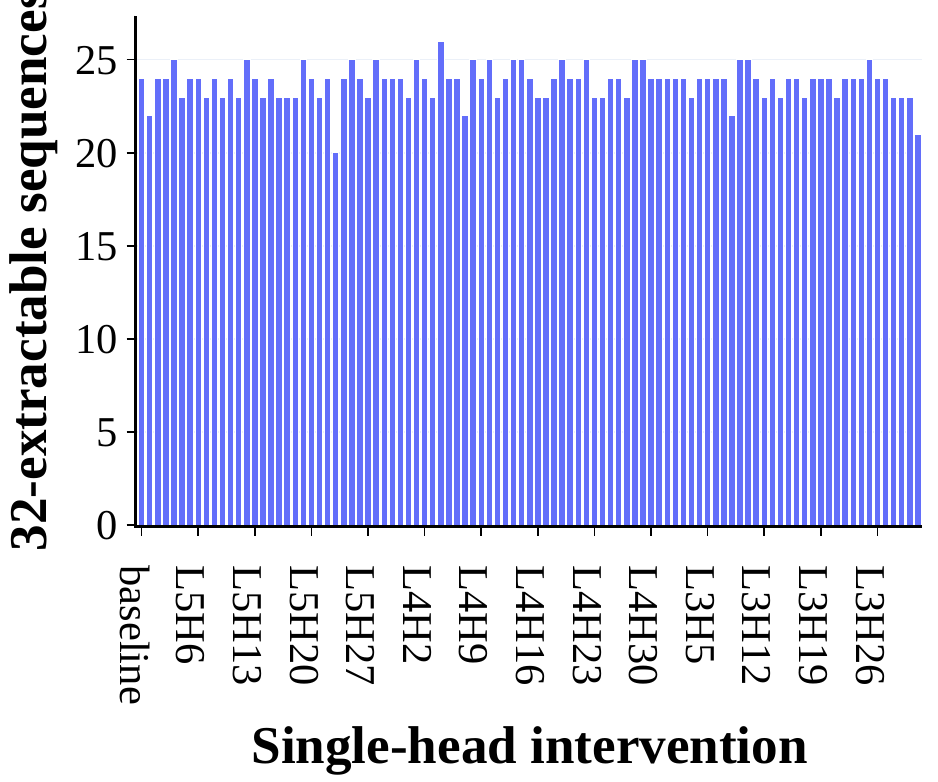}
    \caption{32-extr. recited}
    \end{subfigure}
    \begin{subfigure}{0.3\linewidth}
    \includegraphics[width=1\linewidth]{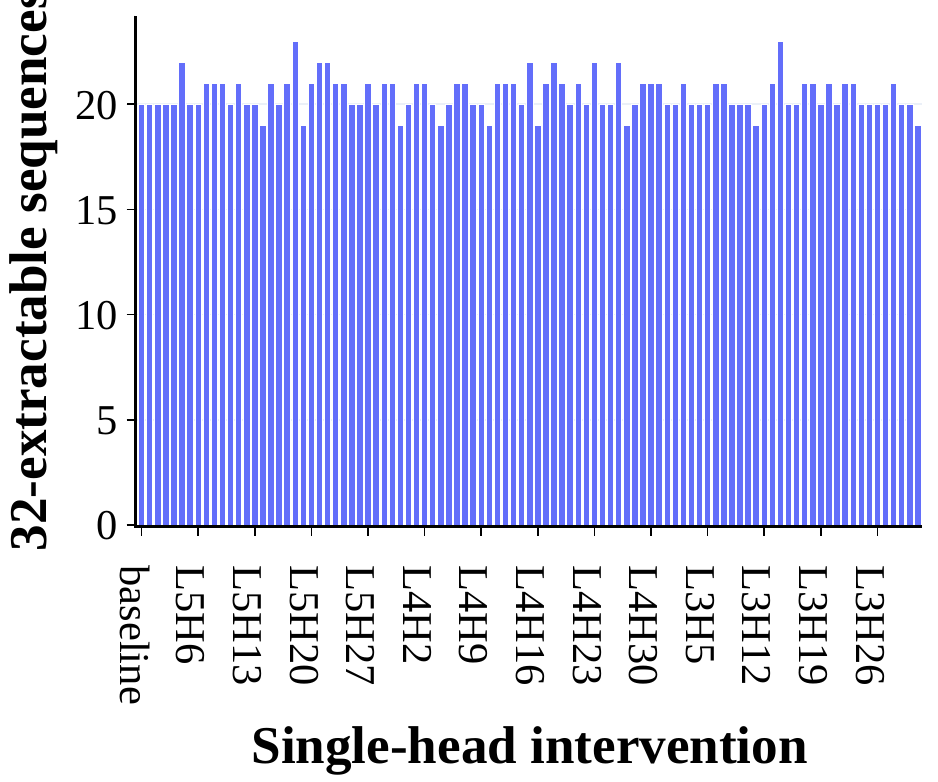}
    \caption{32-extr. recollected}
    \end{subfigure}
    \begin{subfigure}{0.3\linewidth}
    \includegraphics[width=1\linewidth]{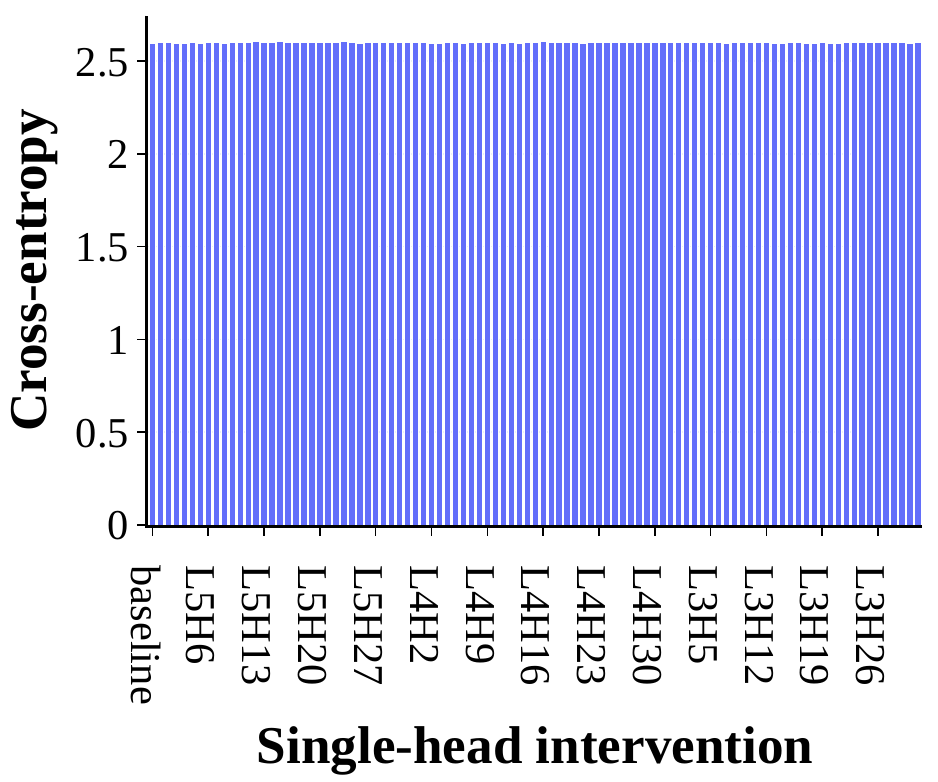}
    \caption{CE on control}
    \end{subfigure} \\
    \textbf{Pythia-6.9B} \\
    \begin{subfigure}{0.3\linewidth}
    \includegraphics[width=1\linewidth]{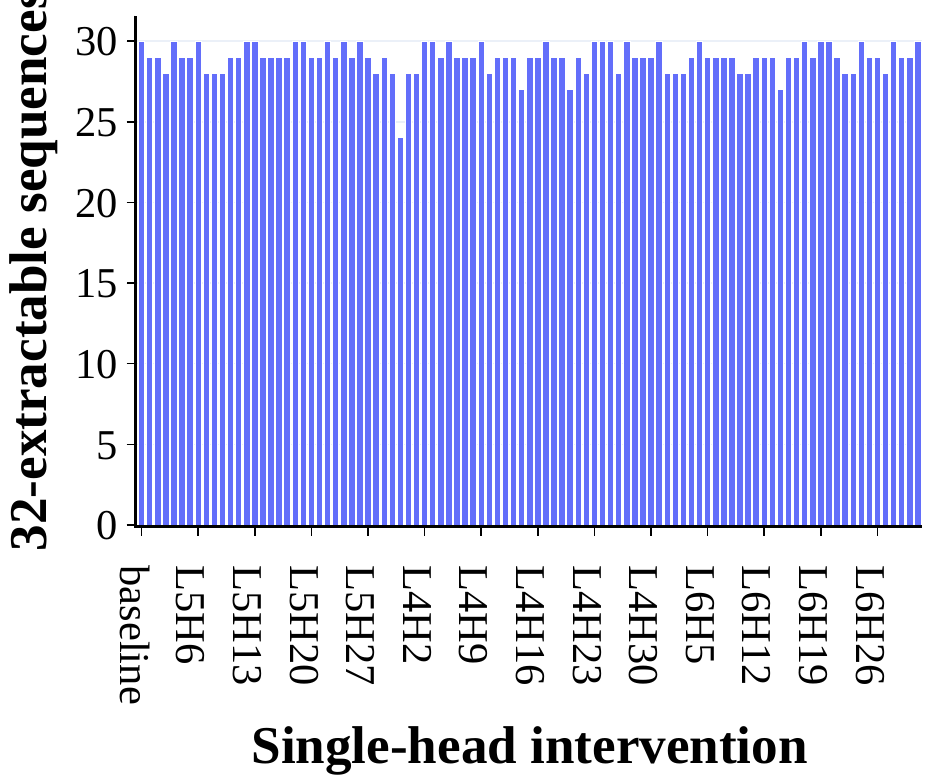}
    \caption{32-extr. recited}
    \end{subfigure}
    \begin{subfigure}{0.3\linewidth}
    \includegraphics[width=1\linewidth]{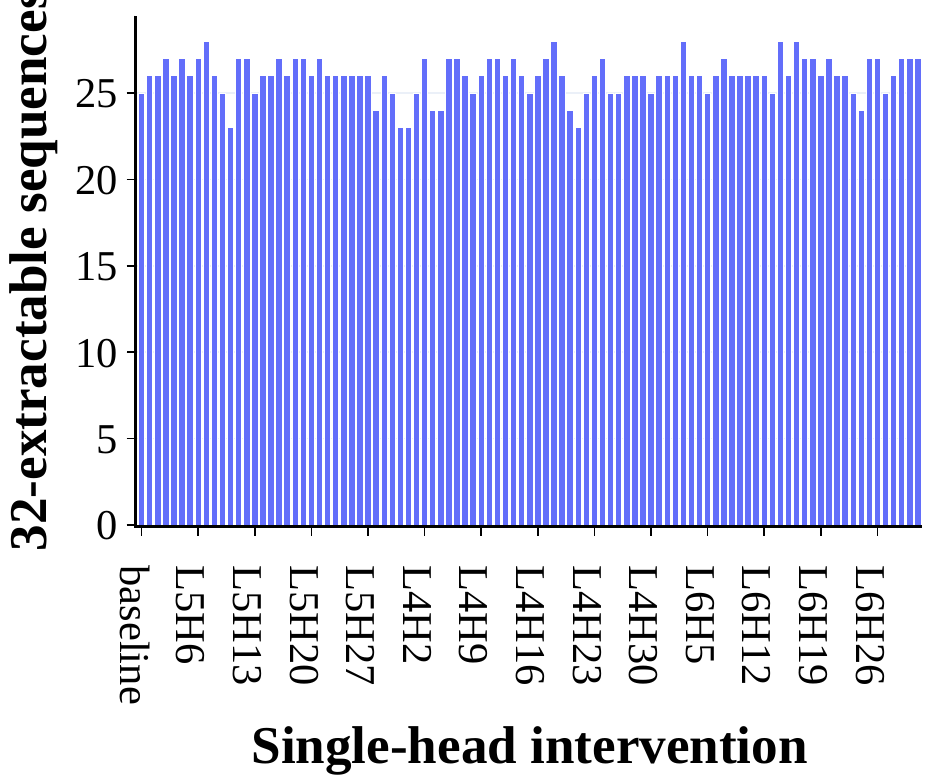}
    \caption{32-extr. recollected}
    \end{subfigure}
    \begin{subfigure}{0.3\linewidth}
    \includegraphics[width=1\linewidth]{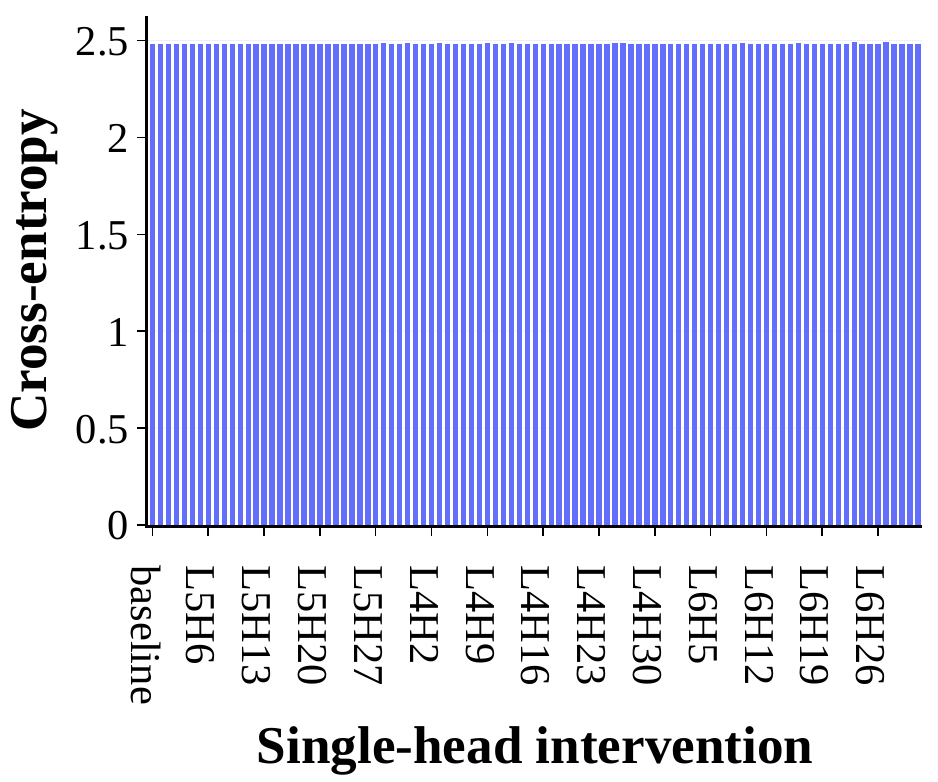}
    \caption{CE on control}
    \end{subfigure} \\
    \end{tabular}
    \caption{\textbf{Attention head ablation in Pythia (2/2).} We ablate each attention head, one by one, and plot number of 32-extractable memorized sequences and CE on control examples in intervened model. }
    \label{fig:attention-ablation2}
\end{figure}

This finding echos results presented by \citet{stoehrLocalizingParagraphMemorization2024}, who find that memorization in a small GPT-Neo model can also be ablated by fine-tuning a single attention head. Our result indicates, that this sub-mechanism for memorization develops early in training.%
Further, we show that this finding generalizes to much larger models around $1$B, indicating that it is not just a phenomenon in smaller, more constricted settings. On the other hand, our naive search and intervention did not yield the same result in the larger models starting from $1.4$B as can be seen in App. Fig.~\ref{fig:attention-ablation2}. There, the maximum attention alignment with other sequences moves upwards in the model (see App. Fig.~\ref{fig:align-all-other}), in sum suggesting a more refined and replicated mechanism there.

\newpage
\newpage

\subsection{Detailed decompositions of other Pythia models}

We plot decompositions analogous to the ones shown for Pythia-$1$B in the main paper.

\end{document}